\documentclass[journal]{IEEEtran}
\usepackage{latexsym}
\usepackage{graphicx}
\usepackage{amsmath}
\usepackage{amssymb}
\usepackage{epsfig}
\usepackage{subfigure}
\usepackage{mathrsfs}
\usepackage{setspace}
\usepackage{epstopdf}
\usepackage{cite}
\usepackage{bm}
\usepackage{algorithm}
\usepackage{algorithmic}
\usepackage{color}
\makeatletter

\newcommand{\Rmnum}[1]{\expandafter\@slowromancap\romannumeral #1@}
\makeatother

\ifCLASSINFOpdf
\else
\fi
\begin{document}
%

\title{Rain Removal By Image Quasi-Sparsity Priors}
%
%
%

\author{Yinglong~Wang,~\IEEEmembership{Student Member,~IEEE},
        Shuaicheng~Liu,~\IEEEmembership{Member,~IEEE},
        Chen~Chen,~\IEEEmembership{Student Member,~IEEE},
        Dehua~Xie,~\IEEEmembership{Student Member,~IEEE},
        and~Bing~Zeng,~\IEEEmembership{Fellow,~IEEE}
\thanks{Manuscript received xxx 2017.}
\thanks{Y. L. Wang, S. C. Liu, D. H. Xie and B. Zeng are with Institute of Image Processing, University of Electronic Science and Technology of China, Chengdu, Sichuan 611731, China.}
\thanks{C. Chen is with Department of Electronic and Computer Engineering, The Hong Kong University of Science and Technology, Kowloon, Hong Kong, China.}
\thanks{All correspondences to Y. L. Wang and B. Zeng (ylwanguestc@gmail.com, eezeng@uestc.edu.cn).}
}

%
%

\markboth{IEEE Transactions on Multimedia,~Vol.~XX, No.~XX, XX~2017}%
{Shell \MakeLowercase{\textit{et al.}}: Rain Removal By Image Quasi-sparsity Priors}
%



\maketitle

\begin{abstract}

   Rain streaks will inevitably be captured by some outdoor vision systems,
   which lowers the image visual quality and also interferes various computer vision applications.
   We present a novel rain removal method in this paper, which consists of two steps,
   i.e., detection of rain streaks and reconstruction of the rain-removed image.
   An accurate detection of rain streaks determines the quality of the overall performance.
   To this end, we first detect rain streaks according to pixel intensities,
   motivated by the observation that rain streaks often possess higher intensities
   compared to other neighboring image structures.
   Some mis-detected locations are then refined through a
   morphological processing and the principal component analysis
   (PCA) such that only locations corresponding to real rain streaks are retained.
   In the second step, we separate image gradients into a background layer
   and a rain streak layer, thanks to the image quasi-sparsity prior,
   so that a rain image can be decomposed into a background layer and a rain layer.
   We validate the effectiveness of our method through quantitative and qualitative evaluations.
   We show that our method can remove rain (even for some relatively bright rain)
   from images robustly and outperforms some state-of-the-art rain removal algorithms.

\end{abstract}

\begin{IEEEkeywords}
Rain removing, sparsity prior, rain detection, image decomposition, feature description, morphology.
\end{IEEEkeywords}

%
\IEEEpeerreviewmaketitle

\section{Introduction}
%
%
%
%
\IEEEPARstart{W}{ith} the development of computer vision techniques,
many learning, detection, matching, and tracking algorithms that are based on small features of images have appeared recently. However, many of these algorithms are quite sensitive to weather conditions under which the image is taken. In this work, we consider the image recovering with good visual quality from a single color image that is spotted by rain during the capturing.

Weather conditions can be classified into steady and dynamic according to the constituent particles \cite{Garg_2004_CVPR}. The former one contains small particles (e.g., fog) and the later one includes large particles (e.g., rain and snow). In the steady condition, small particles cannot be captured by cameras; while in the dynamic condition, droplets of rain and snow can be clearly filmed. He \emph{et al.} proposed a de-haze approach that is based on dark channel priors and has achieved excellent results on various challenging examples \cite{He_2011_PAMI}. Another fast image dehazing work which based on linear transformation are proposed by Wang \emph{et al.} \cite{Wang_2017_TMM}. However, in the case of dynamic weather, the existing rain removal methods still need to be improved. The difficulty lies in two aspects: rain droplets appear in an image randomly and large droplets interfere original image contents.

The earliest work on rain dates back to the study of statistical characteristics of rain in the atmospheric science in 1948 \cite{Marshall_1948_JM}. According to these characteristics,
rain appears in a picture looks quite random and is of different shapes,  which makes it difficult to detect and remove rain streaks from a single image. Therefore, most works pay attentions to rain removal in videos
\cite{Garg_2004_CVPR,Zhang_2006_ICME,Brewer_2008_CS,Bossu_2011_CV,Barnum_2007_PACV,Barnum_2010_CV}, where the rain detection is relatively easier. For example, Barnum \emph{et al.} detected and removed rain streaks for videos in the frequency domain \cite{Barnum_2007_PACV,Barnum_2010_CV}. To the best of our knowledge, dealing with rain removing in a single image started in 2009 when Roser \emph{et al.} detected rain streaks in single image \cite{Roser_2009_CV}. Later on, several other rain removal works based on a single image were proposed, e.g., \cite{Fu_2011_ASSP,Kang_2012_TIP,Chen_2014_CSVT,Xu_2012_CIS,Kim_2013_ICIP,Ding_2015_MTA,Huang_2014_TMM}.

This paper aims at removing rain from a single image. Although the detection and removal of rain in a single image is more challenging as compared with videos, there are still some observations that can be utilized for the rain identification. On one hand, rain streaks are more reflective than other parts of images, leading to higher pixel intensities compared with non-rain pixels. On the other hand, rain streaks usually do not occlude objects completely due to their semi-transparency property. The former one can be utilized for the rain detection while the latter one facilitates the reconstruction in the gradient domain after computing image gradients on all non-rain locations.

There are two main challenges: accurate rain streaks identification and high quality rain-removed image recovery. The detection of rain streaks will not be accurate if only pixel intensities are involved, because other objects with similar or even higher pixel intensities would be mis-classified as rain pixels. After the rain pixels are identified, the final result needs to be reconstructed without introducing noticeable artifacts (e.g., blurring of image contents). Methods such as the weighted mean of neighbouring non-rain pixels and image inpanting~\cite{Bertalmio_2000_CGIT} are all possible candidates that have been considered previously.

In this paper, in order to deal with the first challenge, we over-detect rain pixels in single rain image by the
method in \cite{Wang_2016_ICIP} firstly. To improve the accuracy, we further employ a morphological processing technique\cite{Gonzalez_2002_PUSR} to refine all detected rain pixels. For the second challenge, we decompose the input image into a rain-layer and a non-rain layer in the gradient domain after the rain pixels are identified. We reconstruct the final result by using the non-rain layer under the image quasi-sparsity priors.

\begin{figure}[t]
\begin{minipage}{0.48\linewidth}
\centering{\includegraphics[width=1\linewidth]{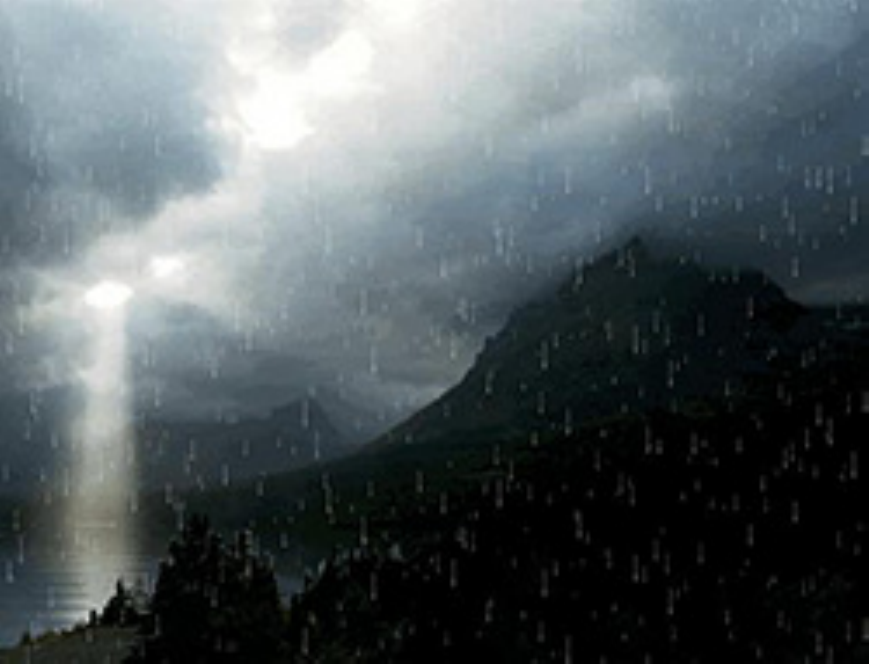}}
\centerline{(a)}
\end{minipage}
\hfill
\begin{minipage}{.48\linewidth}
\centering{\includegraphics[width=1\linewidth]{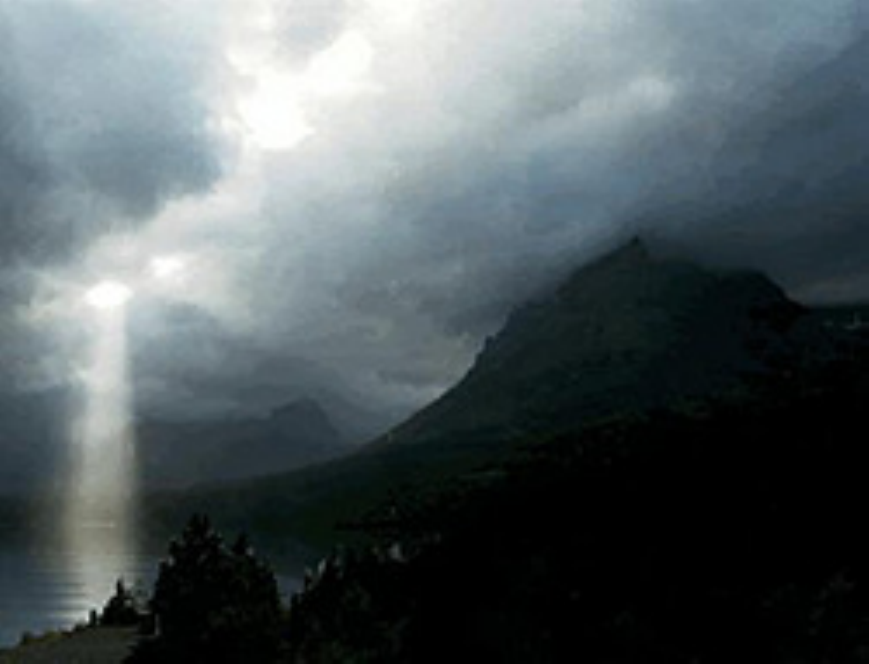}}
\centerline{(b)}
\end{minipage}
\caption{(a) Original rain image. (b) Rain-removed image by our method.}
\label{fig:rain_derain}
\end{figure}

The contributions of our work are the following aspects. (1) We propose a quite unique detection method of rain streaks. (2) We simplify the sparsity \cite{Levin_2007_PAMI} to quasi-sparsity and combine it with the detection of rain to complete rain-removal tasks. After simplifying the sparsity to quasi-sparsity, the loss function is derived into a $L_{1}$-norm minimization problem.  (3) An additional constraint is added to solve the color shift problem that often appears in \cite{Levin_2007_PAMI} successfully. An example of our rain-removed results is shown in Fig. \ref{fig:rain_derain}.

The rest of the paper is organized as follows. We briefly review some related works in Section \ref{sec:RelatedWorks}. We propose the rain streaks detection algorithm in Section \ref{sec:RainStreaksDetection}. The reconstruction of rain-removed images is presented in Section \ref{sec:ImageReconstruction}. The results and comparisons are presented and discussed in Section \ref{sec:ExperimentalResults}. Finally, some conclusions are drawn in Section \ref{sec:Conclusion}.


%

\section{Related Works}
\label{sec:RelatedWorks}

Rain removal can be performed in the spatial domain or the frequency domain, and some are focused on the single-image scenario. A brief review of the existing algorithms is presented in the following.

\textbf{Rain removal from videos in the spatial domain:} Garg and Nayar analyzed the visual effect of rain streaks comprehensively \cite{Garg_2004_CVPR} by developing a correlation model to describe rain's dynamics and a motion blur model to explain the photometry of rain. Through these two models, rain streaks can be detected efficiently and then removed in videos. To make the study more complete, Garg and Nayar further built a rain appearance model based on a rain oscillation model that was developed in the atmospheric science in \cite{Garg_2006_TG}. They also developed an image-based rain-rendering algorithm by creating a database to describe different kinds of rain appearances under various lighting and viewing directions. In \cite{Garg_2007_CV}, Garg and Nayar analyzed various factors that influence the visual effect of rain. Based on these analyses, an efficient algorithm was developed to control rain. Besides, by modeling the distortion of raindrop, they accomplished photorealistic rain-rendering.

Another rain removal algorithm that is based on both temporal and chromatic characteristics of rain streaks in video was proposed by Zhang \emph{et al.} \cite{Zhang_2006_ICME}. This work shows that a certain area is not always infected by rain streaks. On the other hand, when indeed affected by rain, the intensity changes of chromatic components (namely, R, G, B) of a pixel approximately equal to each other. These two properties have been utilized to detect and then remove rain streaks in videos. However, constrained by temporal properties, this method can only deal with the videos that are obtained by using a stationary camera.

In \cite{Brewer_2008_CS}, Brewer and Liu suggested that (1) a region with instantaneous intensity spike be probably affected by rain streaks and (2) streak-like objects in a region with a nearly consistent range of aspect ratios be considered as rain streaks. Once detected, rain streaks can be removed by calculating the mean value of two neighbouring frames. A rain streaks detection method that uses a histogram of orientation of streaks (HOS) was introduced by Bossu \emph{et al.} in \cite{Bossu_2011_CV}. This method proposes to decompose an image sequence into foreground and background, while potential rain streaks are detected in foreground. Then, HOS is calculated, which follows a model of
Gaussian-uniform mixture. Finally, the Gaussian distribution whose amplitude stands for rain presence and the uniform distribution standing for noise are separated by an algorithm of expectation maximization.

\textbf{Rain removal from videos in the frequency domain:}
In \cite{Barnum_2007_PACV}, Barnum \emph{et al.} combined
a physical model of rain streaks (for determining the general shape and brightness of rain) and some statistical properties of rain streaks to show the influence of rain on image sequences in the frequency domain.
Once detected, the spectrum of rain streaks can be suppressed to obtain rain-removed image sequences. Later on, they combined a shape model
with statistical properties of rain streaks to detect and remove rain
streaks, also in the frequency domain, and demonstrated a better accuracy  \cite{Barnum_2010_CV}.

\begin{figure*}
\centering
\begin{minipage}{1\linewidth}
\centering{\includegraphics[width=.9\linewidth]{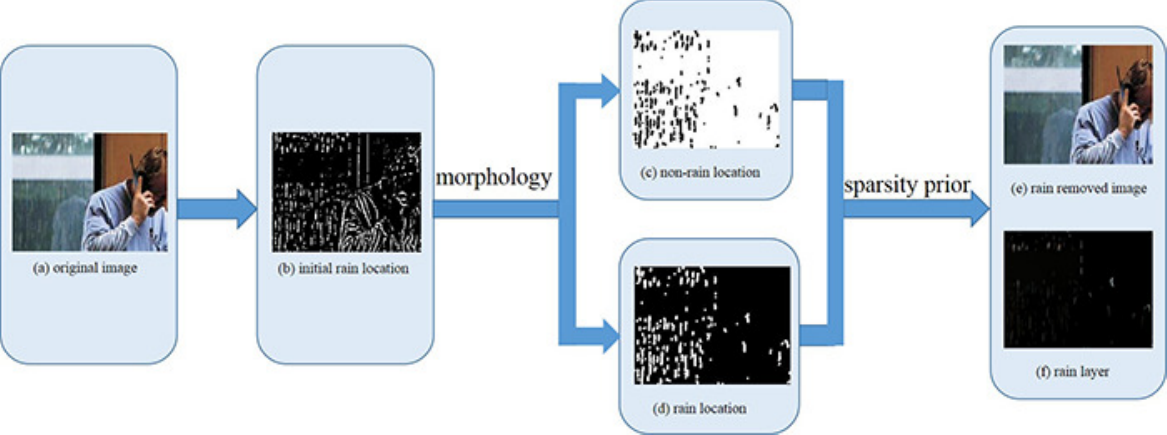}}
\centerline{}
\end{minipage}
\caption{Pipeline of our method. We first identify rain locations from the input rain image (a) to generate an initial rain map (b). Then, a morphological processing is used to refine these initial locations, producing a new non-rain map (c) and rain locations (d) (white pixels). The final result (e) and optionally a rain layer (f) can be reconstructed based on (c) and (d), respectively.}
\label{fig:pipeline}
\end{figure*}

\textbf{Single image rain removal:} Roser \emph{et al.} detected rain streaks in a single image monocularly, based on a photometric raindrop model \cite{Roser_2009_CV}. Meanwhile, Halimeh \emph{et al.} detected raindrops on car windshield by utilizing a model that describes the shape of raindrop and a relationship between raindrops and the environment \cite{Halimeh_2009_VS}.

For the first time, Fu \emph{et al.} accomplished the rain-removal task for a single image by utilizing morphological component analysis (MCA) \cite{Fu_2011_ASSP}. Some improved or extended versions have been proposed by Kang \emph{et al.} \cite{Kang_2012_TIP} ,Chen \emph{et al.} \cite{Chen_2014_CSVT}, Wang \emph{et al.} \cite{Wang_2016_ICIP} and Wang \emph{et al.}
\cite{Wang_2017_TIP}. In particular, Kang \emph{et al.} used the histogram of oriented gradients (HOG) \cite{Dalal_2005_CVPR} to separate rain and non-rain dictionary atoms, while Chen \emph{et al.} extended rain removal task to a single color image.
The denosing paper \cite{Huang_2014_TMM} on TMM which treats rain as a kind of noise removes rain streak by a self-learning based image decomposition method.
In \cite{Wang_2016_ICIP,Wang_2017_TIP}, Wang \emph{et al.} developed a rough detection method of rain to remove bright rain streaks.

More recently, Luo \emph{et al.} proposed that a rain image be decomposed into the rain layer and non-rain layer by a highly discriminative code on a learned dictionary that is based on a screen blend model \cite{Luo_2015_ICCV}. On the other hand, a novel rain removal method based on the guided filter was proposed by Xu \emph{et al.} \cite{Xu_2012_CIS}, in which a rain-free guidance image is constructed and a guided filter \cite{He_2013_PAMI} is used to remove rain in a single image. In \cite{Kim_2013_ICIP}, Kim \emph{et al.} assumed that rain streaks have an elliptical shape. Then, a kernel regression method \cite{Takeda_2007_TIP} is used to extract elliptical components in the image to detect rain streaks. Once detected, rain streaks are removed by non-local mean filter \cite{Buades_2005_CVPR}. In the meantime, Chen \emph{et al.} proposed a low-rank model of rain streaks to capture the spatio-temporally correlated rain streaks \cite{Chen_2013_ICCV}.

Lately, a rain removal method based on the $L_0$ gradient minimization was proposed by Ding \emph{et al.} \cite{Ding_2015_MTA}. By this method, majority of rain streaks can be restrained, but a lot of image details also vanish with the rain streaks removal. Another novel rain removal method was developed by Li \emph{et al.}  \cite{Li_2016_CVPR}, in which some patch-based priors for both the background layer and rain layer are used to accomplish the rain removal task. Because these priors are based on Gaussian mixture models and can accommodate the rain streaks with multiple orientations and scales, this method obtains the state-of-the-art effectiveness.

\textbf{Deep learning based methods:} In recent years, deep learning is utilized in many computer vision tasks, including rain
removal. In \cite{Fu_2017_TIP}, Fu \emph{et al.} designed a DerainNet to learn the mapping relationship between rain and clean images. They also proposed a deep detail network which
can directly reduces the mapping range to simplify the learning
process, then remove rain streaks in single color images \cite{Fu_2017_CVPR}. Yang \emph{et al.} built a new model for rain images and designed a multi-task deep learning architecture
to remove rain streaks in single images \cite{Yang_2017_CVPR}.
A DID-MDN net tried to estimate the density of rain
first, then remove rain streaks \cite{Zhang_2018_CVPR}.

\section{Rain Streaks Detection}
\label{sec:RainStreaksDetection}

Fig. \ref{fig:pipeline} shows the pipeline of our method.
Given an input rain image (Fig. \ref{fig:pipeline}(a)), we first detect the rain
locations according to pixel intensities. Since the initial
locations (Fig. \ref{fig:pipeline}(b)) are usually inaccurate,
they will be refined using the proposed morphology approach, generating a
refined rain location map (Fig. \ref{fig:pipeline}(d)) as well as a non-rain location map (Fig. \ref{fig:pipeline}(c)).
The final result (Fig. \ref{fig:pipeline}(e)) is reconstructed from the image gradients
on all non-rain locations.
Optionally, we can also reconstruct an image that contains rain only (Fig. \ref{fig:pipeline}(f)).

In this section, we utilize the rain image in Fig. \ref{fig:pipeline}(a)
as an example to present the details of rain streaks detection.
Firstly, an initial rain location is obtained by the method in \cite{Wang_2016_ICIP}.
Then, the mis-detections are refined by a morphological processing and the principal component analysis (PCA).

\subsection{Initial detection of rain streaks}

Rain pixels often possess higher values than
their neighbouring non-rain pixels. Therefore, Wang \emph{et al.} \cite{Wang_2016_ICIP}
over-detected rain location by this characteristic.

\begin{figure}[t]
\centering
\begin{minipage}{0.48\linewidth}
\centering{\includegraphics[width=.9\linewidth]{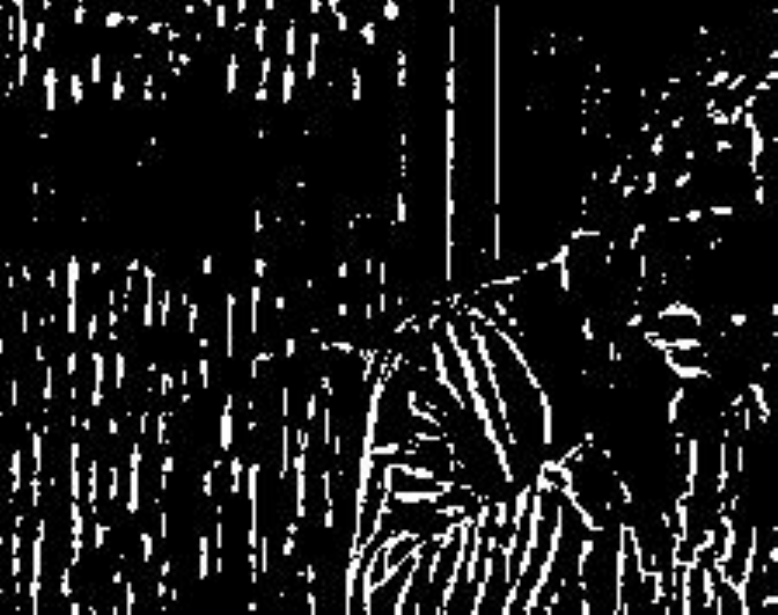}}
\centerline{(a)}
\end{minipage}
\caption{Initial rain's location map.}
\label{fig:initial_location}
\end{figure}

For each pixel $I(i,j)$ in a given rain image $I$,
Wang \emph{et al.} calculate 5 mean values $\bar{I}^{(k)}$ $(k=1,2,3,4,5)$
in the windows $w^{(k)}$ with pixel $I(i,j)$ located in the center,
top-left, top-right, bottom-left, and bottom-right of the window, respectively.
If the following inequalities
\begin{equation}\label{eq:detect_condition}
 I(i, j)>\bar{I}^{(k)}=\frac{\sum_{\{m,n\}\in w^{(k)}}I(m,n)}{|w^{(k)}|}, k\!=\!\{1,2,3,4,5\},
\end{equation}
where $|w^{(k)}|$ stands for the window size,
are satisfied for all color channels, $I(i,j)$ is recognized as a rain pixel
and the corresponding term $S_R(i,j)$ in the so-called binary location map $S_R$ is set to be 1;
otherwise $S_R(i,j)$ is assigned as 0.
Detection result $S_{R}$ is a binary image as shown in Fig. \ref{fig:initial_location}.

\subsection{An analysis of mis-detections}

It can be seen from Fig. \ref{fig:initial_location}
that not only rain streaks but also some non-rain components
appear in rain detection result. How to recognize those
non-rain components and eliminate their influence is
thus very critical - a lot of image details and useful
information would otherwise get lost after the removal of rain streaks.

In order to separate rain from non-rain objects,
some characteristics of rain can be useful.
We describe them as follows:
\begin{itemize}
\item rain streaks usually do not have too large size in width,
\item the directions of all rain streaks in a scene are nearly consistent,
\item the color of a rain streak is usually shallow white, and
\item the length of a rain streak is usually larger than its width.
\end{itemize}
These characteristics are very robust to describe rain, and
some of them have been utilized in some existing rain-removal
works, such as \cite{Chen_2014_CSVT}, \cite{Kim_2013_ICIP} and so on.
Later on, we will see that when these characteristics are
combined with our proposed morphological processing,
the error detection will be reduced largely.

\subsection{Refining of initial locations of rain streaks}

\noindent \textbf{First}, all connected components shown in Fig. \ref{fig:initial_location}
are extracted by the morphology method and the details can be referred to \cite{Gonzalez_2002_PUSR}.

\begin{figure}[t]
\begin{minipage}{0.48\linewidth}
\centering{\includegraphics[width=.9\linewidth]{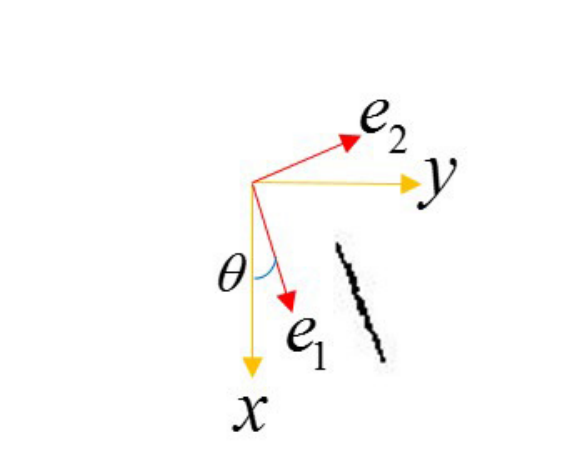}}
\centerline{(a)}
\end{minipage}
\hfill
\begin{minipage}{.48\linewidth}
\centering{\includegraphics[width=.9\linewidth]{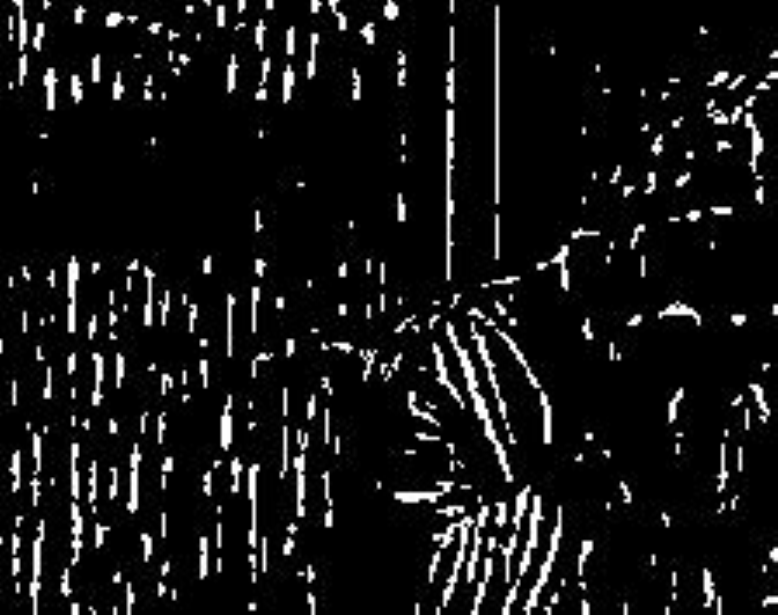}}
\centerline{(b)}
\end{minipage}
\caption{(a) An example of PCA description. (b) Refined result by connected component width.}
\label{fig:single_streak}
\end{figure}

\noindent \textbf{Second}, PCA is used to describe the shape of every connected component.
In order to describe this step more visually, we select one connected component
from Fig. \ref{fig:initial_location} as an example
to show the refining process, the selected component is in Fig. \ref{fig:single_streak}(a).
Because some colors can not be seen clearly on a black background,
we have changed the selected streak to black and the background to white in Fig. \ref{fig:single_streak}(a).

\begin{figure*}[t]
\begin{minipage}{0.24\linewidth}
\centering{\includegraphics[width=.9\linewidth]{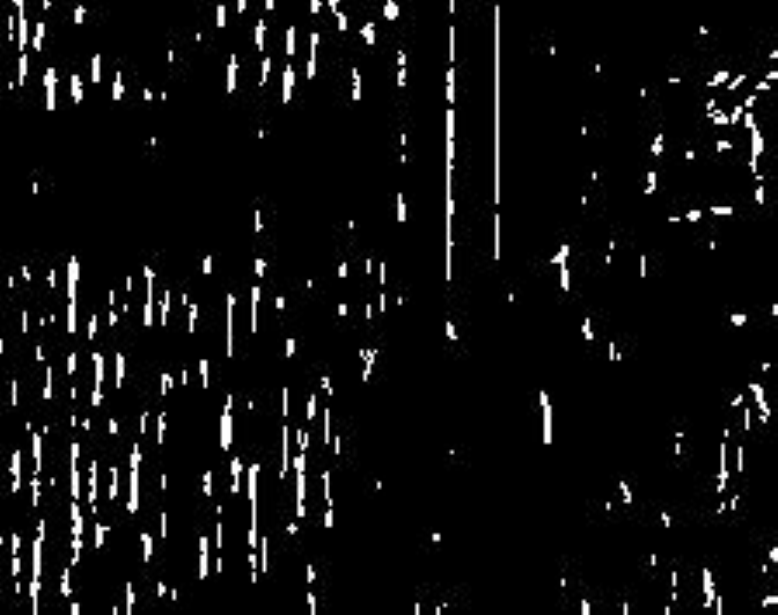}}
\centerline{(a)}
\end{minipage}
\hfill
\begin{minipage}{.24\linewidth}
\centering{\includegraphics[width=.9\linewidth]{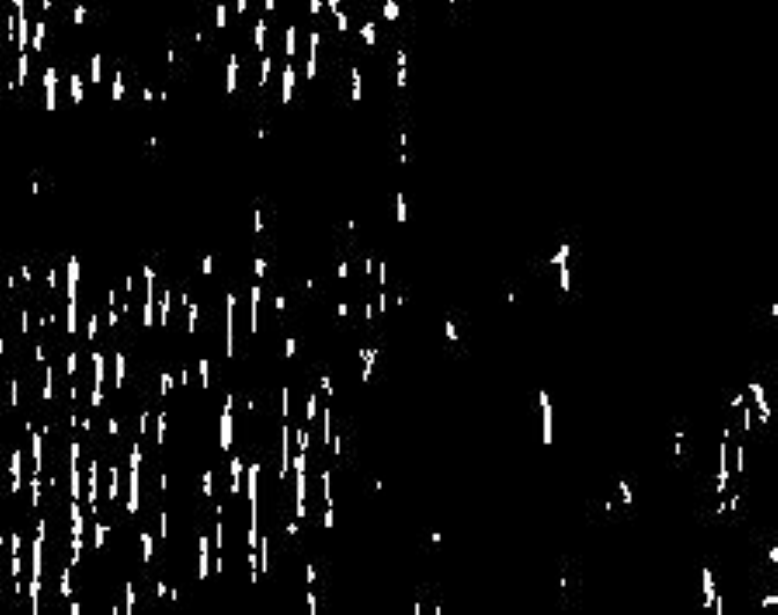}}
\centerline{(b)}
\end{minipage}
\hfill
\begin{minipage}{.24\linewidth}
\centering{\includegraphics[width=.9\linewidth]{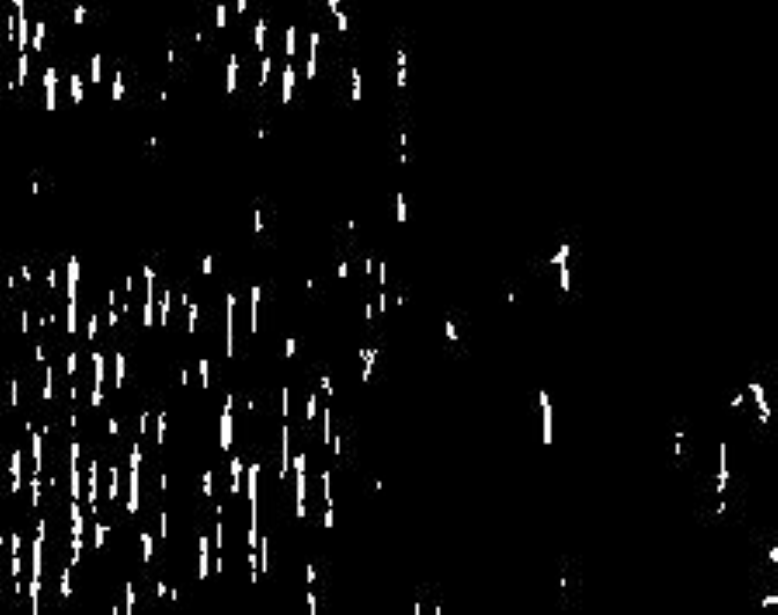}}
\centerline{(c)}
\end{minipage}
\hfill
\begin{minipage}{.24\linewidth}
\centering{\includegraphics[width=.9\linewidth]{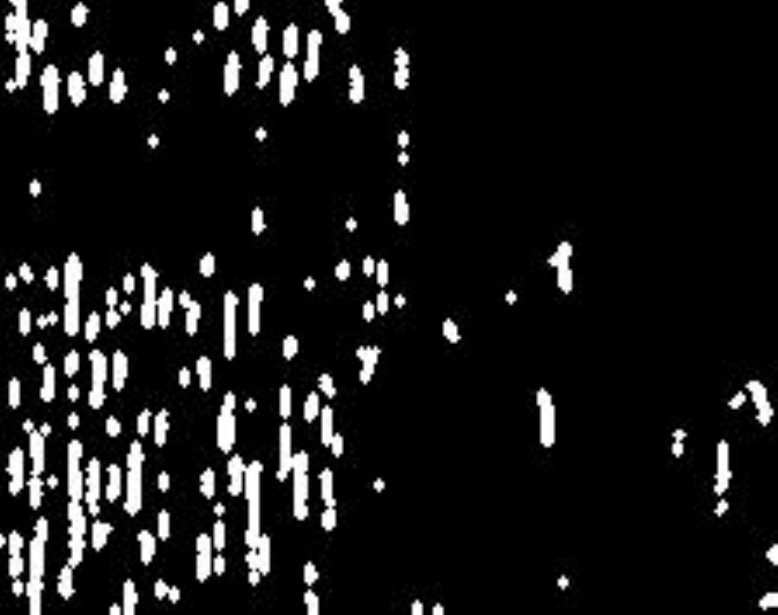}}
\centerline{(d)}
\end{minipage}
\caption{(a) Refined result by the connected component angle. (b) Refined result by the connected component color. (c) Refined result by the connected component aspect ratio. (d) Dilation of rain streaks.}
\label{fig:revision}
\end{figure*}

For $p^{th}( p=1, 2, ..., P)$ connected component, we calculate the covariance matrix of location vectors of all pixels in it.
Suppose that there are $N$ pixels in $p^{th}$ connected component.
Hence, there are $N$ sample vectors of pixel locations so that the mean location
vector $\bm{m}_{\bm{z}}$ and covariance matrix $\bm{C}_{\bm{z}}$ can be calculated as
\begin{equation} \label{eq:mean_approximation}
\bm{m}_{\bm{z}} = \frac {1}{N} \sum^{N}_{n=1} \bm{z}_{n}
\end{equation}
\begin{equation} \label{eq:covariance_matrix2}
\bm{C}_{\bm{z}} = \frac {1}{N} \sum^{N}_{n=1} \bm{z}_{n}\bm{z}_{n}^{T} - \bm{m}_{\bm{z}}\bm{m}_{\bm{z}}^{T}
\end{equation}
where $\bm{z}_{n}= [x_{n}, y_{n}]^{T}$, and $x_n$ and $y_n$ are respectively
the corresponding coordinates of the $n^{th}$ pixel ($n=1, 2, \cdots, N$).

After the covariance matrix $\bm{C}_{\bm{z}}$ of $p^{th}$ connected component is obtained,
we perform the eigenvalue decomposition of $\bm{C}_{\bm{z}}$
and obtain the eigenvalues $\lambda_{1}$, $\lambda_{2}$ and
their corresponding eigenvector $\bm{e}_{1}$, $\bm{e}_{2}$ ($\lambda_{1}$ is the larger eigenvalue).
The description of PCA to the shape of connected
components are shown in Fig. \ref{fig:single_streak}(a).
The red arrows stand for two eigenvectors, while two yellow arrows denote
the coordinate axes. Here, $\theta$ is the angle between $x$-axis and eigenvector
$\bm{e}_{1}$ and it can be calculated as $\theta=\arctan(\frac {\bm{e}_{1}(2)}{\bm{e}_{1}(1)})$.
Notice that in order to avoid the red direction arrow from occluding the connected component,
the origin of the coordinate system is not placed on the connected component.

From Fig. \ref{fig:single_streak}(a), we learn that $\bm{e}_{1}$
(corresponding to the larger eigenvalue $\lambda_{1}$) points to the direction
where the location variance has the maximum value; whereas $\bm{e}_{2}$
(corresponding to the smaller eigenvalue $\lambda_{2}$) is perpendicular to the maximum variance direction.

Accordingly, we define the length of a connected component as
\begin{equation} \label{eq:length}
L=c\lambda_{1}
\end{equation}
and its width as
\begin{equation} \label{eq:width}
W=c\lambda_{2}
\end{equation}
where $c$ is a proportional parameter. We assume that $c$ is a constant in an image.
The specific value of $c$ is not important, because it does not affect the ratio of the
length and width of a connected component.
The more important quantity is the direction angle of a connected components,
which is denoted as $\theta$ in Fig. \ref{fig:single_streak}(a), but is now re-defined as
\begin{equation} \label{eq:angle}
D=\theta
\end{equation}
and name $D$ as the direction of a connected component.

In our experiment, the values $\lambda_{1}$, $\lambda_{2}$, $\bm{e}_{1}$,
$\bm{e}_{2}$ and $D$ of all $p^{th}(p=1, 2, ..., P)$ connected components are calculated,
$P$ is the number of connected components.
As an example, these values of the given connected components
in Fig. \ref{fig:single_streak}(a) are
$\lambda_{1}=172.8949$, $\lambda_{2}=0.5852$,
$\bm{e}_{1}=(0.9309, 0.3653)^{T}$,
$\bm{e}_{2}=(-0.3653, 0.9309)^{T}$ and $D=21.4286^{\circ}$ respectively.

\noindent \textbf{Third}, after obtaining the quantified characteristics of all connected components,
we recognise non-rain connected components as follows.
\begin{itemize}

\item As we said above, rain streaks usually do not have large width
as compared to some non-rain objects.
Hence, the $K$-means is used here to classify the connected components by their width $W$.
The connected components with larger width are mis-detected non-rain components and
we set their corresponding values in location map $S_{R}$ as $0$.
The refined result in this way is shown in Fig. \ref{fig:single_streak}(b).

There are not so many wide non-rain objects in this given image, hence
the refinement by width is not too apparent.
We can see that some non-rain components at right
bottom corner disappear in Fig. \ref{fig:single_streak}(b).
This is because when textures of an image are complex,
some non-rain streaks combine together and form
a larger connected component so that the width becomes large.

\item An apparent characteristic of rain streaks is that they follow
nearly the same falling direction and the angle will not be too large generally.
If we use the direction angle $D$ of connected component defined in Equation
(\ref{eq:angle}) to describe this characteristic, $\vert D \vert$
of rain components must be less than a threshold $T1$ ($\vert D \vert$ is
the absolute value of $D$).
Hence, by the threshold $T1$, we can recognize the mis-detected non-rain connected components
in Fig. \ref{fig:single_streak} (b). Then the non-rain connected
components are set to be 0, and the refined result is shown in Fig. \ref{fig:revision}(a).

\item After refining by the width and direction constraints,
majority of non-rain components are recognized. However, some non-rain components that
are similar in shape to the rain streaks still remain.
Rain streaks usually possess neutral color.
According to this feature, Chen \emph{et al.} \cite{Chen_2014_CSVT} proposed to
identify rain dictionary atoms by the eigen color feature \cite{Tsai_2008_IET_CV}.
In our work, we utilize the color characteristics of
rain to revise the mis-detected non-rain connected components.

For $p^{th}$ connected component in Fig. \ref{fig:revision}(a),
we calculate the mean color vector of all pixels in it,
and denote as $[\bar{R}, \bar{G}, \bar{B}]$.
Then we transform this 3-D RGB color vector into a 2-D vector as follows:
\begin{small}
\begin{equation}\label{eq:color_transform}
\begin{split}
 u & =\frac{2\Phi-\bar{G}-\bar{B}} {\Phi} \\
 v & =max \left \{
               \frac{\Phi-\bar{G}}{\Phi}, \frac{\Phi-\bar{B}}{\Phi}
            \right \}
\end{split}
\end{equation}
\end{small}
where $\Phi=\frac{1}{3}(\bar{R}+\bar{G}+\bar{B})$.
It is clear from (\ref{eq:color_transform}) that,
after the transform any connected components having neutral color will
be clustered around $(0, 0)$ in the $u$-$v$ space.
Hence, we calculate the magnitude of this 2-D vector $(u, v)$
(i.e.,the Euclidean distance to the origin of the $u$-$v$ space),
if the magnitude is larger than a pre-set value $T2$,
the $p^{th}$ connected component is recognized as a mis-detected non-rain connected component.
For all remaining connected components in Fig. \ref{fig:revision}(a),
we repeat this process and revise the mis-detected non-rain connected components.
The refined result is shown in Fig. \ref{fig:revision}(b).

\item According to \cite{Kim_2013_ICIP}, a rain streak has a larger size
in length than in width. Hence, we classify the connected components
whose aspect ratio are less than $\mu$ as non-rain components.
By excluding the connected components that have small aspect ratios,
the refined result is shown in Fig. \ref{fig:revision}(c).

\item Finally, in order to avoid some slim rain edges
from remaining in our final rain-removed result,
we dilate the connected components in
Fig. \ref{fig:revision}(c) by a $3\times 3$ 'disk' mask,
to obtain the final result for rain streaks detection,
as shown in Fig. \ref{fig:revision}(d).
\end{itemize}

Our rain detection is a stepwise revision method.
By utilizing morphology and PCA, we quantify the rain's
characteristics and detect rain streaks relatively accurately.

\section{Image Reconstruction}
\label{sec:ImageReconstruction}
In this section, we try to verify the sparsity
of natural rain images and utilize one Laplacian distribution
to approximate the sparsity prior of natural rain image,
we name the approximate prior as \emph{quasi-sparsity prior}.
Then, based on the quasi-sparsity and several constraints,
the rain-removed result is obtained by
separating a rain image into rain layer and background layer.

\subsection{Quasi-sparsity of rain images}

In \cite{Levin_2007_PAMI}, Levin and Weiss tried to separate
the background and reflection from an image by sparsity prior of natural images.
We also utilize image sparsity in our rain removal task.
The sparsity of an image mentioned in \cite{Levin_2007_PAMI} can
be depicted as: when a derivative filter is applied on an image,
the logarithm of the histogram of the obtained gradient image reaches
peak value at zero and falls off much faster than a Gaussian.
Levin \emph{et al.} demonstrated that sparse distributions
will lead to a better image decomposition \cite{Levin_2002_NIPS}.
Hence, the sparsity of a natural
image is crucial to its decomposition into several layers.

\begin{figure}[t]
\centering
\begin{minipage}{0.48\linewidth}
\centering{\includegraphics[width=.9\linewidth]{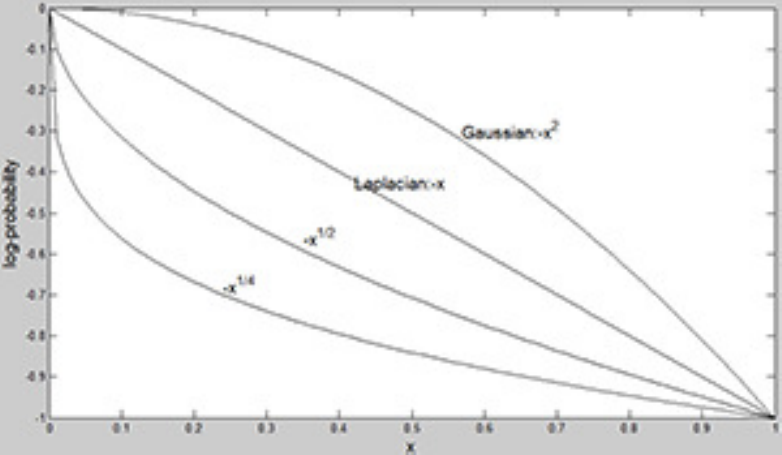}}
\centerline{(a)}
\end{minipage}
\hfill
\begin{minipage}{.48\linewidth}
\centering{\includegraphics[width=.9\linewidth]{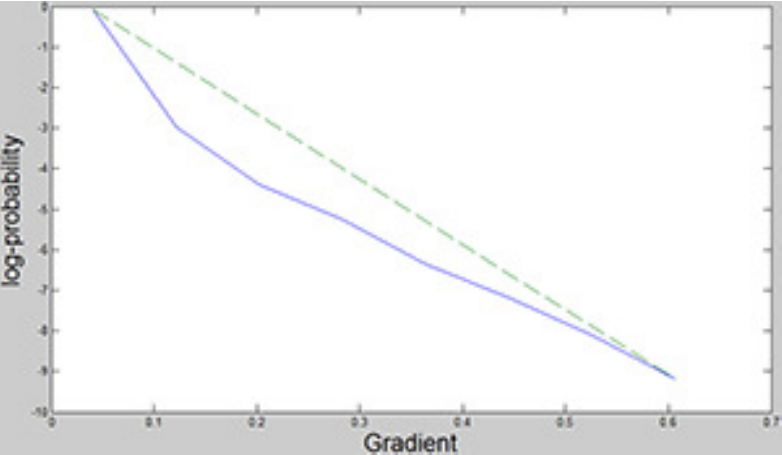}}
\centerline{(b)}
\end{minipage}
\caption{(a) Log-probability of several distributions. (b) Sparsity verification on one rain image.}
\label{fig:sparsity}
\end{figure}

\begin{figure*}[t]
\centering
\begin{minipage}{0.24\linewidth}
\centering{\includegraphics[width=.9\linewidth]{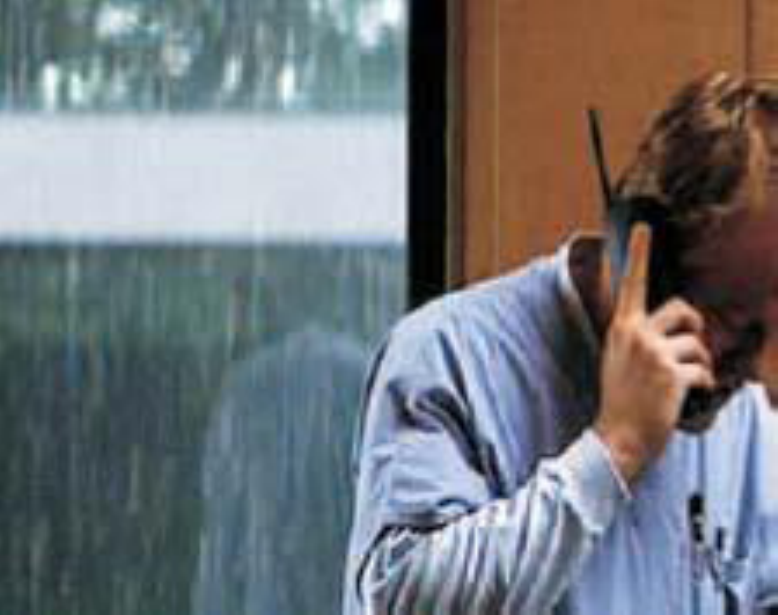}}
\centerline{(a)}
\end{minipage}
\hfill
\begin{minipage}{0.24\linewidth}
\centering{\includegraphics[width=.9\linewidth]{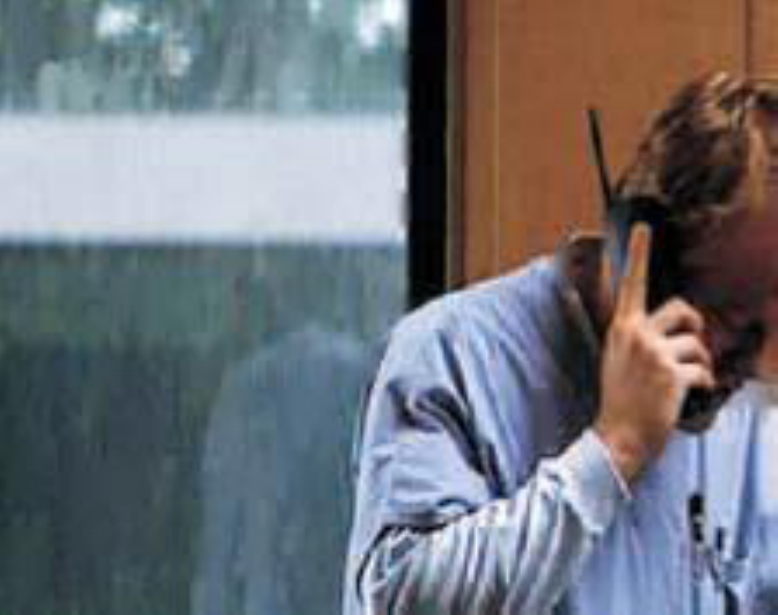}}
\centerline{(b)}
\end{minipage}
\hfill
\begin{minipage}{.24\linewidth}
\centering{\includegraphics[width=.9\linewidth]{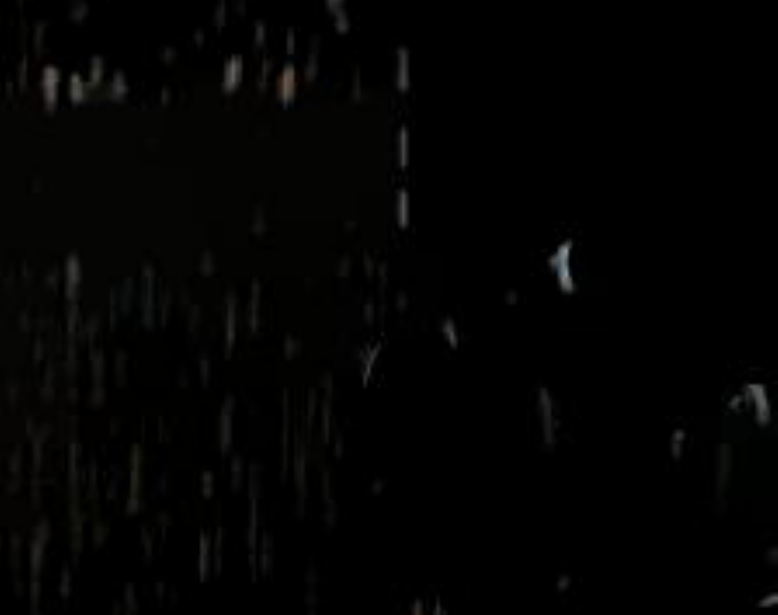}}
\centerline{(c)}
\end{minipage}
\hfill
\begin{minipage}{.24\linewidth}
\centering{\includegraphics[width=.9\linewidth]{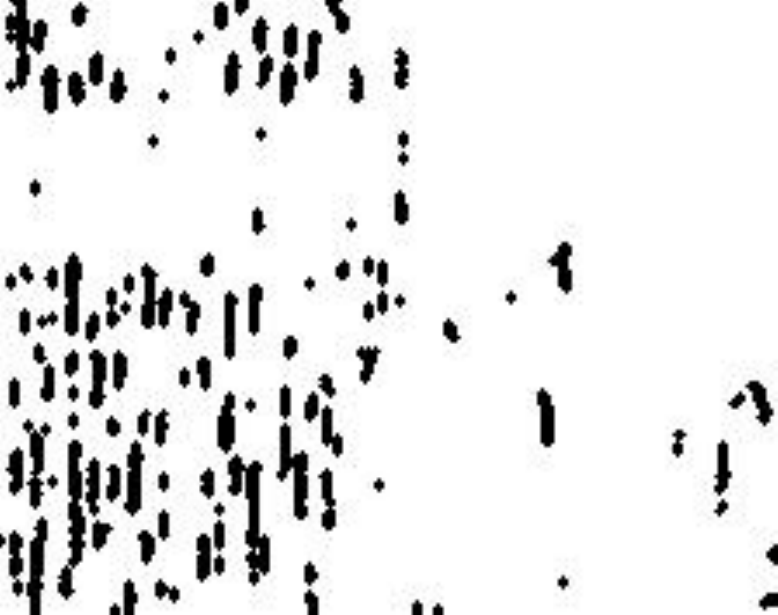}}
\centerline{(d)}
\end{minipage}
\caption{(a) Original rain image. (b) Rain-removed image.
(c) Rain component removed from (a).
(d) Non-rain location $S_{NR}$ that is obtained by $S_{NR}=1-S_{R}$ (the white area).}
\label{fig:results}
\end{figure*}

\begin{figure*}[t]
\centering
\begin{minipage}{0.195\linewidth}
\centering{\includegraphics[width=.9\linewidth]{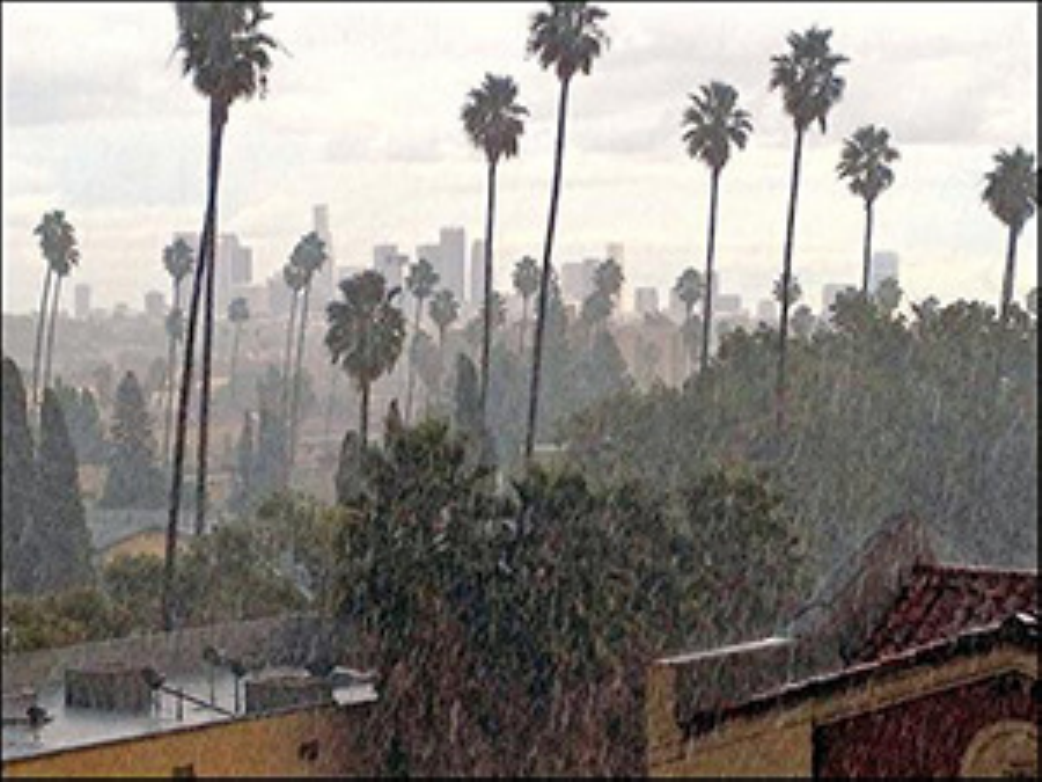}}
\centerline{(a)}
\end{minipage}
\hfill
\begin{minipage}{0.195\linewidth}
\centering{\includegraphics[width=.9\linewidth]{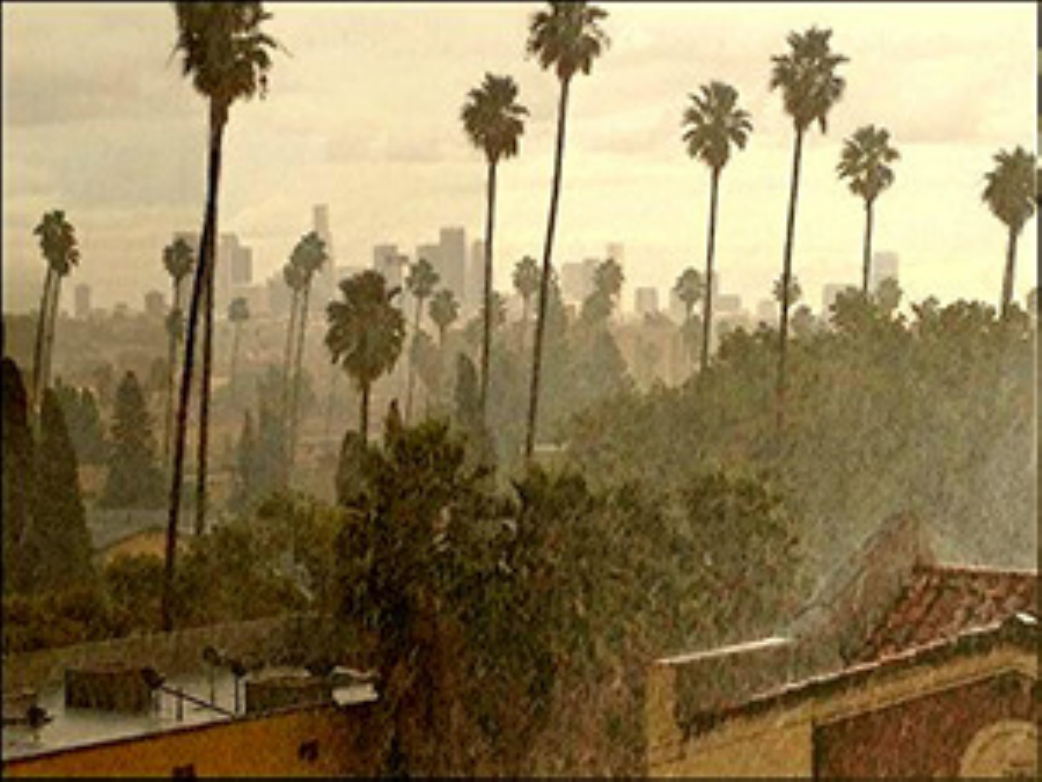}}
\centerline{(b)}
\end{minipage}
\hfill
\begin{minipage}{.195\linewidth}
\centering{\includegraphics[width=.9\linewidth]{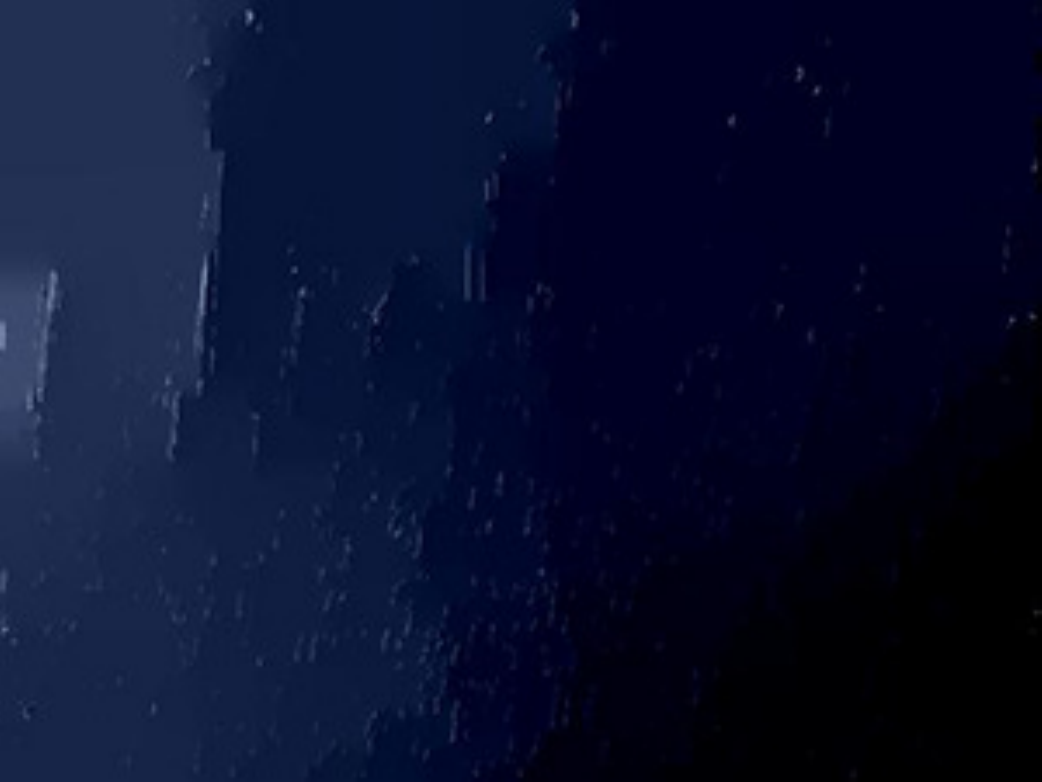}}
\centerline{(c)}
\end{minipage}
\hfill
\begin{minipage}{.195\linewidth}
\centering{\includegraphics[width=.9\linewidth]{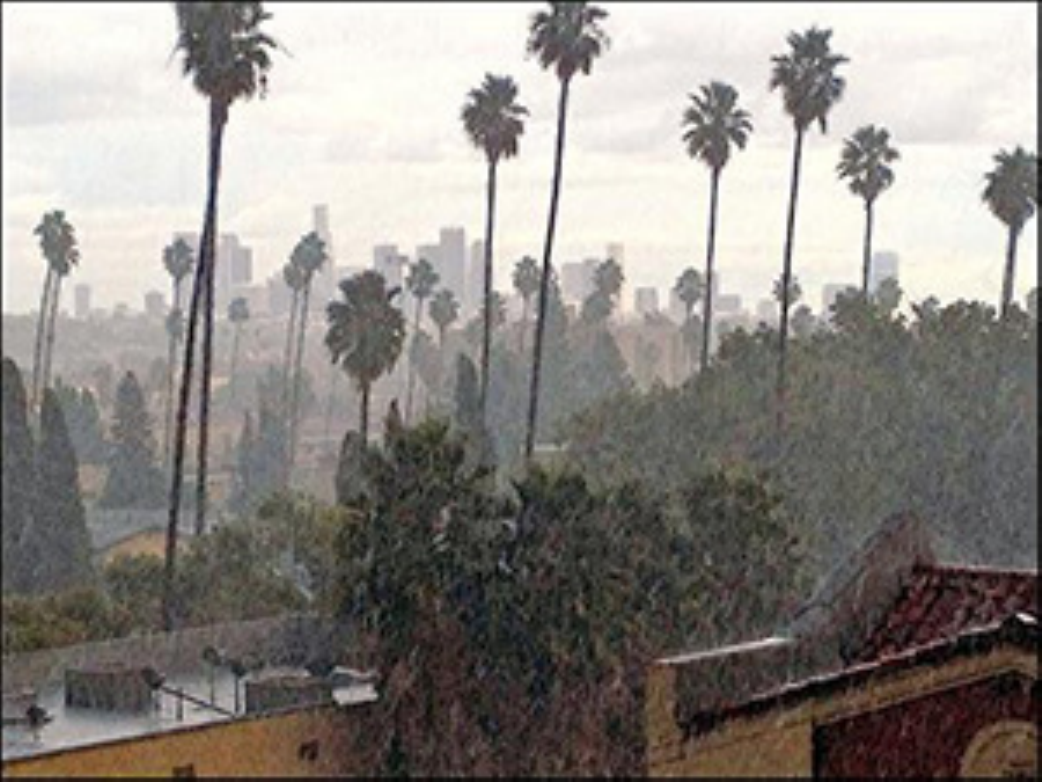}}
\centerline{(d)}
\end{minipage}
\hfill
\begin{minipage}{.195\linewidth}
\centering{\includegraphics[width=.9\linewidth]{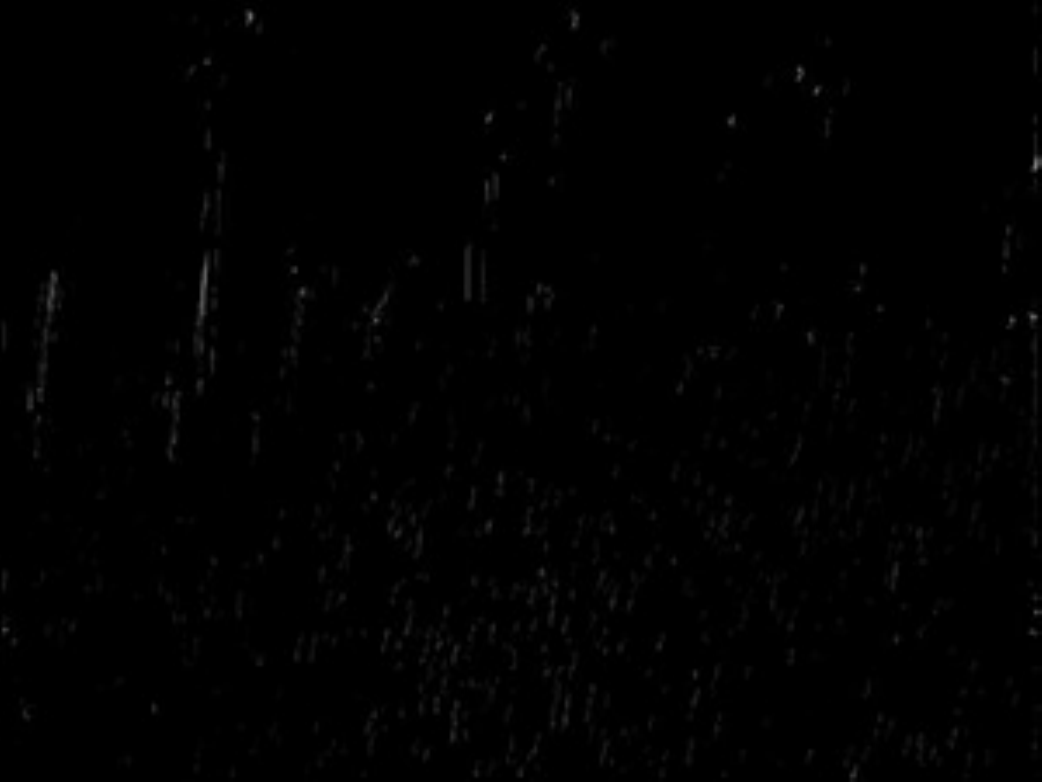}}
\centerline{(e)}
\end{minipage}
\caption{(a) One rain image; (b) background layer without the third constraint;
(c) rain layer without the third constraint;
(d) background layer with the third constraint; (e) rain layer with the third constraint.}
\label{fig:correct_separate}
\end{figure*}

Fig. \ref{fig:sparsity}(a) illustrates the logarithm probabilities of
several distributions. Laplacian distribution exactly results in a straight line and
connects the maximum and minimum values.
We can see that Gaussian distribution falls off the slowest and is above the
straight line so that it is viewed as non-sparse. The other two distributions
below the straight line are classified as sparse according to \cite{Levin_2007_PAMI}.
Laplacian distribution is on the border of
sparsity and non-sparsity.

In order to verify the sparsity of rain images,
we conduct an experiment on nearly 200 rain images and part of these
images are also used in the experiment section. Here, we use the image
in Fig. \ref{fig:results}(a) as an example to illustrate the sparsity of rain images.
Fig. \ref{fig:sparsity}(b) shows the logarithm curve (the blue curve)
of histogram after applying an horizontal derivative filter on it.
Obviously, the result reveals that the rain image satisfies the sparsity requirement.

However, decomposing a rain image $I$ into the rain layer $I_{R}$ and background layer $I_{NR}$ as
\begin{equation} \label{eq:image_decomposition}
I=I_{R}+I_{NR}
\end{equation}
is a massively ill-posed problem. To simplify this kind of problem,
Levin \emph{et al.} proposed that users can label some edges or areas that belong
to $I_{R}$ and some other edge or areas that belong to $I_{NR}$
to increase the constraint for this kind of problem \cite{Levin_2007_PAMI}.

Sparsity ensures that an edge of unit contrast will not be split,
and will appear in one layer \cite{Levin_2007_PAMI}.
In our task, we have detected nearly all rain locations and
the remaining region is labelled as the non-rain area.
Our detection offers better constraints to this ill-posed problem
than the manually-labeled operation in \cite{Levin_2007_PAMI},
and also realize the role of sparsity in certain degree.
Unlike in \cite{Levin_2007_PAMI}, we relax the probability constraint and utilize single
Laplacian function to approximate the sparsity of rain images
and named as \emph{quasi-sparse distribution}:
\begin{equation} \label{eq:histogram_approximation}
P(x)=e^{- \vert x \vert}
\end{equation}
Hence, the quasi-sparsity prior over the whole image $I$ is as follows:
\begin{equation}\label{eq:laplacian_approximation}
P(I)=\prod_{i, k}e^{- \vert \omega_{i, k} \cdot I \vert}
\end{equation}
where $\omega_{i,k}$ is the $k^{th}$ filter which centered
at $i^{th}$ pixel.  The filters
with two orientations (horizontal and vertical) and two degrees (the
first derivative and the second derivative) are used here.

\subsection{Optimization}

For an given rain image $I$, $S_{R}$ is the detected rain location,
and the non-rain location can be obtained by $S_{NR}=1-S_{R}$.
The following constraints are satisfied to separate an rain image
into rain layer $I_{R}$ and background (non-rain) layer $I_{NR}$:
\begin{enumerate}
\item $I=I_{R}+I_{NR}$;
\item the gradients of $I_{R}$ and $I_{NR}$ at their corresponding locations
in $S_{R}$ and $S_{NR}$ respectively agree with the gradient of image $I$;
\item the values of $I_{NR}$ at location $S_{NR}$ are close to the value of $I$.
\end{enumerate}
The first two constraints are also utilized in \cite{Levin_2007_PAMI}.
As shown later, this will lead some non-normal separation for some specific images.
To improve the separation, we add the third constraint and it plays a role
as boundary condition.

As the work in \cite{Weiss_2001_ICCV}, we assume that
derivative filters are independent over space and orientation;
rain layer $I_{R}$ and background layer $I_{NR}$ are independent.
Then the quasi-sparsity prior can be written as follows according
to the first constraint:
\begin{equation} \label{eq:prior_define}
P(I)=P(I_{R})P(I_{NR})=\prod_{i, k}e^{- ( \vert \omega_{i, k} \cdot I_{R} \vert +  \vert \omega_{i, k} \cdot I_{NR} \vert)}
\end{equation}
We would like to obtain $I_{R}$ and $I_{NR}$ which maximize the
above likelihood function. It is equal to minimize the following loss function:
\begin{equation}\label{eq:loss_function}
J(I_{R}, I_{NR})=\sum_{i, k} \vert \omega_{i, k} \cdot I_{R} \vert + \vert \omega_{i, k} \cdot I_{NR} \vert
\end{equation}
Combined with the second and third constraints,
we rewrite Equation (\ref{eq:loss_function}) as
\begin{equation}\label{eq:loss_function1}
\begin{split}
&J_{1}(I_{R})=\sum_{i, k} \vert \omega_{i, k} \cdot I_{R} \vert + \vert \omega_{i, k} \cdot (I-I_{R}) \vert \\
& \qquad \quad + \lambda \sum_{i \in S_{R}, k} \vert \omega_{i,k} \cdot I_{R} - \omega_{i, k} \cdot I \vert \\
& \qquad \quad +\lambda \sum_{i \in S_{NR}, k} \vert \omega_{i, k} \cdot I_{R} \vert  \\
& \qquad \quad +\eta \sum_{i \in S_{NR}} \vert I_{R} \vert
\end{split}
\end{equation}
where $\lambda$ and $\eta$ are regularization parameters.

If $v$ is defined as the vectorized version of image $I_{R}$,
Equation (\ref{eq:loss_function1}) becomes
\begin{equation}\label{eq:loss_function2}
J_{2}(v)= \Vert Av-b \Vert_{1}
\end{equation}
where $\Vert \cdot \Vert_{1}$ is the $L_{1}$ norm, $A$
is relative to the derivative filters, $\lambda$ and $\eta$, and $b$ is relative to the image derivative,
the values of $I$ at location $S_{NR}$, zero and $\lambda$, $\eta$.

This is a $L_{1}$-norm optimization problem,
and it can be solved by iterative reweighted least squares (IRLS) \cite{Burrus_2009_CPAM}.
We summarize the process in Algorithm \ref{alg:whole_algorithm}.
Once $v$ is obtained, we resize it back to the rain-layer image $I_{R}$.
Then, the rain-removed image $I_{NR}$ can be obtained as
\begin{equation}\label{eq:rain_remove}
I_{NR}=I-I_{R}
\end{equation}
One example of the rain layer and rain-removed image is shown in Fig. \ref{fig:results}(a) and (b), respectively.
In Fig. \ref{fig:results}(c)(d), we show the constructed rain layer and the non-rain location $S_{NR}$.

As mentioned above, the third constraint plays an important role
in the correct separation of rain images. Here, we show
an example in Fig. \ref{fig:correct_separate} to suggest the role of this constraint.
In Fig. \ref{fig:correct_separate}(b)(c), we can see that serious
color shift (means that the colors of non-rain details in (b) are abnormal) will
appear without the third constraint.
The reason is that some colors go to the rain layer (c) in under-determined conditions.
By adding the third constraint, the separation quality
can be improved and we can obtain a natural rain-removed image.

\begin{algorithm}
\renewcommand{\algorithmicrequire}{\textbf{Input:}}
\caption{IRLS}
\label{alg:whole_algorithm}
\begin{algorithmic}
\REQUIRE $A$, $b$, $Iter$
\renewcommand{\algorithmicrequire}{Initialization}
\REQUIRE $v=[ A^{T}A ]^{-1}Ab$
\FOR{$t$=1 to $Iter$ }
\STATE $e=abs(Av-b)$
\STATE $z(i) = e(i)^{-0.5}, i=1, 2, ...$
\STATE $\Omega = diag(z)$
\STATE $v =[ A^{T}\Omega^{T}\Omega A ]^{-1}A^{T}\Omega^{T}\Omega b$
\ENDFOR
\renewcommand{\algorithmicensure}{\textbf{Output:}}
\ENSURE $v$
\end{algorithmic}
\end{algorithm}


\section{Experimental Results}
\label{sec:ExperimentalResults}

\begin{table*} [t]
\centering
\caption{The Average Time Consumed by Selected Methods on $256 \times 256$ Images.}
\begin{tabular}{lccccccc}
\hline
Method     &  \cite{Ding_2015_MTA}  & \cite{Chen_2014_CSVT}   & \cite{Luo_2015_ICCV}  & \cite{Li_2016_CVPR}  & \cite{Fu_2017_CVPR}  & \cite{Zhang_2018_CVPR}  & Ours   \\
Time(s) &  1.25s & 97.15s & 69.69s & 1260.40s & 5.30s   & 0.20s   & 28.01s   \\
\hline
\end{tabular}
\label{tab:time}
\end{table*}

\begin{table*} [t]
\small
\newcommand{\tabincell}[2]{\begin{tabular}{@{}#1@{}}#2\end{tabular}}
\centering
\caption{Image Performances (Top: \textbf{PSNR}, Bottom: \textbf{SSIM}) of Different Methods (Rows) on $11$ Synthesized Rain Images (Columns) against Ground-truth.}
\begin{tabular}{l|c|c|c|c|c|c|c|c|c|c|c}
\hline
        & Image 1                       & Image 2                       & Image 3                       & Image 4                       & Image 5                       & Image 6                       & Image 7                       & Image 8                       & Image 9                       & Image 10                      & Image 11                      \\
\hline
\cite{Ding_2015_MTA} & \tabincell{c}{34.65 \\ 0.867} & \tabincell{c}{33.70 \\ 0.889} & \tabincell{c}{33.89 \\ 0.802} & \tabincell{c}{34.17 \\ 0.805} & \tabincell{c}{35.16 \\ 0.861} & \tabincell{c}{35.93 \\ 0.835} & \tabincell{c}{41.29 \\ 0.796} & \tabincell{c}{31.77 \\ 0.811} & \tabincell{c}{32.50 \\ 0.874} & \tabincell{c}{34.58 \\ 0.907} & \tabincell{c}{33.22 \\0.832 } \\
\hline
\cite{Chen_2014_CSVT}   & \tabincell{c}{34.31 \\ 0.803} & \tabincell{c}{32.36 \\ 0.759} & \tabincell{c}{34.92 \\ 0.750} & \tabincell{c}{34.68 \\ 0.738} & \tabincell{c}{34.95 \\ 0.774} & \tabincell{c}{32.55 \\ 0.824} & \tabincell{c}{38.58 \\ 0.775} & \tabincell{c}{31.84 \\ 0.602} & \tabincell{c}{32.11 \\ 0.704} & \tabincell{c}{34.59 \\ 0.854} & \tabincell{c}{34.15 \\ 0.784} \\
\hline
\cite{Luo_2015_ICCV} & \tabincell{c}{32.69 \\ 0.767} & \tabincell{c}{30.23 \\ 0.703} & \tabincell{c}{31.53 \\ 0.748} & \tabincell{c}{32.43 \\ 0.820} & \tabincell{c}{33.73 \\ 0.888} & \tabincell{c}{29.45 \\ 0.841} & \tabincell{c}{35.95 \\ 0.784} & \tabincell{c}{29.45 \\ 0.790} & \tabincell{c}{30.43 \\ 0.879} & \tabincell{c}{31.63 \\ 0.864} & \tabincell{c}{32.99 \\ 0.843} \\
\hline
\cite{Li_2016_CVPR} & \tabincell{c}{31.55 \\ 0.701} & \tabincell{c}{30.45 \\ 0.686} & \tabincell{c}{31.23 \\ 0.789} & \tabincell{c}{32.27 \\ 0.691} & \tabincell{c}{33.34 \\ 0.748} & \tabincell{c}{31.13 \\ 0.754} & \tabincell{c}{36.39 \\ 0.681} & \tabincell{c}{29.54 \\ 0.570} & \tabincell{c}{30.32 \\ 0.686} & \tabincell{c}{32.35 \\ 0.786} & \tabincell{c}{32.42 \\ 0.749} \\
\hline
Ours   & \tabincell{c}{\textbf{35.46} \\ \textbf{0.886}} & \tabincell{c}{\textbf{35.30} \\ \textbf{0.901}} & \tabincell{c}{\textbf{35.04} \\ \textbf{0.827}} & \tabincell{c}{\textbf{34.86} \\ \textbf{0.832}} & \tabincell{c}{\textbf{35.38} \\ \textbf{0.897}} & \tabincell{c}{\textbf{36.03} \\ \textbf{0.842}} & \tabincell{c}{\textbf{41.31} \\ \textbf{0.846}} & \tabincell{c}{\textbf{31.94} \\ \textbf{0.854}} & \tabincell{c}{\textbf{33.42} \\ \textbf{0.883}} & \tabincell{c}{\textbf{34.91} \\ \textbf{0.916}} & \tabincell{c}{\textbf{34.53} \\ \textbf{0.866}} \\
\hline
\end{tabular}
\label{tab:psnrssim}
\end{table*}

\begin{figure*}[t]
\centering
\begin{minipage}{0.1\linewidth}
\centering{\includegraphics[width=.99\linewidth]{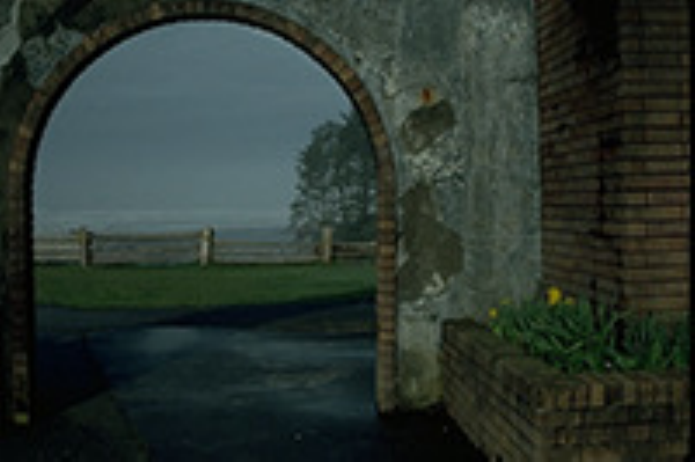}}
\end{minipage}
\begin{minipage}{0.1\linewidth}
\centering{\includegraphics[width=.99\linewidth]{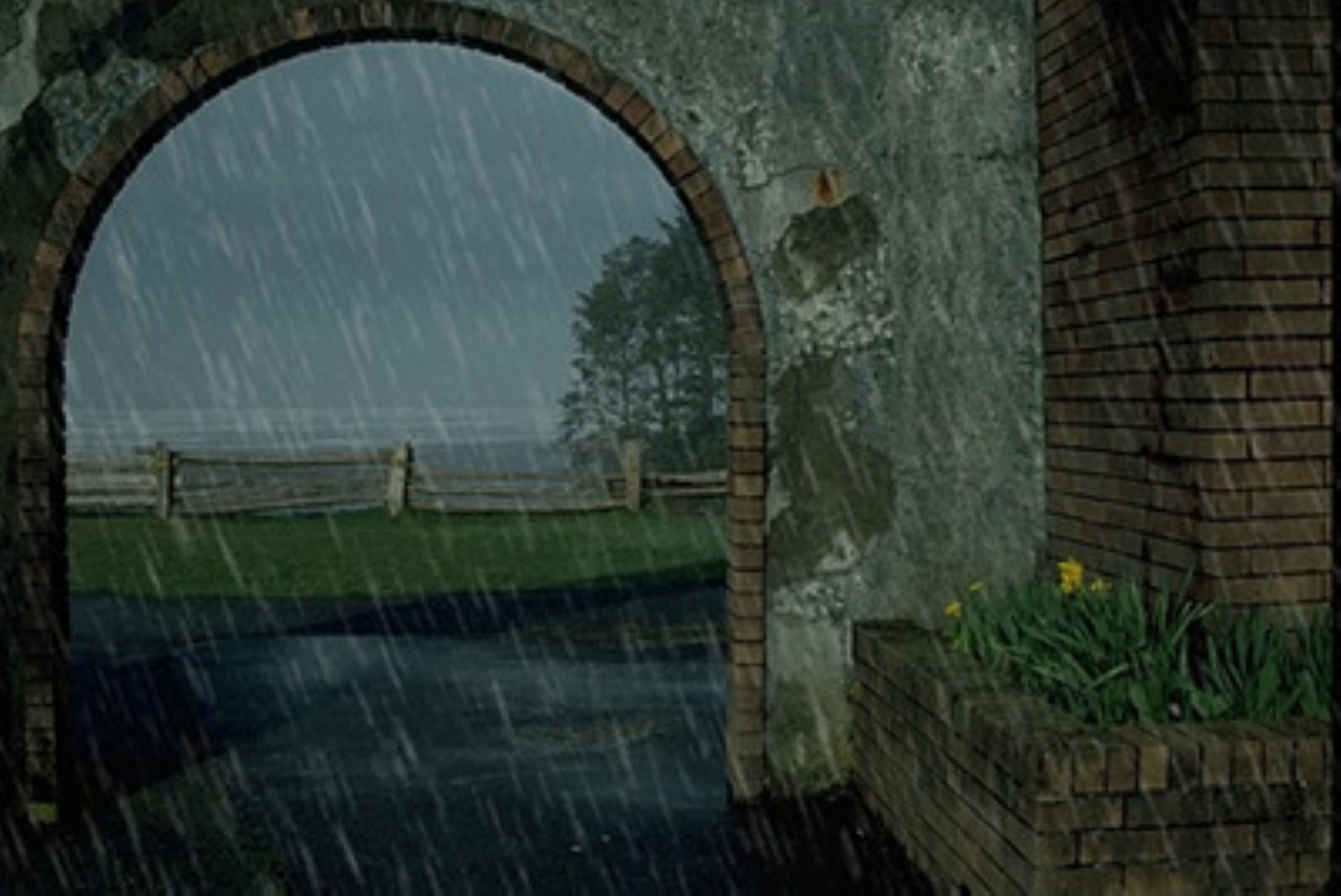}}
\end{minipage}
\begin{minipage}{.1\linewidth}
\centering{\includegraphics[width=.99\linewidth]{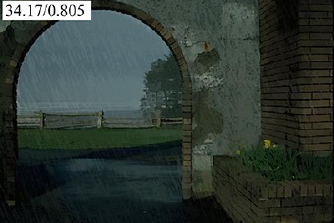}}
\end{minipage}
\begin{minipage}{.1\linewidth}
\centering{\includegraphics[width=.99\linewidth]{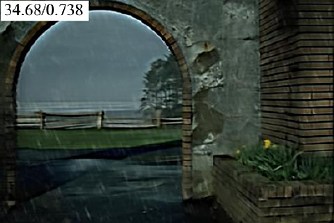}}
\end{minipage}
\begin{minipage}{.1\linewidth}
\centering{\includegraphics[width=.99\linewidth]{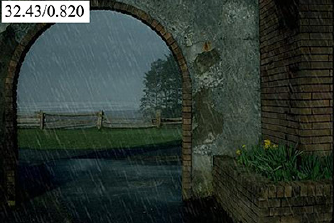}}
\end{minipage}
\begin{minipage}{.1\linewidth}
\centering{\includegraphics[width=.99\linewidth]{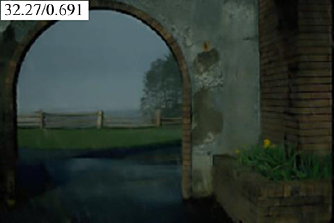}}
\end{minipage}
\begin{minipage}{.1\linewidth}
\centering{\includegraphics[width=.99\linewidth]{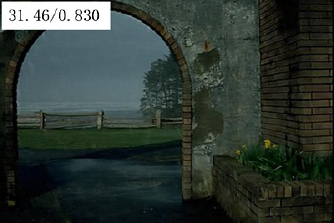}}
\end{minipage}
\begin{minipage}{.1\linewidth}
\centering{\includegraphics[width=.99\linewidth]{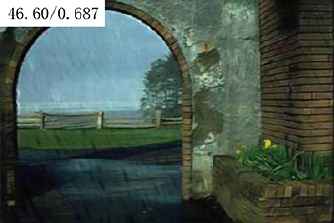}}
\end{minipage}
\begin{minipage}{.1\linewidth}
\centering{\includegraphics[width=.99\linewidth]{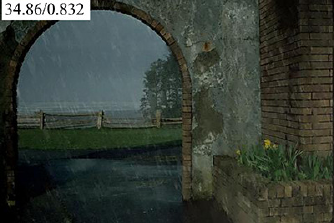}}
\end{minipage} \\
\vspace{0.5mm}
\begin{minipage}{0.1\linewidth}
\centering{\includegraphics[width=.99\linewidth]{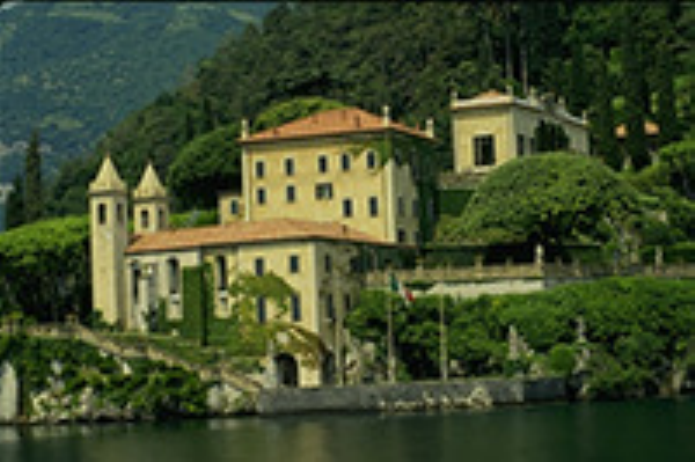}}
\centerline{(a)}
\end{minipage}
\begin{minipage}{0.1\linewidth}
\centering{\includegraphics[width=.99\linewidth]{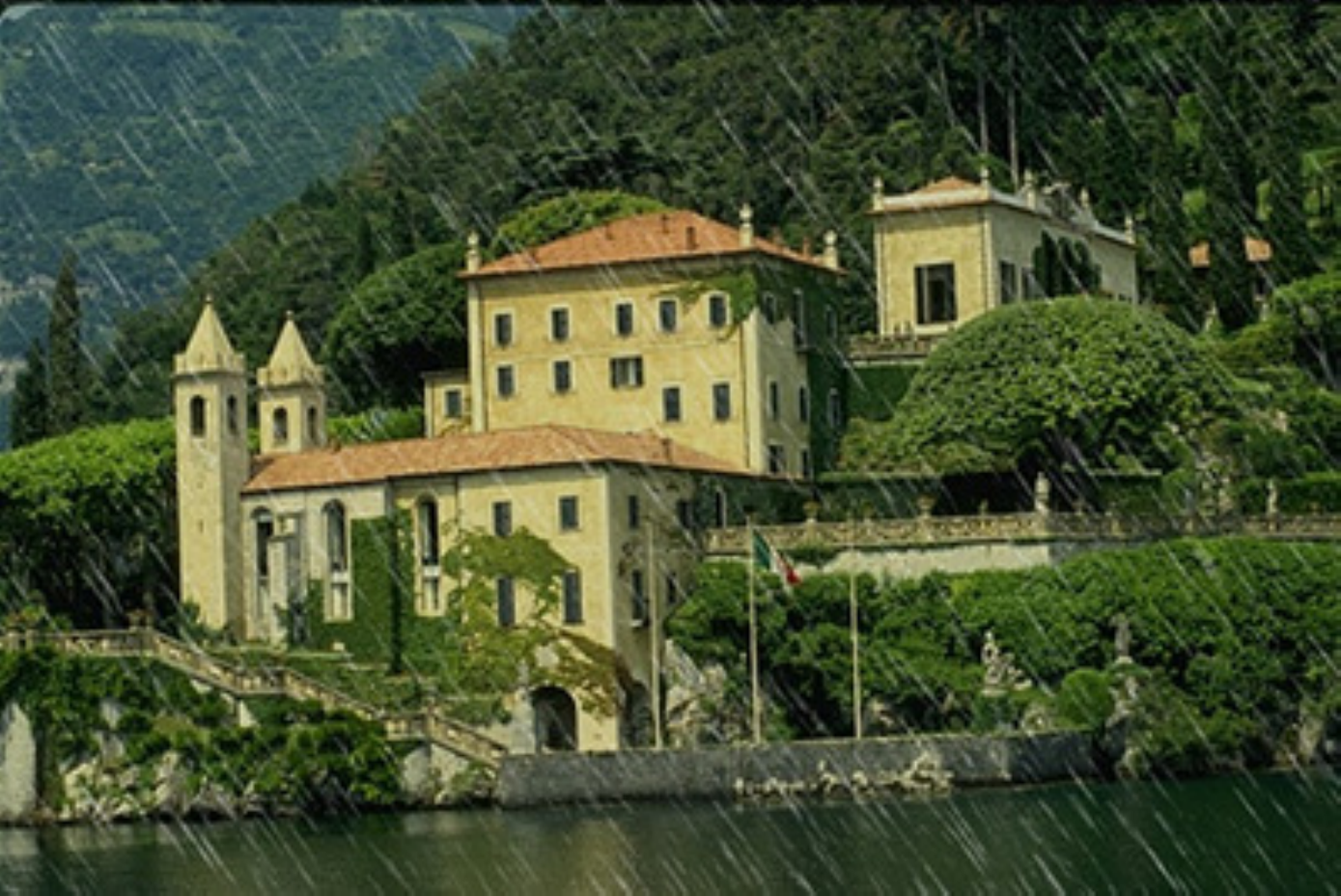}}
\centerline{(b)}
\end{minipage}
\begin{minipage}{.1\linewidth}
\centering{\includegraphics[width=.99\linewidth]{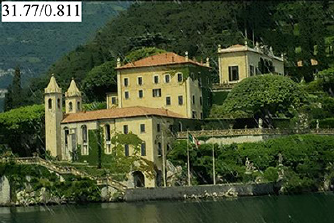}}
\centerline{(c)}
\end{minipage}
\begin{minipage}{.1\linewidth}
\centering{\includegraphics[width=.99\linewidth]{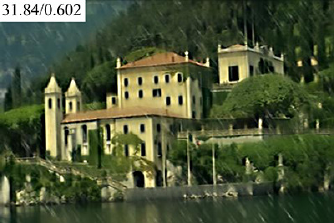}}
\centerline{(d)}
\end{minipage}
\begin{minipage}{.1\linewidth}
\centering{\includegraphics[width=.99\linewidth]{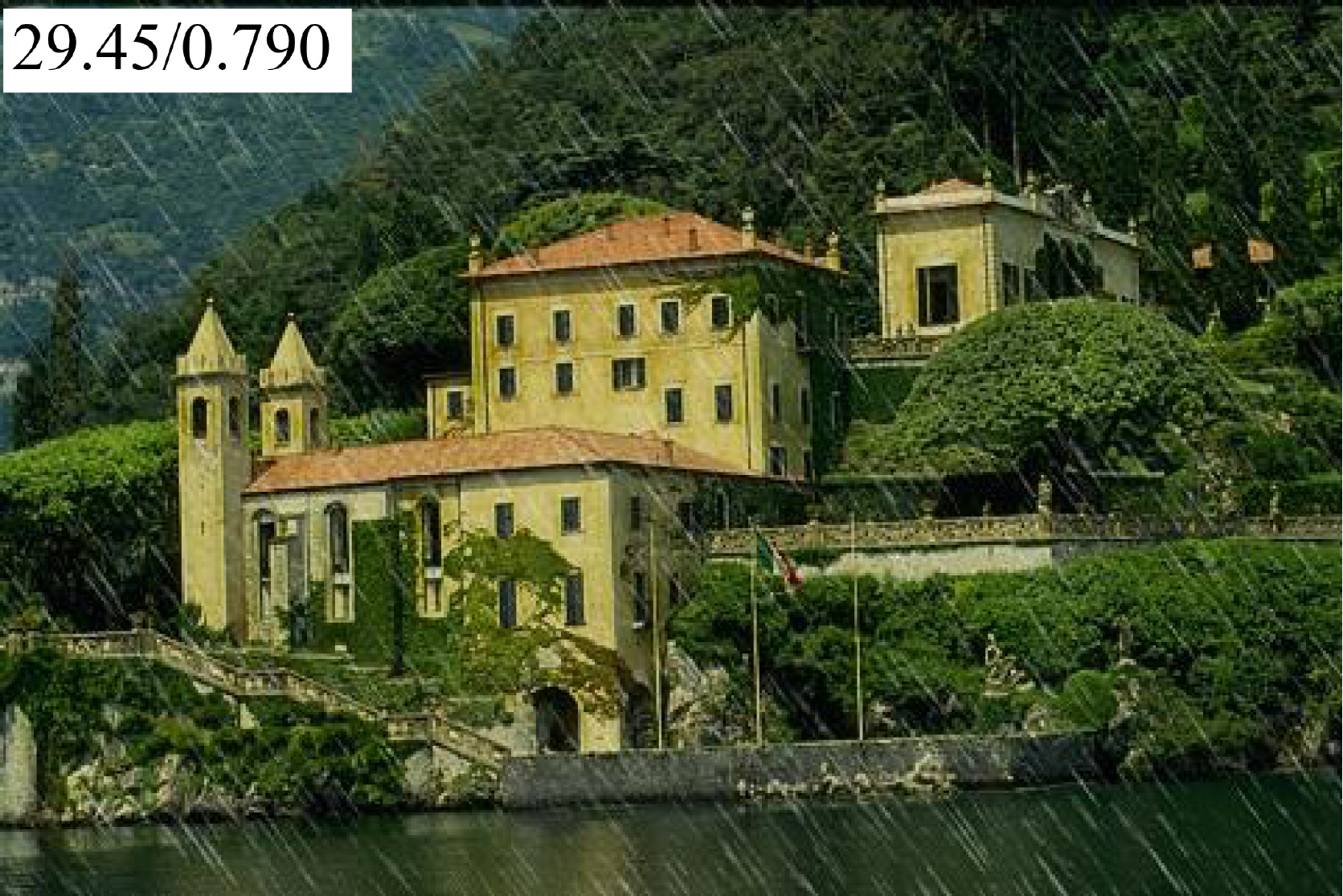}}
\centerline{(e)}
\end{minipage}
\begin{minipage}{.1\linewidth}
\centering{\includegraphics[width=.99\linewidth]{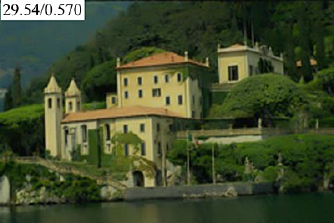}}
\centerline{(f)}
\end{minipage}
\begin{minipage}{.1\linewidth}
\centering{\includegraphics[width=.99\linewidth]{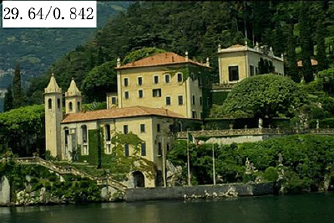}}
\centerline{(g)}
\end{minipage}
\begin{minipage}{.1\linewidth}
\centering{\includegraphics[width=.99\linewidth]{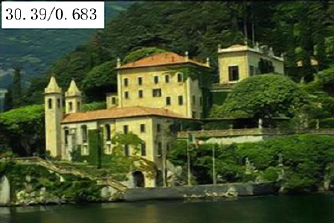}}
\centerline{(h)}
\end{minipage}
\begin{minipage}{.1\linewidth}
\centering{\includegraphics[width=.99\linewidth]{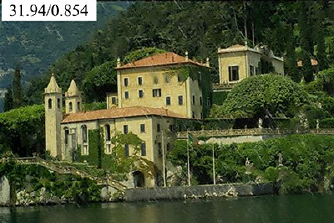}}
\centerline{(i)}
\end{minipage}
\caption{(a) Groudtruth. (b) Original synthesized rain images. (c) Results by Ding \emph{et al.} in \cite{Ding_2015_MTA}.
(d) Results by Chen \emph{et al.} in \cite{Chen_2014_CSVT}. (e) Results by Luo \emph{et al.} in \cite{Luo_2015_ICCV}.
(f) Results by Li \emph{et al.} in \cite{Li_2016_CVPR}. (g) Results by Fu \emph{et al.} in \cite{Fu_2017_CVPR}.
(h) Results by Zhang \emph{et al.} in \cite{Zhang_2018_CVPR}. (i) Results by our method.}
\label{fig:result_render_compare}
\end{figure*}

\begin{figure}[t]
\begin{minipage}{0.48\linewidth}
\centering{\includegraphics[width=1\linewidth]{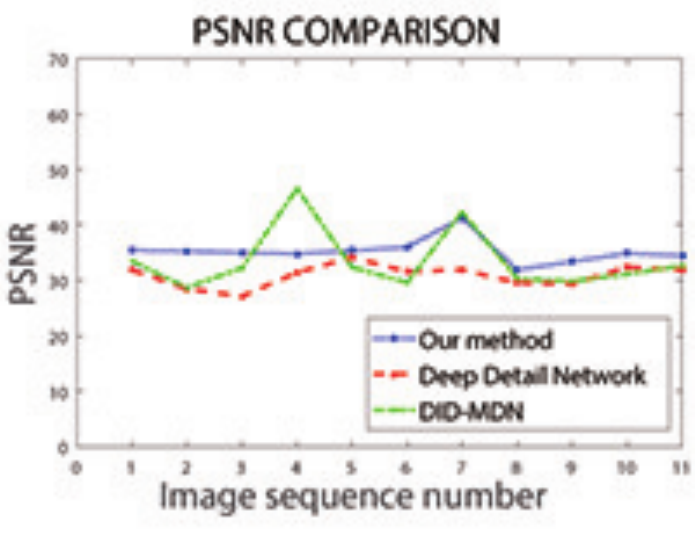}}
\centerline{(a)}
\end{minipage}
\hfill
\begin{minipage}{.48\linewidth}
\centering{\includegraphics[width=1\linewidth]{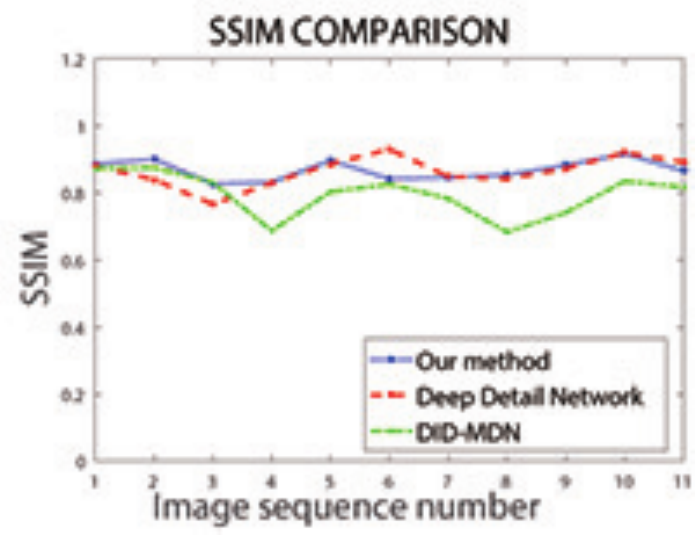}}
\centerline{(b)}
\end{minipage}
\caption{Objective comparisons with two state-of-the-art deep learning works: (a) PSNR comparison. (b) SSIM comparison.}
\label{fig:PSNR_SSIM}
\end{figure}

In order to verify the effectiveness of our method, several state-of-the-art traditional and deep learning based rain removal works are selected for comparisons. The method by Ding \emph{et al.} \cite{Ding_2015_MTA} removes rain streaks in a single image by developing an $L_0$ smoothing filter that is derived from the guided filter by He \emph{et al.} \cite{He_2013_PAMI}. This work produces excellent rain removal results for some kinds of images and keeps good visual quality. Meanwhile, several rain removal works that are based on dictionary learning have appeared in recent years \cite{Fu_2011_ASSP,Kang_2012_TIP,Chen_2014_CSVT}. Among them, the work by Chen \emph{et al.} \cite{Chen_2014_CSVT} produces the best rain removal effect. In addition, two most recent works, by Luo
\emph{et al.} \cite{Luo_2015_ICCV} and Li \emph{et al.} \cite{Li_2016_CVPR}, respectively, are also selected in our comparisons.
For deep learning based rain-removed methods, we select the most recent two works \cite{Fu_2017_CVPR} and \cite{Zhang_2018_CVPR}.
Compared with other deep learning based works, these two works are more robust and can obtain better rain-removed visual quality.

\begin{figure*}[!htb]
\centering
\begin{minipage}{0.115\linewidth}
\centering{\includegraphics[width=.995\linewidth]{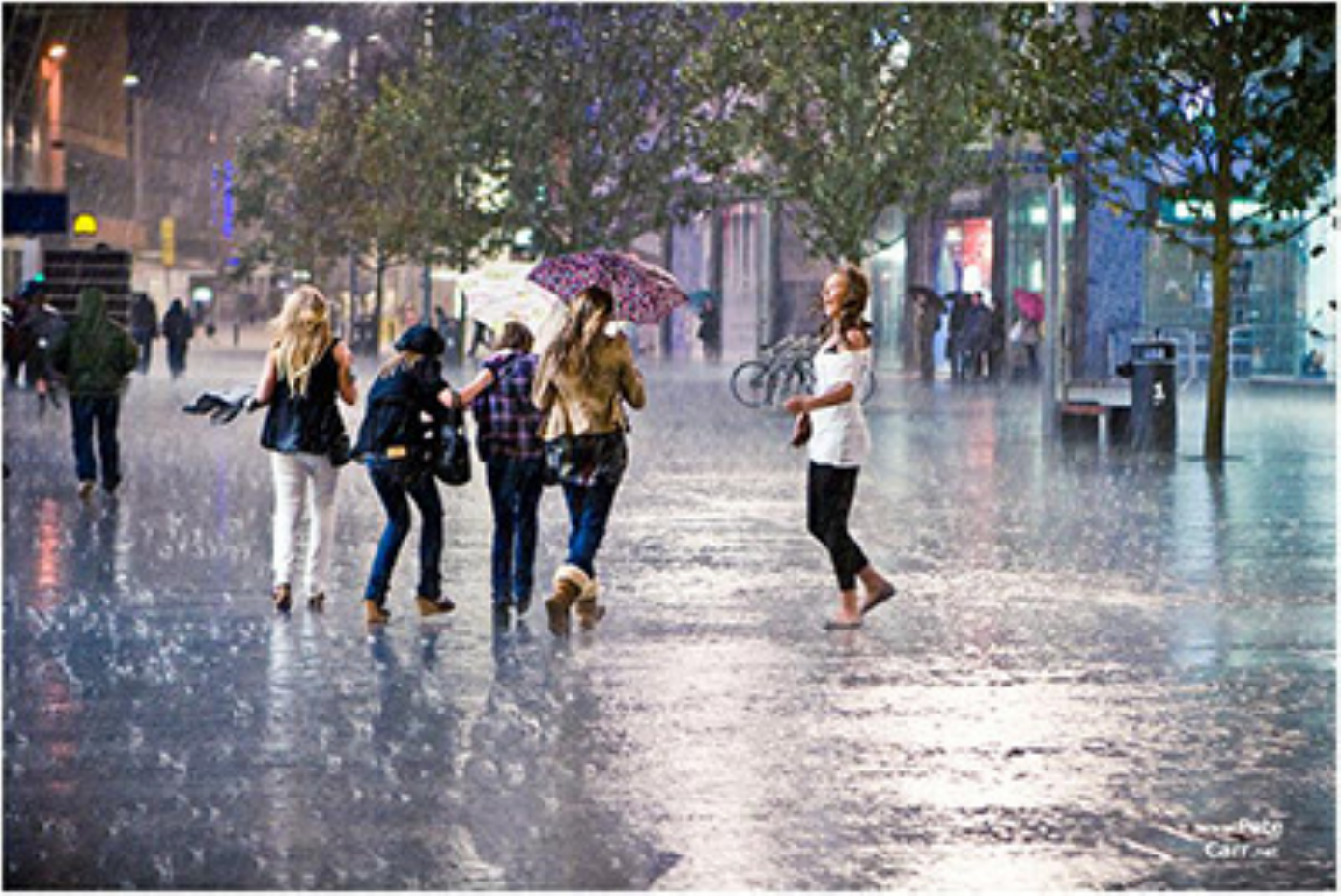}}
\end{minipage}
\hfill
\begin{minipage}{.115\linewidth}
\centering{\includegraphics[width=.995\linewidth]{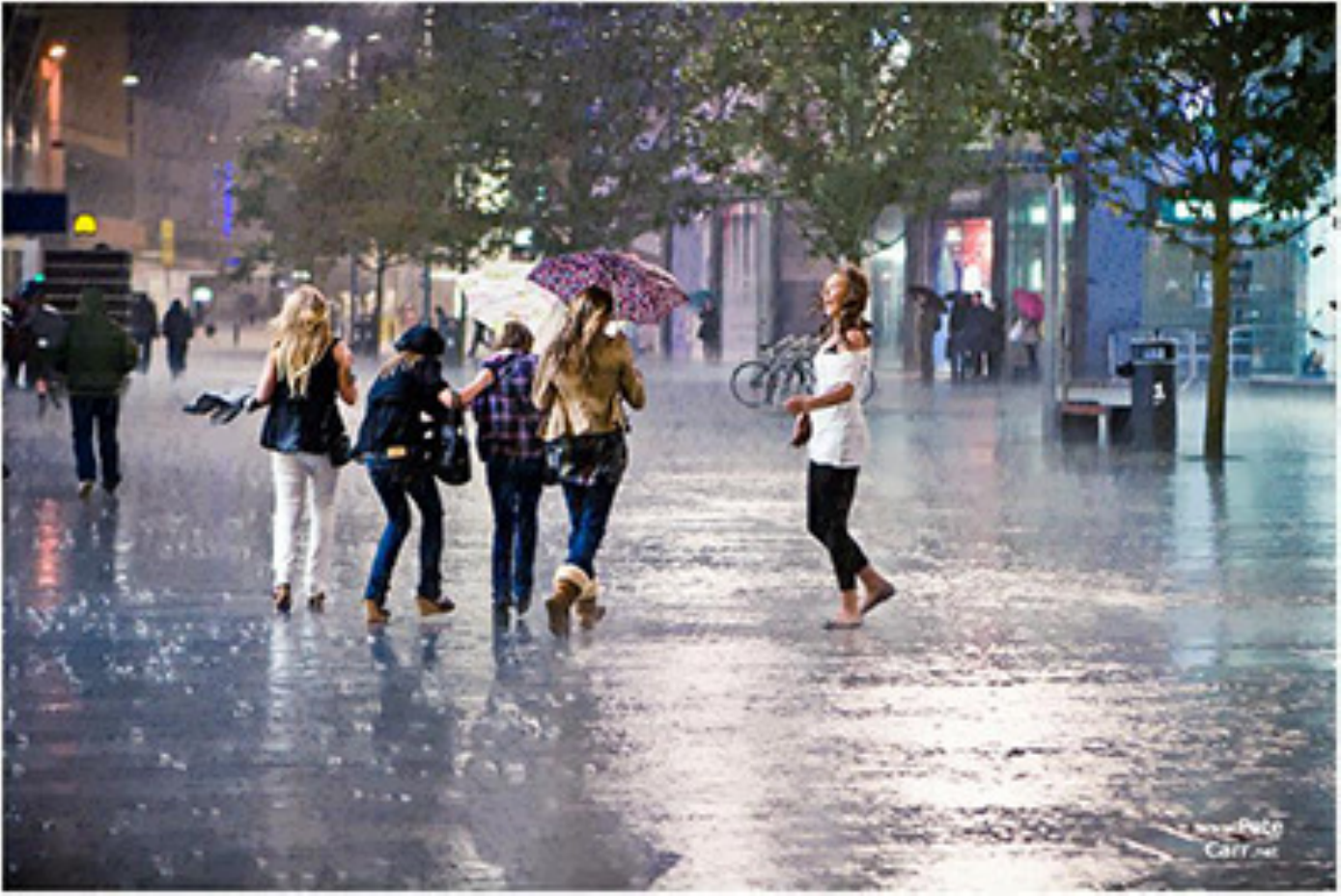}}
\end{minipage}
\hfill
\begin{minipage}{.115\linewidth}
\centering{\includegraphics[width=.995\linewidth]{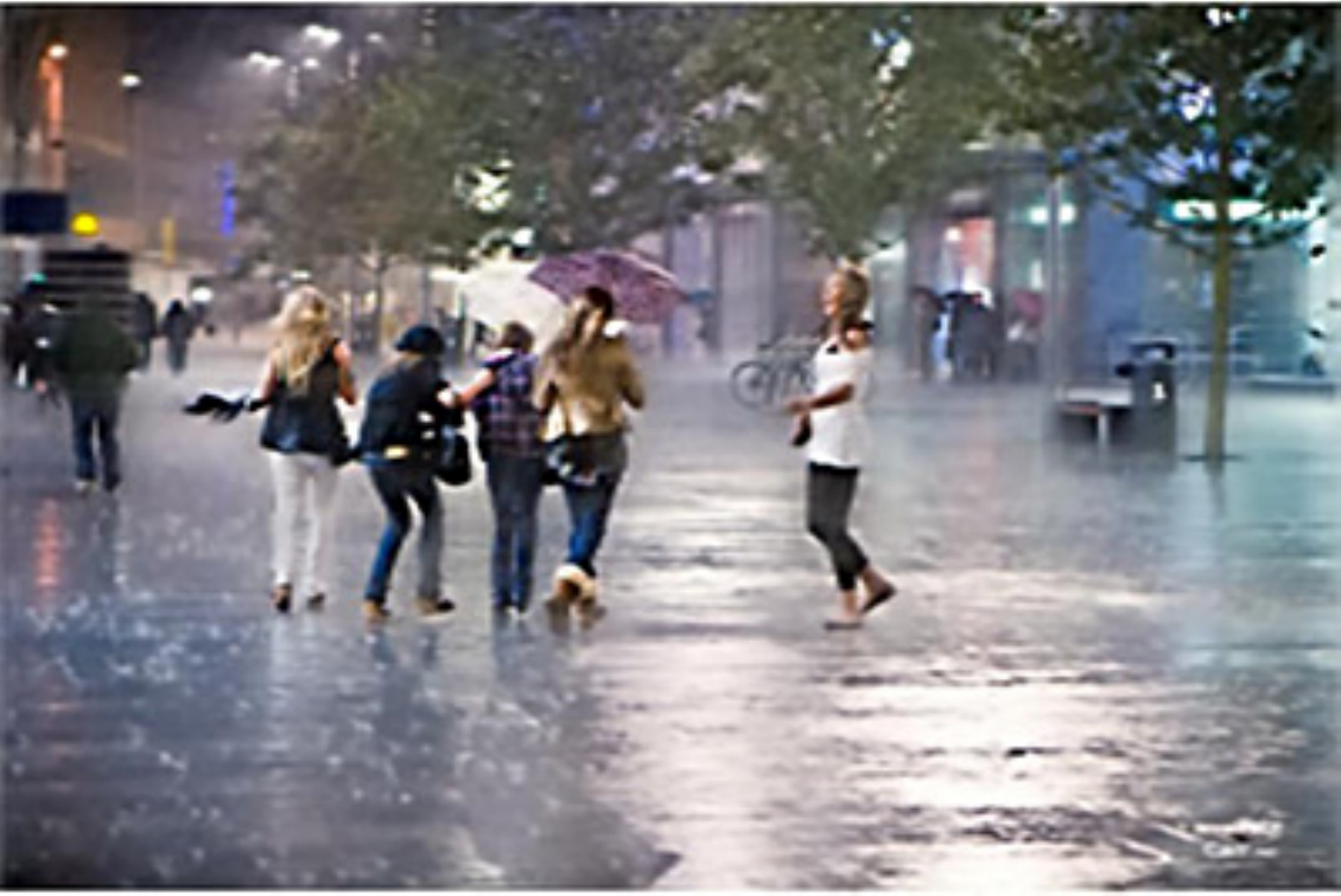}}
\end{minipage}
\hfill
\begin{minipage}{.115\linewidth}
\centering{\includegraphics[width=.995\linewidth]{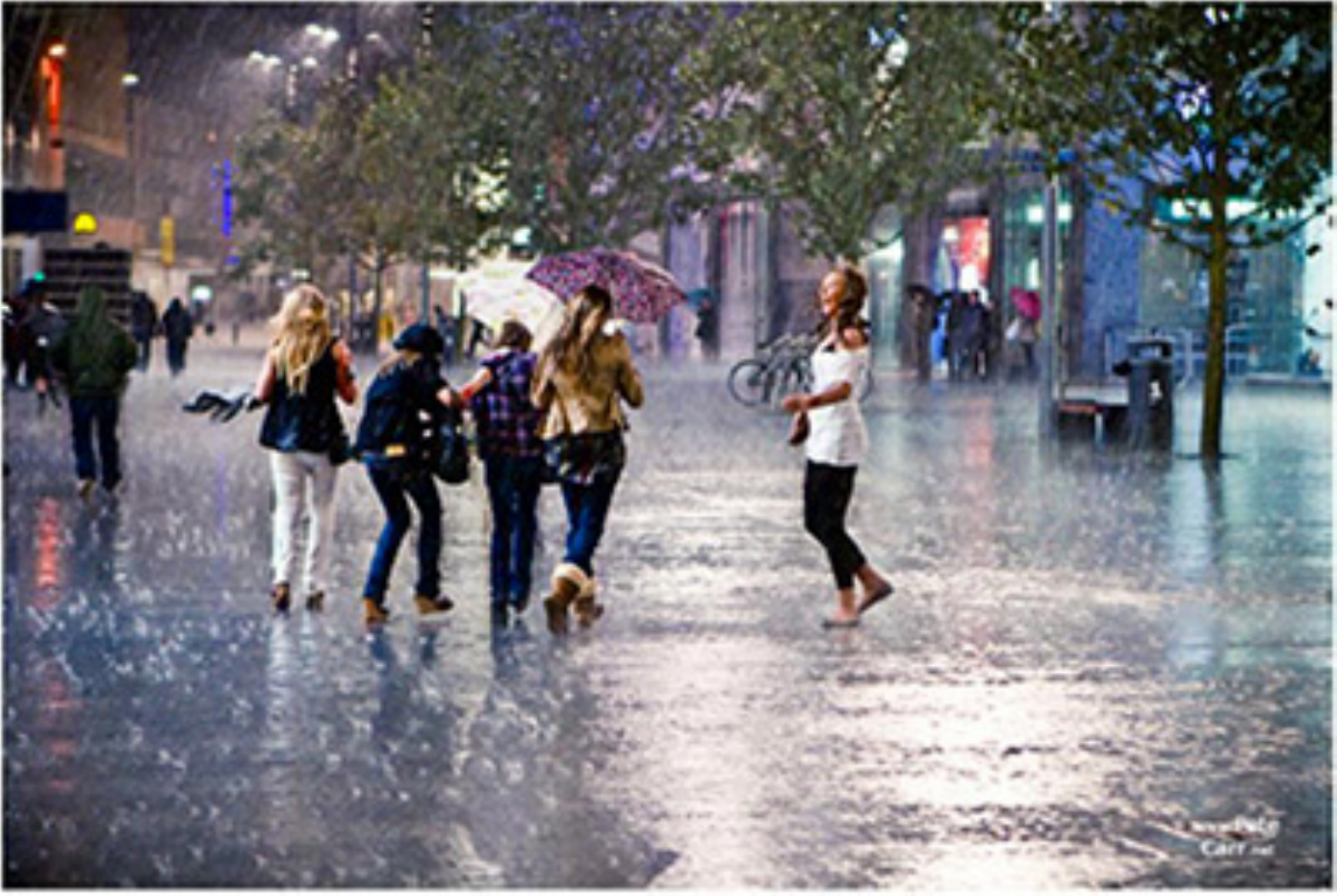}}
\end{minipage}
\hfill
\begin{minipage}{.115\linewidth}
\centering{\includegraphics[width=.995\linewidth]{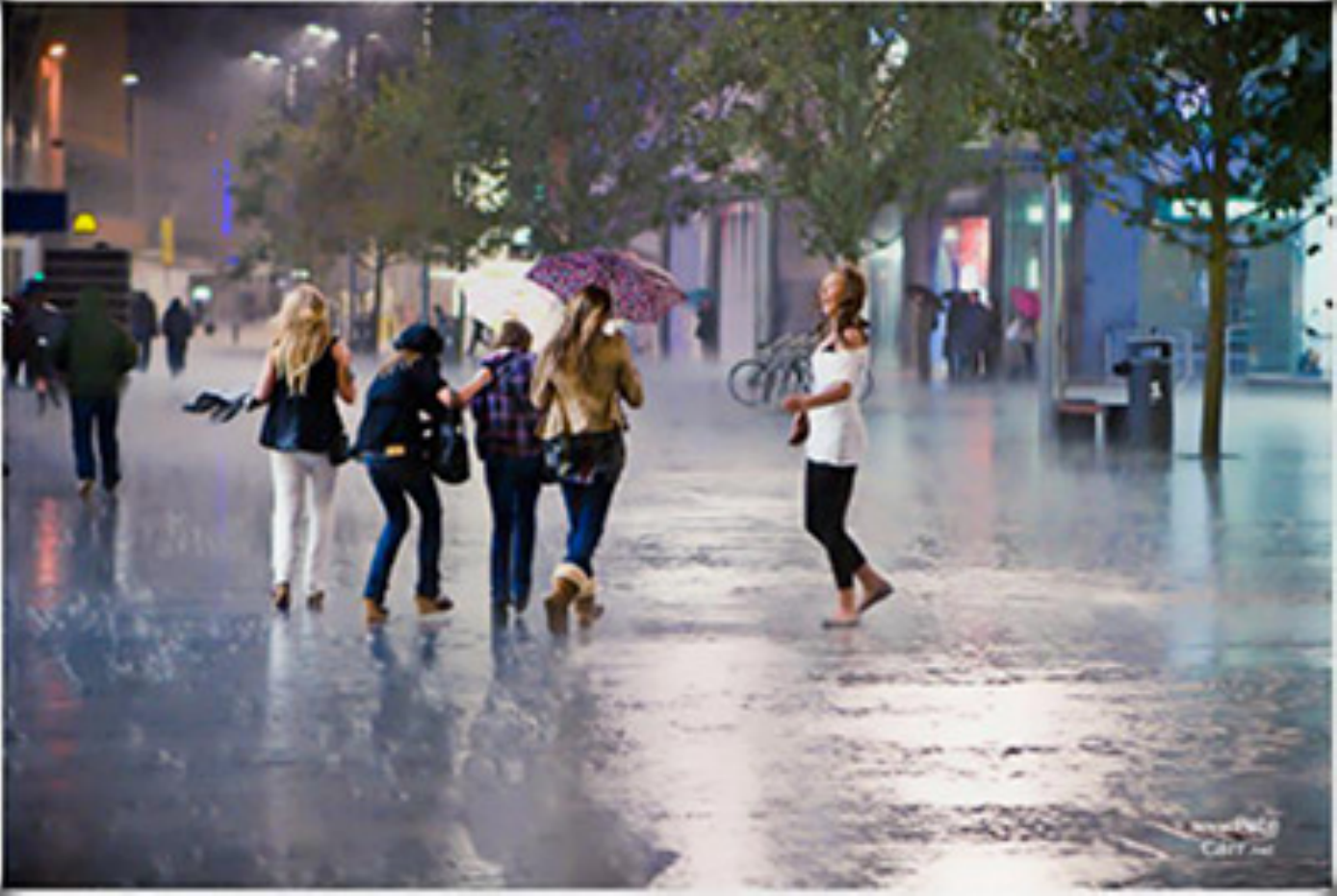}}
\end{minipage}
\hfill
\begin{minipage}{.115\linewidth}
\centering{\includegraphics[width=.995\linewidth]{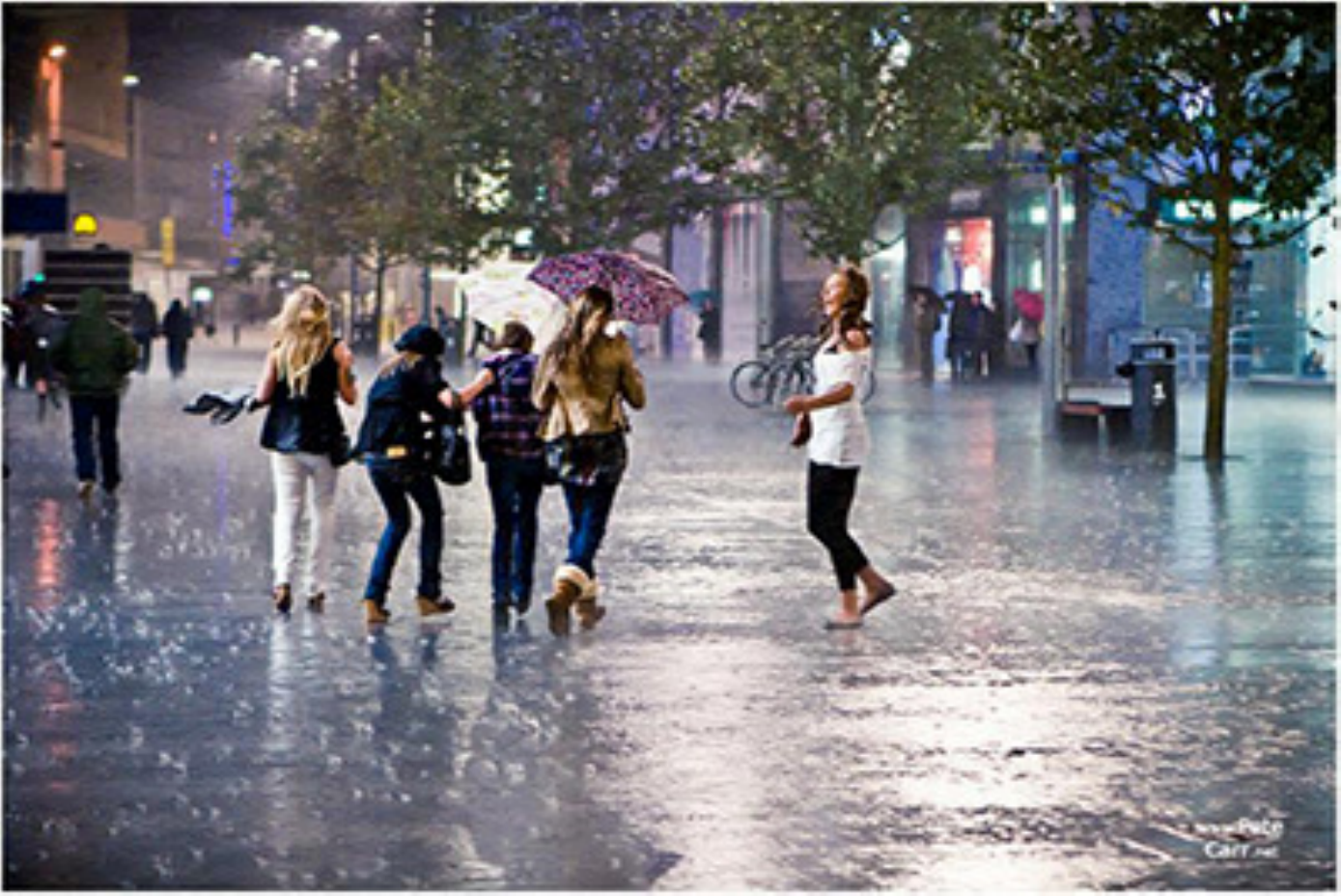}}
\end{minipage}
\hfill
\begin{minipage}{.115\linewidth}
\centering{\includegraphics[width=.995\linewidth]{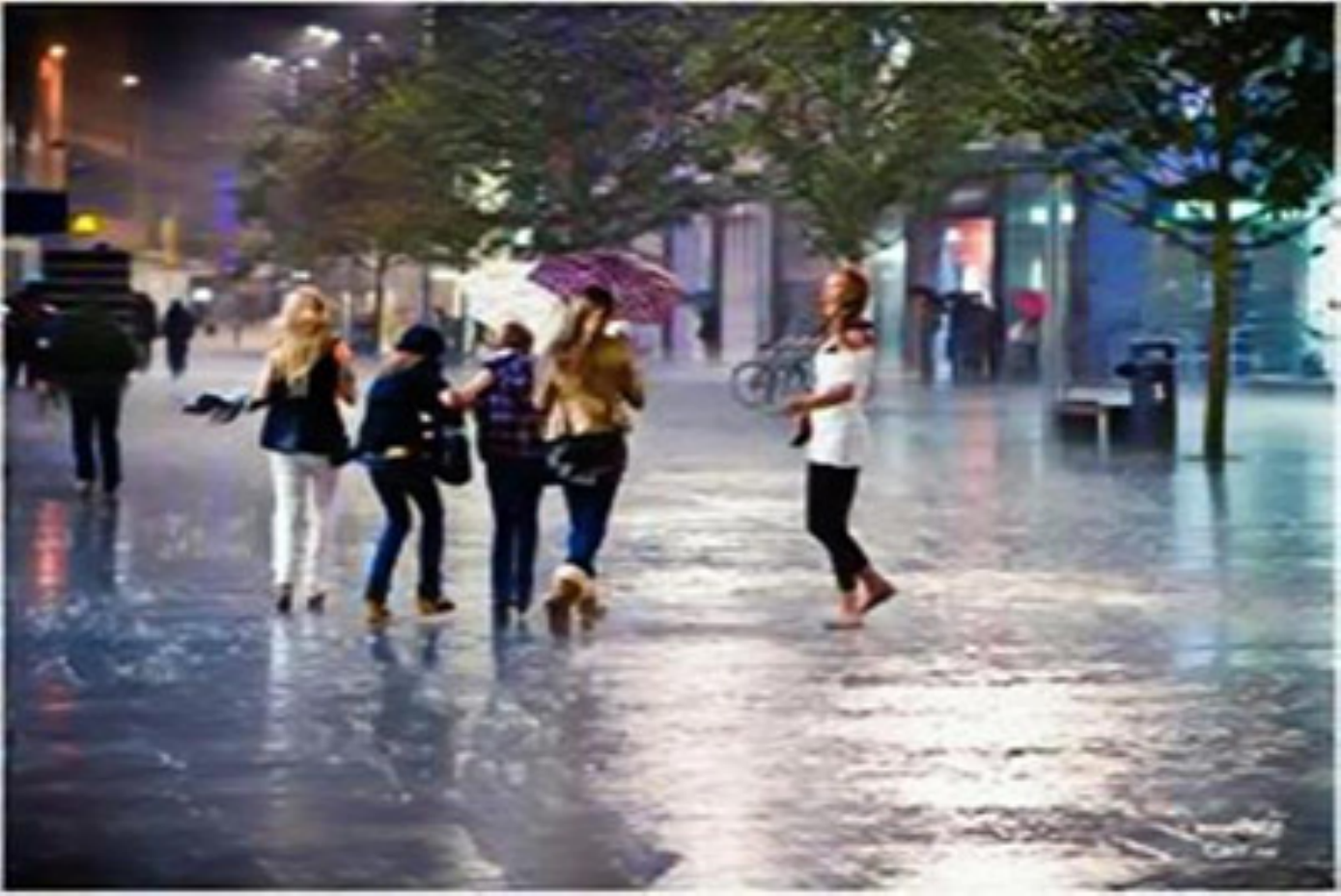}}
\end{minipage}
\hfill
\begin{minipage}{.115\linewidth}
\centering{\includegraphics[width=.995\linewidth]{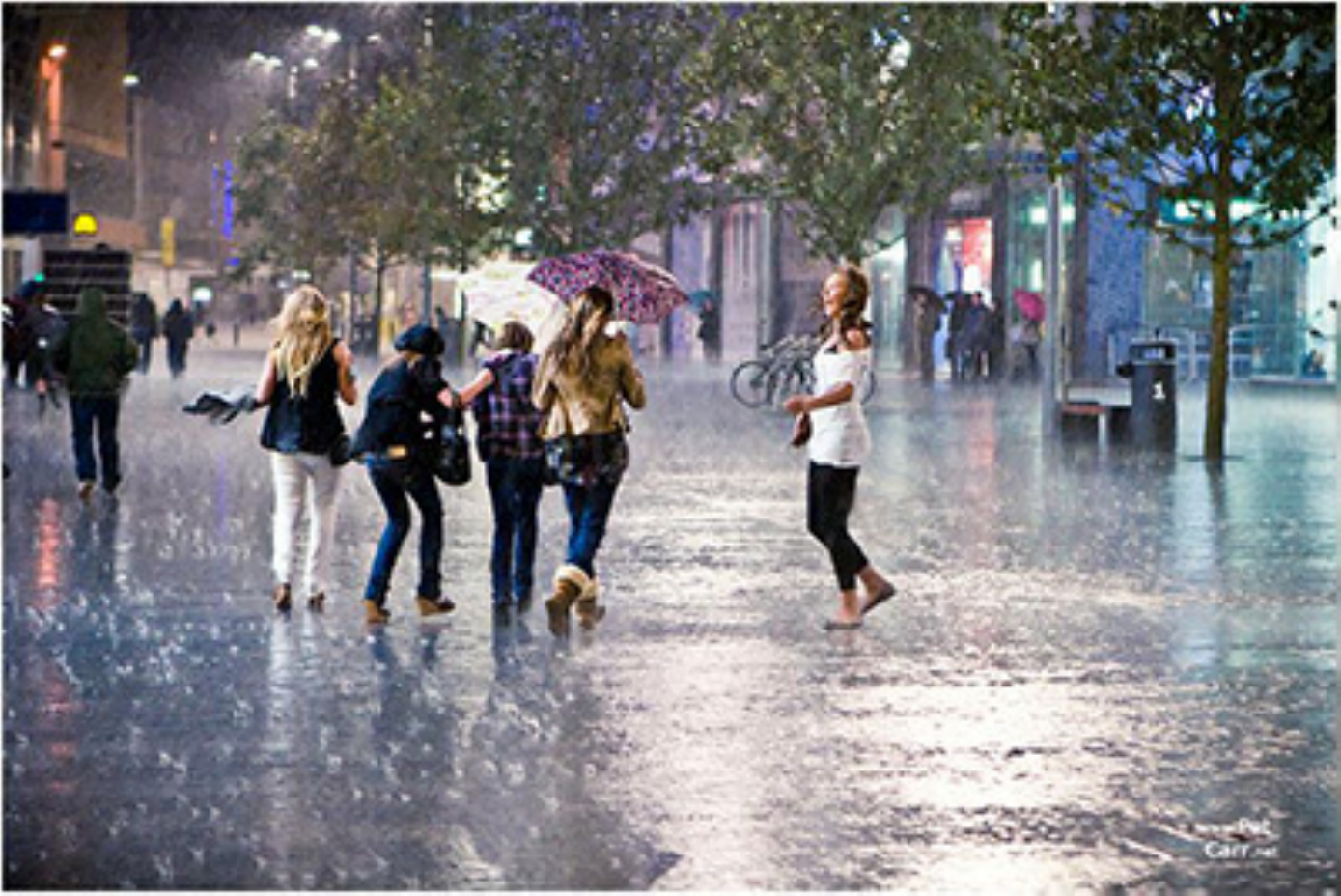}}
\end{minipage}
\vspace{0.5mm}
\vfill
\begin{minipage}{0.115\linewidth}
\centering{\includegraphics[width=.995\linewidth]{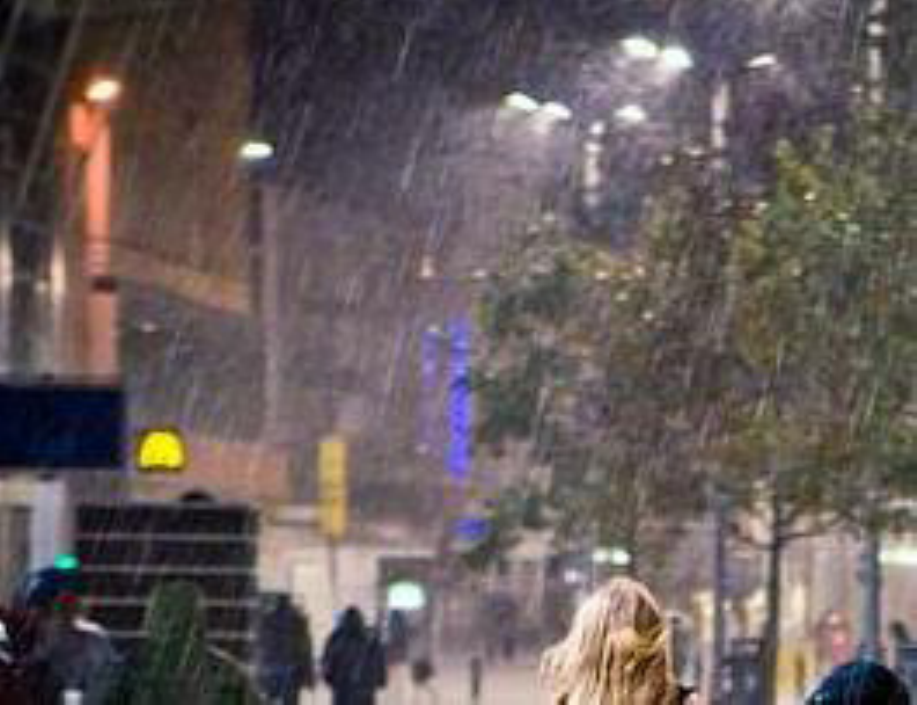}}
\end{minipage}
\hfill
\begin{minipage}{.115\linewidth}
\centering{\includegraphics[width=.995\linewidth]{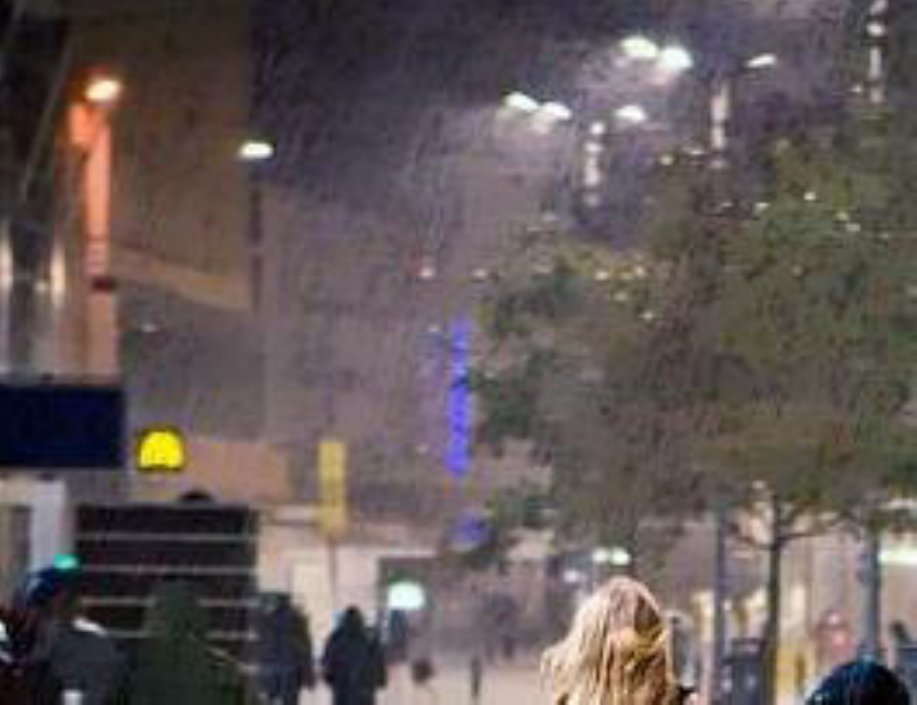}}
\end{minipage}
\hfill
\begin{minipage}{.115\linewidth}
\centering{\includegraphics[width=.995\linewidth]{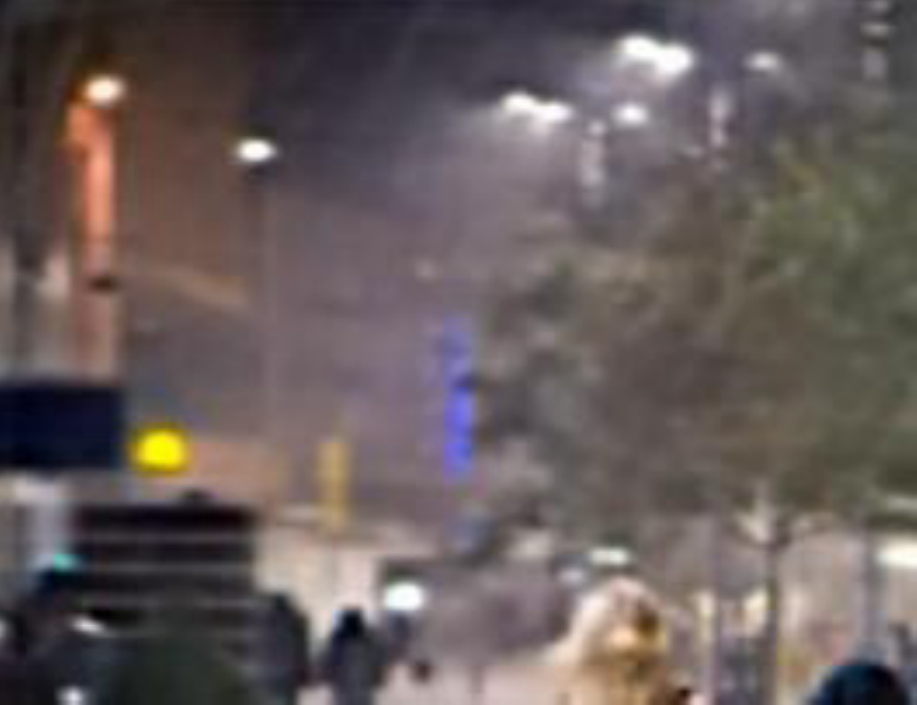}}
\end{minipage}
\hfill
\begin{minipage}{.115\linewidth}
\centering{\includegraphics[width=.995\linewidth]{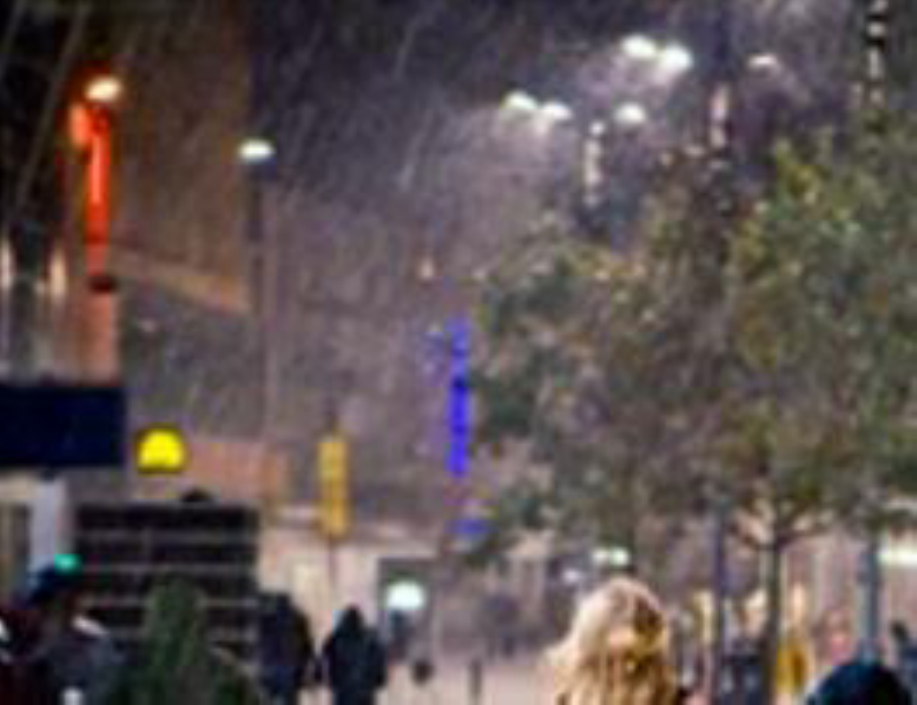}}
\end{minipage}
\hfill
\begin{minipage}{.115\linewidth}
\centering{\includegraphics[width=.995\linewidth]{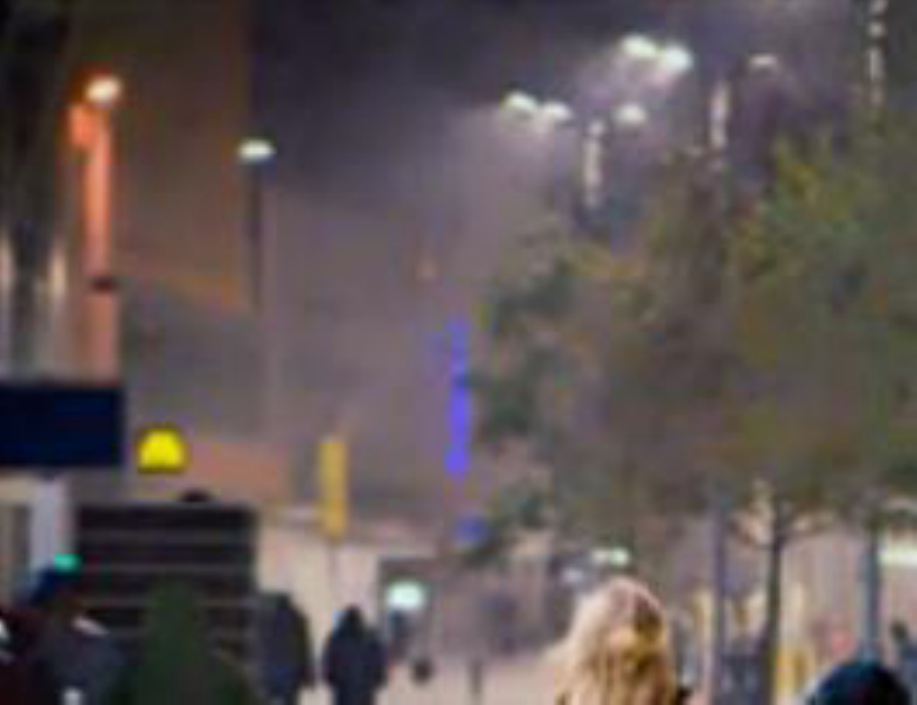}}
\end{minipage}
\hfill
\begin{minipage}{.115\linewidth}
\centering{\includegraphics[width=.995\linewidth]{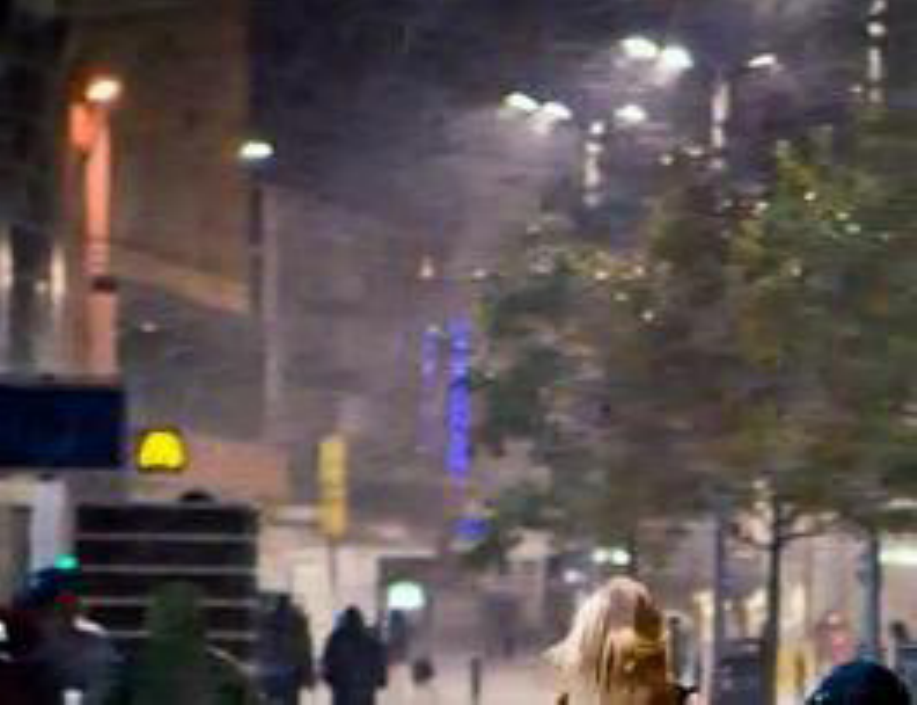}}
\end{minipage}
\hfill
\begin{minipage}{.115\linewidth}
\centering{\includegraphics[width=.995\linewidth]{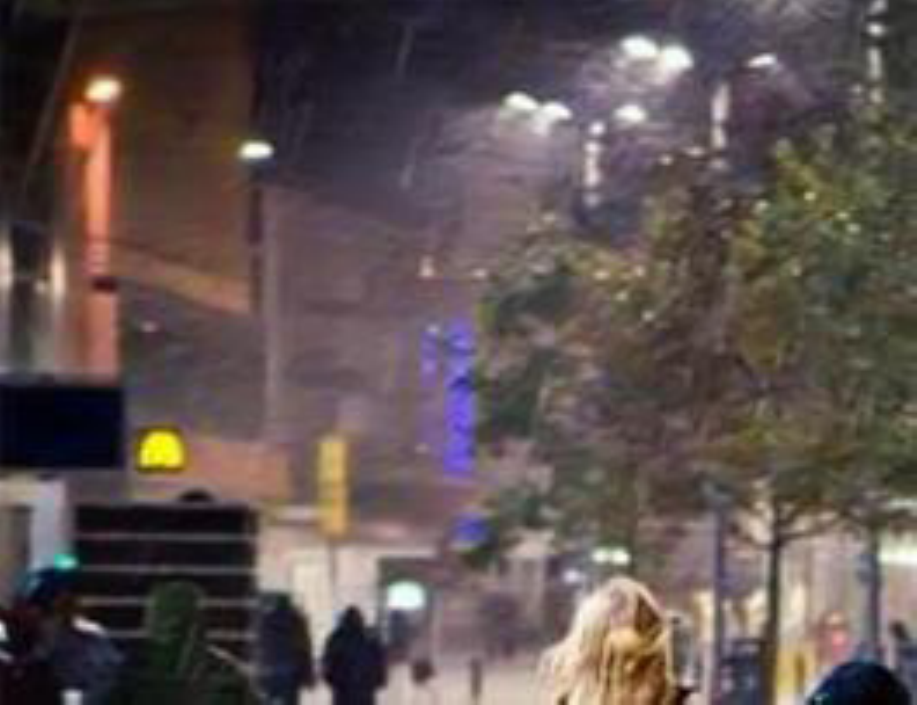}}
\end{minipage}
\hfill
\begin{minipage}{.115\linewidth}
\centering{\includegraphics[width=.995\linewidth]{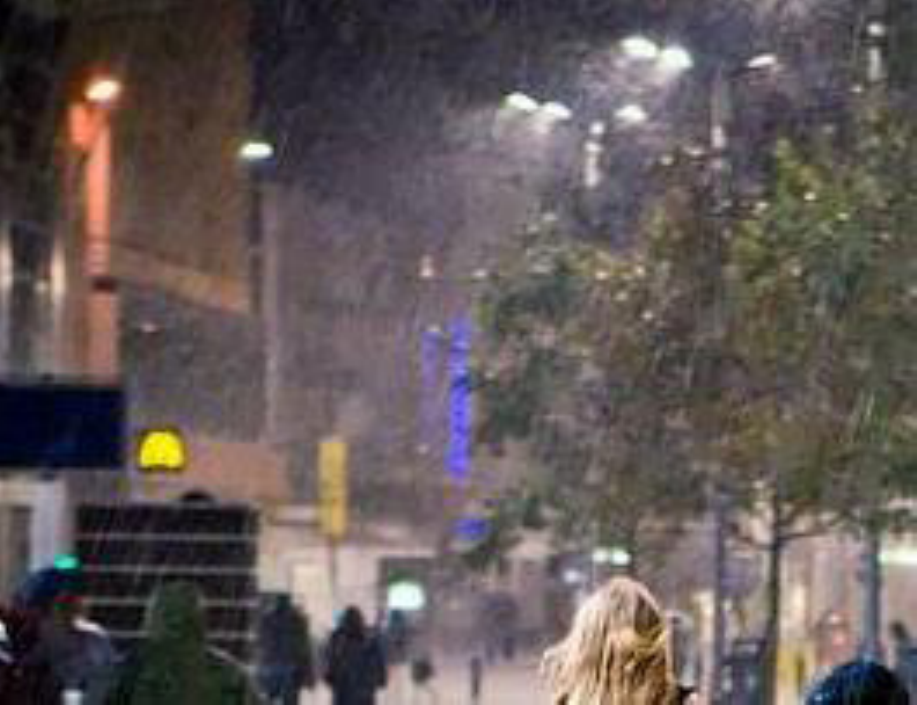}}
\end{minipage}
\vspace{0.5mm}
\vfill
\begin{minipage}{0.115\linewidth}
\centering{\includegraphics[width=.995\linewidth]{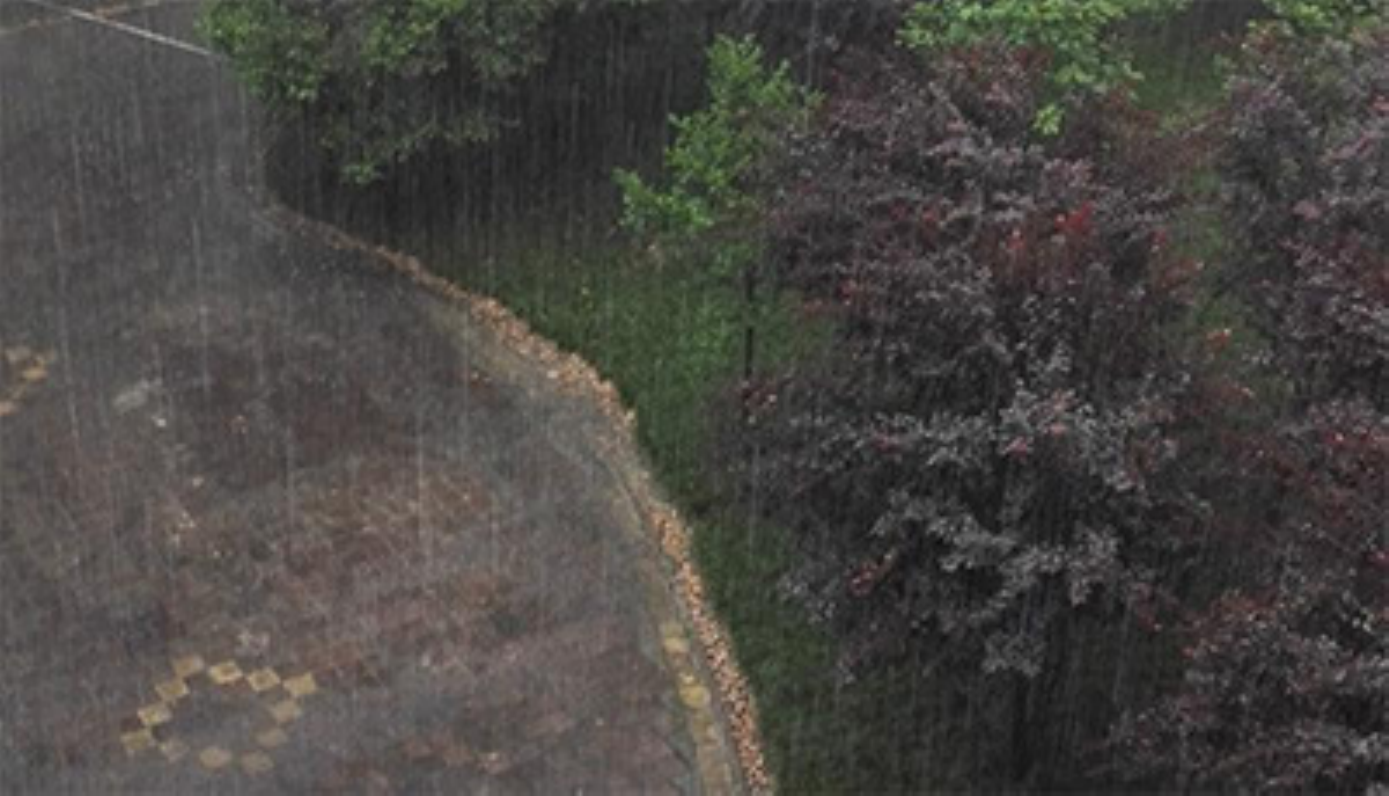}}
\end{minipage}
\hfill
\begin{minipage}{.115\linewidth}
\centering{\includegraphics[width=.995\linewidth]{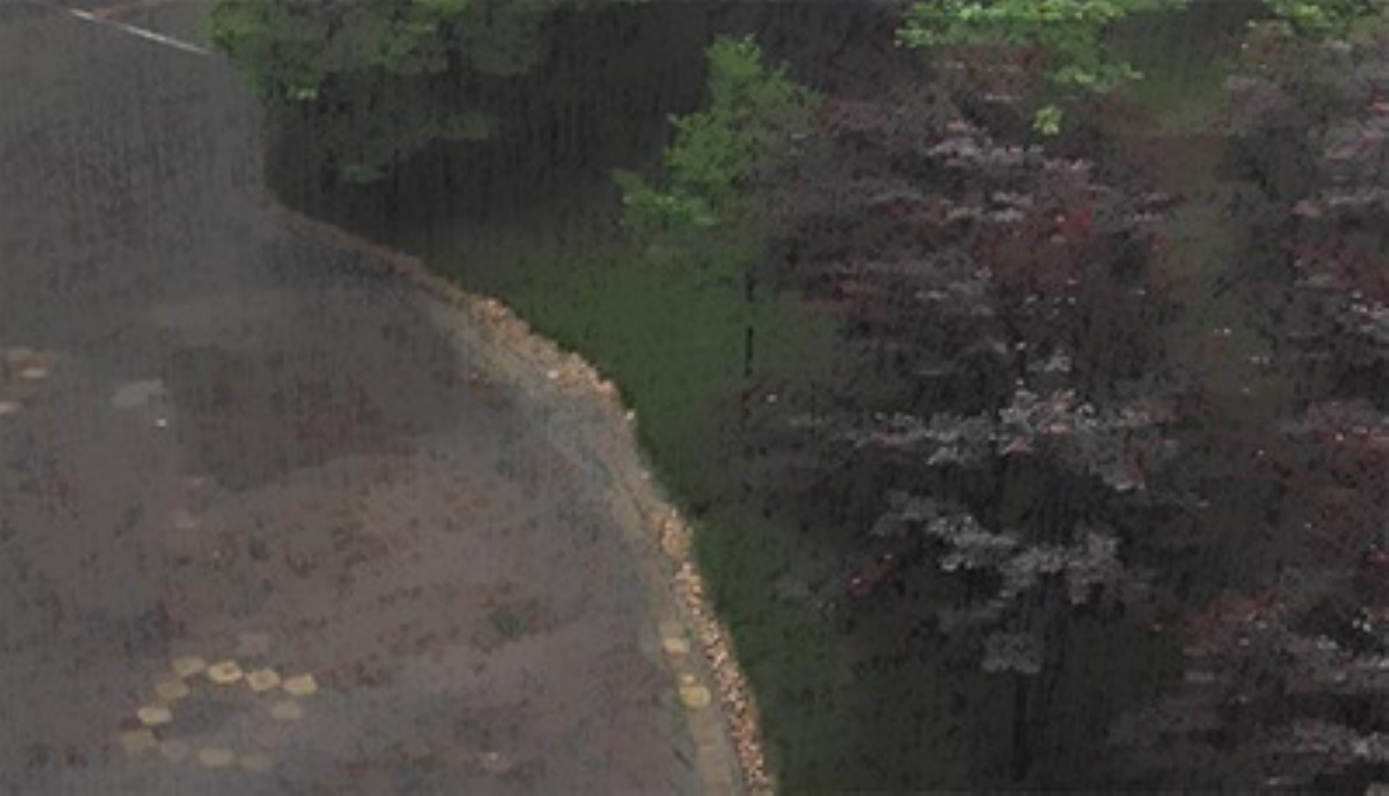}}
\end{minipage}
\hfill
\begin{minipage}{.115\linewidth}
\centering{\includegraphics[width=.995\linewidth]{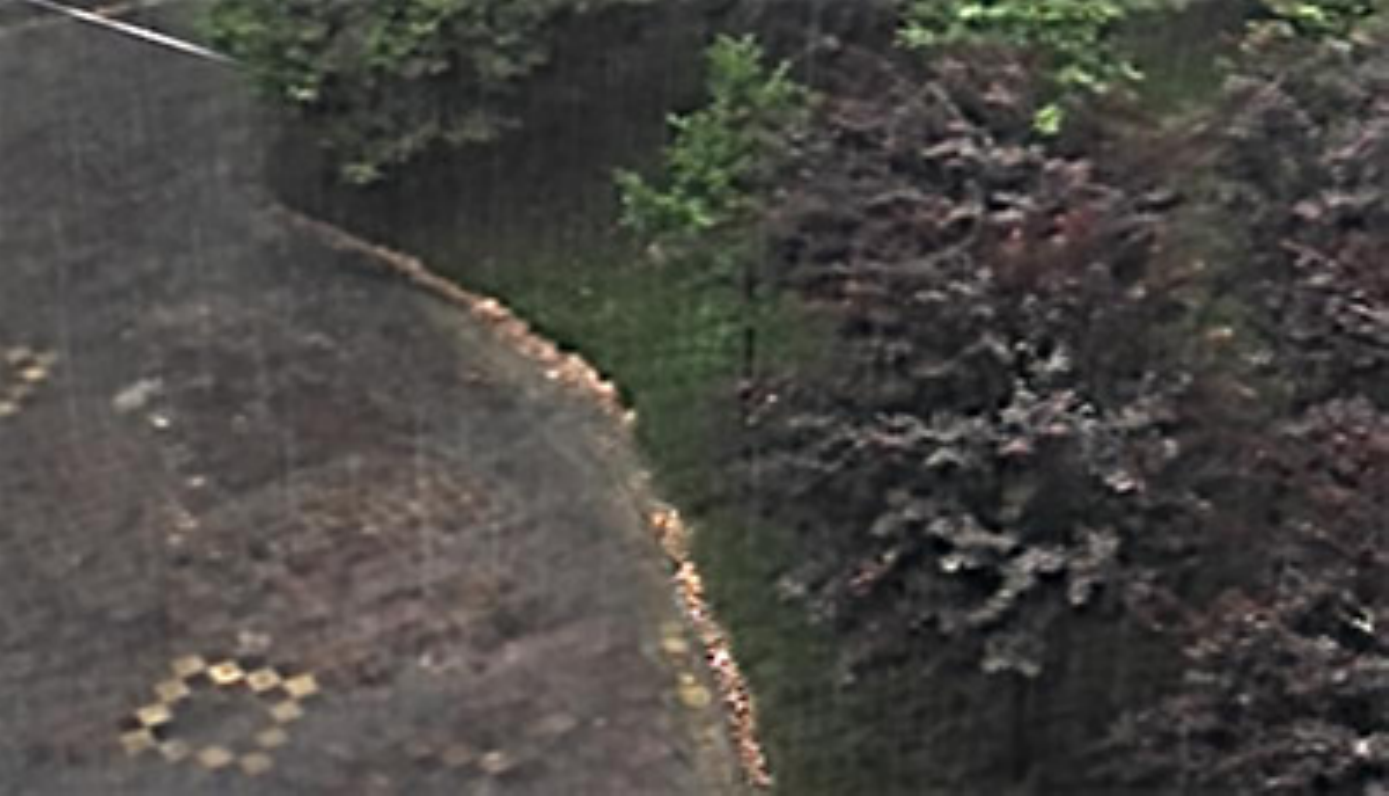}}
\end{minipage}
\hfill
\begin{minipage}{.115\linewidth}
\centering{\includegraphics[width=.995\linewidth]{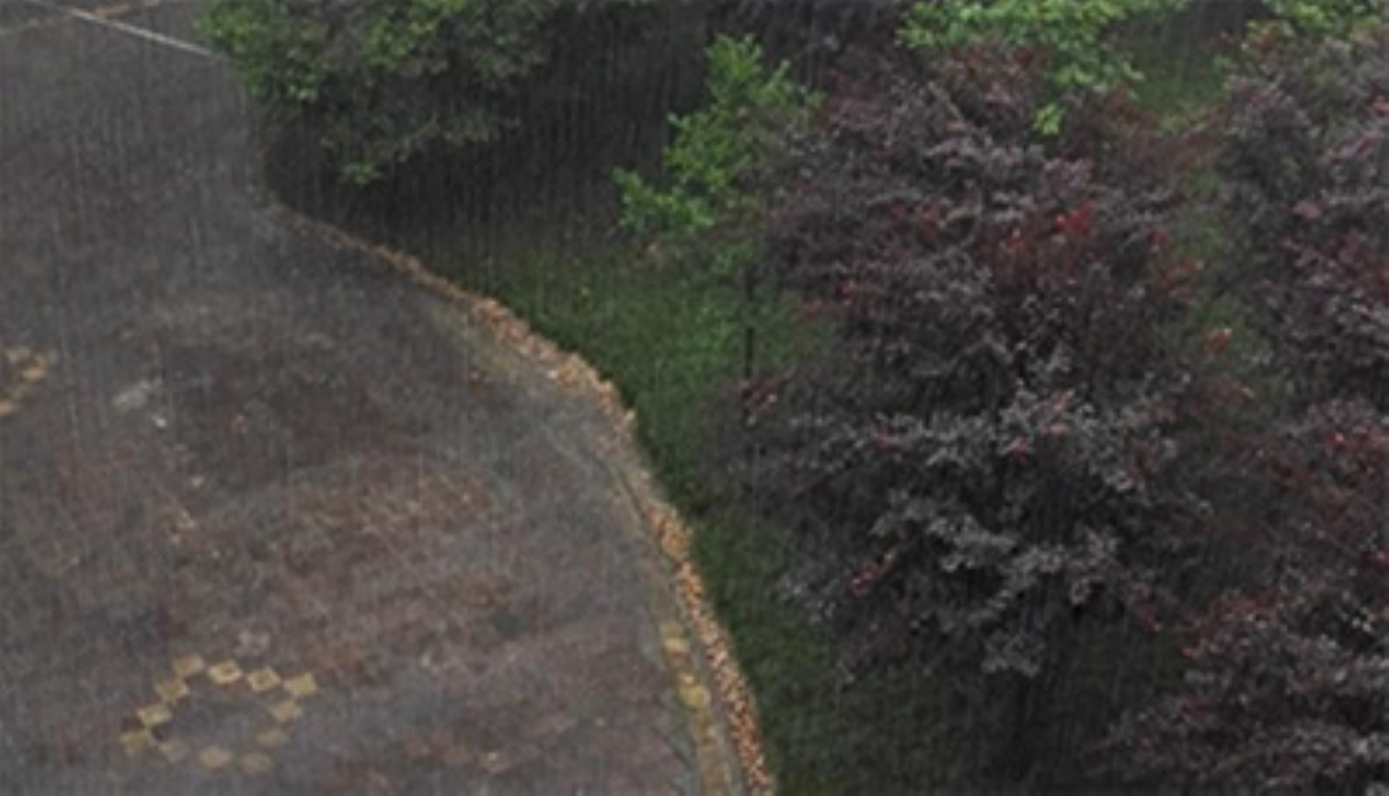}}
\end{minipage}
\hfill
\begin{minipage}{.115\linewidth}
\centering{\includegraphics[width=.995\linewidth]{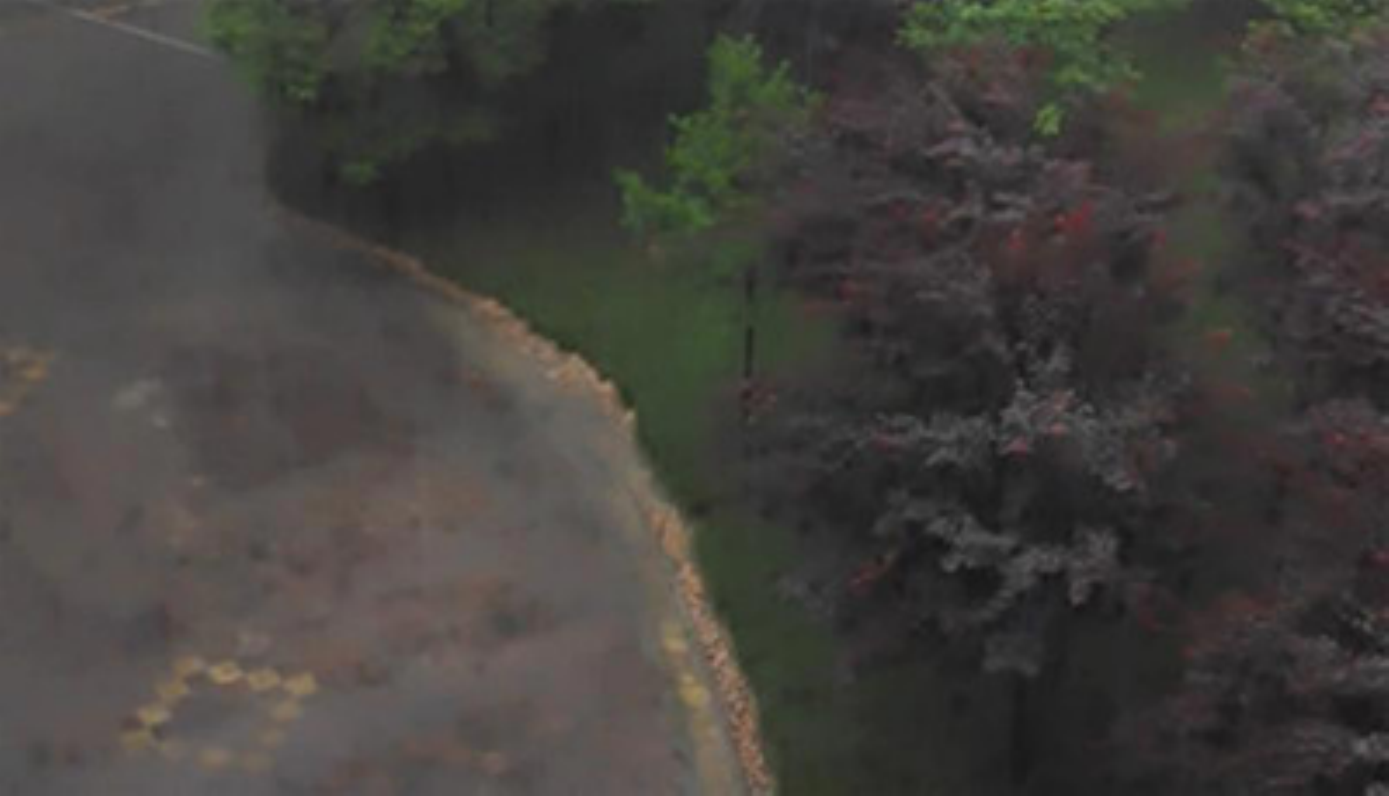}}
\end{minipage}
\hfill
\begin{minipage}{.115\linewidth}
\centering{\includegraphics[width=.995\linewidth]{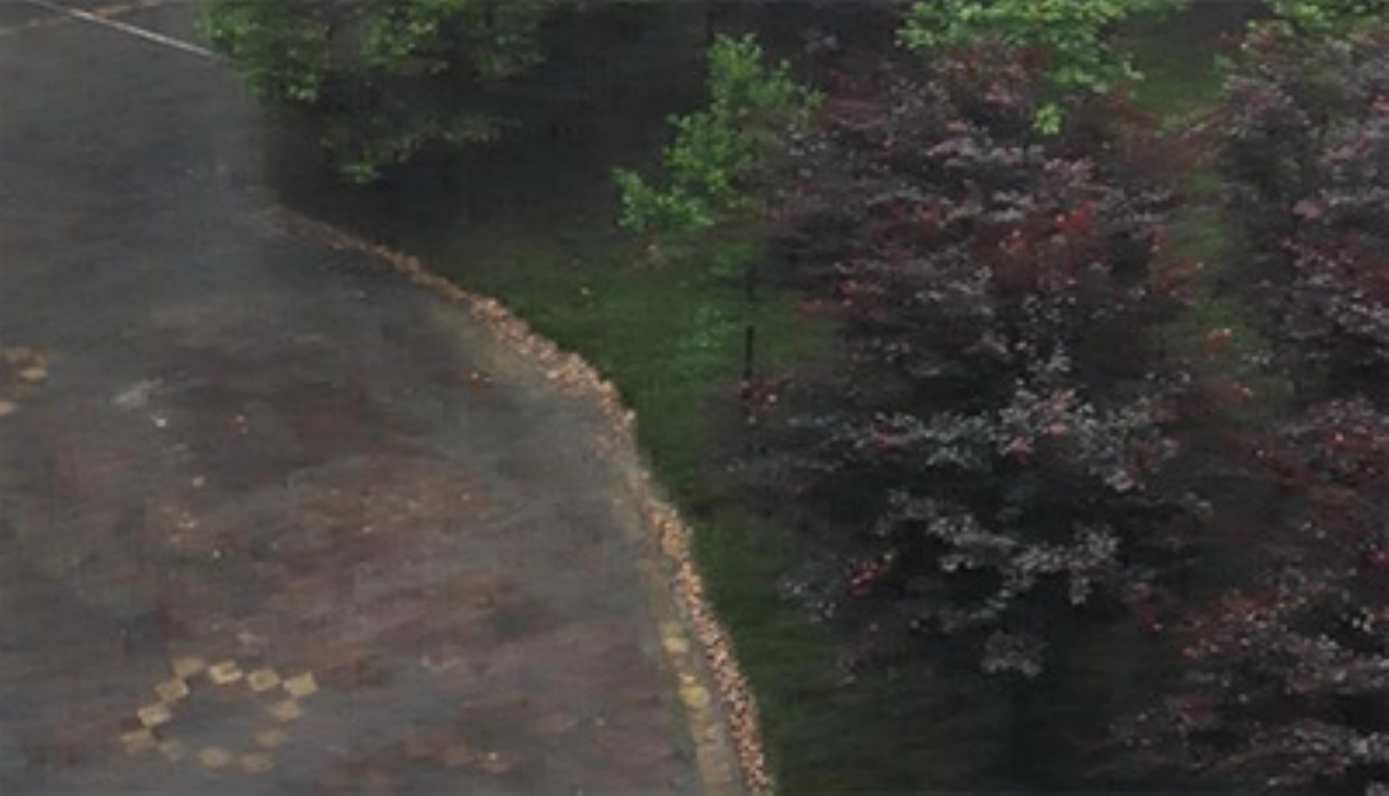}}
\end{minipage}
\hfill
\begin{minipage}{.115\linewidth}
\centering{\includegraphics[width=.995\linewidth]{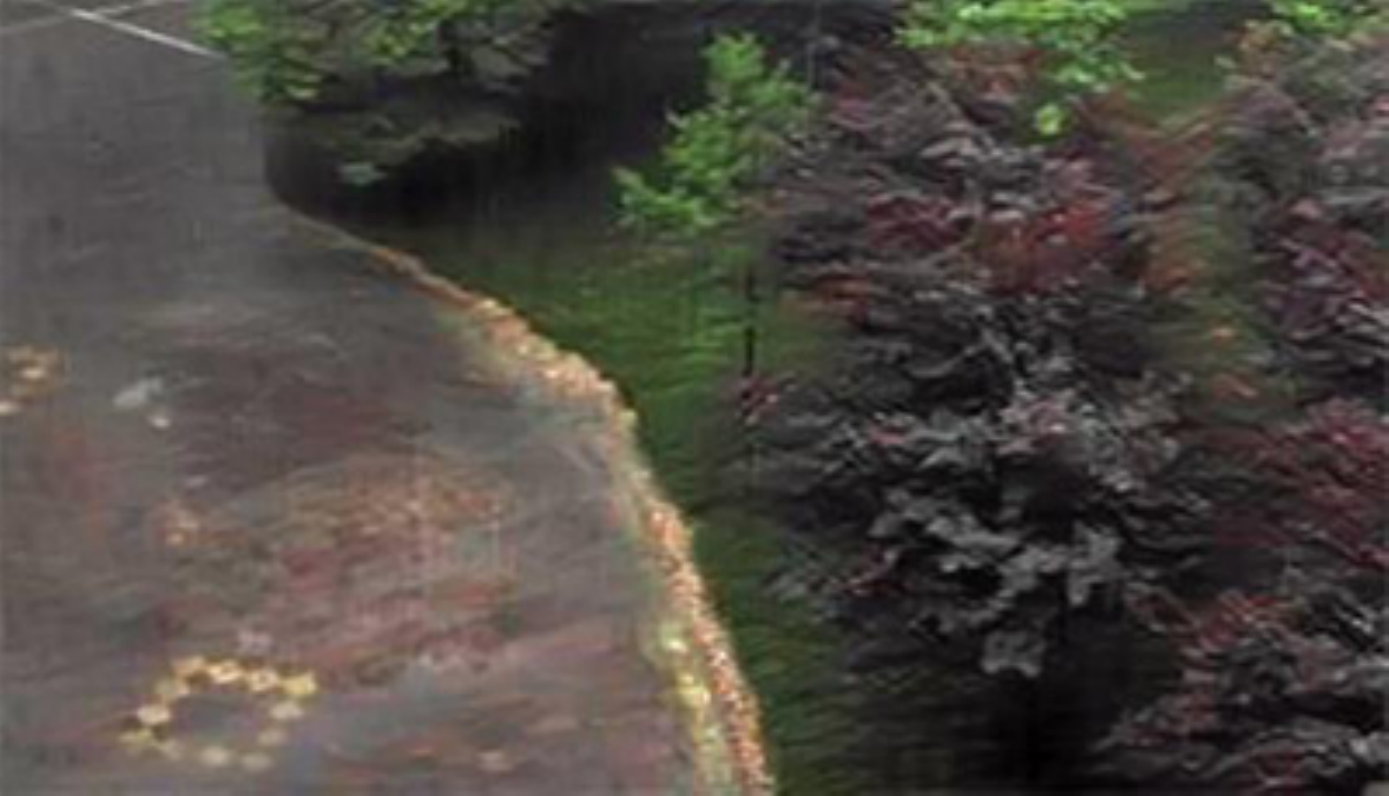}}
\end{minipage}
\hfill
\begin{minipage}{.115\linewidth}
\centering{\includegraphics[width=.995\linewidth]{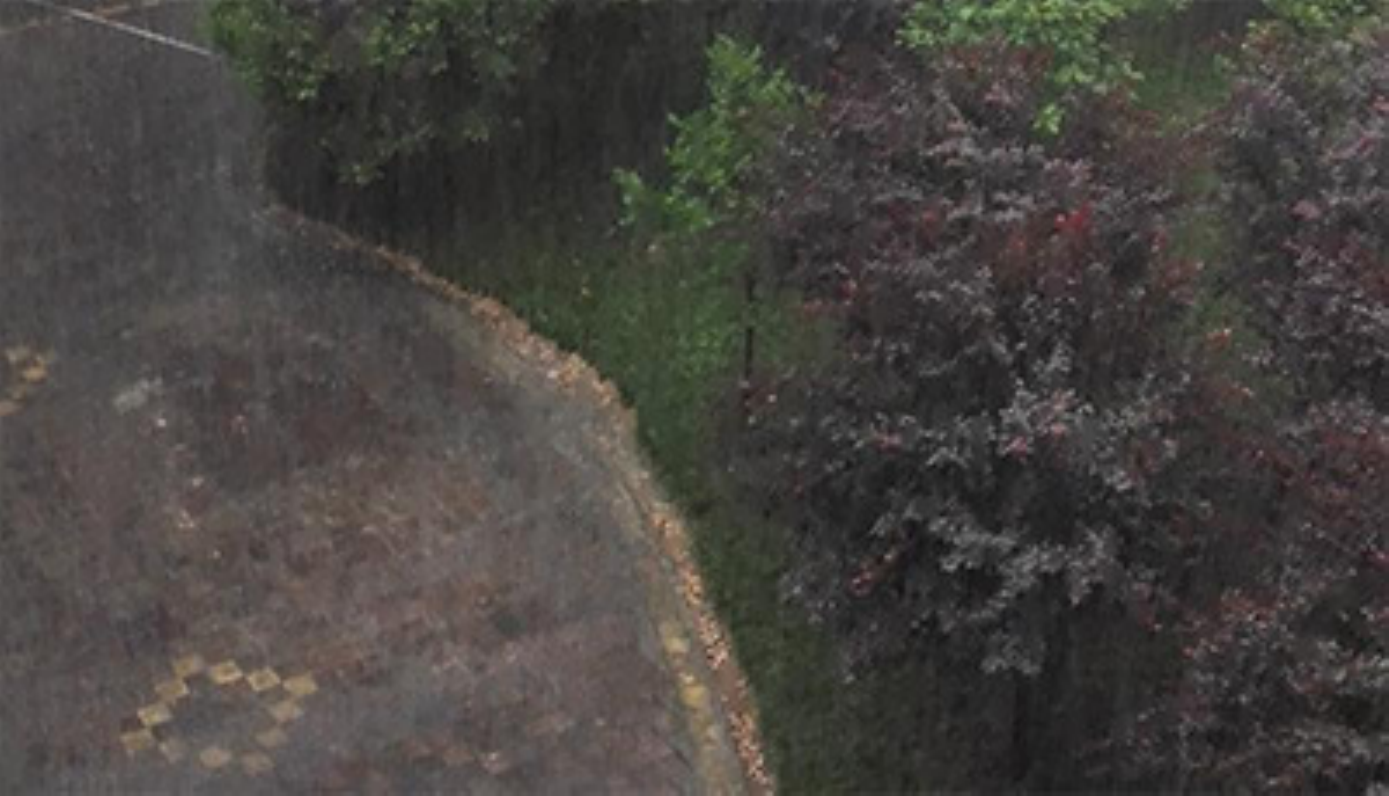}}
\end{minipage}
\vspace{0.5mm}
\vfill
\begin{minipage}{0.115\linewidth}
\centering{\includegraphics[width=.995\linewidth]{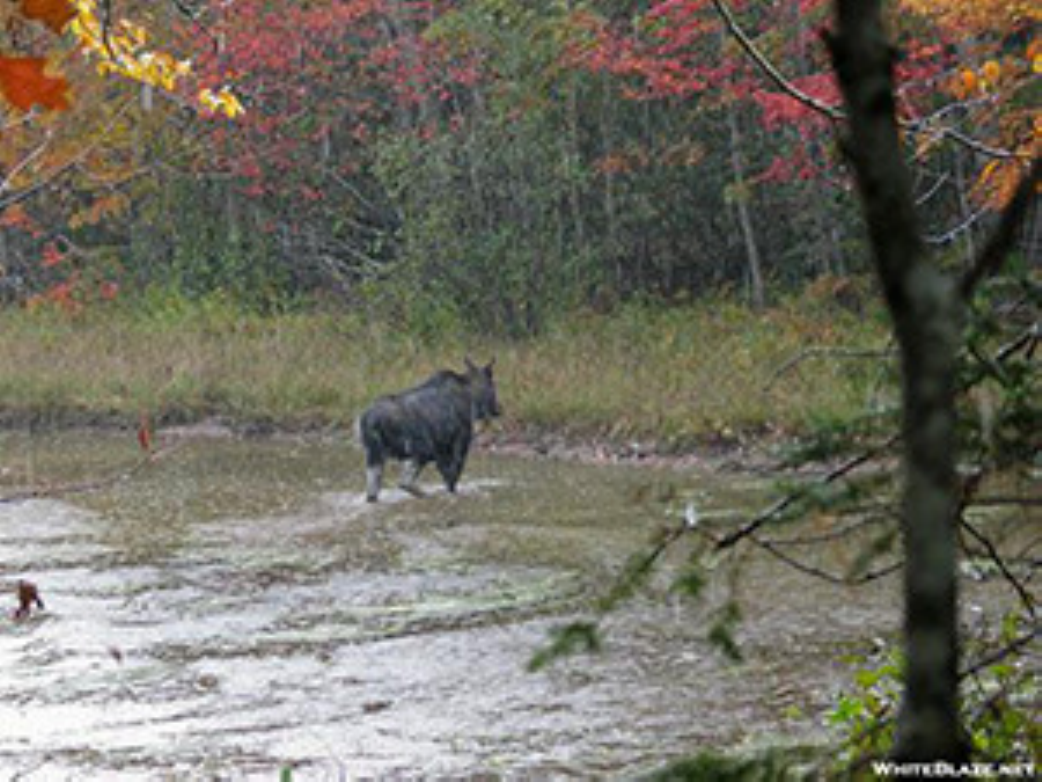}}
\end{minipage}
\hfill
\begin{minipage}{.115\linewidth}
\centering{\includegraphics[width=.995\linewidth]{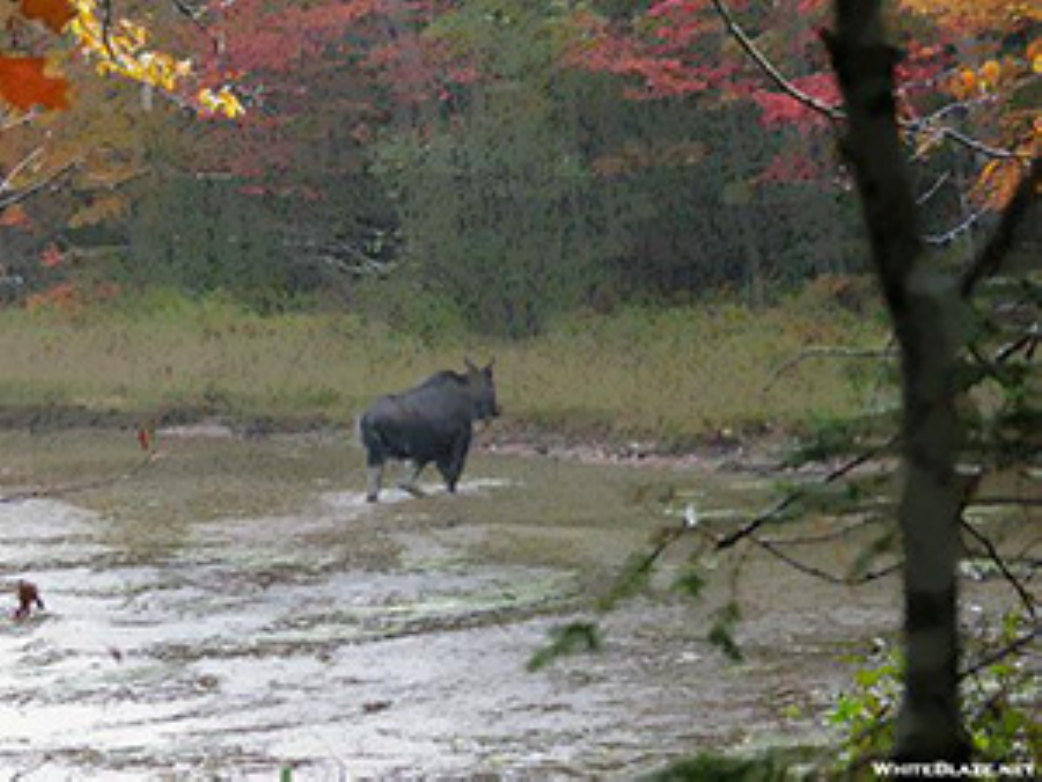}}
\end{minipage}
\hfill
\begin{minipage}{.115\linewidth}
\centering{\includegraphics[width=.995\linewidth]{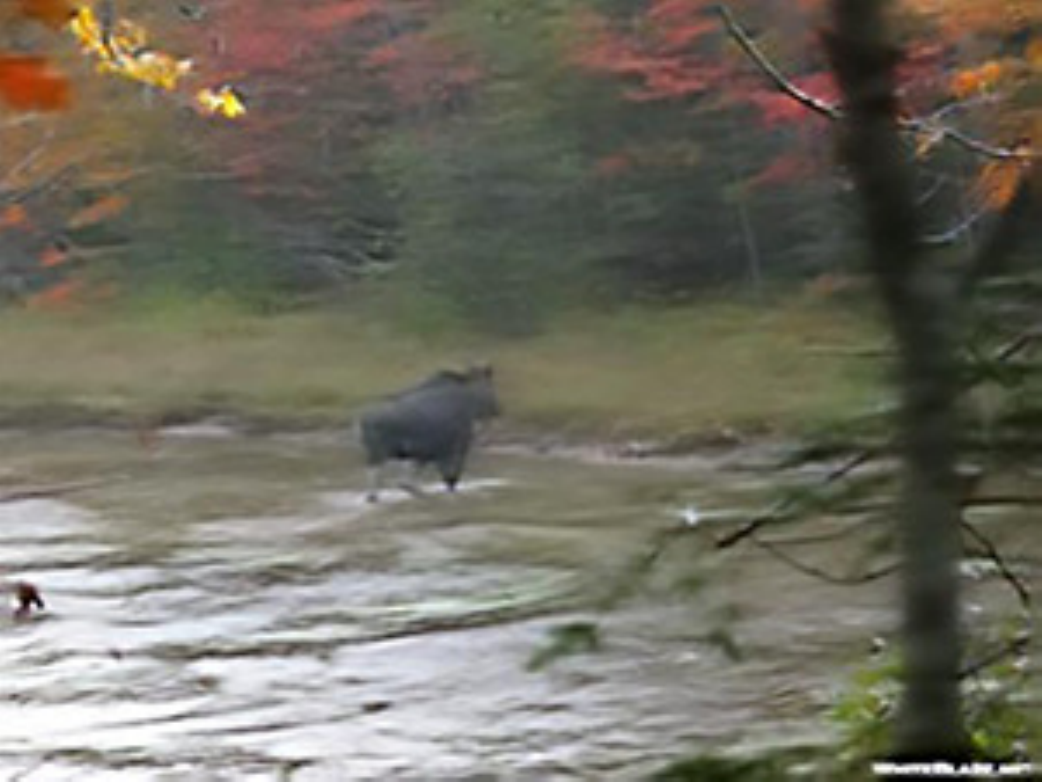}}
\end{minipage}
\hfill
\begin{minipage}{.115\linewidth}
\centering{\includegraphics[width=.995\linewidth]{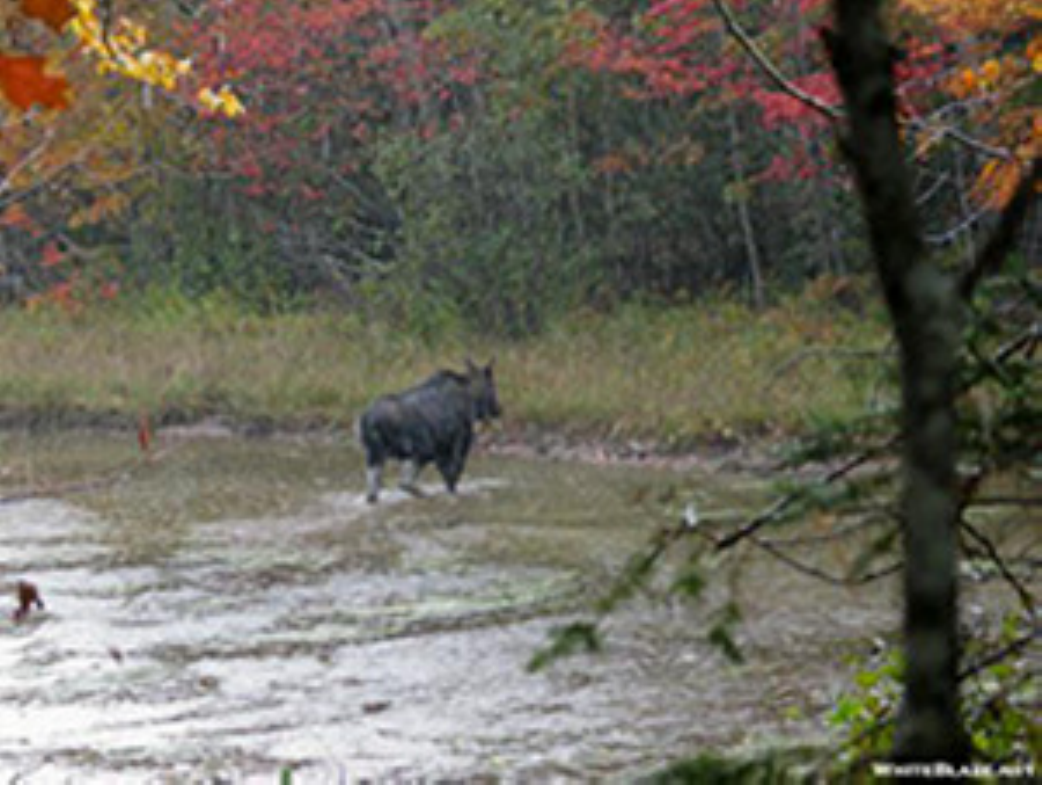}}
\end{minipage}
\hfill
\begin{minipage}{.115\linewidth}
\centering{\includegraphics[width=.995\linewidth]{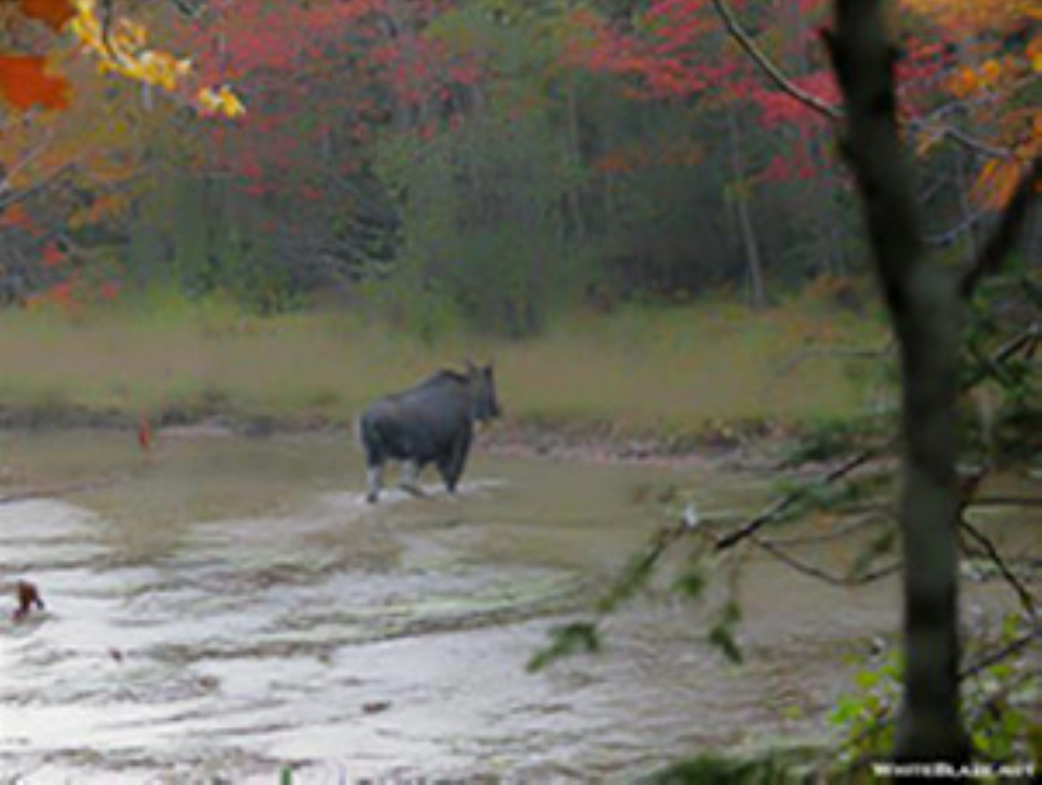}}
\end{minipage}
\hfill
\begin{minipage}{.115\linewidth}
\centering{\includegraphics[width=.995\linewidth]{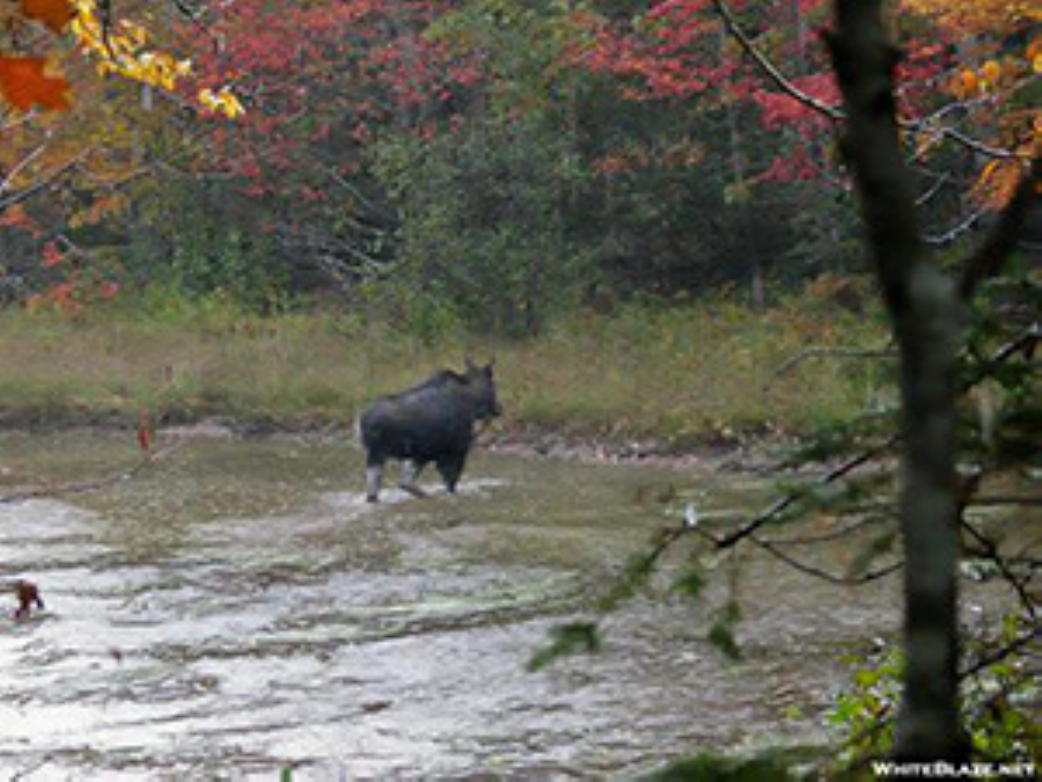}}
\end{minipage}
\hfill
\begin{minipage}{.115\linewidth}
\centering{\includegraphics[width=.995\linewidth]{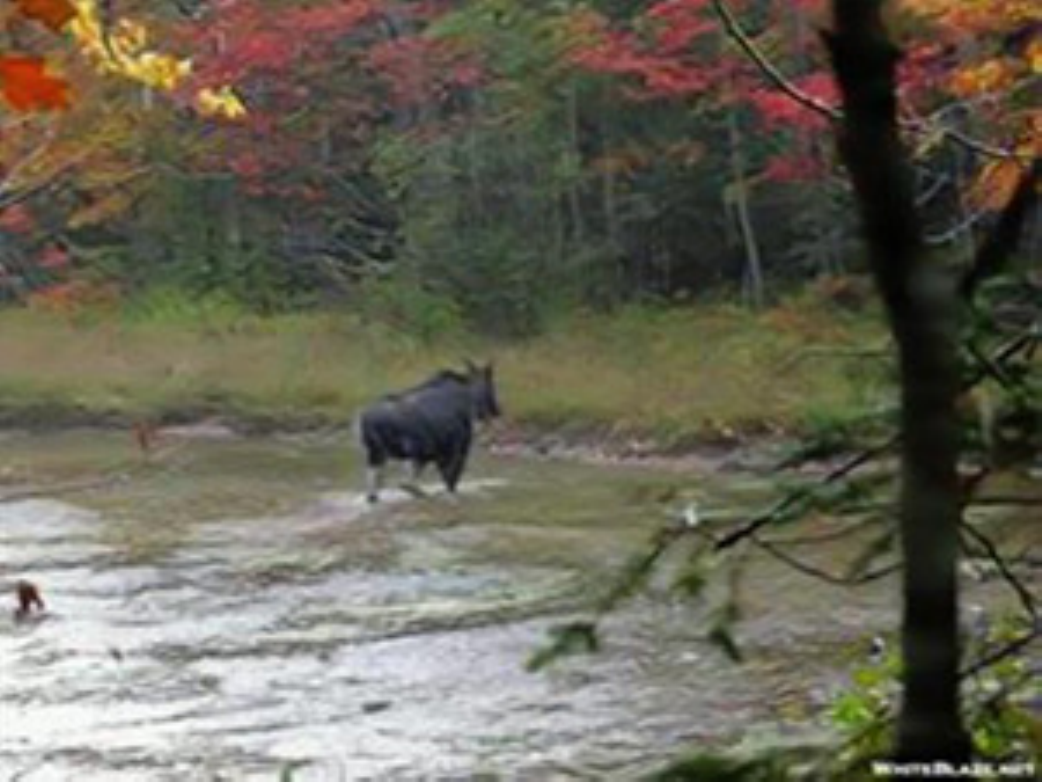}}
\end{minipage}
\hfill
\begin{minipage}{.115\linewidth}
\centering{\includegraphics[width=.995\linewidth]{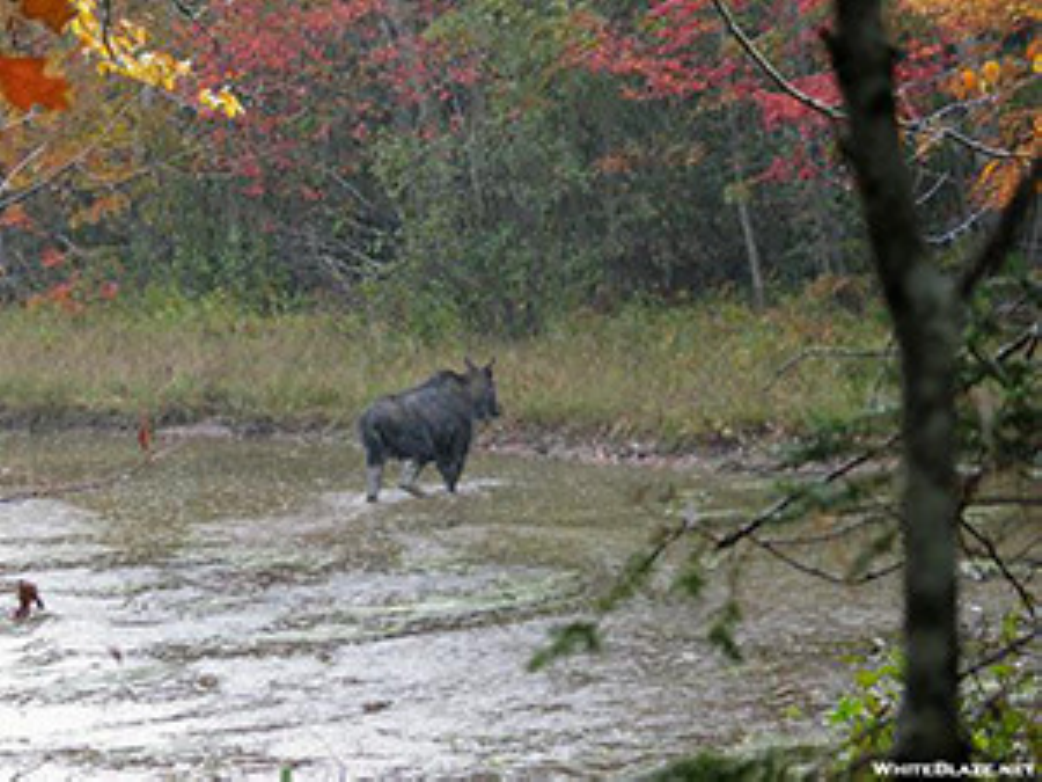}}
\end{minipage}
\vspace{0.5mm}
\vfill
\begin{minipage}{0.115\linewidth}
\centering{\includegraphics[width=.995\linewidth]{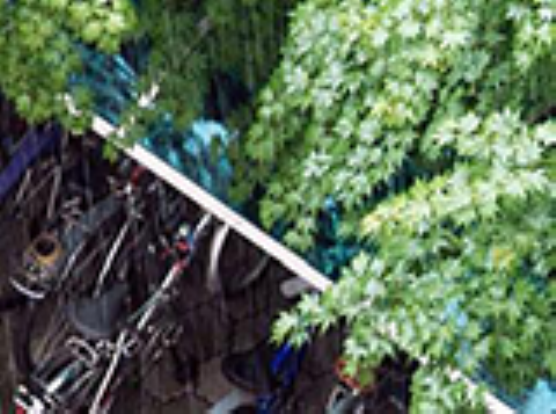}}
\end{minipage}
\hfill
\begin{minipage}{.115\linewidth}
\centering{\includegraphics[width=.995\linewidth]{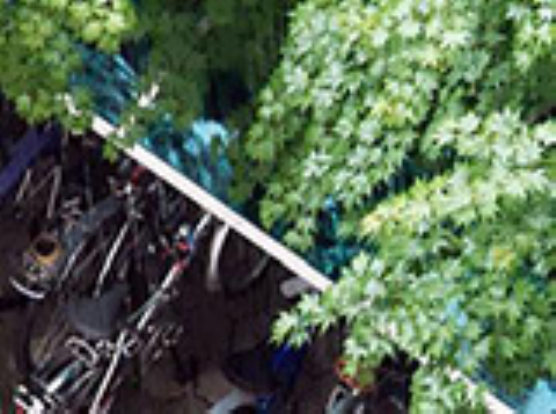}}
\end{minipage}
\hfill
\begin{minipage}{.115\linewidth}
\centering{\includegraphics[width=.995\linewidth]{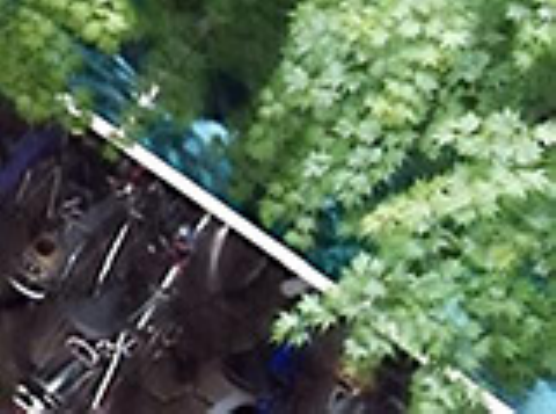}}
\end{minipage}
\hfill
\begin{minipage}{.115\linewidth}
\centering{\includegraphics[width=.995\linewidth]{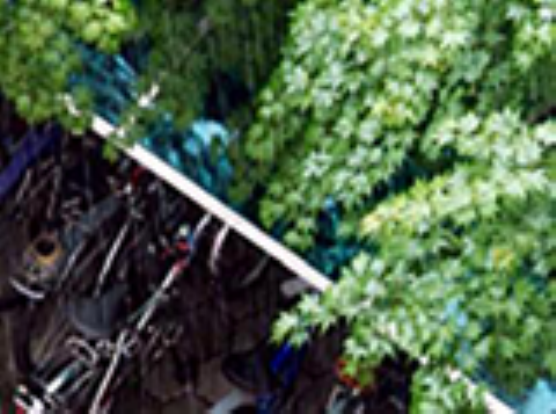}}
\end{minipage}
\hfill
\begin{minipage}{.115\linewidth}
\centering{\includegraphics[width=.995\linewidth]{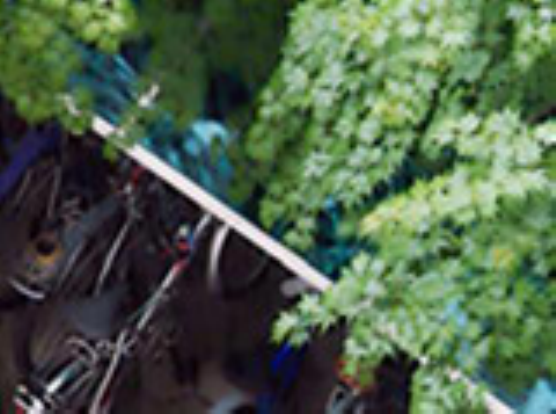}}
\end{minipage}
\hfill
\begin{minipage}{.115\linewidth}
\centering{\includegraphics[width=.995\linewidth]{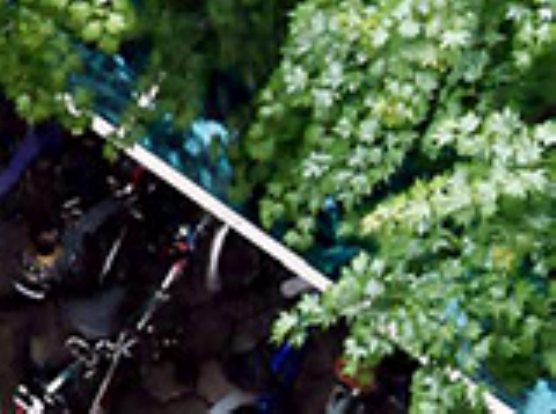}}
\end{minipage}
\hfill
\begin{minipage}{.115\linewidth}
\centering{\includegraphics[width=.995\linewidth]{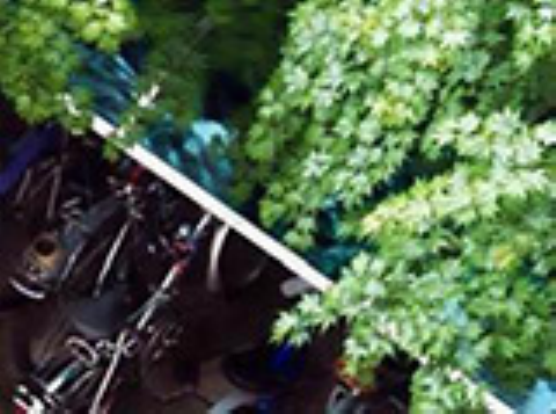}}
\end{minipage}
\hfill
\begin{minipage}{.115\linewidth}
\centering{\includegraphics[width=.995\linewidth]{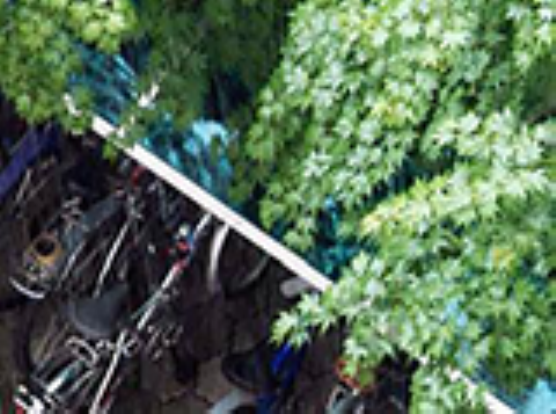}}
\end{minipage}
\vspace{0.5mm}
\vfill
\begin{minipage}{0.115\linewidth}
\centering{\includegraphics[width=.995\linewidth]{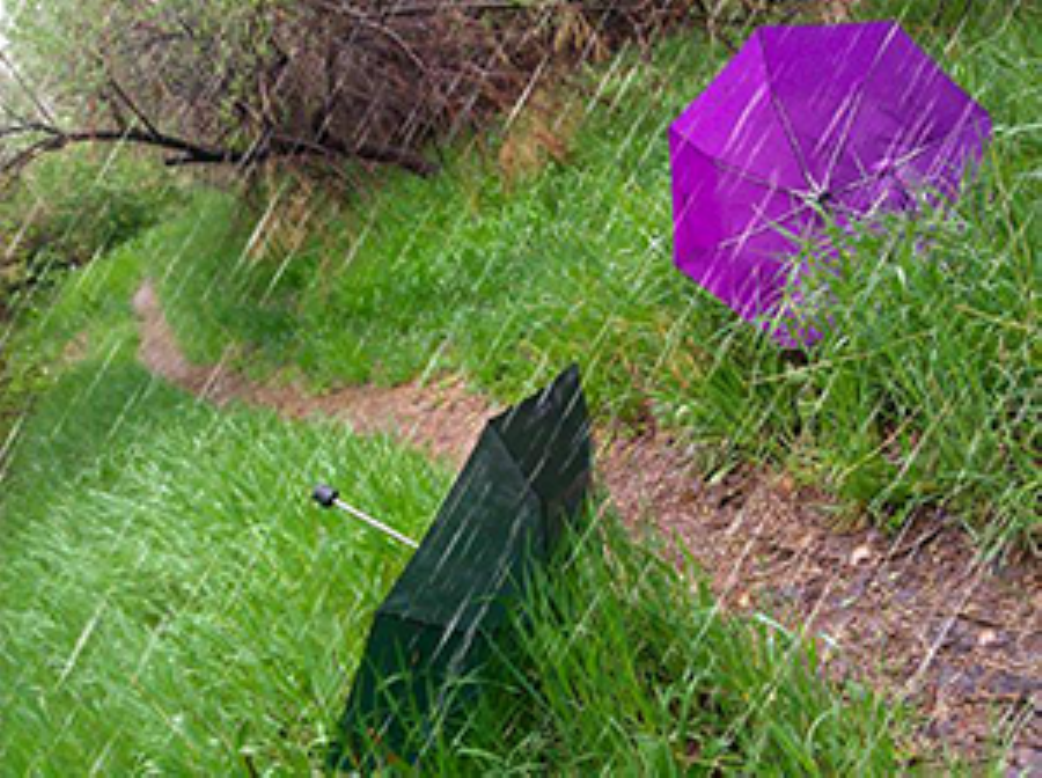}}
\end{minipage}
\hfill
\begin{minipage}{.115\linewidth}
\centering{\includegraphics[width=.995\linewidth]{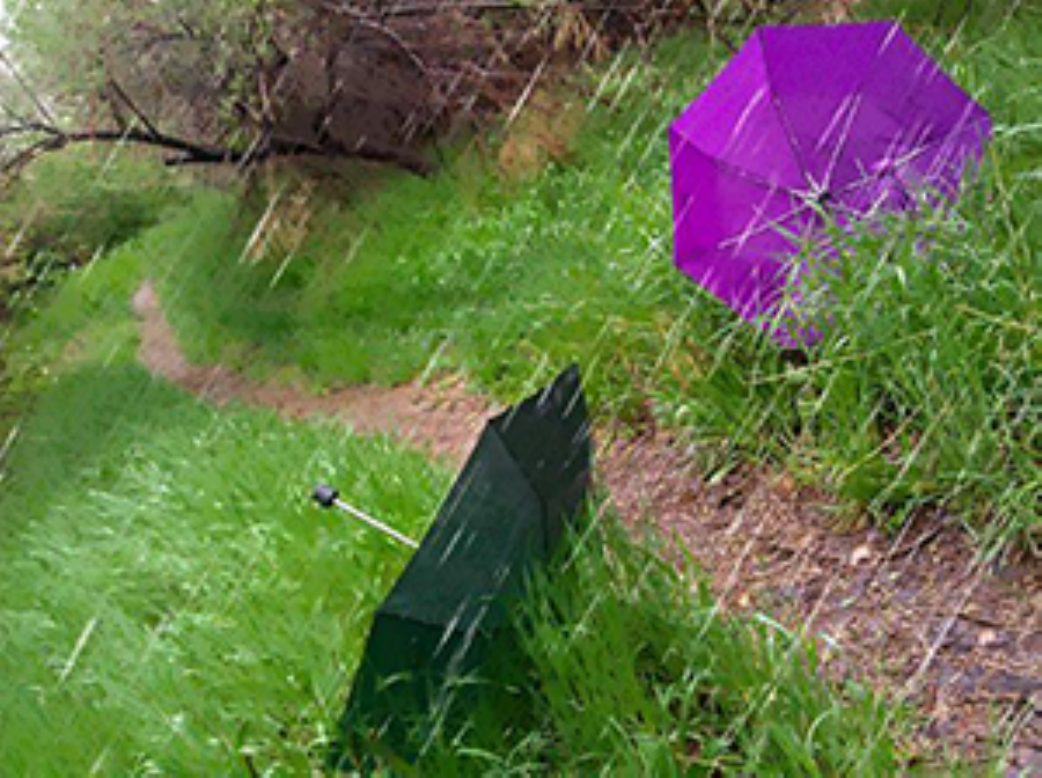}}
\end{minipage}
\hfill
\begin{minipage}{.115\linewidth}
\centering{\includegraphics[width=.995\linewidth]{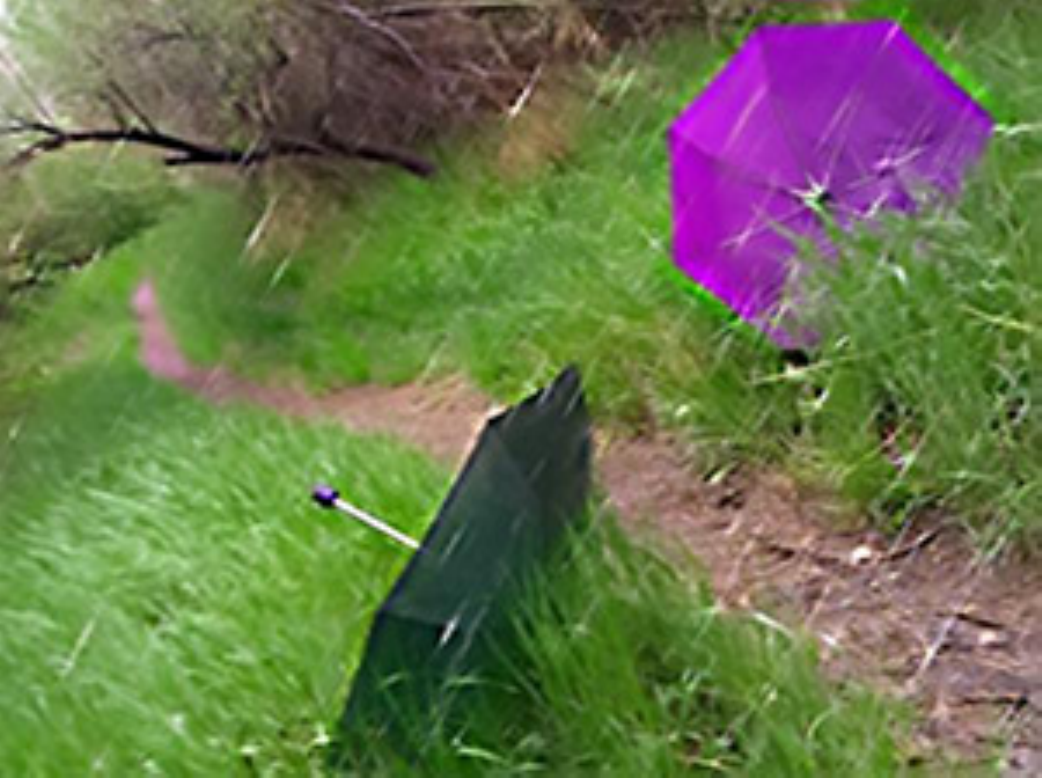}}
\end{minipage}
\hfill
\begin{minipage}{.115\linewidth}
\centering{\includegraphics[width=.995\linewidth]{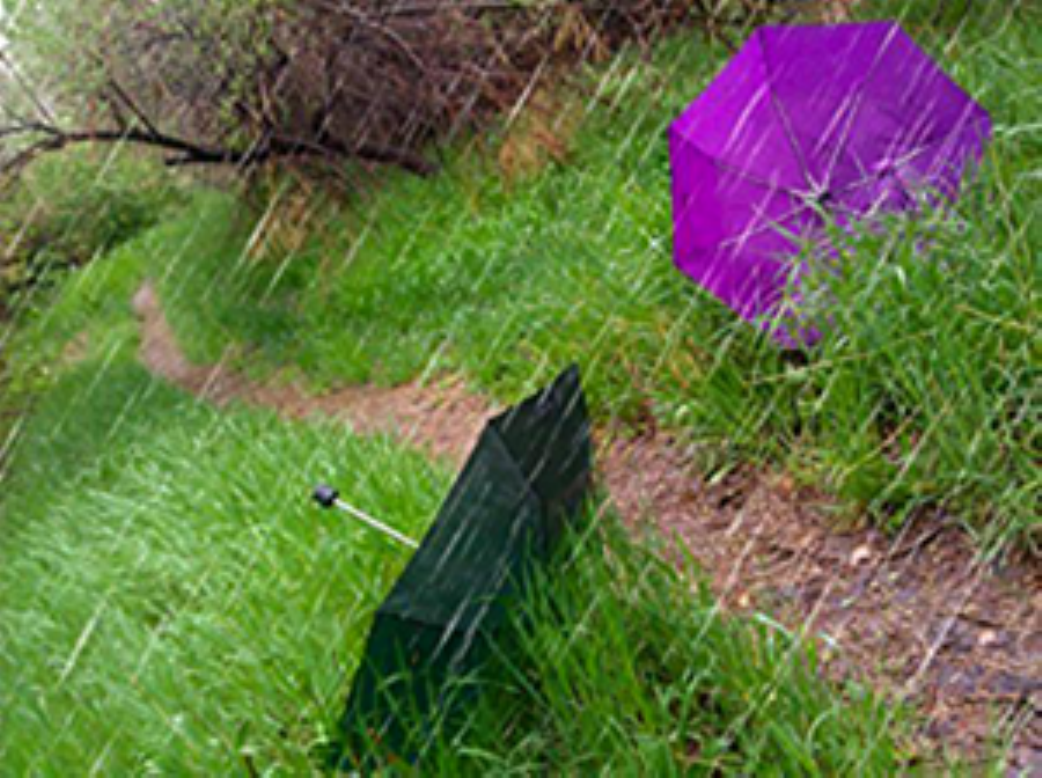}}
\end{minipage}
\hfill
\begin{minipage}{.115\linewidth}
\centering{\includegraphics[width=.995\linewidth]{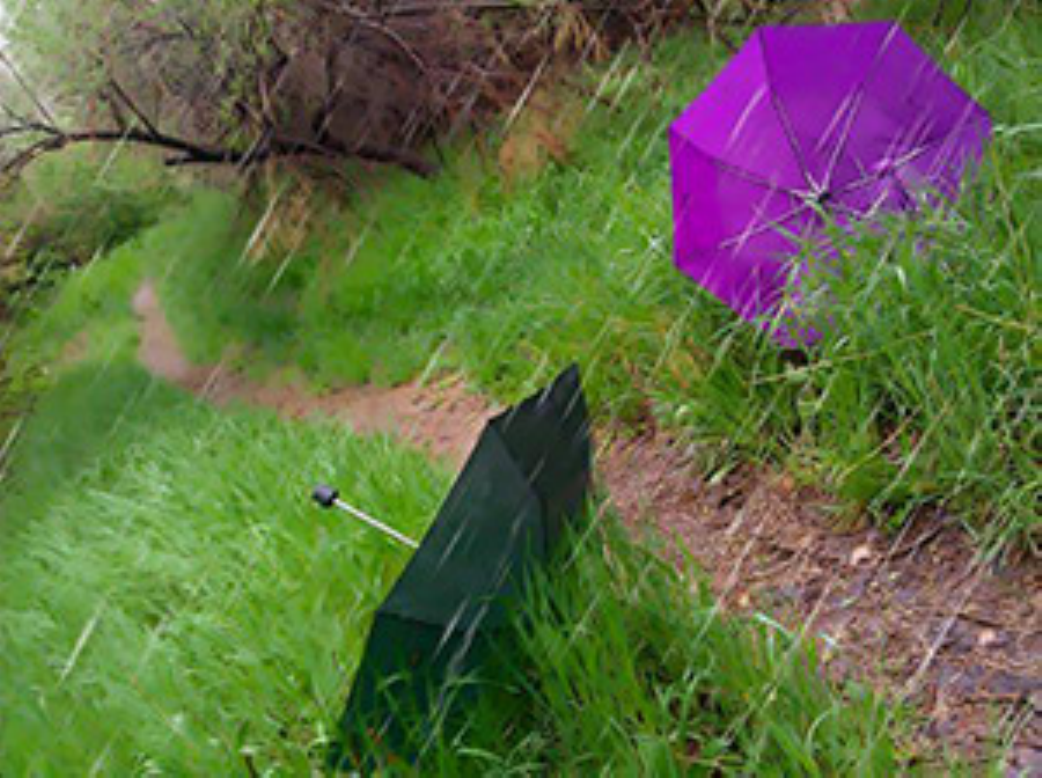}}
\end{minipage}
\hfill
\begin{minipage}{.115\linewidth}
\centering{\includegraphics[width=.995\linewidth]{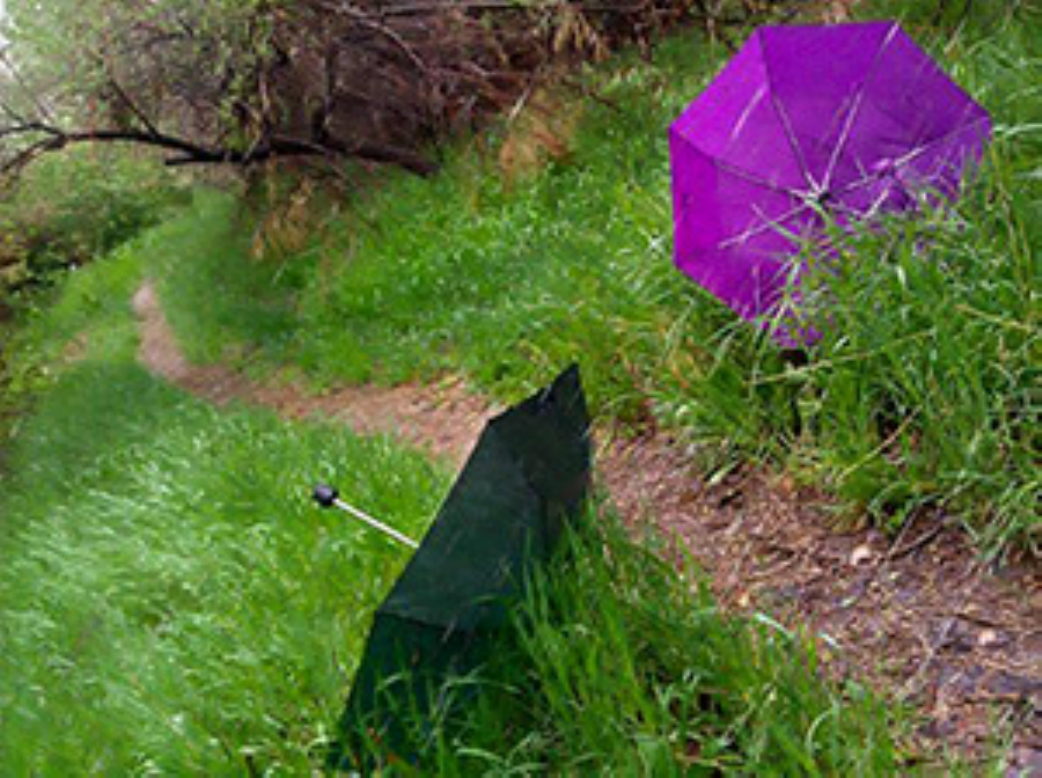}}
\end{minipage}
\hfill
\begin{minipage}{.115\linewidth}
\centering{\includegraphics[width=.995\linewidth]{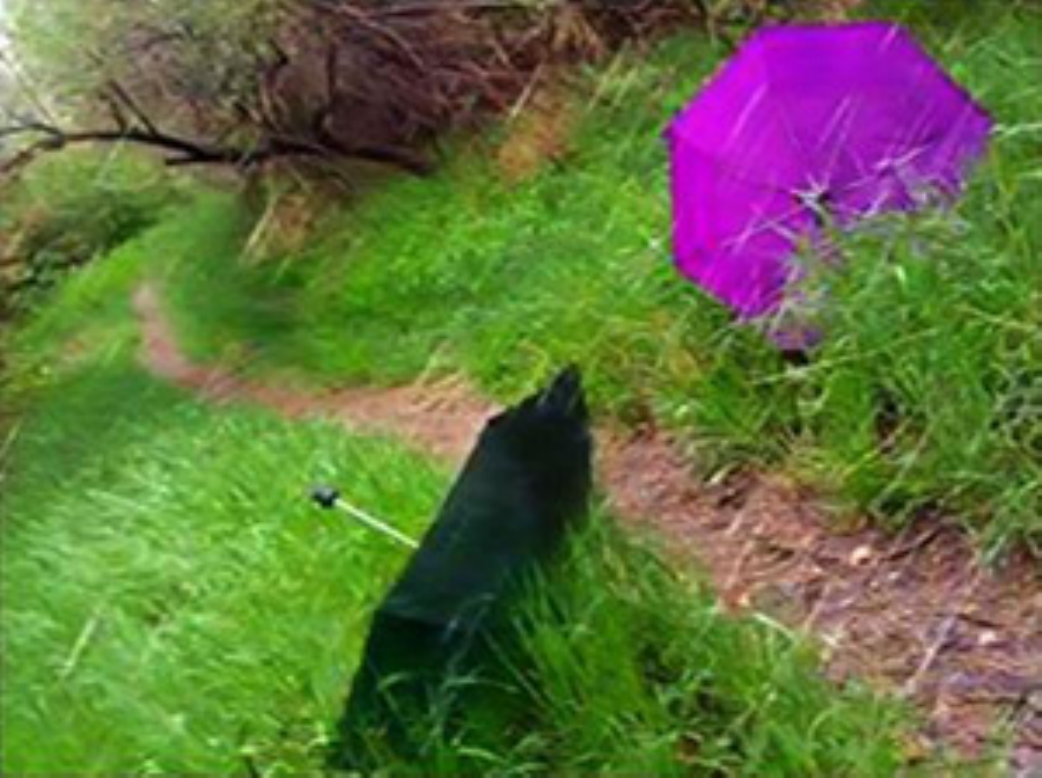}}
\end{minipage}
\hfill
\begin{minipage}{.115\linewidth}
\centering{\includegraphics[width=.995\linewidth]{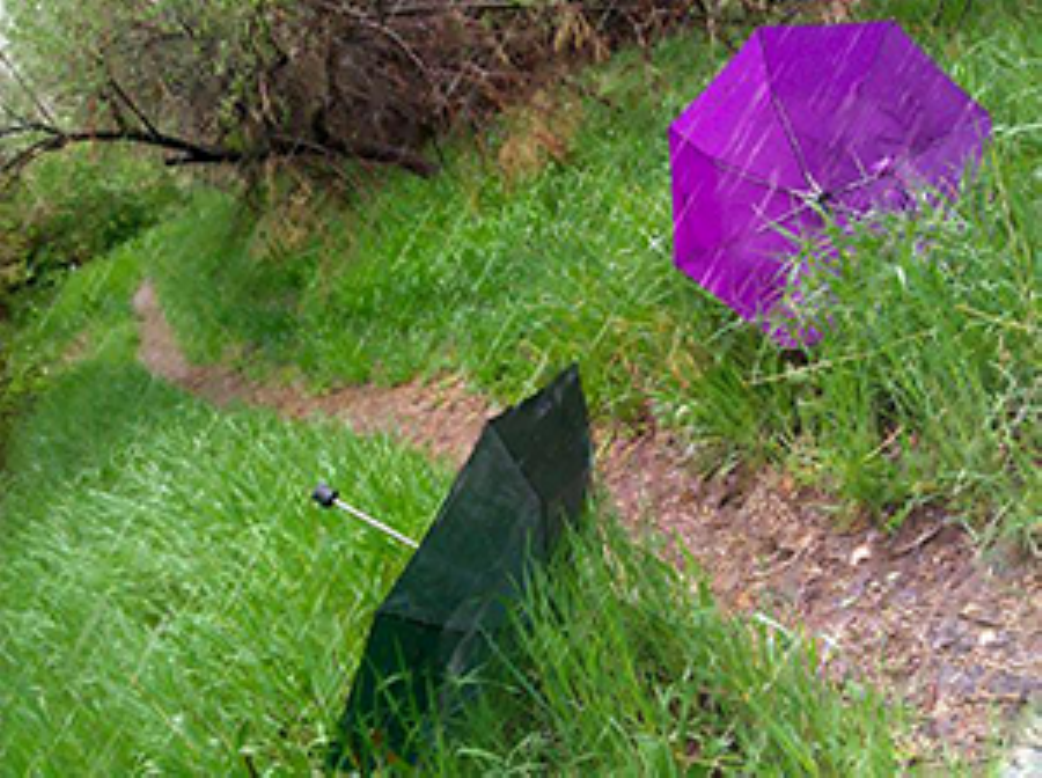}}
\end{minipage}
\vspace{0.5mm}
\vfill
\begin{minipage}{0.115\linewidth}
\centering{\includegraphics[width=.995\linewidth]{images/test25}}
\end{minipage}
\hfill
\begin{minipage}{.115\linewidth}
\centering{\includegraphics[width=.995\linewidth]{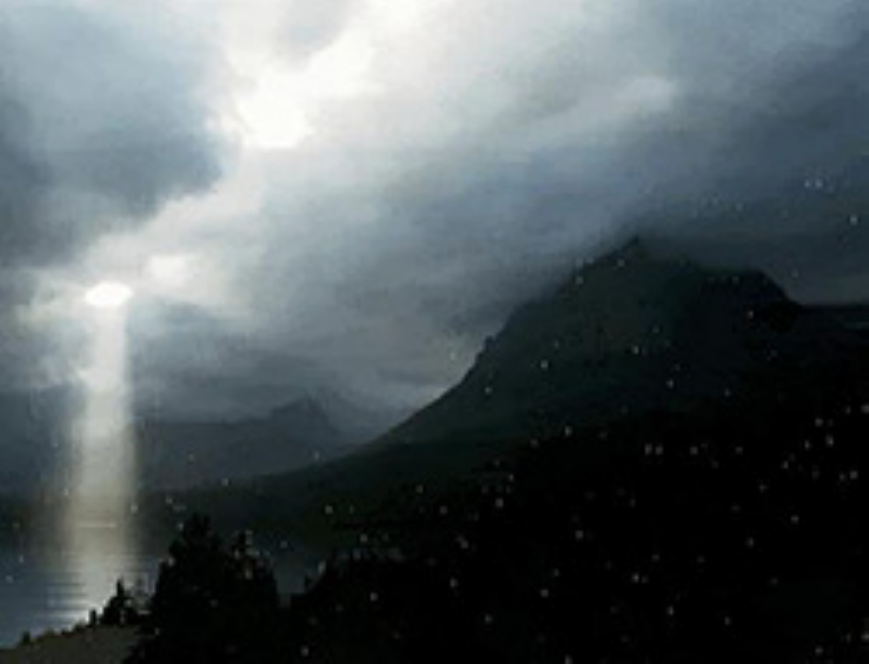}}
\end{minipage}
\hfill
\begin{minipage}{.115\linewidth}
\centering{\includegraphics[width=.995\linewidth]{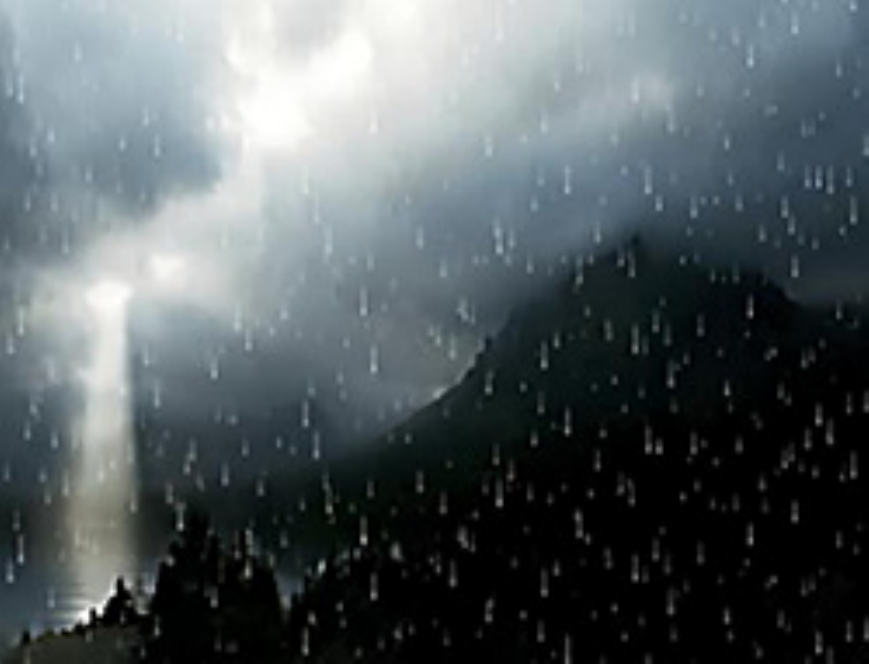}}
\end{minipage}
\hfill
\begin{minipage}{.115\linewidth}
\centering{\includegraphics[width=.995\linewidth]{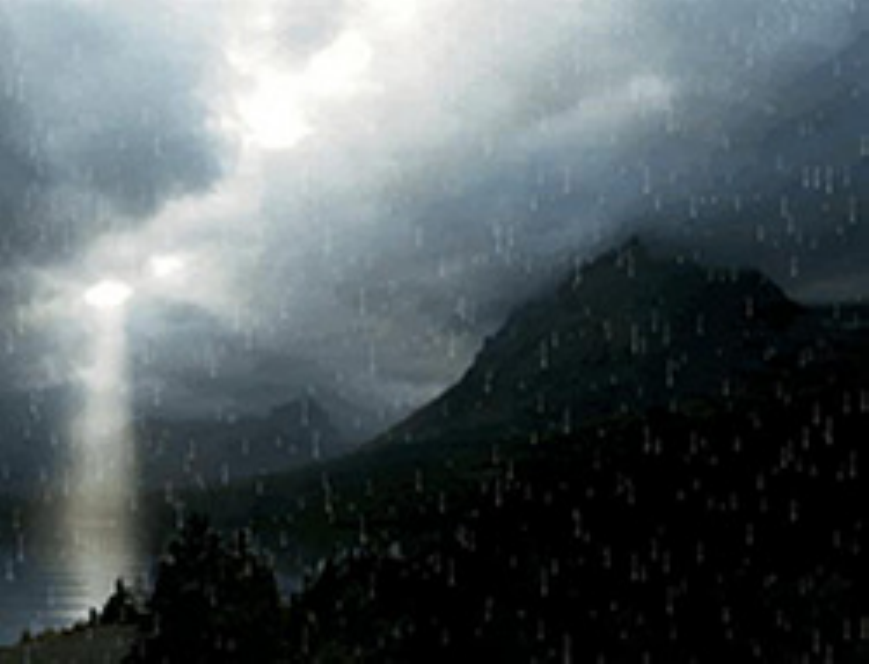}}
\end{minipage}
\hfill
\begin{minipage}{.115\linewidth}
\centering{\includegraphics[width=.995\linewidth]{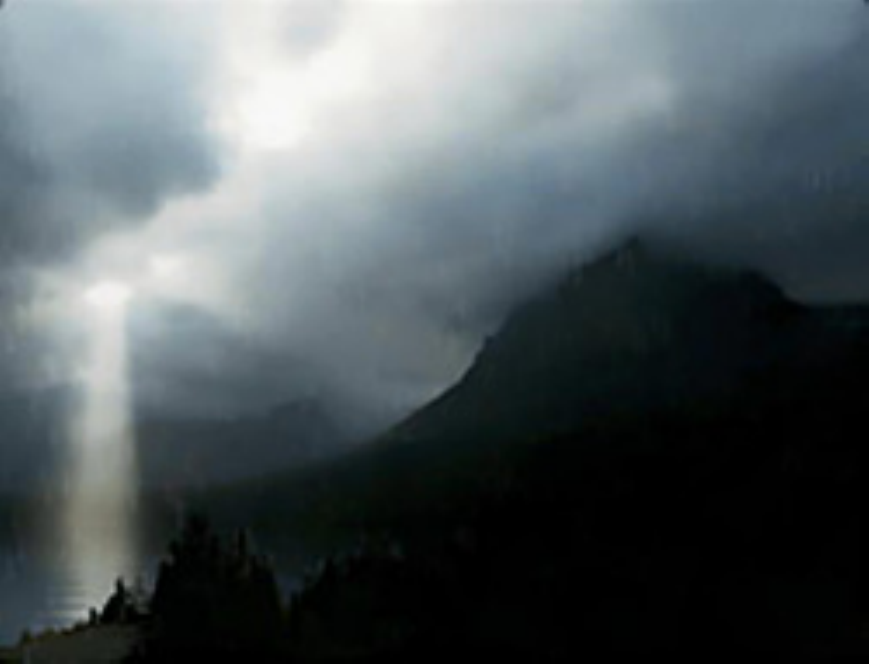}}
\end{minipage}
\hfill
\begin{minipage}{.115\linewidth}
\centering{\includegraphics[width=.995\linewidth]{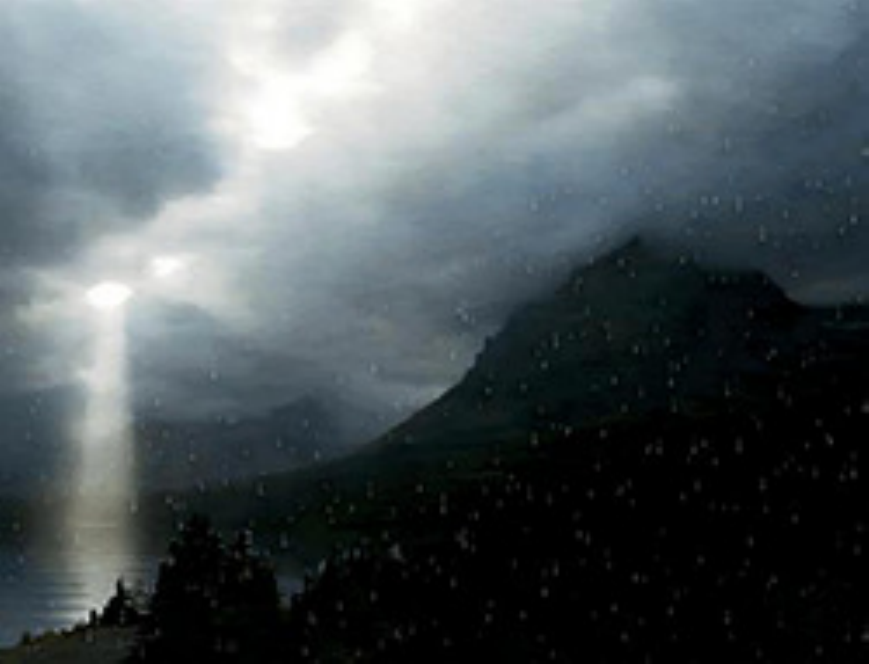}}
\end{minipage}
\hfill
\begin{minipage}{.115\linewidth}
\centering{\includegraphics[width=.995\linewidth]{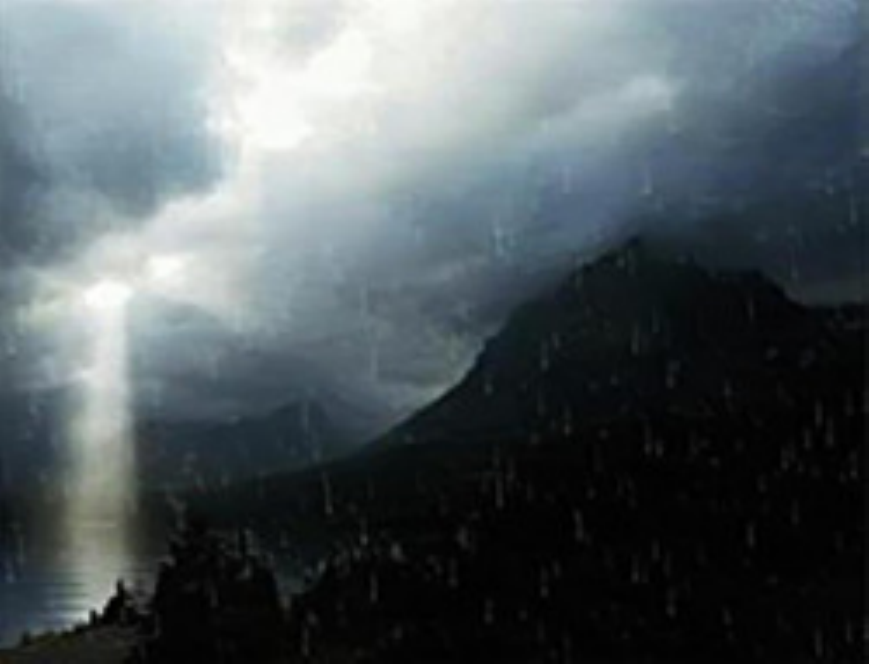}}
\end{minipage}
\hfill
\begin{minipage}{.115\linewidth}
\centering{\includegraphics[width=.995\linewidth]{images/test25_I_nd}}
\end{minipage}
\vspace{0.5mm}
\vfill
\begin{minipage}{0.115\linewidth}
\centering{\includegraphics[width=.995\linewidth]{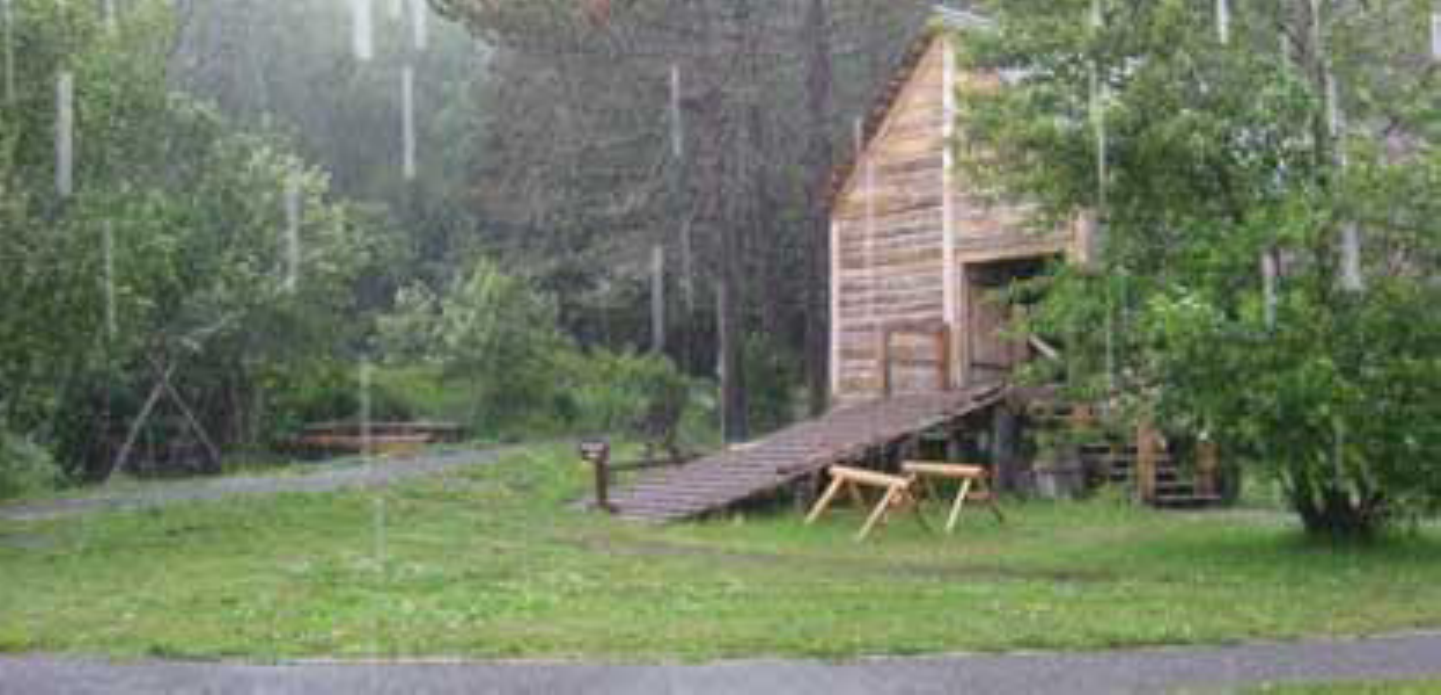}}
\end{minipage}
\hfill
\begin{minipage}{.115\linewidth}
\centering{\includegraphics[width=.995\linewidth]{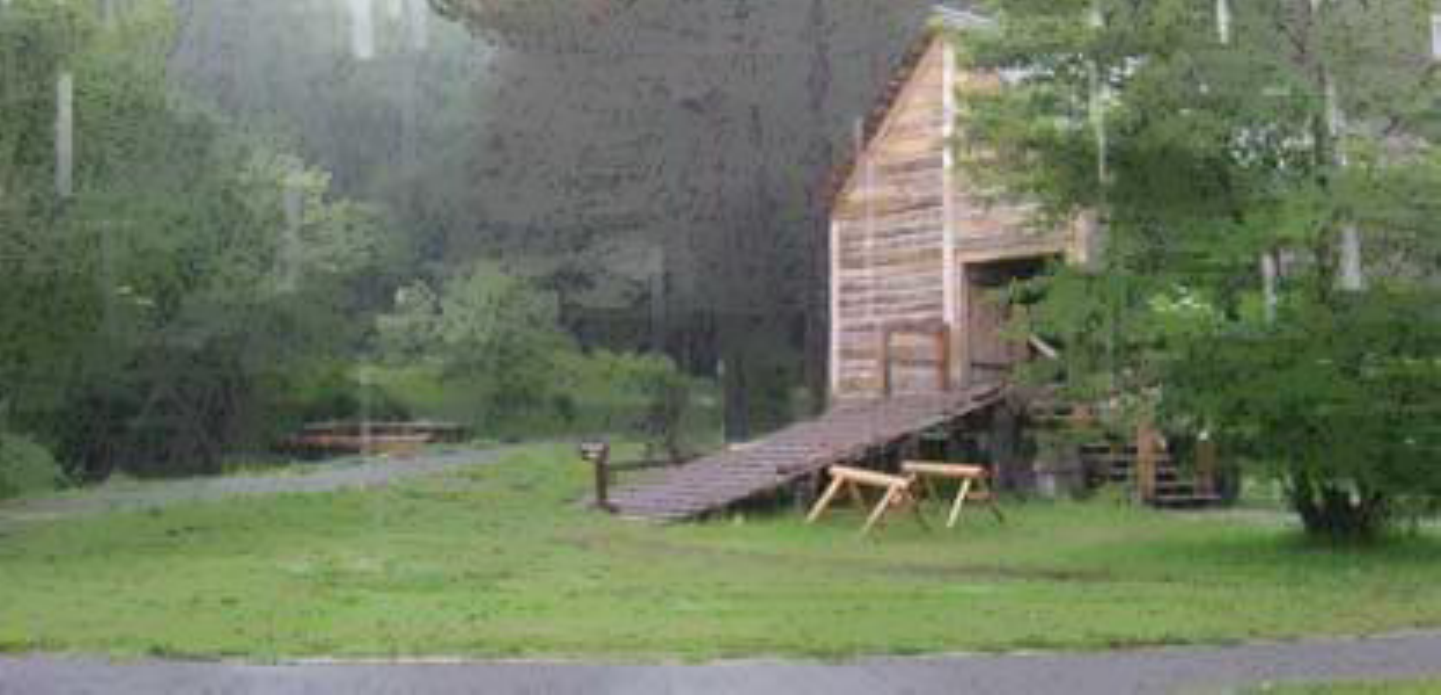}}
\end{minipage}
\hfill
\begin{minipage}{.115\linewidth}
\centering{\includegraphics[width=.995\linewidth]{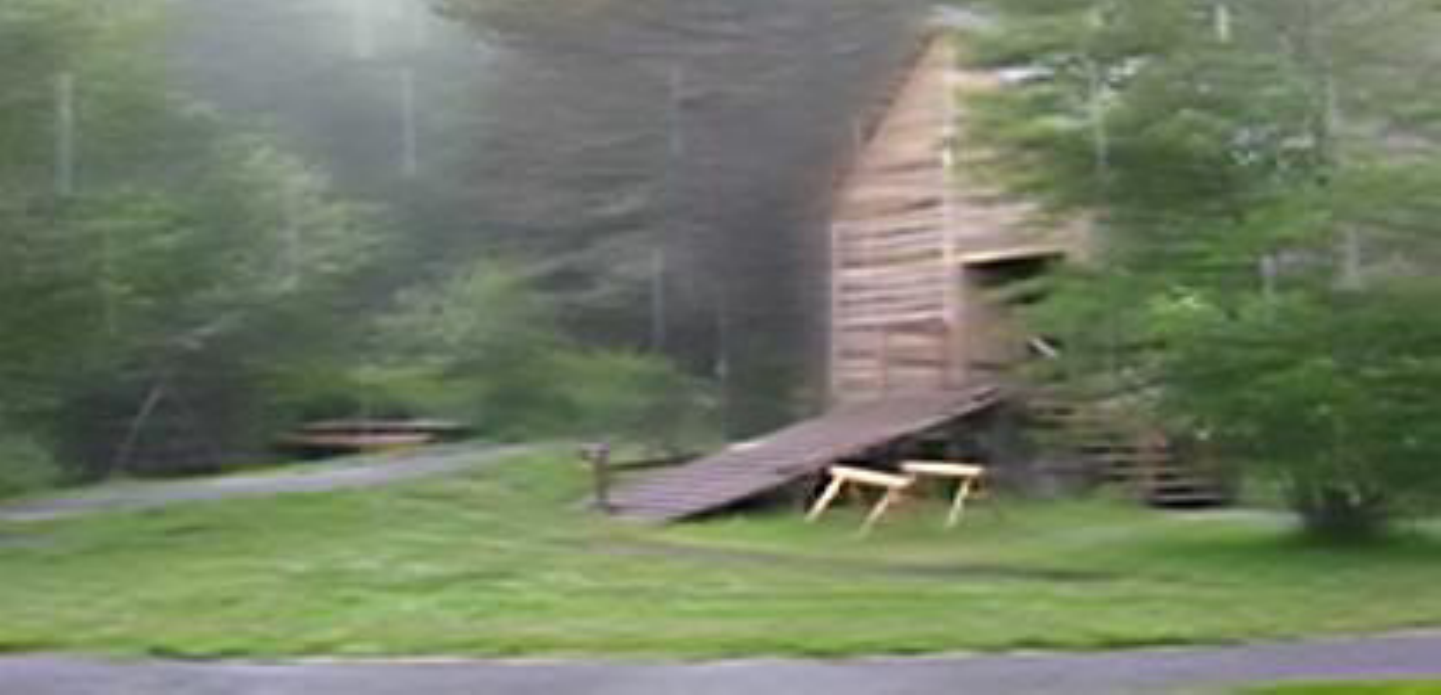}}
\end{minipage}
\hfill
\begin{minipage}{.115\linewidth}
\centering{\includegraphics[width=.995\linewidth]{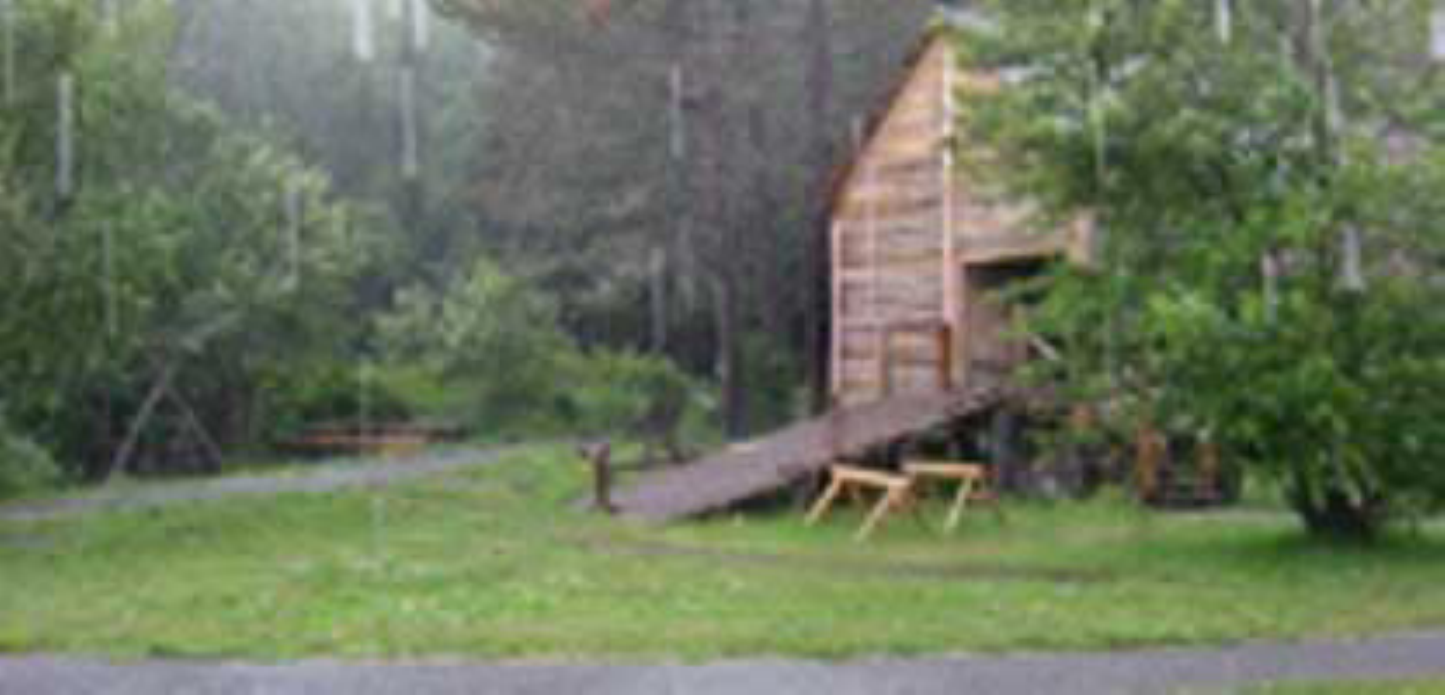}}
\end{minipage}
\hfill
\begin{minipage}{.115\linewidth}
\centering{\includegraphics[width=.995\linewidth]{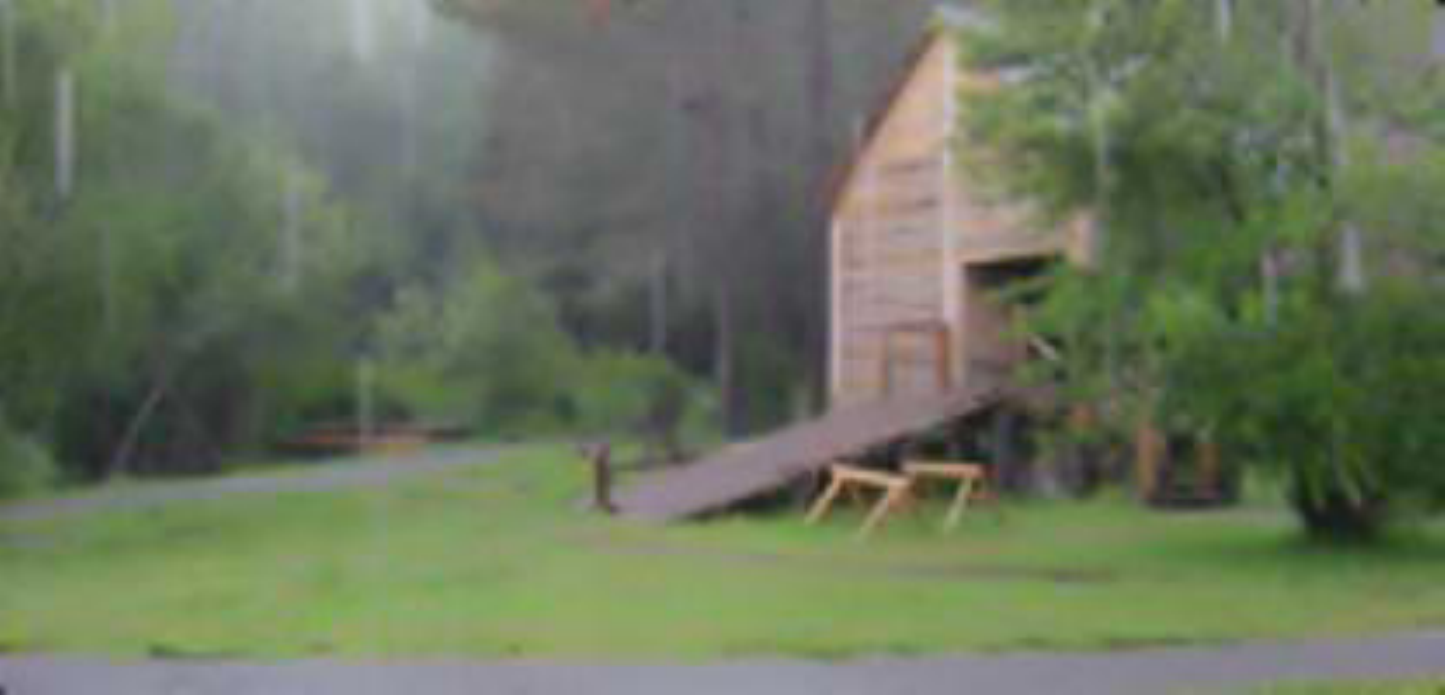}}
\end{minipage}
\hfill
\begin{minipage}{.115\linewidth}
\centering{\includegraphics[width=.995\linewidth]{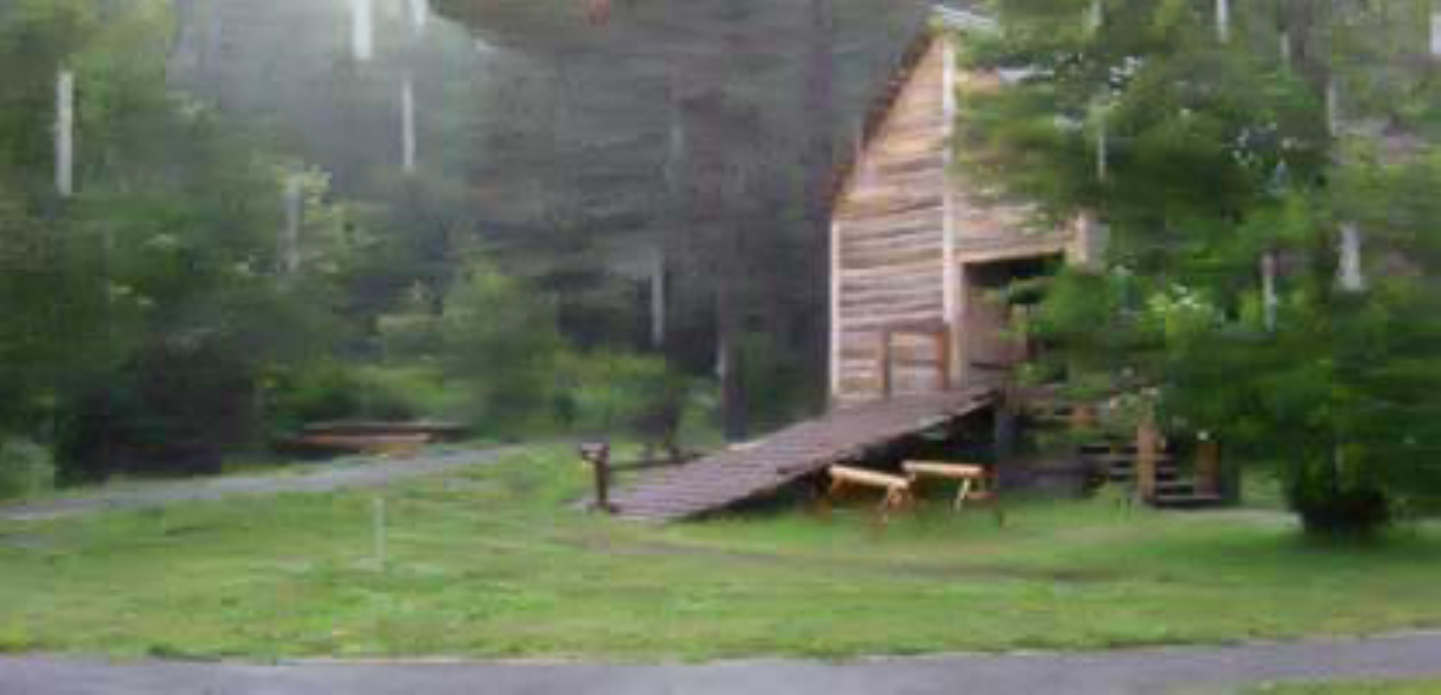}}
\end{minipage}
\hfill
\begin{minipage}{.115\linewidth}
\centering{\includegraphics[width=.995\linewidth]{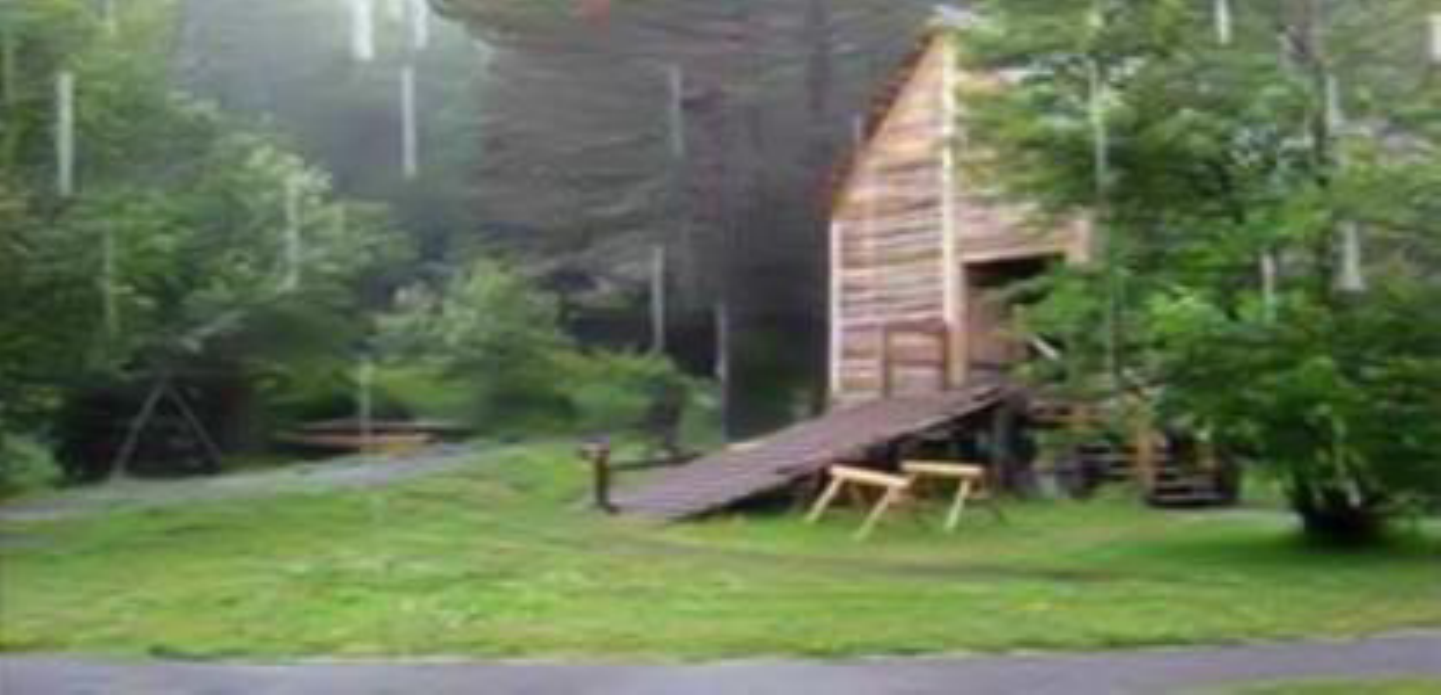}}
\end{minipage}
\hfill
\begin{minipage}{.115\linewidth}
\centering{\includegraphics[width=.995\linewidth]{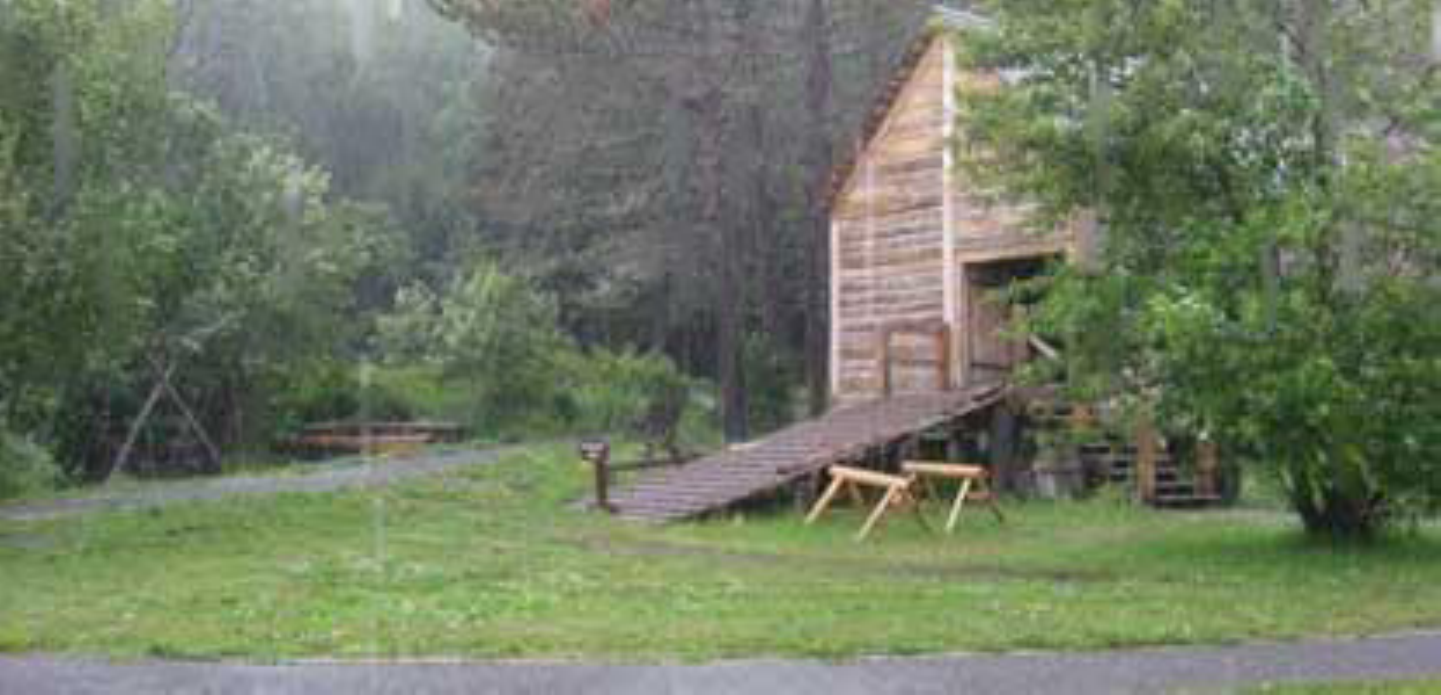}}
\end{minipage}
\vspace{0.5mm}
\vfill
\begin{minipage}{0.115\linewidth}
\centering{\includegraphics[width=.995\linewidth]{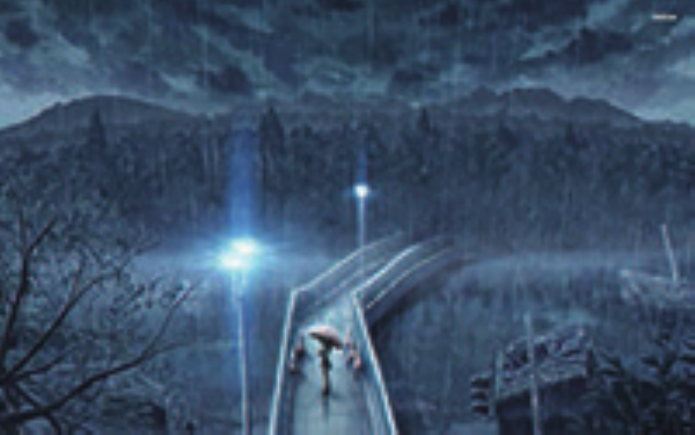}}
\end{minipage}
\hfill
\begin{minipage}{.115\linewidth}
\centering{\includegraphics[width=.995\linewidth]{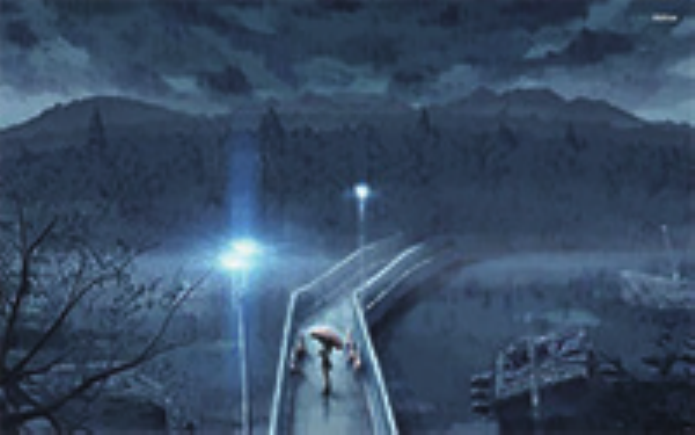}}
\end{minipage}
\hfill
\begin{minipage}{.115\linewidth}
\centering{\includegraphics[width=.995\linewidth]{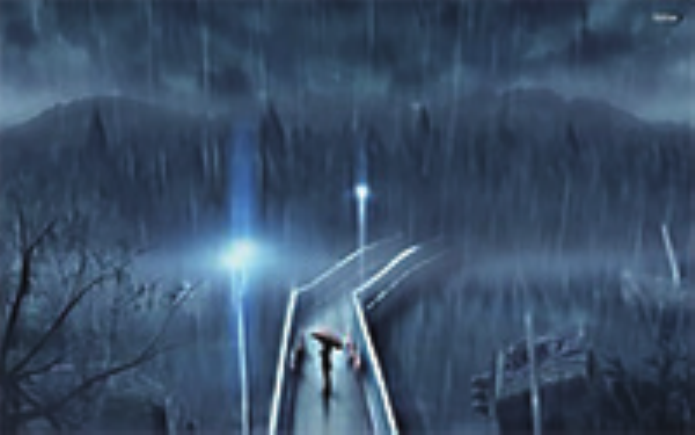}}
\end{minipage}
\hfill
\begin{minipage}{.115\linewidth}
\centering{\includegraphics[width=.995\linewidth]{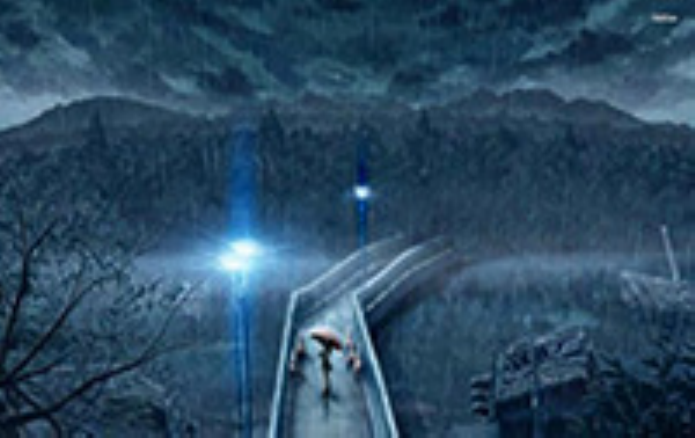}}
\end{minipage}
\hfill
\begin{minipage}{.115\linewidth}
\centering{\includegraphics[width=.995\linewidth]{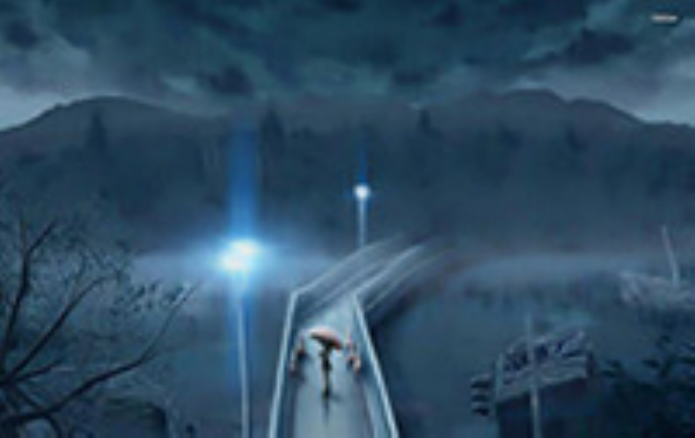}}
\end{minipage}
\hfill
\begin{minipage}{.115\linewidth}
\centering{\includegraphics[width=.995\linewidth]{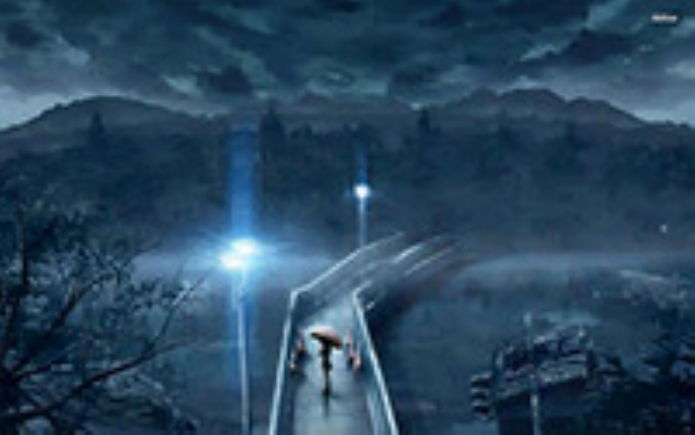}}
\end{minipage}
\hfill
\begin{minipage}{.115\linewidth}
\centering{\includegraphics[width=.995\linewidth]{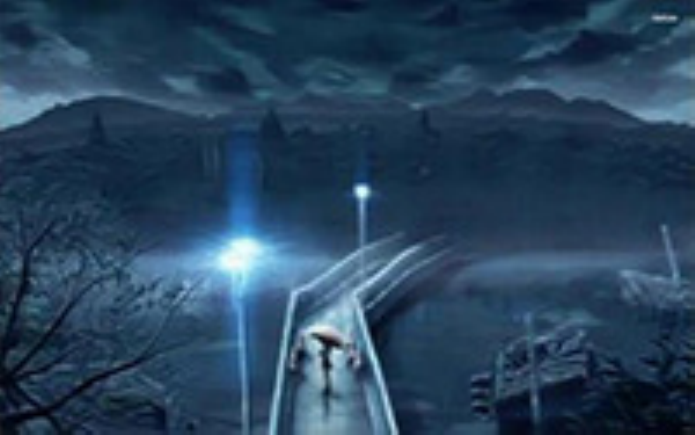}}
\end{minipage}
\hfill
\begin{minipage}{.115\linewidth}
\centering{\includegraphics[width=.995\linewidth]{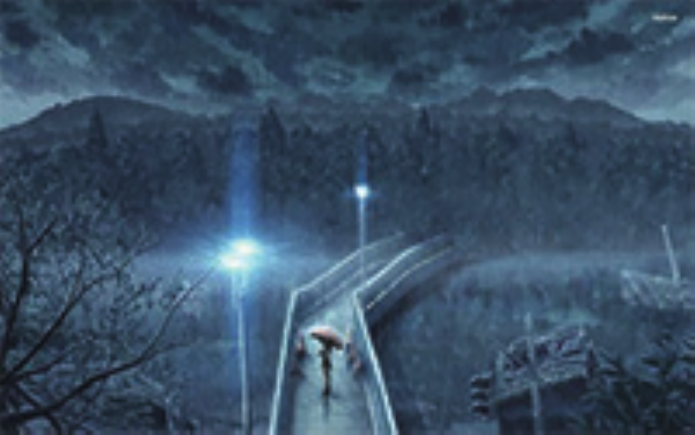}}
\end{minipage}
\vspace{0.5mm}
\vfill
\begin{minipage}{0.115\linewidth}
\centering{\includegraphics[width=.995\linewidth]{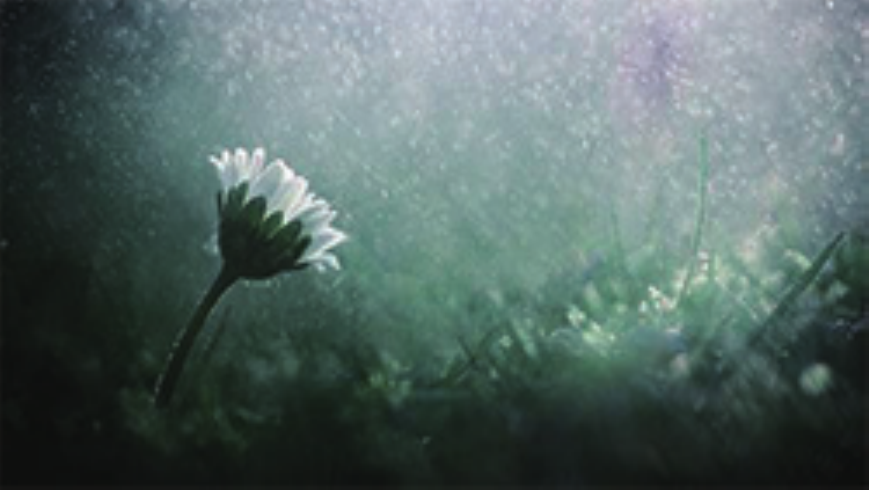}}
\end{minipage}
\hfill
\begin{minipage}{.115\linewidth}
\centering{\includegraphics[width=.995\linewidth]{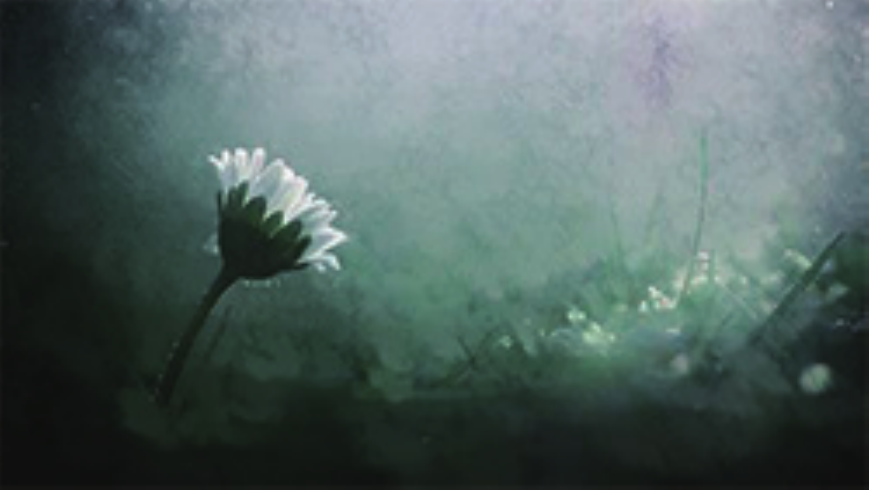}}
\end{minipage}
\hfill
\begin{minipage}{.115\linewidth}
\centering{\includegraphics[width=.995\linewidth]{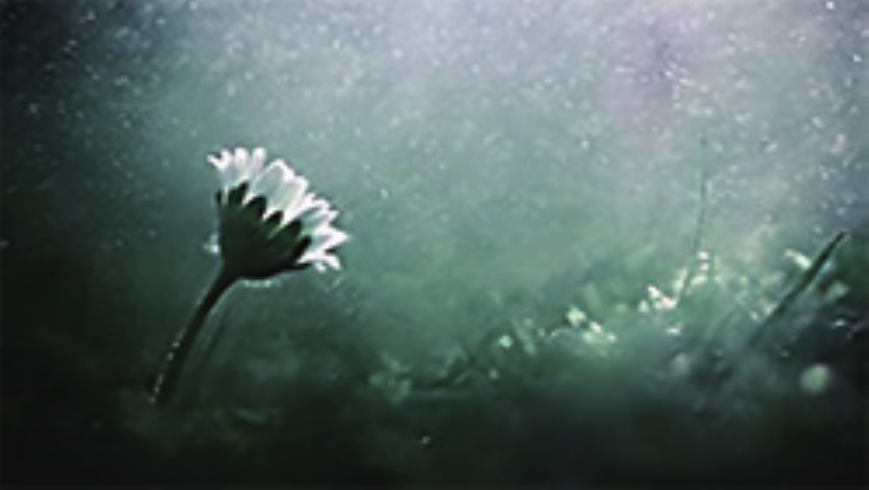}}
\end{minipage}
\hfill
\begin{minipage}{.115\linewidth}
\centering{\includegraphics[width=.995\linewidth]{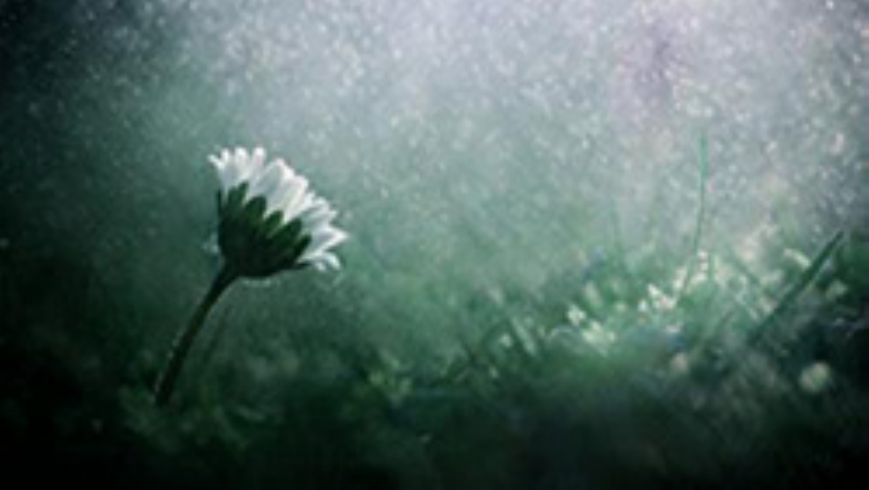}}
\end{minipage}
\hfill
\begin{minipage}{.115\linewidth}
\centering{\includegraphics[width=.995\linewidth]{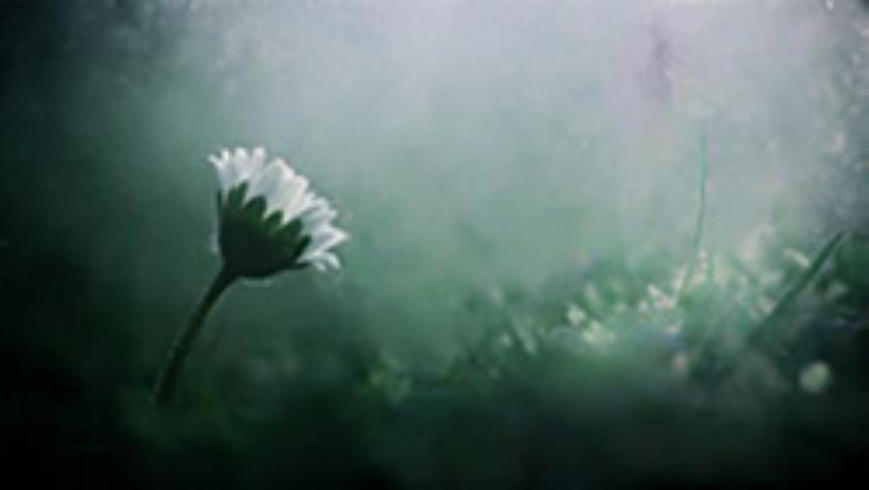}}
\end{minipage}
\hfill
\begin{minipage}{.115\linewidth}
\centering{\includegraphics[width=.995\linewidth]{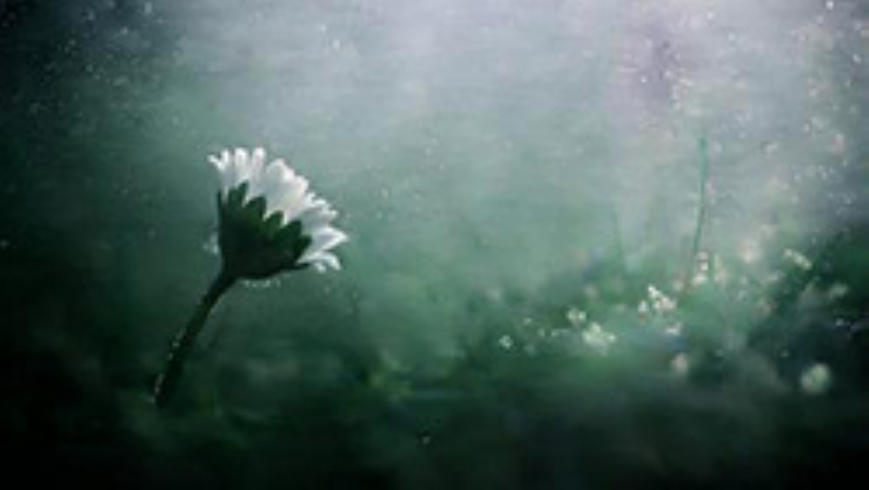}}
\end{minipage}
\hfill
\begin{minipage}{.115\linewidth}
\centering{\includegraphics[width=.995\linewidth]{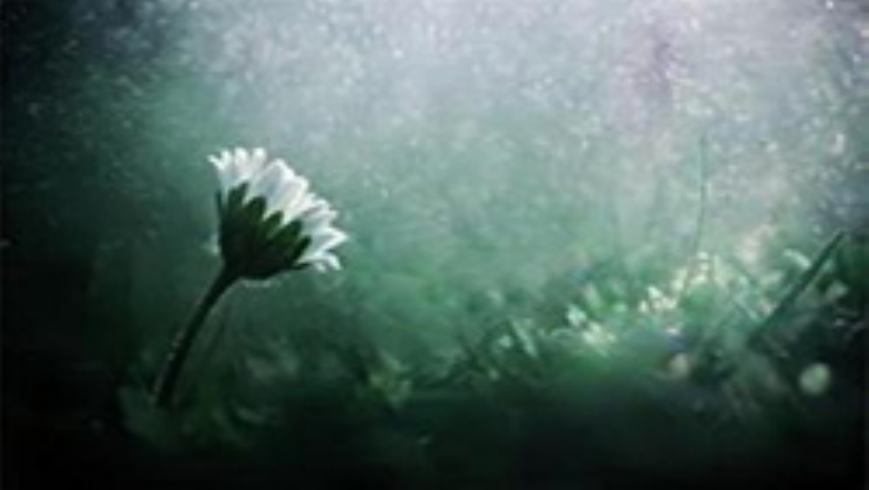}}
\end{minipage}
\hfill
\begin{minipage}{.115\linewidth}
\centering{\includegraphics[width=.995\linewidth]{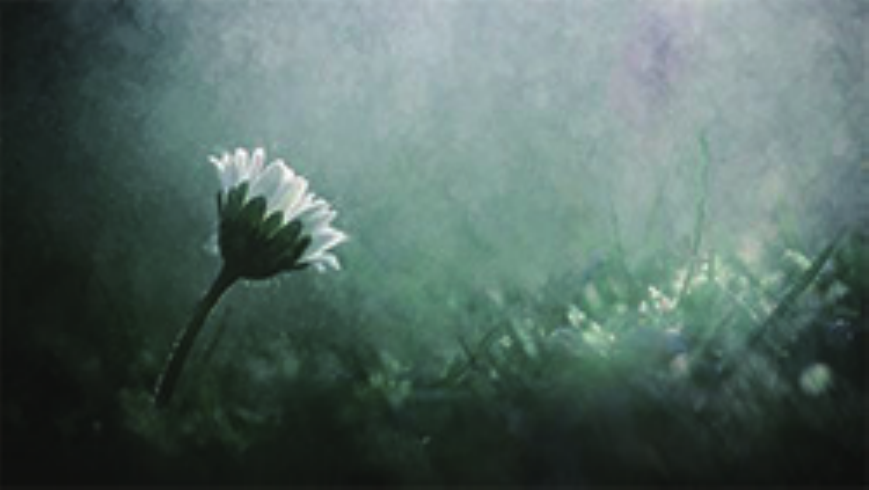}}
\end{minipage}
\vspace{0.5mm}
\vfill
\begin{minipage}{0.115\linewidth}
\centering{\includegraphics[width=.995\linewidth]{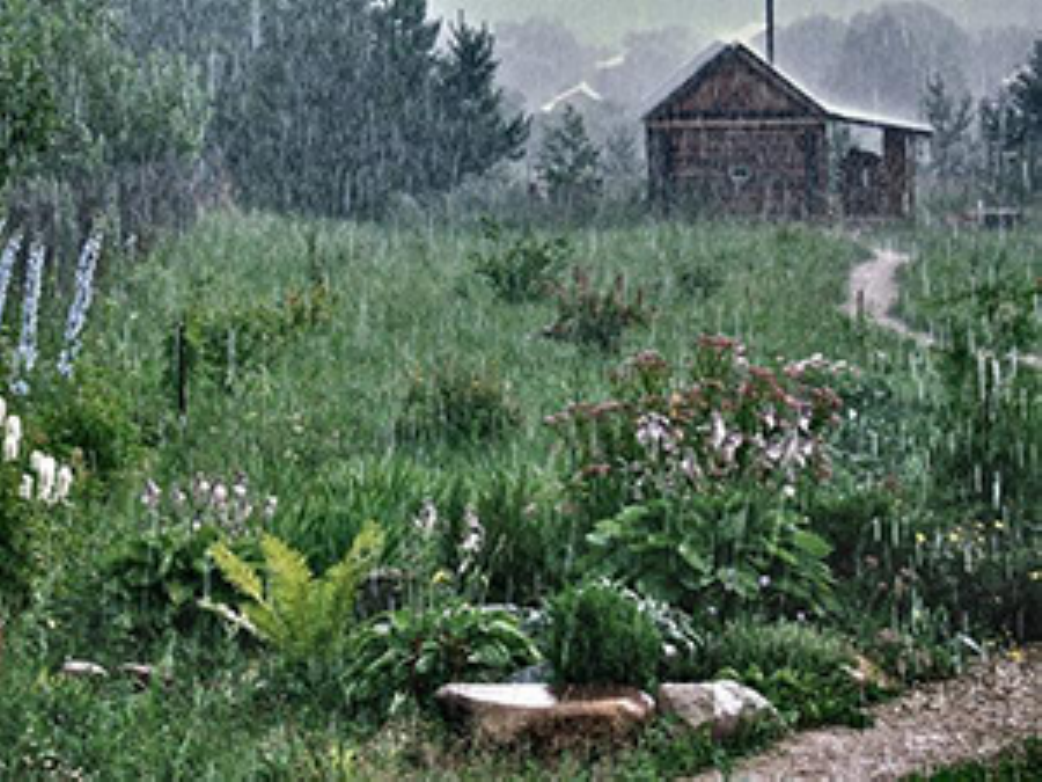}}
\centerline{(a)}
\end{minipage}
\hfill
\begin{minipage}{.115\linewidth}
\centering{\includegraphics[width=.995\linewidth]{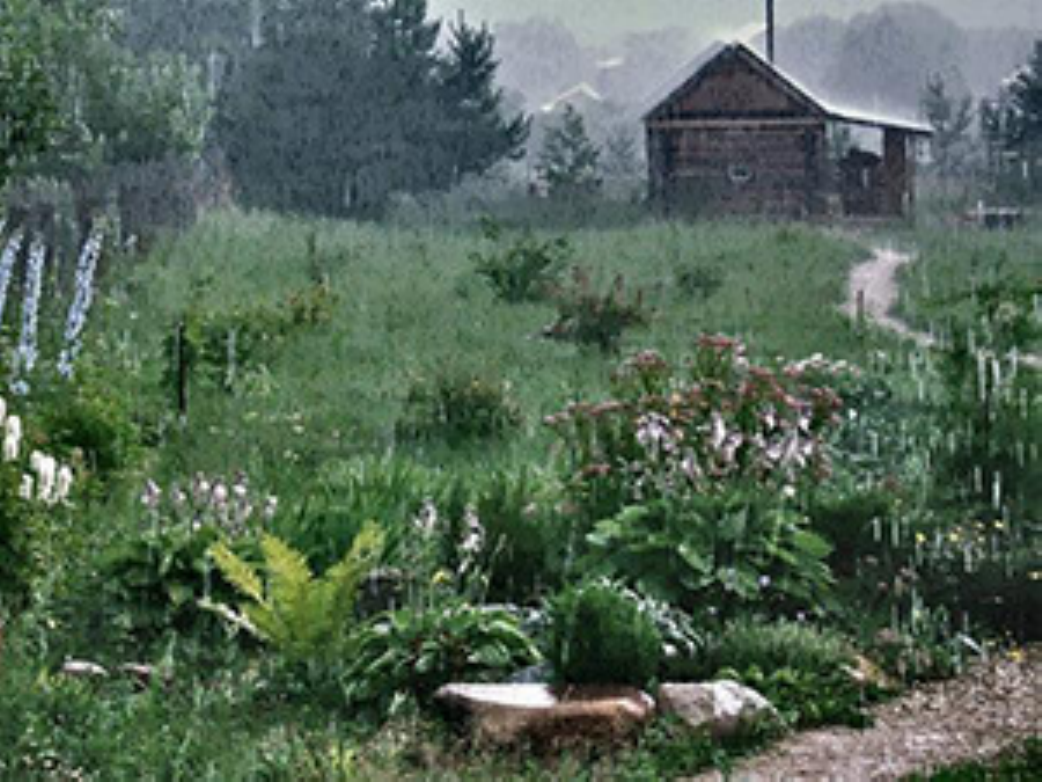}}
\centerline{(b)}
\end{minipage}
\hfill
\begin{minipage}{.115\linewidth}
\centering{\includegraphics[width=.995\linewidth]{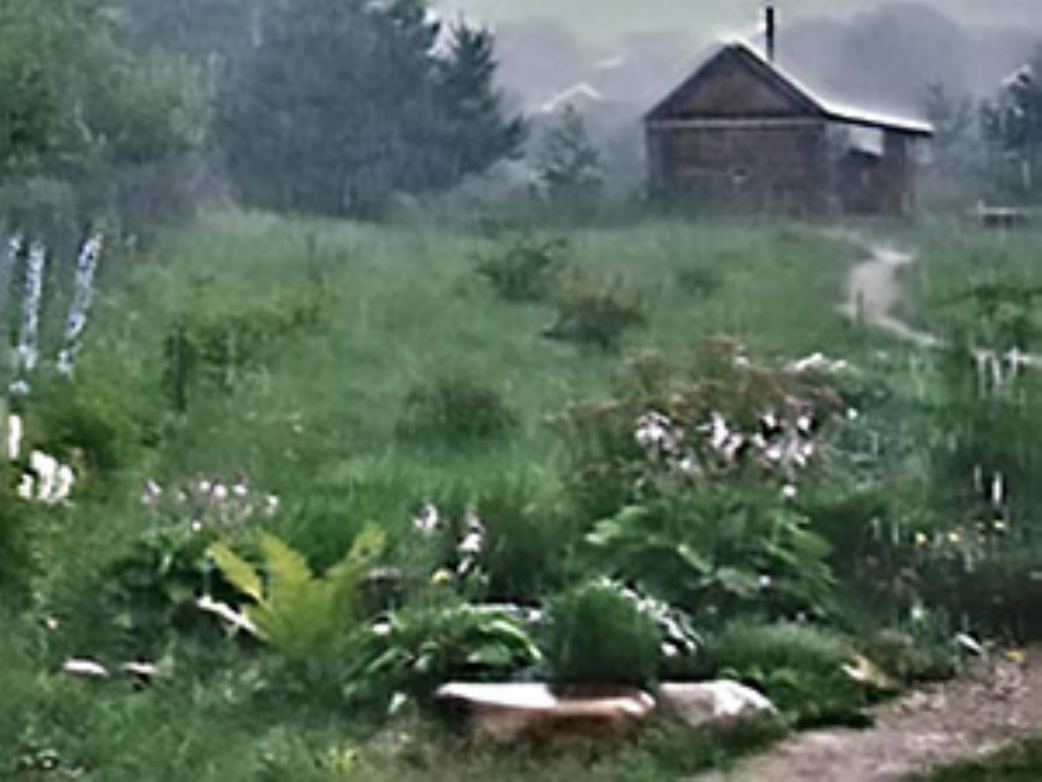}}
\centerline{(c)}
\end{minipage}
\hfill
\begin{minipage}{.115\linewidth}
\centering{\includegraphics[width=.995\linewidth]{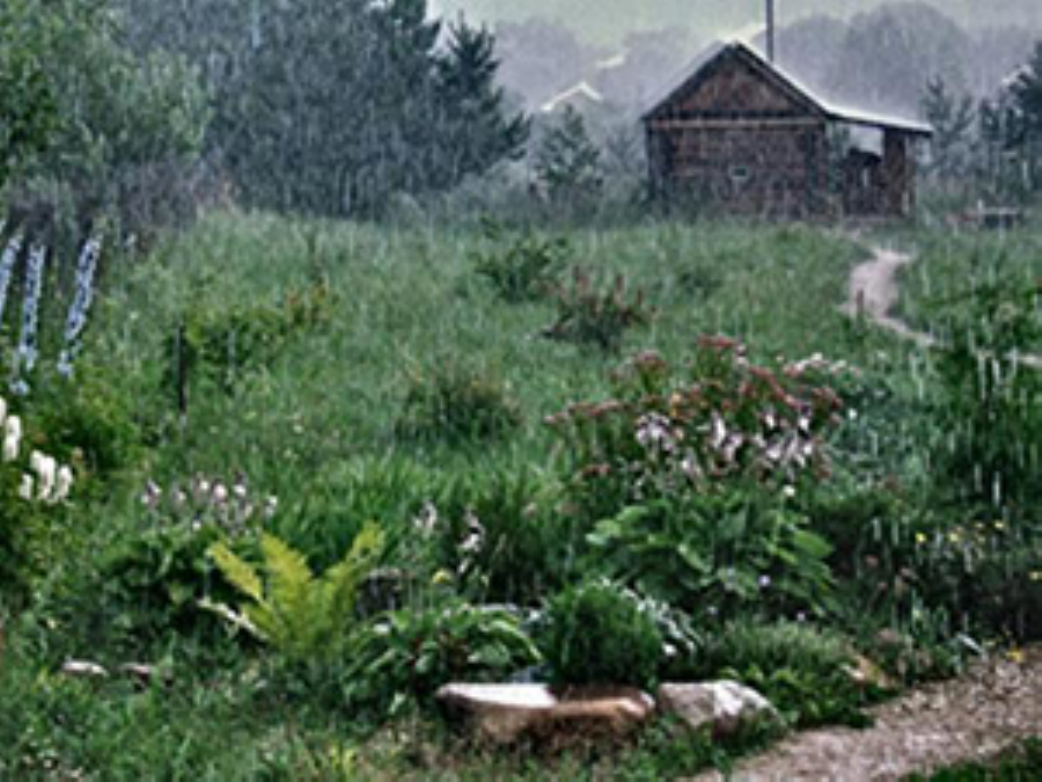}}
\centerline{(d)}
\end{minipage}
\hfill
\begin{minipage}{.115\linewidth}
\centering{\includegraphics[width=.995\linewidth]{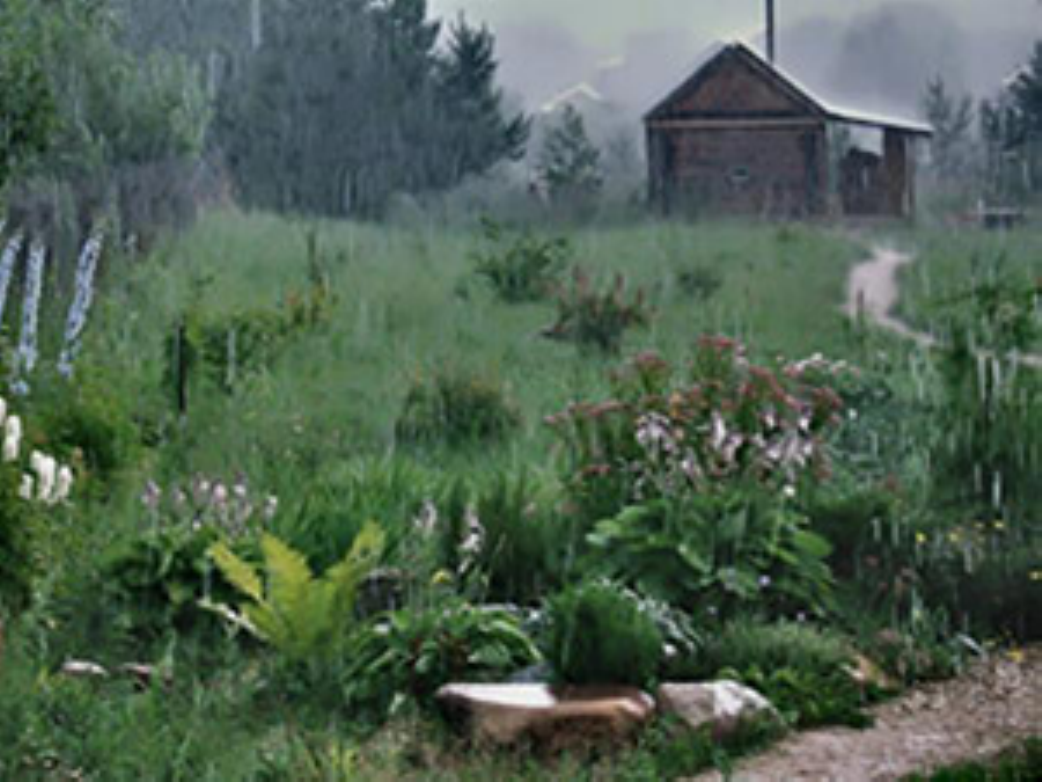}}
\centerline{(e)}
\end{minipage}
\hfill
\begin{minipage}{.115\linewidth}
\centering{\includegraphics[width=.995\linewidth]{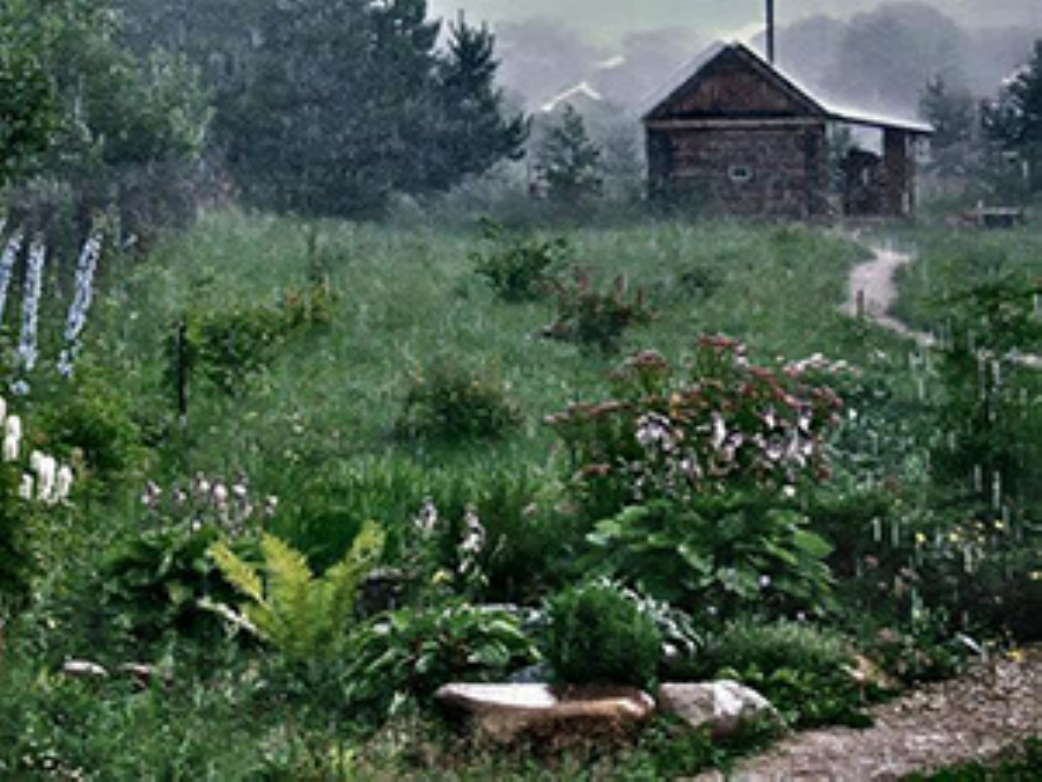}}
\centerline{(f)}
\end{minipage}
\hfill
\begin{minipage}{.115\linewidth}
\centering{\includegraphics[width=.995\linewidth]{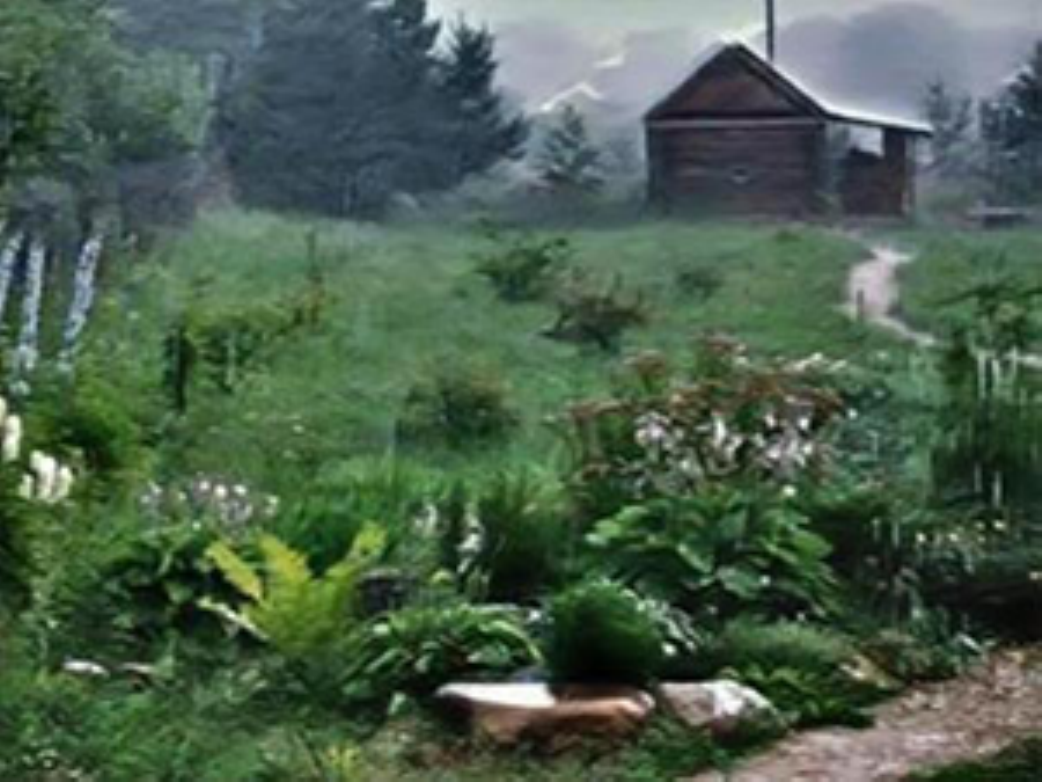}}
\centerline{(g)}
\end{minipage}
\hfill
\begin{minipage}{.115\linewidth}
\centering{\includegraphics[width=.995\linewidth]{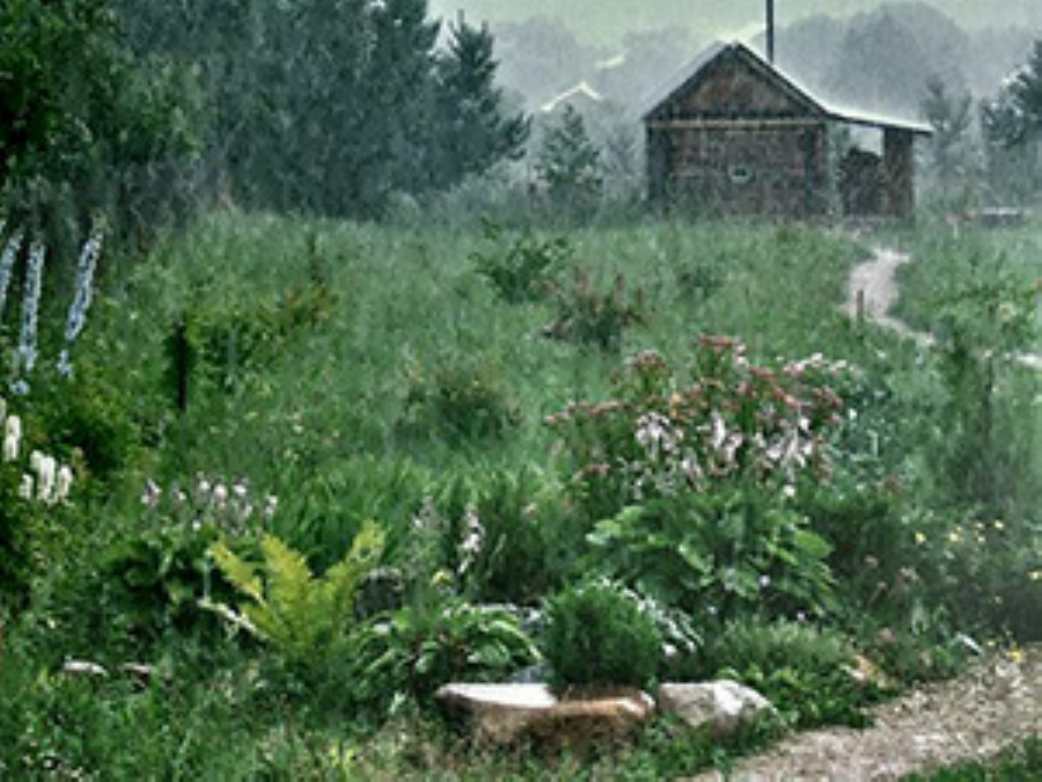}}
\centerline{(h)}
\end{minipage}
\vspace{0.5mm}
\caption{(a) Original rain images. (b) Results by Ding \emph{et al.} in \cite{Ding_2015_MTA}.
(c) Results by Chen \emph{et al.} in \cite{Chen_2014_CSVT}. (d) Results by Luo \emph{et al.} in \cite{Luo_2015_ICCV}.
(e) Results by Li \emph{et al.} in \cite{Li_2016_CVPR}. (f) Results by Fu \emph{et al.} in \cite{Fu_2017_CVPR}.
(g) Results by Zhang \emph{et al.} in \cite{Zhang_2018_CVPR}. (h) Results by our proposed method.}
\label{fig:result_compare}
\end{figure*}

We implement our rain removal algorithm using MATLAB on
an Intel (R) Xeon (R) CPU E5-2643 v2 @ 3.5 GHz 3.5 GHz (2 processors) with 64G RAM.
Some parameters used in our work are: the size of the window in
Equation (\ref{eq:detect_condition}) is $7 \times 7$;
the iteration time of $K$-means in Section \ref{sec:RainStreaksDetection} is 100;
the thresholds $T1$, $T2$ and $\mu$ are 10, 0.08, 2 respectively;
regularization parameter $\lambda$ and $\eta$ in loss function (\ref{eq:loss_function1}) is $0.25$ and $0.1$,
and the iteration time in IRLS is $3$.
The parameters here are robust in our experiments.
While for the parameter $T1$, it can be slightly changed for different images.
Because, rain's direction is downward
and its value $D$ is close to $0$ in most image,
we set the threshold $T1$ as $10$ in our paper.
Rain's direction $D$ can be approximately evaluated by user easily.
For some rain which has large falling direction (e.g. the sixth row in Fig. \ref{fig:result_compare}),
the threshold $T1$ can be changed to a larger value.

We first test the run-time consumed by the selected methods on images with size $256 \times 256$. Our method takes $28.01$ seconds. Specifically, the initial detection of rain streaks takes $5.30$ seconds; the rain streaks refining by morphology needs $2.19$ seconds; and the majority of time is spent on rain and non-rain layer separation by using quasi-sparsity prior, which is $20.02$ seconds. Upon the same image, the time consumed by other selected methods are listed in Table \ref{tab:time}. By comparison, our algorithm is the fourth fastest one in selected methods.

Because the task in this work is to remove rain streaks in single images,
we need to evaluate the effectiveness of our algorithm subjectively and objectively. For the purpose of objective evaluations, we synthesize rain images from clean images. Two such ground-truth images and synthesized rain images are shown in Fig. \ref{fig:result_render_compare}(a) and (b), respectively, and the other columns are the corresponding rain-removed results by different state-of-the-art algorithms and our method.

We also collect many real rain images and present the corresponding rain removal results as shown in Figs. \ref{fig:result_compare} and \ref{fig:result_compare1} for subjective assessments.

\begin{table*} [t]
\centering
\caption{User Study Result. The Numbers Are The Percentages of Votes Which Are Obtained by Each Method.}
\begin{tabular}{lccccccc}
\hline
Method     &  \cite{Ding_2015_MTA}  & \cite{Chen_2014_CSVT}   & \cite{Luo_2015_ICCV}  & \cite{Li_2016_CVPR} & \cite{Fu_2017_CVPR}  & \cite{Zhang_2018_CVPR} & Ours   \\
Percentage &  5.50\% & 1.25\% & 2.50\% & 3.75\% & 21.00\% & 9.50\%  & 56.50\%   \\
\hline
\end{tabular}
\label{tab:statistics}
\end{table*}

\subsection{Objective assessment}

In order to evaluate the performances of different methods more completely
and accurately, we synthesize rain images by the method in \cite{Luo_2015_ICCV} and implement different rain removal algorithms on these synthesized images. Then, we calculate the PSNR and SSIM \cite{Wang_2004_TIP} values between the rain-removed images and the ground-truth images.

Fig. \ref{fig:result_render_compare} shows two examples where each row presents a ground-truth image, the rain image (obtained by synthesis),
and the rain-removed results by different methods. Note that we show the corresponding PSNR/SSIM values at the top-left corner of each rain-removed image. The PSNR/SSIM values of more examples by selected traditional methods are shown in Table \ref{tab:psnrssim}. The comparisons of PSNR/SSIM with deep learning methods are shown in Fig. \ref{fig:PSNR_SSIM}.

According to the PSNR/SSIM values, the method by Ding \emph{et al.} \cite{Ding_2015_MTA} produces very good results compared with the other traditional methods. Because of the use of an $L_0$ threshold, objects with large structures in the image will usually be preserved well, thus leading to higher SSIM values. In the meantime, rain streaks below the $L_0$ threshold will be removed, leading to higher PSNR values.

The method by Chen \emph{et al.} \cite{Chen_2014_CSVT} can remove the rain steaks that possess lower pixel intensities but the rain streaks with higher intensities will remain (the reason will be described latter).
Furthermore, because the HOG descriptor used in this method can not identify rain streaks from tenuous details well, it would lose many details (the second image in Fig. \ref{fig:result_render_compare}).
For the above two reasons, the PSNR/SSIM values are relatively lower than the method by Ding \emph{et al.}

The work by Luo \emph{et al.} \cite{Luo_2015_ICCV} can not remove rain streaks well. It makes rain streaks more tenuous in size and weaker in intensity. We show the results of Li \emph{et al.} \cite{Li_2016_CVPR} in the sixth column of Fig. \ref{fig:result_render_compare}. This method removes rain streaks quite well. However, a lot of image details have also been removed at the same time. It can be seen from Table \ref{tab:psnrssim} that these two methods produce lower PSNR and SSIM values.

Finally, it is seen from Table \ref{tab:psnrssim} that our proposed method produces the best PSNR/SSIM results consistently for all 11 test images compared with the selected traditional methods. For some test images (5 out of 11), the PSNR value of our method is about 1 dB higher than the second best method (i.e., Ding's method).

We can see from Fig. \ref{fig:PSNR_SSIM} that our method produces comparable PSNR/SSIM values compared with the state-of-the-art
deep learning methods. Only for the two rendering rain images shown in Fig. \ref{fig:result_render_compare}, the work by Fu \emph{et al.} \cite{Fu_2017_CVPR} removes nearly all rain streaks and keeps image details relatively well. But its PSNR/SSIM values are slightly low compared with ours. The reason is that our method only remove rain streaks on
the detected rain pixels and the non-rain pixels nearly unchanged.
Though a few of light rain streaks remain in our results in these selected two images, our method keeps good image details in majority part of images. The work by Zhang \emph{et al.} removes rain streaks well, but the image details loss seriously.
That is why this method has high PSNR values, while the SSIM values are low.
Besides, these two methods also can not remove all rain streaks in some rendering rain images, but our method can achieve
good results, especially, for practical images which will be shown later. As we know, deep learning methods
are very good at dealing with rendering rain images, because they are trained from them. Real-world images are real challenging
for them.

\subsection{User study}

To conduct a visual (subjective) evaluation on the performances of selected methods, we invited 20 viewers (14 males and 6 females, they all are undergraduate, master or Ph.D students in computer vision field) to evaluate the visual quality of different methods in terms of the following three aspects:
\begin{itemize}
\item less rain residual,
\item the maintenance of the image details, and
\item overall perception.
\end{itemize}
In the evaluation, $20$ groups of results are selected and every group involves the results by Ding \emph{et al.}, Chen \emph{et al.}, Luo \emph{et al.}, Li \emph{et al.}, Fu \emph{et al.}, Zhang \emph{et al.} and our method. To ensure fairness, the results in each group are arranged randomly. For each group, the viewers are asked to select only one result which they like most by considering the three criterions together.

The evaluation result is shown in Table \ref{tab:statistics}. It is clear that our rain removal results are favored by a majority of viewers (56.50\%).

\begin{figure*}[!htb]
\centering
\begin{minipage}{0.115\linewidth}
\centering{\includegraphics[width=.995\linewidth]{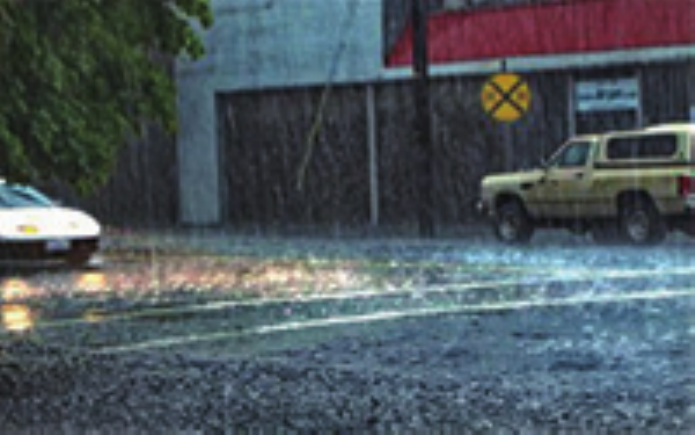}}
\end{minipage}
\hfill
\begin{minipage}{.115\linewidth}
\centering{\includegraphics[width=.995\linewidth]{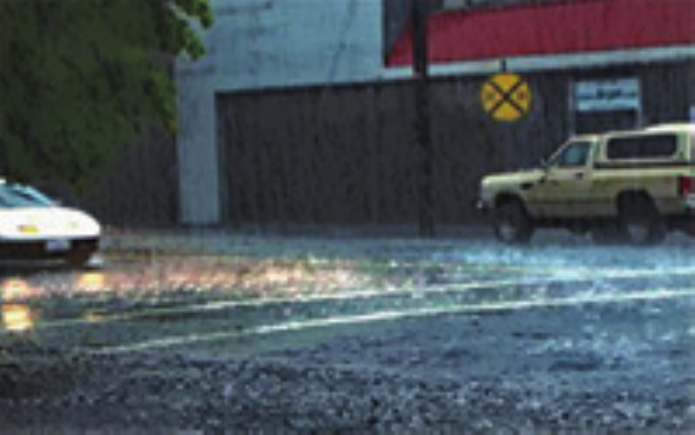}}
\end{minipage}
\hfill
\begin{minipage}{.115\linewidth}
\centering{\includegraphics[width=.995\linewidth]{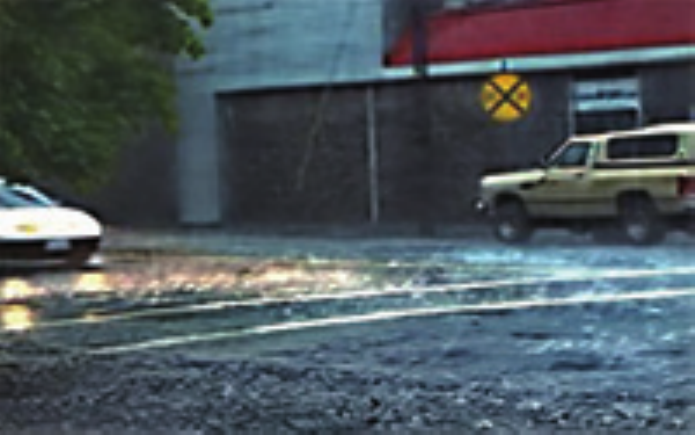}}
\end{minipage}
\hfill
\begin{minipage}{.115\linewidth}
\centering{\includegraphics[width=.995\linewidth]{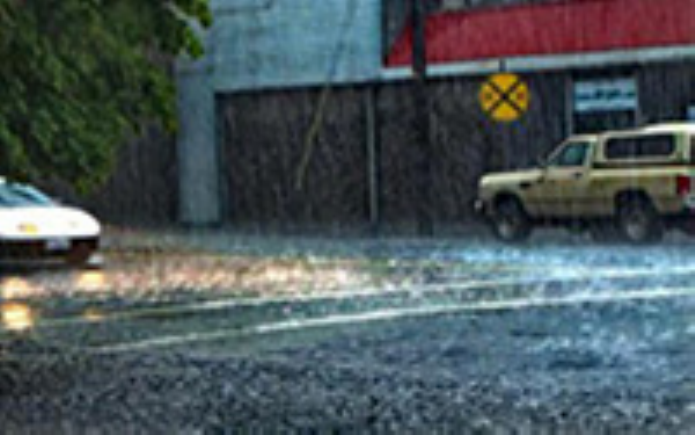}}
\end{minipage}
\hfill
\begin{minipage}{.115\linewidth}
\centering{\includegraphics[width=.995\linewidth]{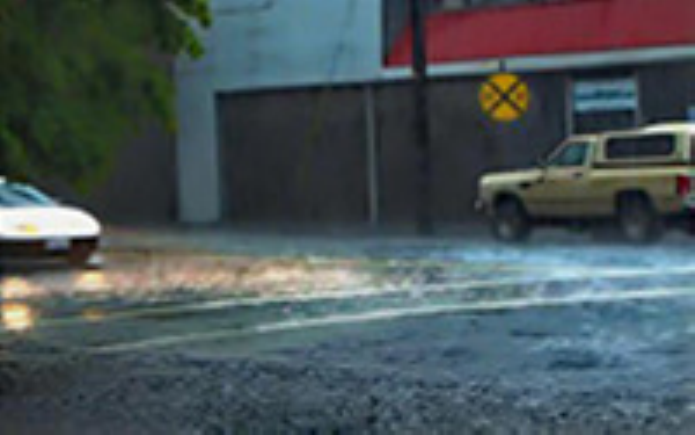}}
\end{minipage}
\hfill
\begin{minipage}{.115\linewidth}
\centering{\includegraphics[width=.995\linewidth]{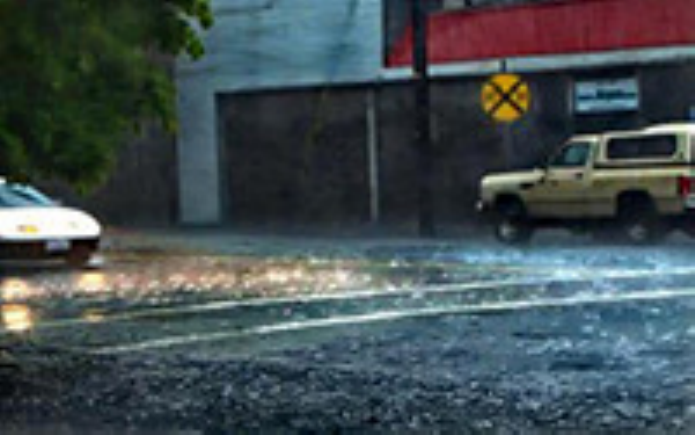}}
\end{minipage}
\hfill
\begin{minipage}{.115\linewidth}
\centering{\includegraphics[width=.995\linewidth]{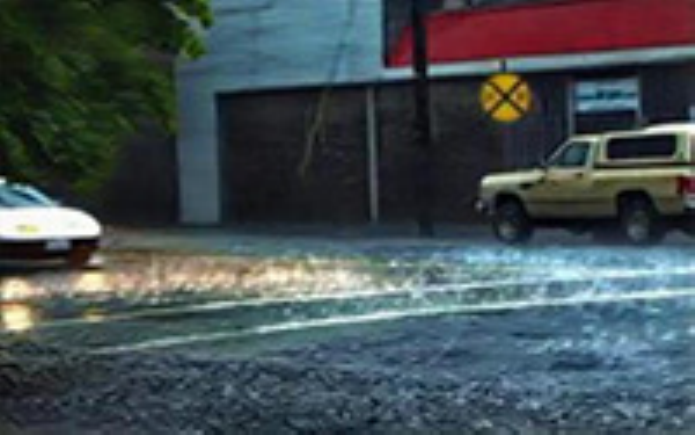}}
\end{minipage}
\hfill
\begin{minipage}{.115\linewidth}
\centering{\includegraphics[width=.995\linewidth]{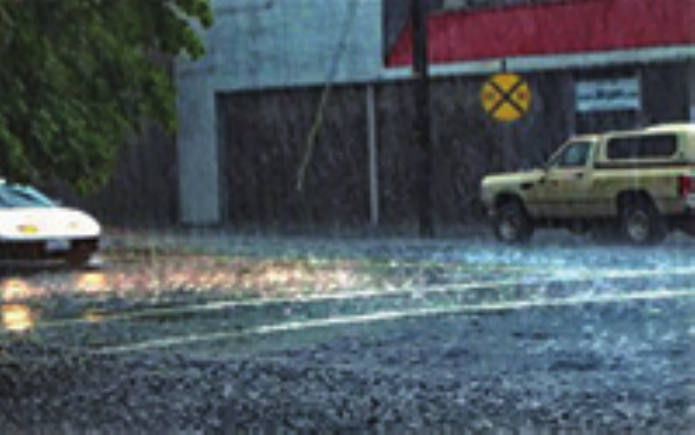}}
\end{minipage}
\vspace{0.5mm}
\vfill
\begin{minipage}{0.115\linewidth}
\centering{\includegraphics[width=.995\linewidth]{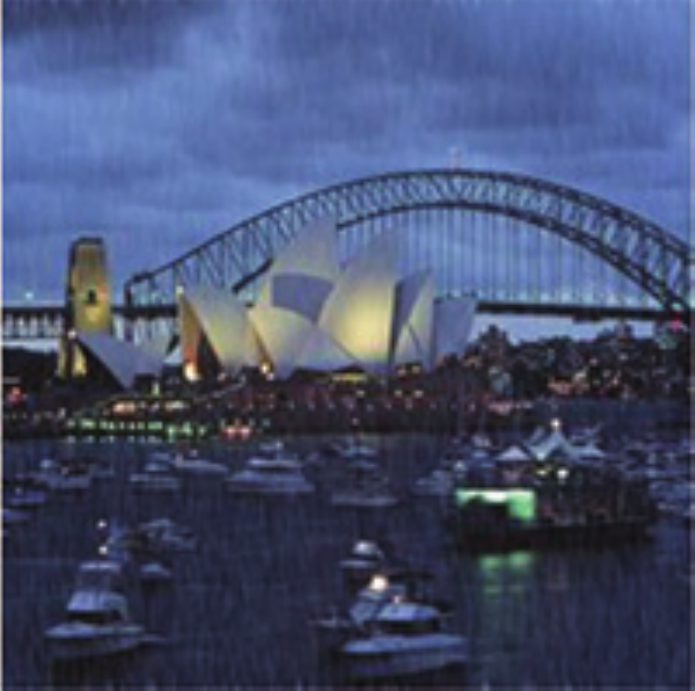}}
\end{minipage}
\hfill
\begin{minipage}{.115\linewidth}
\centering{\includegraphics[width=.995\linewidth]{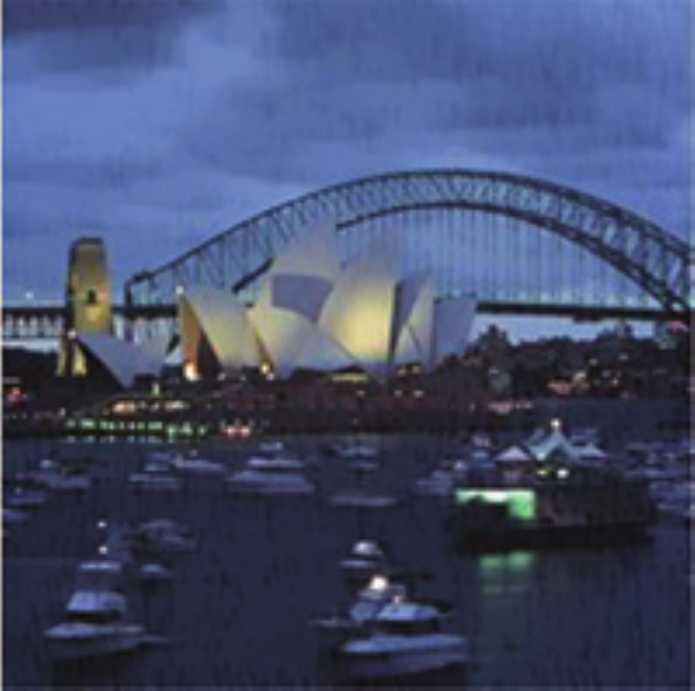}}
\end{minipage}
\hfill
\begin{minipage}{.115\linewidth}
\centering{\includegraphics[width=.995\linewidth]{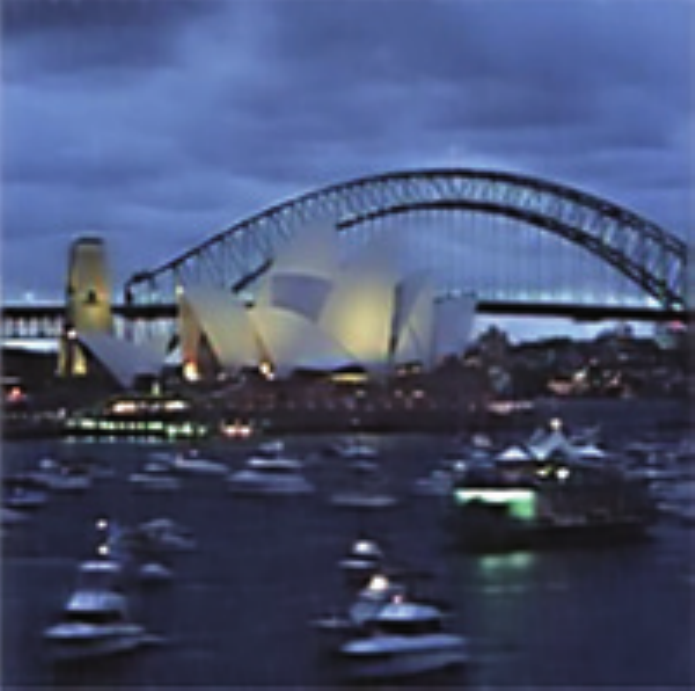}}
\end{minipage}
\hfill
\begin{minipage}{.115\linewidth}
\centering{\includegraphics[width=.995\linewidth]{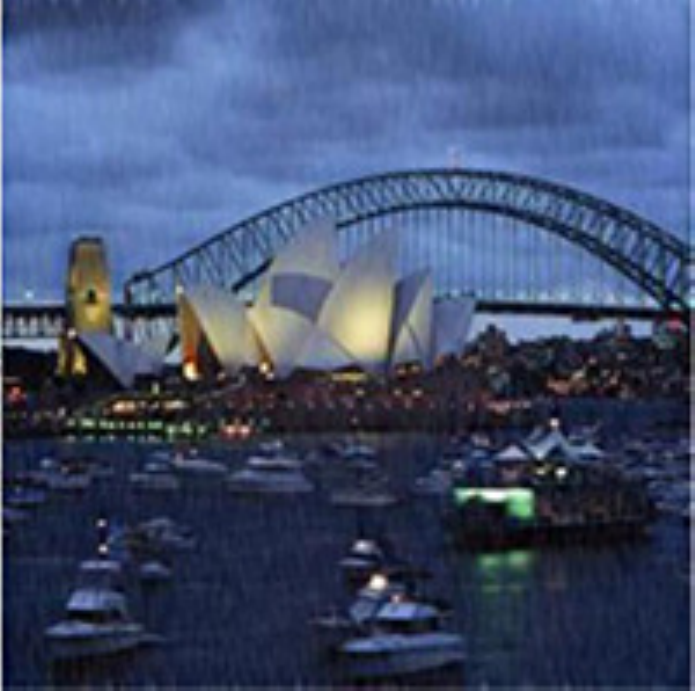}}
\end{minipage}
\hfill
\begin{minipage}{.115\linewidth}
\centering{\includegraphics[width=.995\linewidth]{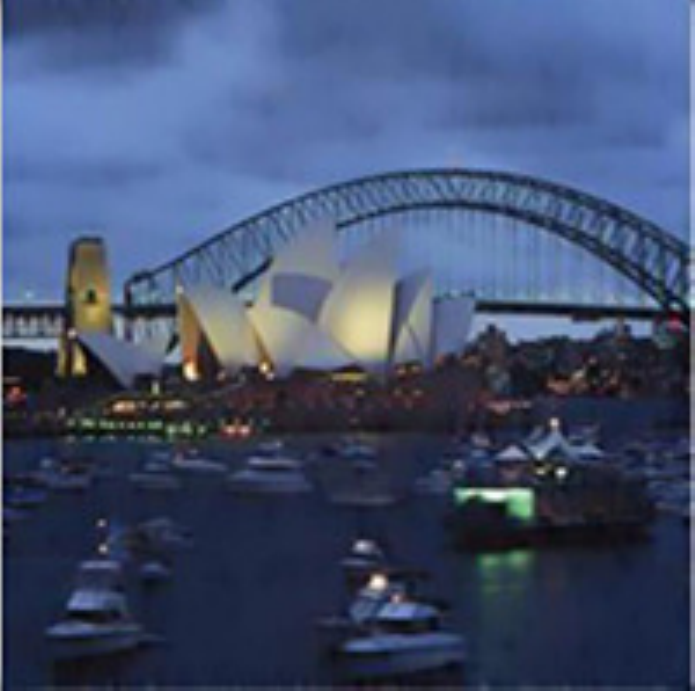}}
\end{minipage}
\hfill
\begin{minipage}{.115\linewidth}
\centering{\includegraphics[width=.995\linewidth]{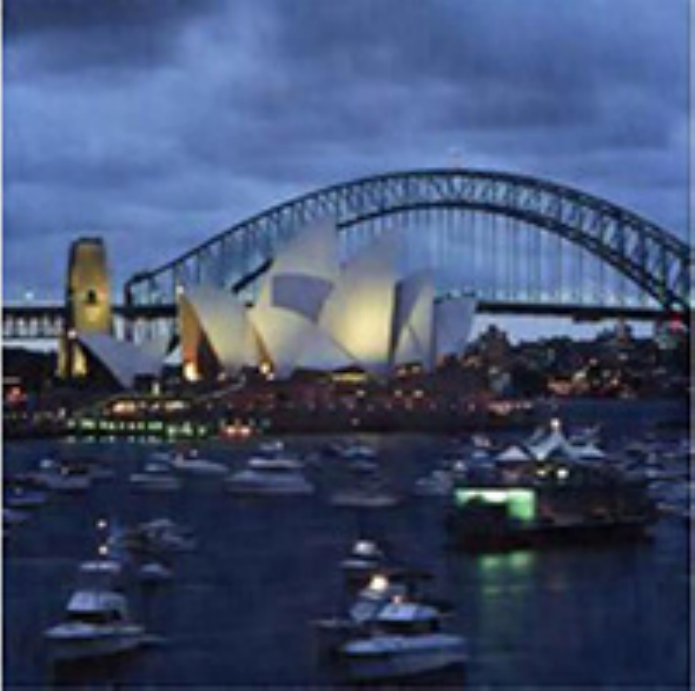}}
\end{minipage}
\hfill
\begin{minipage}{.115\linewidth}
\centering{\includegraphics[width=.995\linewidth]{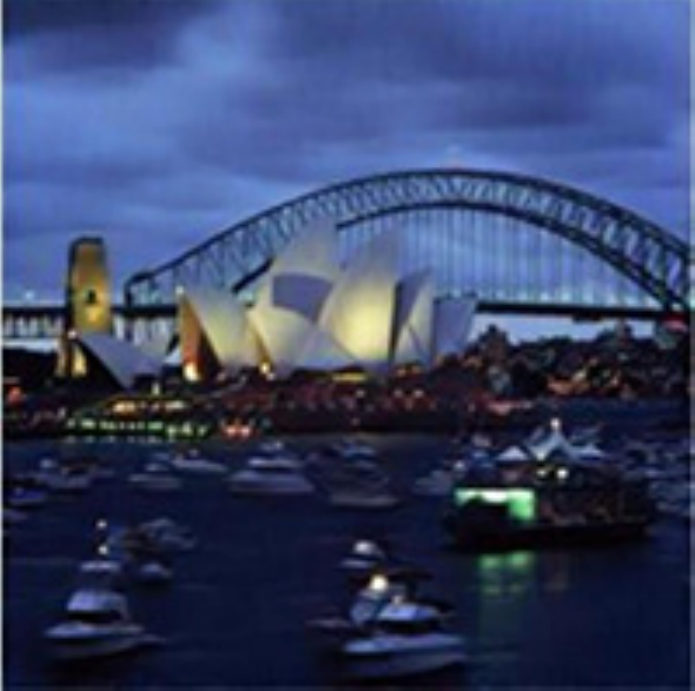}}
\end{minipage}
\hfill
\begin{minipage}{.115\linewidth}
\centering{\includegraphics[width=.995\linewidth]{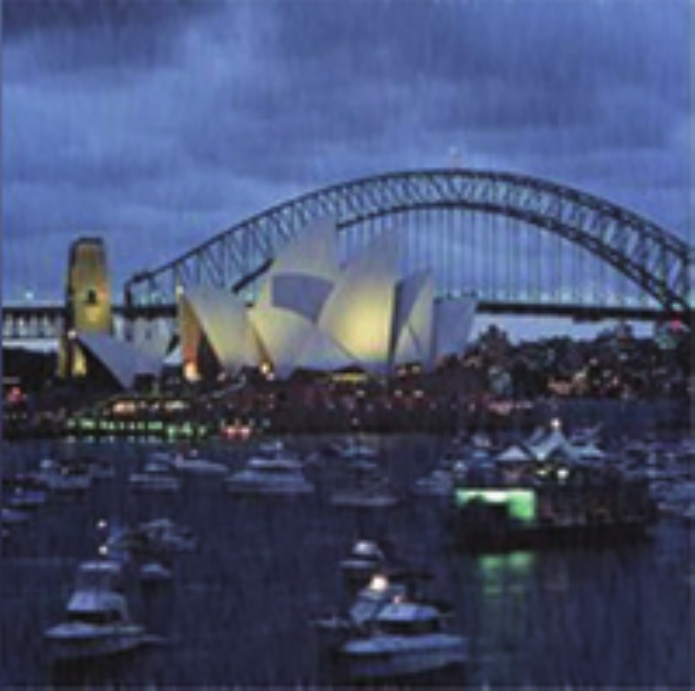}}
\end{minipage}
\vspace{0.5mm}
\vfill
\begin{minipage}{0.115\linewidth}
\centering{\includegraphics[width=.995\linewidth]{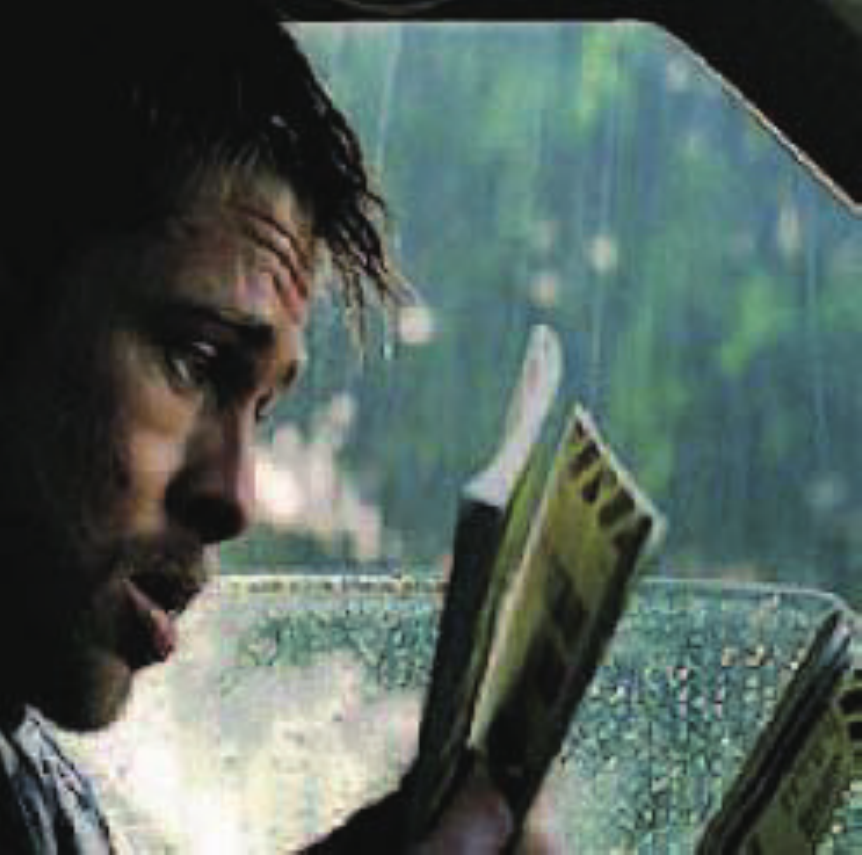}}
\end{minipage}
\hfill
\begin{minipage}{.115\linewidth}
\centering{\includegraphics[width=.995\linewidth]{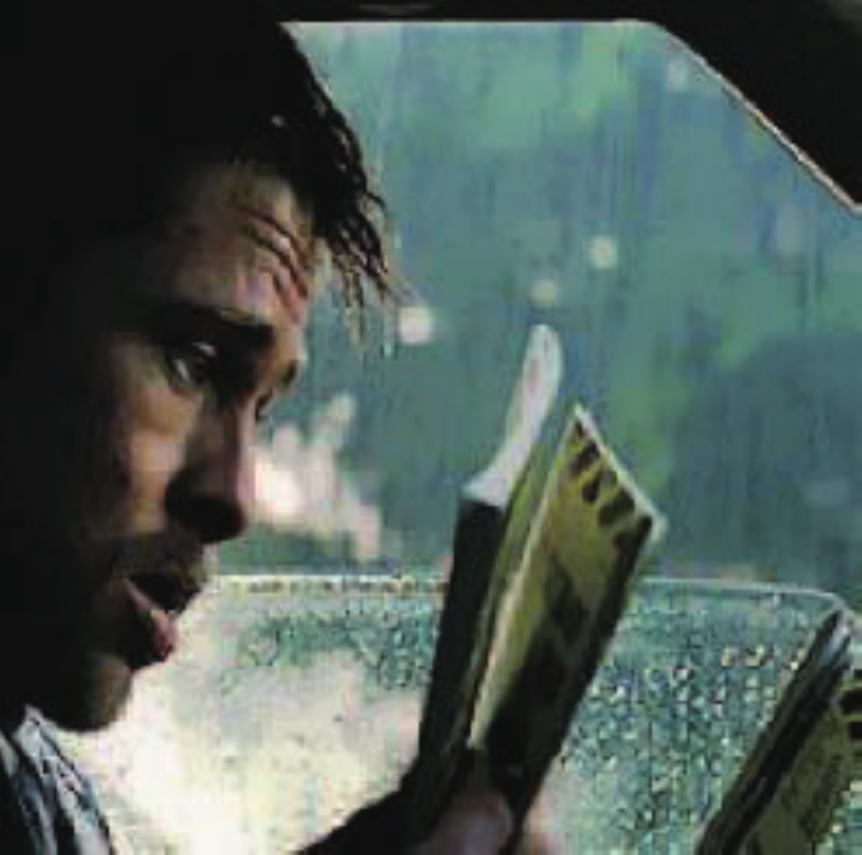}}
\end{minipage}
\hfill
\begin{minipage}{.115\linewidth}
\centering{\includegraphics[width=.995\linewidth]{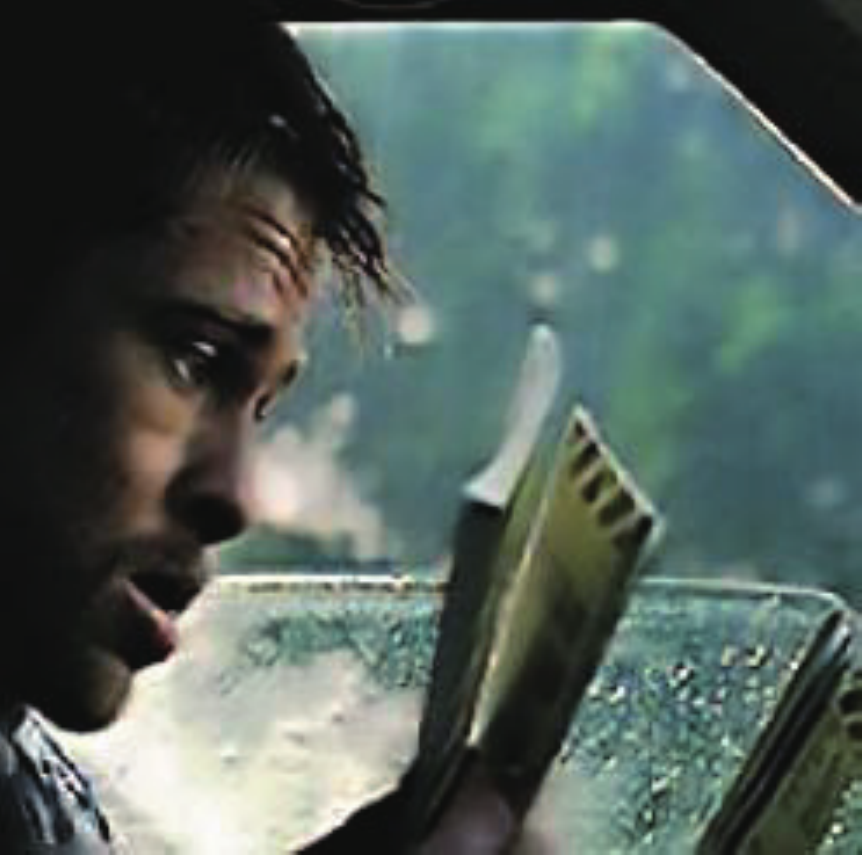}}
\end{minipage}
\hfill
\begin{minipage}{.115\linewidth}
\centering{\includegraphics[width=.995\linewidth]{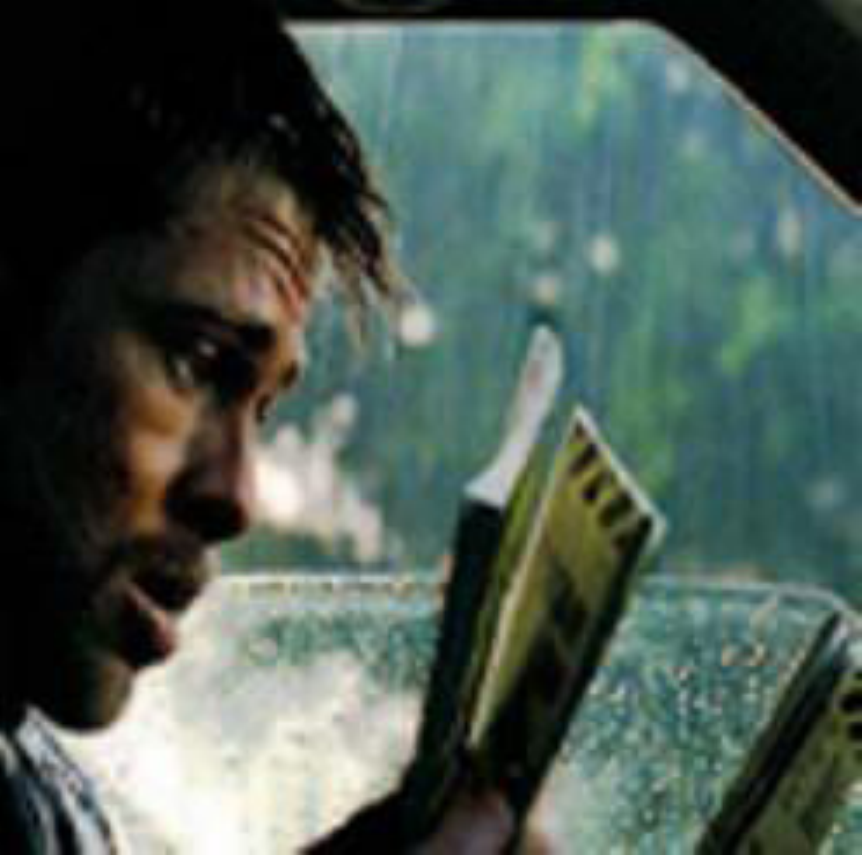}}
\end{minipage}
\hfill
\begin{minipage}{.115\linewidth}
\centering{\includegraphics[width=.995\linewidth]{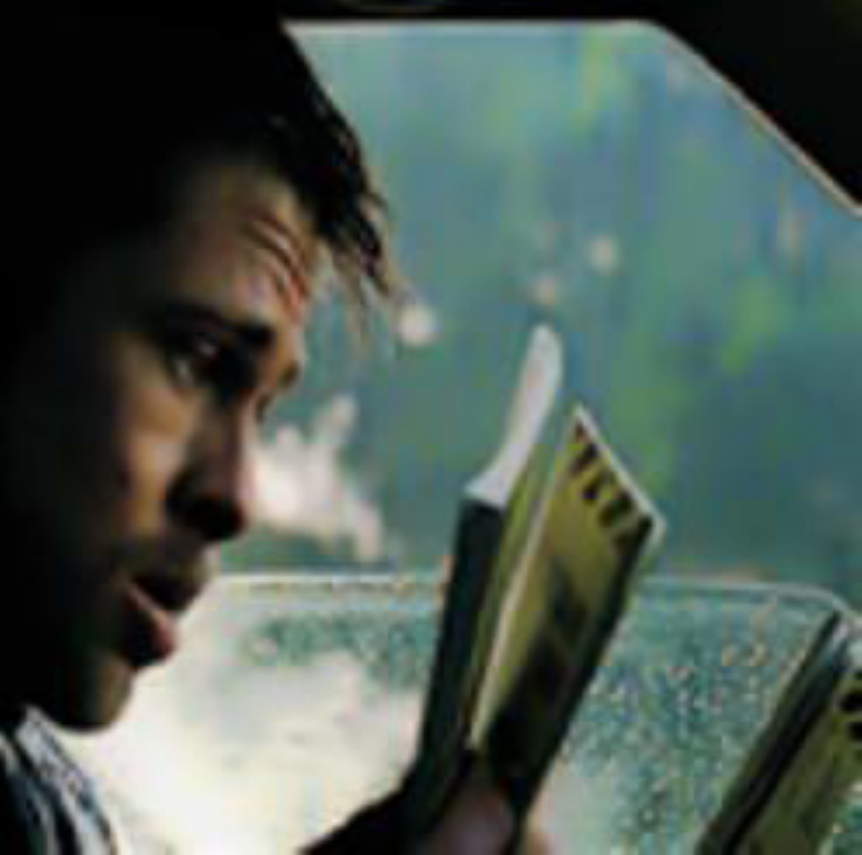}}
\end{minipage}
\hfill
\begin{minipage}{.115\linewidth}
\centering{\includegraphics[width=.995\linewidth]{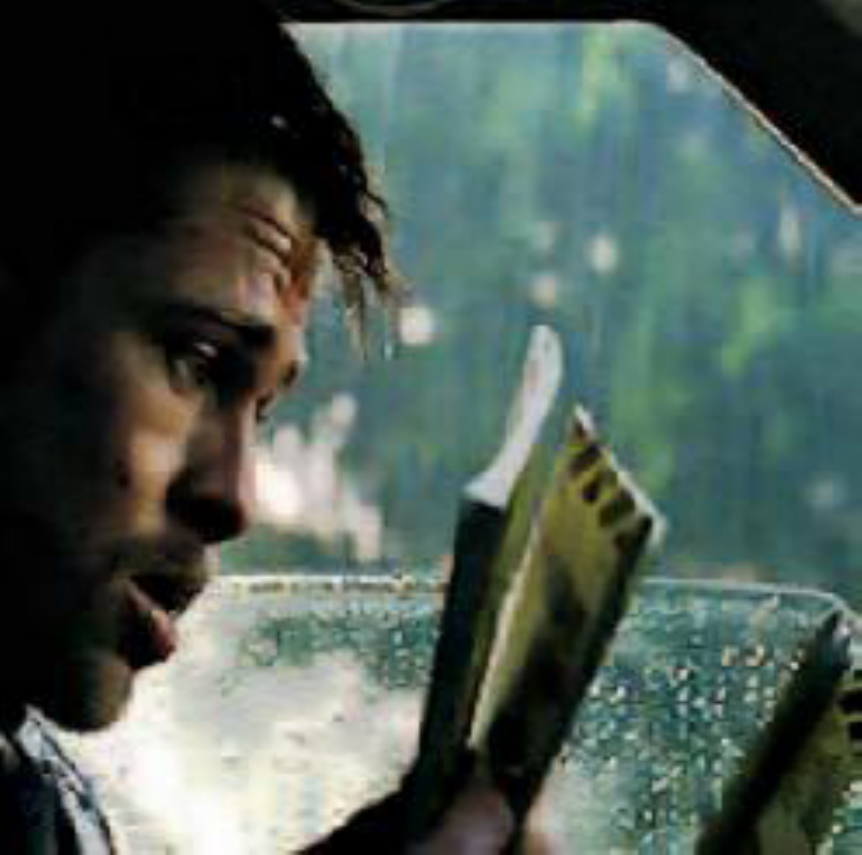}}
\end{minipage}
\hfill
\begin{minipage}{.115\linewidth}
\centering{\includegraphics[width=.995\linewidth]{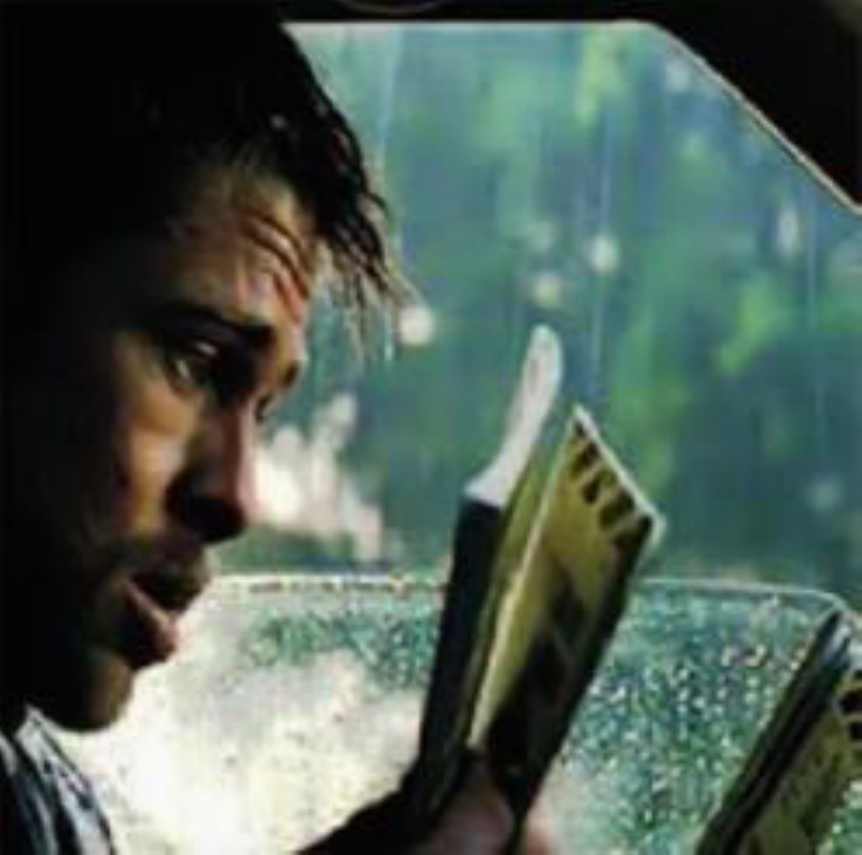}}
\end{minipage}
\hfill
\begin{minipage}{.115\linewidth}
\centering{\includegraphics[width=.995\linewidth]{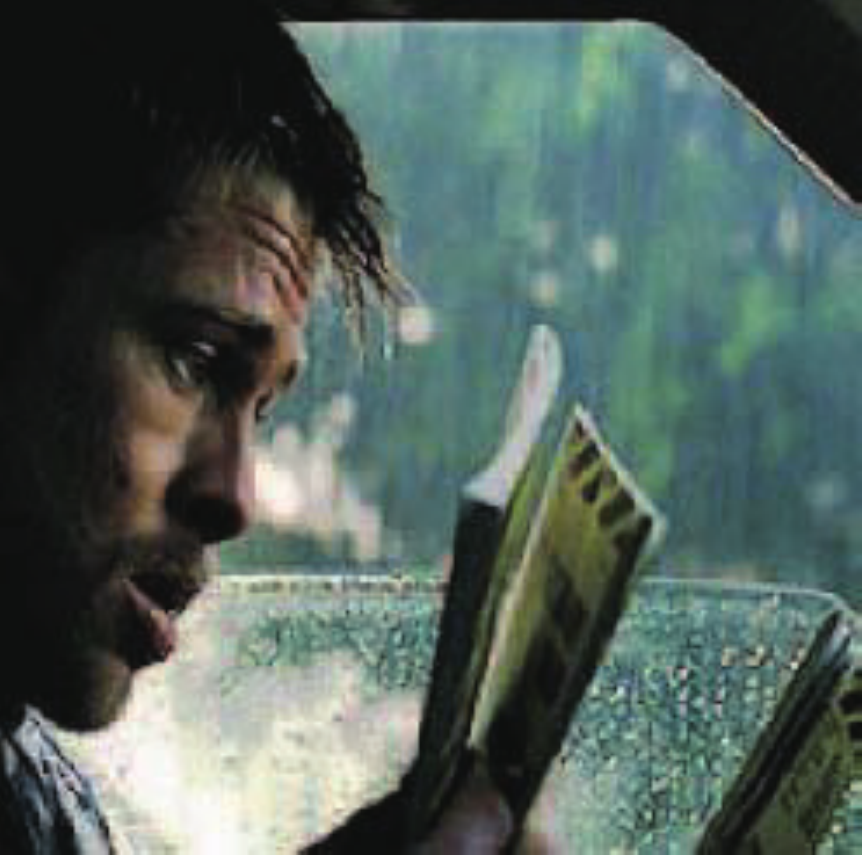}}
\end{minipage}
\vspace{0.5mm}
\vfill
\begin{minipage}{0.115\linewidth}
\centering{\includegraphics[width=.995\linewidth]{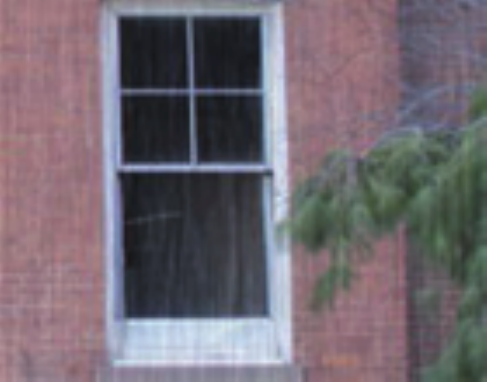}}
\end{minipage}
\hfill
\begin{minipage}{.115\linewidth}
\centering{\includegraphics[width=.995\linewidth]{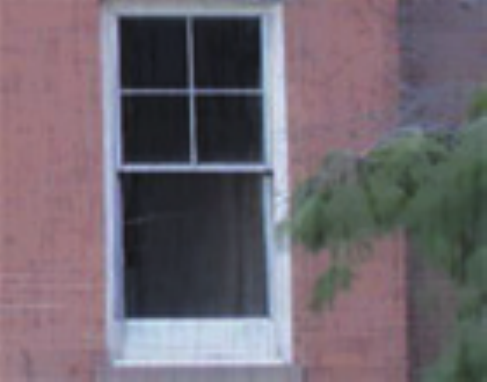}}
\end{minipage}
\hfill
\begin{minipage}{.115\linewidth}
\centering{\includegraphics[width=.995\linewidth]{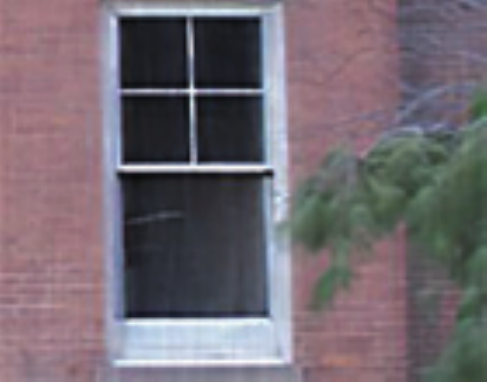}}
\end{minipage}
\hfill
\begin{minipage}{.115\linewidth}
\centering{\includegraphics[width=.995\linewidth]{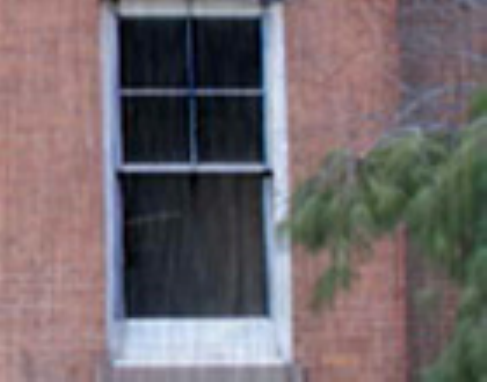}}
\end{minipage}
\hfill
\begin{minipage}{.115\linewidth}
\centering{\includegraphics[width=.995\linewidth]{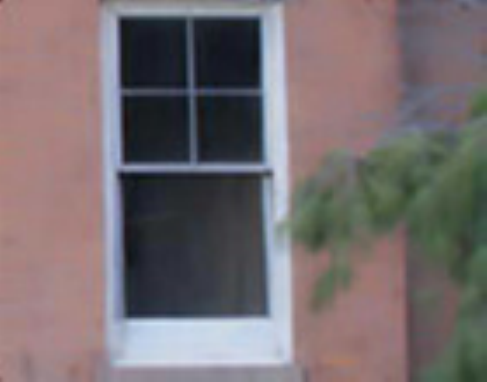}}
\end{minipage}
\hfill
\begin{minipage}{.115\linewidth}
\centering{\includegraphics[width=.995\linewidth]{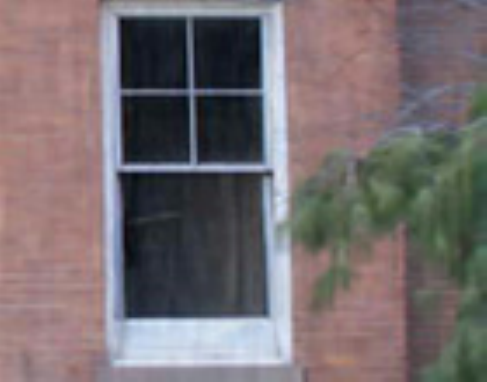}}
\end{minipage}
\hfill
\begin{minipage}{.115\linewidth}
\centering{\includegraphics[width=.995\linewidth]{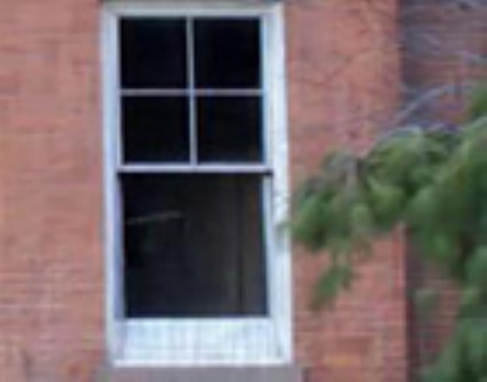}}
\end{minipage}
\hfill
\begin{minipage}{.115\linewidth}
\centering{\includegraphics[width=.995\linewidth]{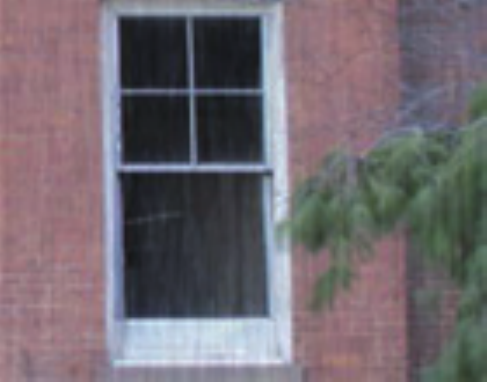}}
\end{minipage}
\vspace{0.5mm}
\vfill
\begin{minipage}{0.115\linewidth}
\centering{\includegraphics[width=.995\linewidth]{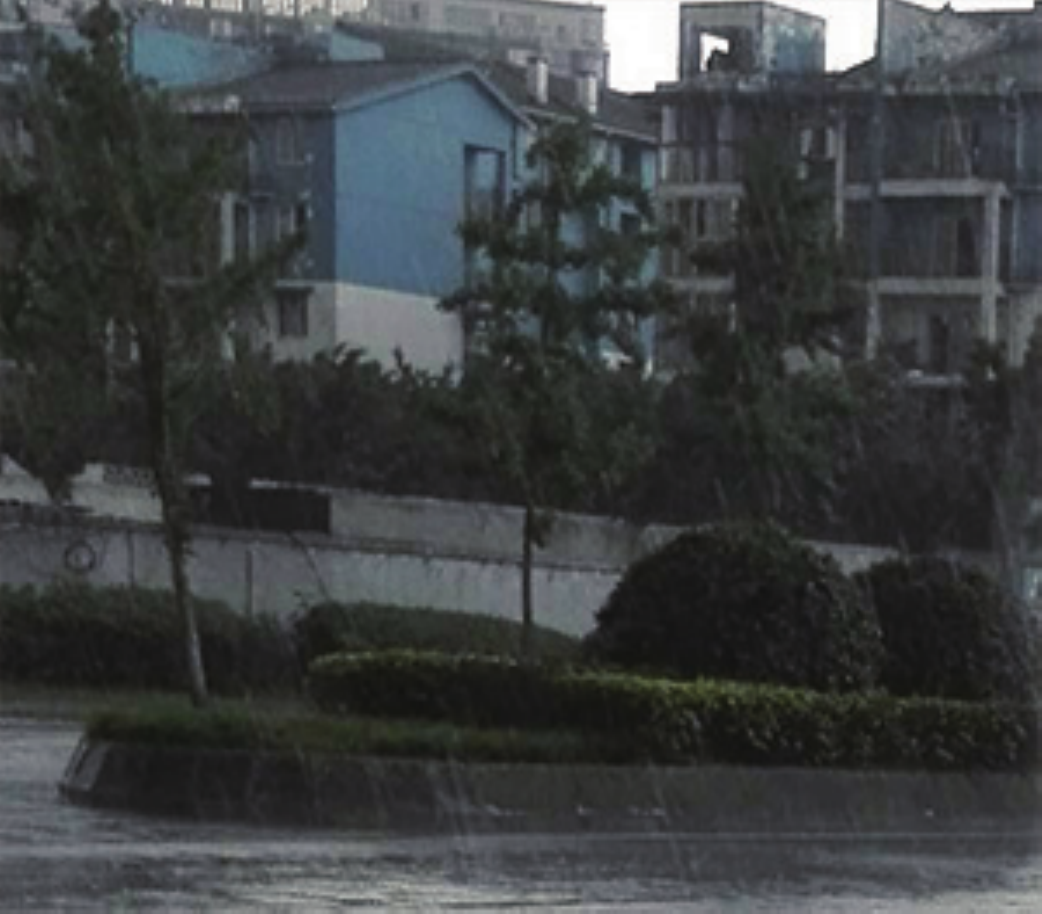}}
\end{minipage}
\hfill
\begin{minipage}{.115\linewidth}
\centering{\includegraphics[width=.995\linewidth]{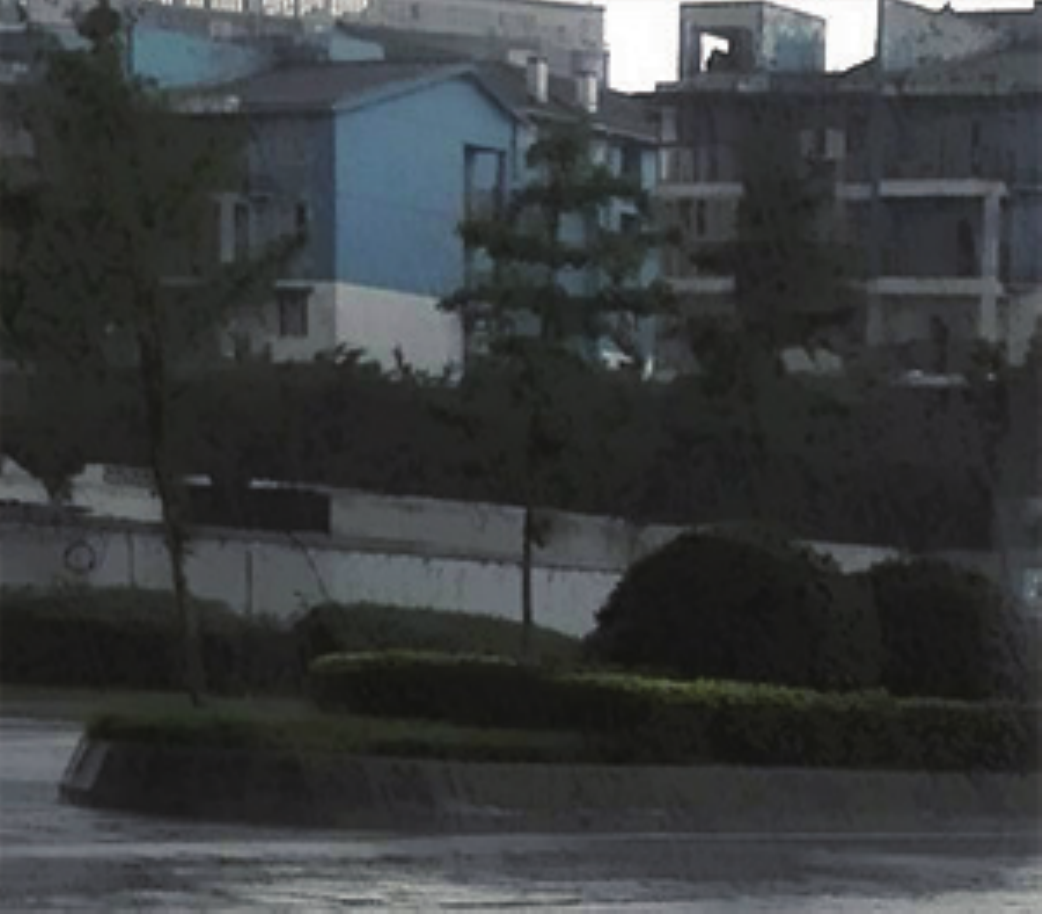}}
\end{minipage}
\hfill
\begin{minipage}{.115\linewidth}
\centering{\includegraphics[width=.995\linewidth]{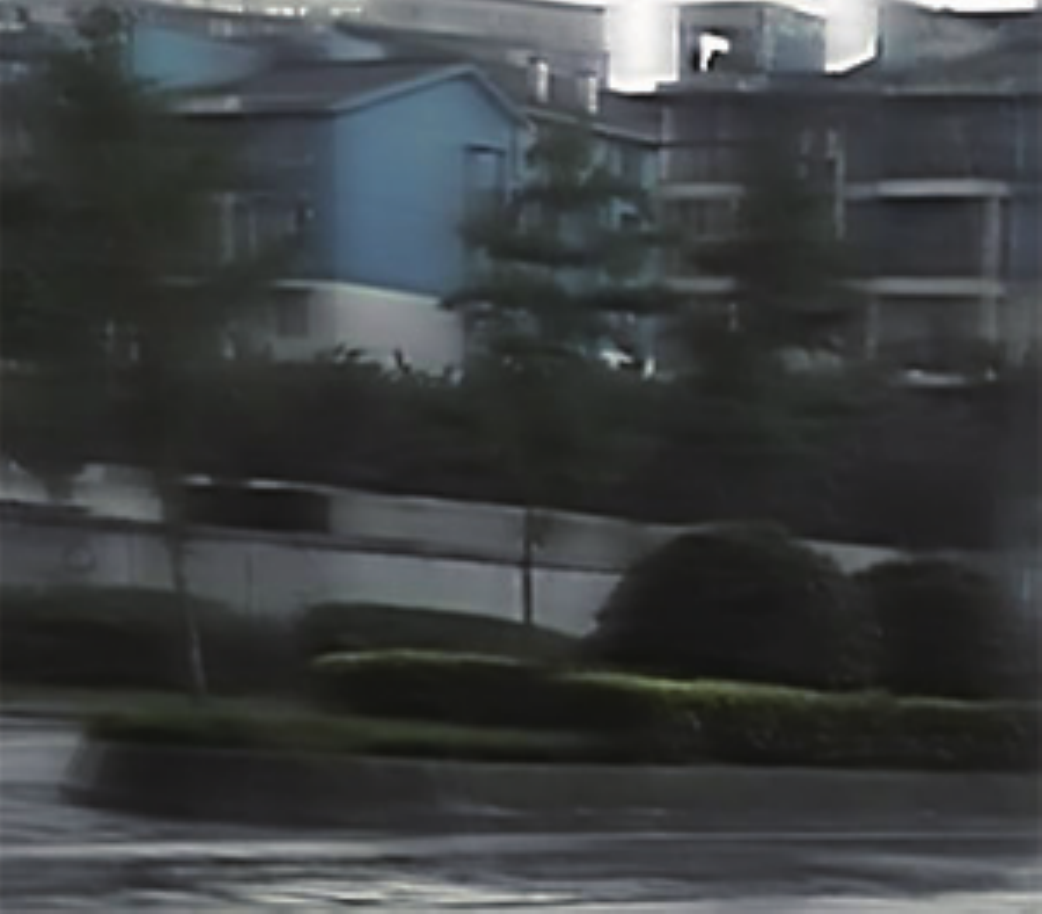}}
\end{minipage}
\hfill
\begin{minipage}{.115\linewidth}
\centering{\includegraphics[width=.995\linewidth]{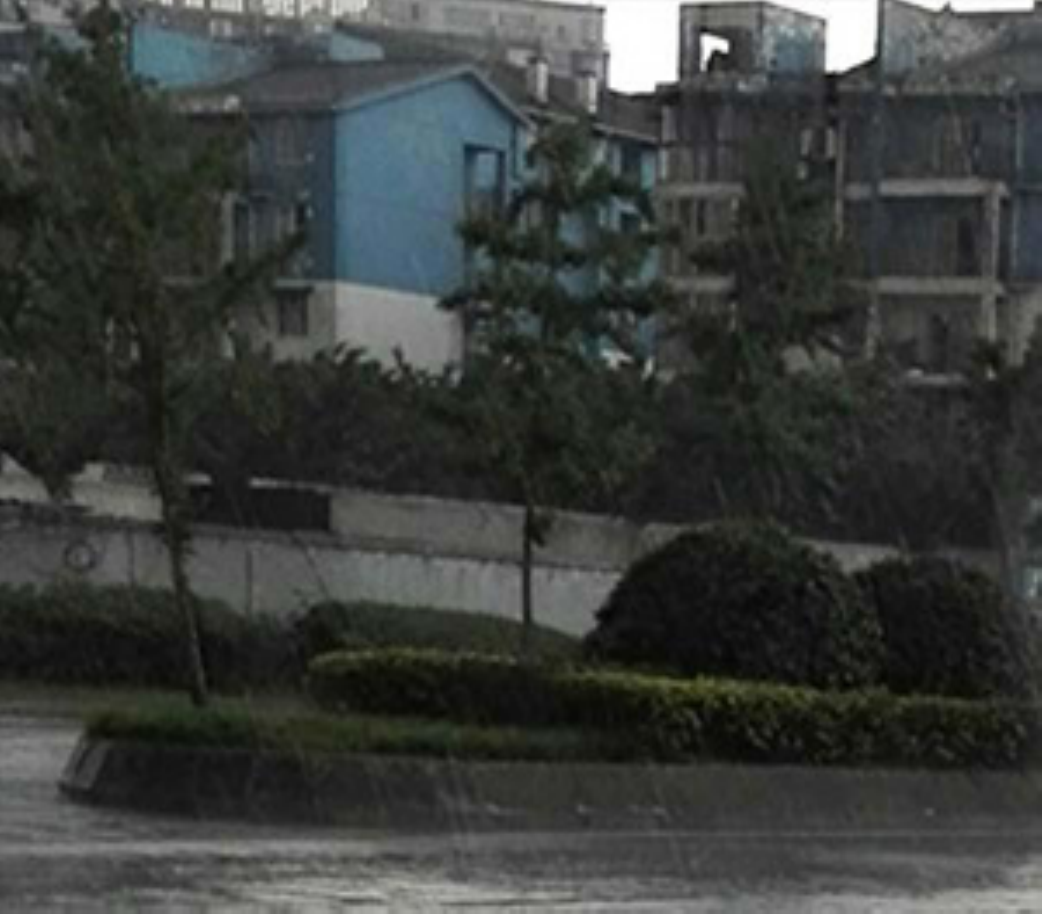}}
\end{minipage}
\hfill
\begin{minipage}{.115\linewidth}
\centering{\includegraphics[width=.995\linewidth]{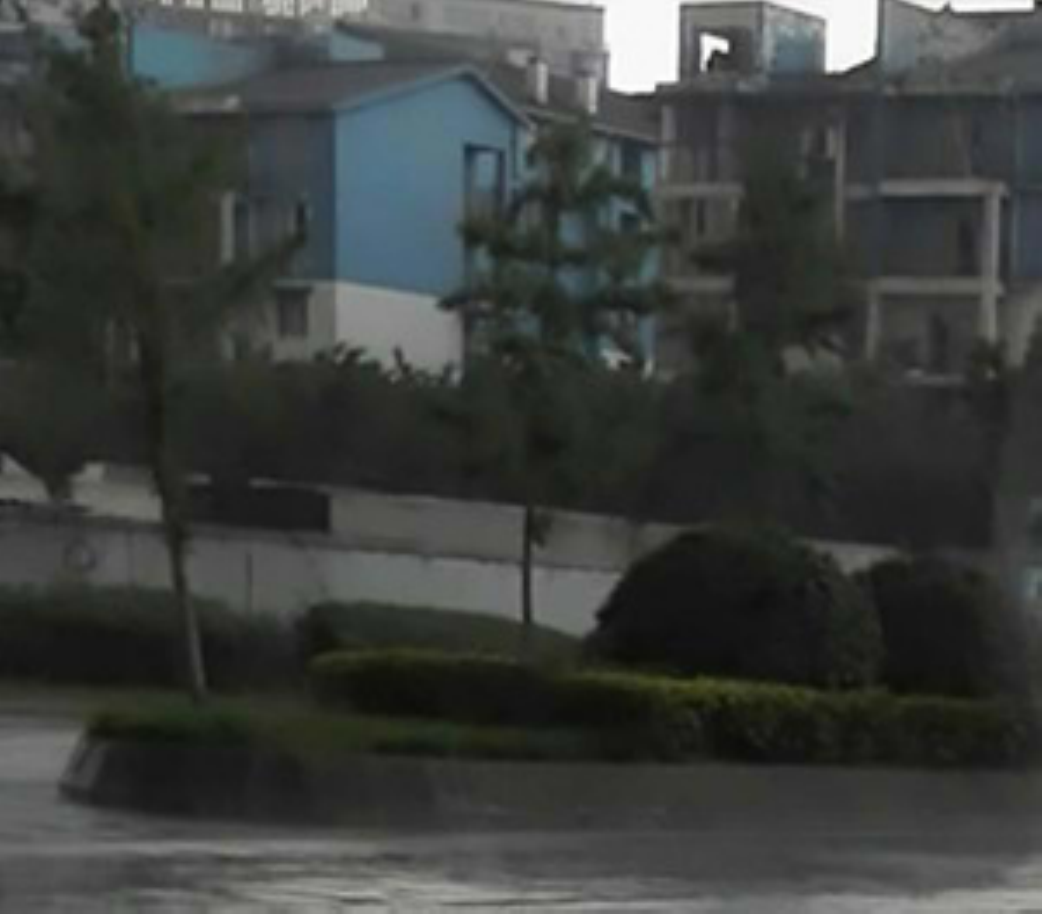}}
\end{minipage}
\hfill
\begin{minipage}{.115\linewidth}
\centering{\includegraphics[width=.995\linewidth]{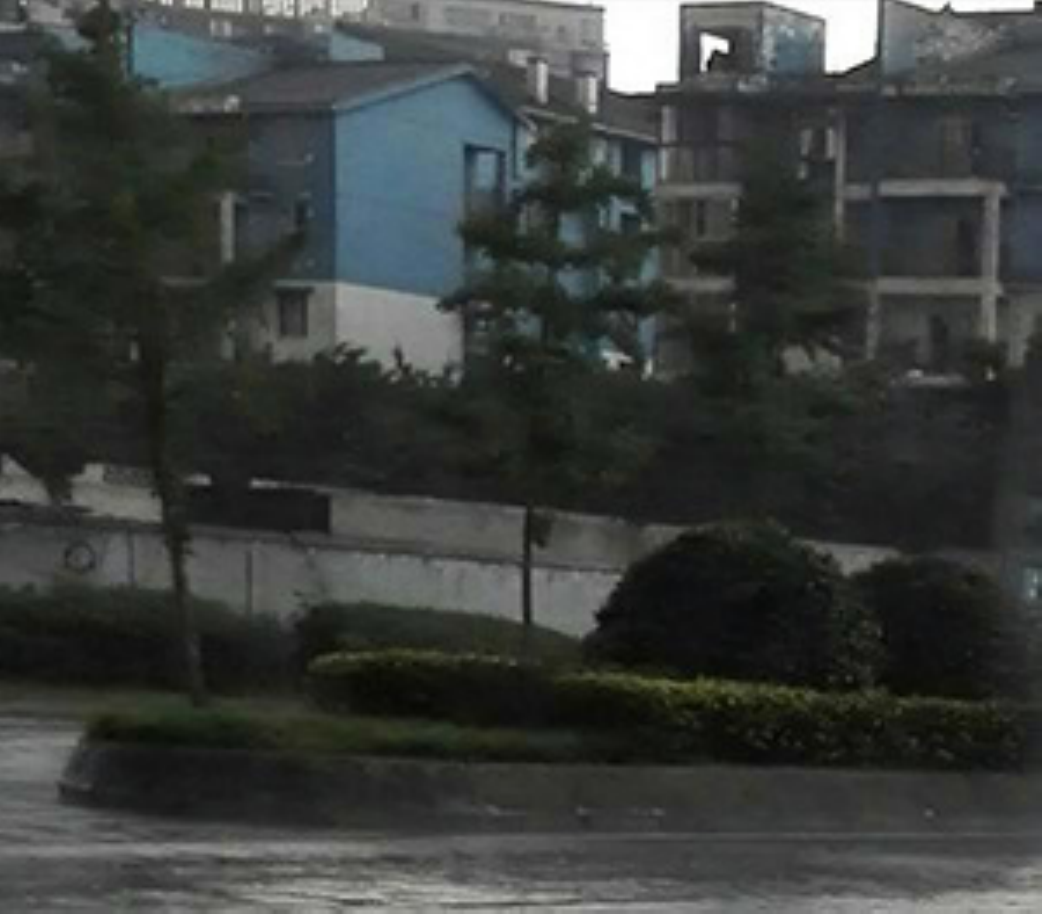}}
\end{minipage}
\hfill
\begin{minipage}{.115\linewidth}
\centering{\includegraphics[width=.995\linewidth]{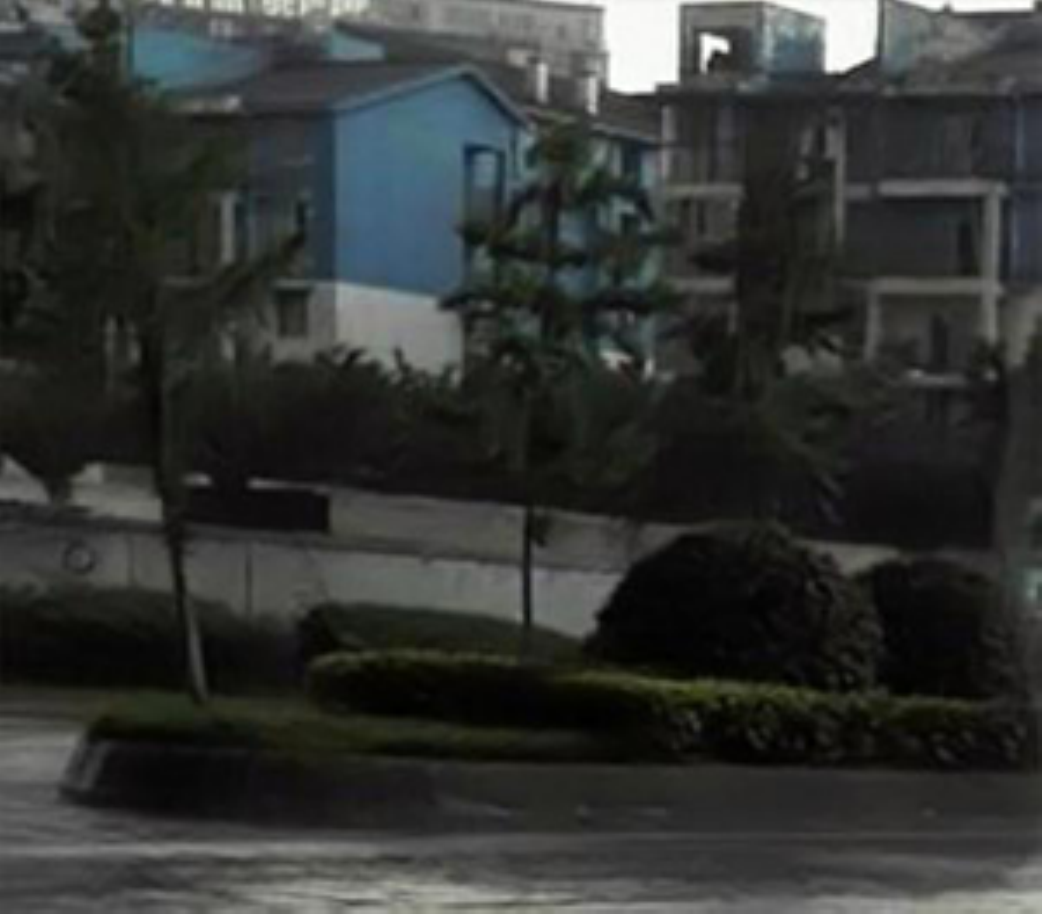}}
\end{minipage}
\hfill
\begin{minipage}{.115\linewidth}
\centering{\includegraphics[width=.995\linewidth]{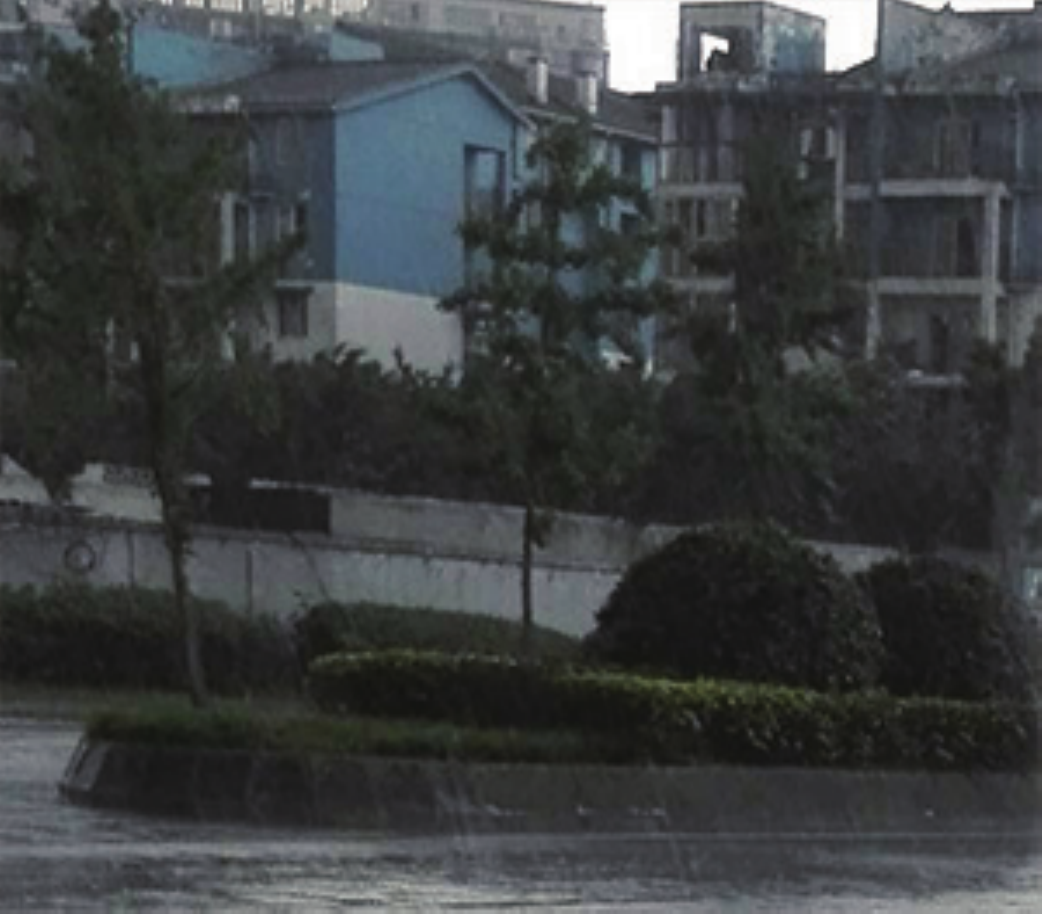}}
\end{minipage}
\vspace{0.5mm}
\vfill
\begin{minipage}{0.115\linewidth}
\centering{\includegraphics[width=.995\linewidth]{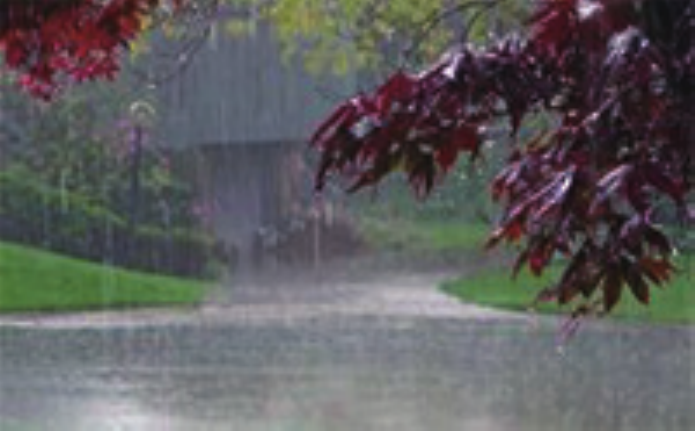}}
\end{minipage}
\hfill
\begin{minipage}{.115\linewidth}
\centering{\includegraphics[width=.995\linewidth]{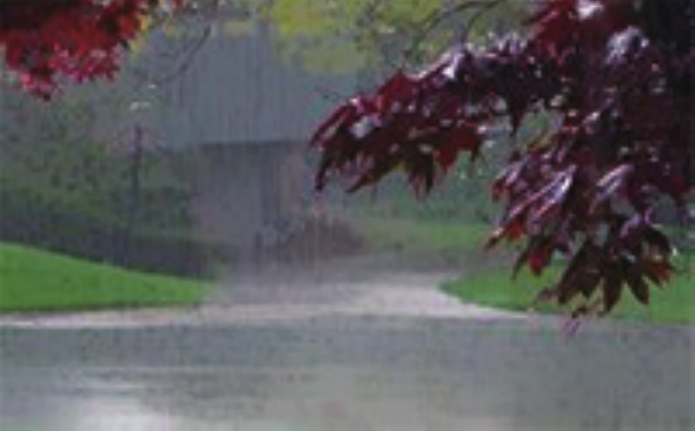}}
\end{minipage}
\hfill
\begin{minipage}{.115\linewidth}
\centering{\includegraphics[width=.995\linewidth]{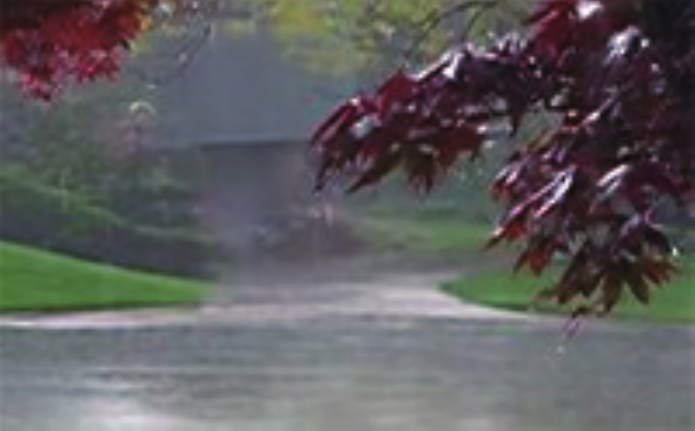}}
\end{minipage}
\hfill
\begin{minipage}{.115\linewidth}
\centering{\includegraphics[width=.995\linewidth]{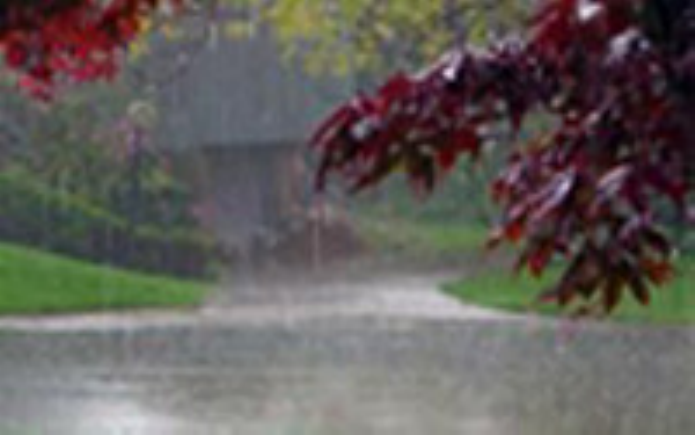}}
\end{minipage}
\hfill
\begin{minipage}{.115\linewidth}
\centering{\includegraphics[width=.995\linewidth]{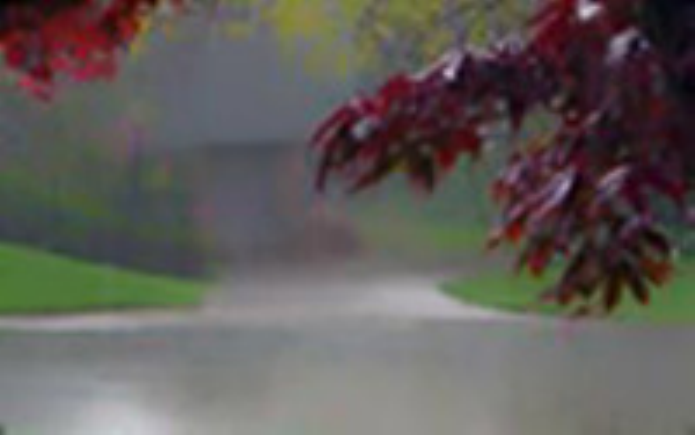}}
\end{minipage}
\hfill
\begin{minipage}{.115\linewidth}
\centering{\includegraphics[width=.995\linewidth]{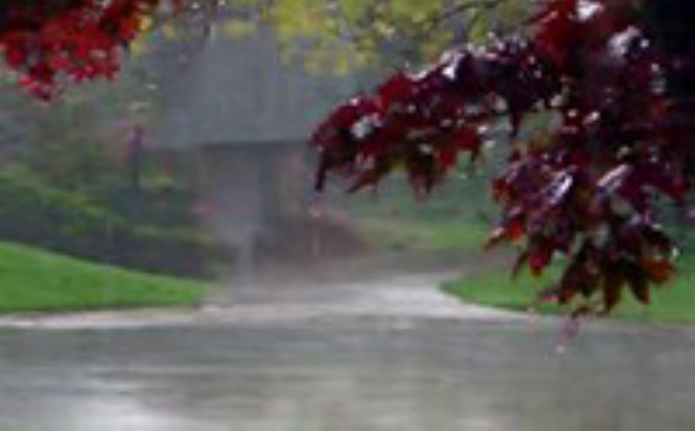}}
\end{minipage}
\hfill
\begin{minipage}{.115\linewidth}
\centering{\includegraphics[width=.995\linewidth]{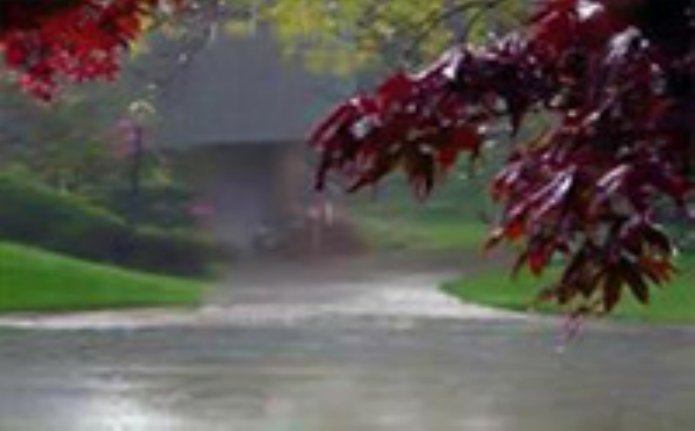}}
\end{minipage}
\hfill
\begin{minipage}{.115\linewidth}
\centering{\includegraphics[width=.995\linewidth]{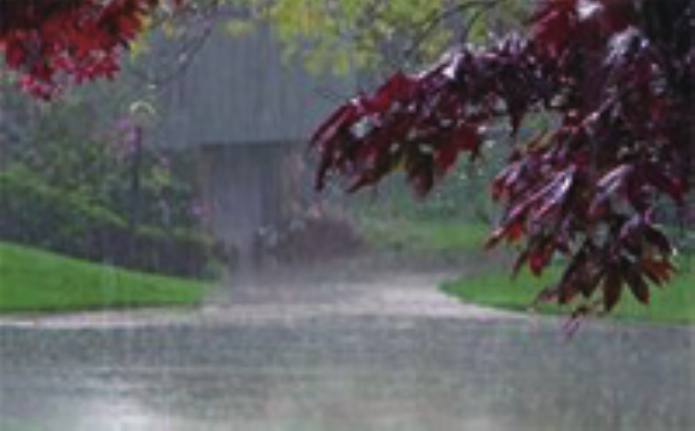}}
\end{minipage}
\vspace{0.5mm}
\vfill
\begin{minipage}{0.115\linewidth}
\centering{\includegraphics[width=.995\linewidth]{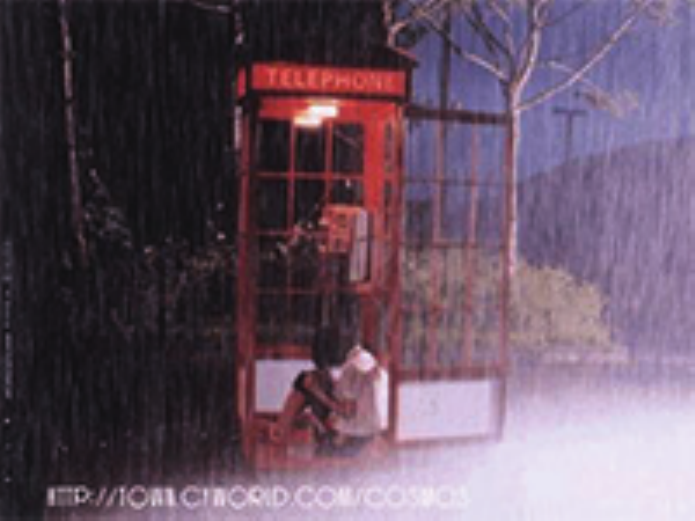}}
\end{minipage}
\hfill
\begin{minipage}{.115\linewidth}
\centering{\includegraphics[width=.995\linewidth]{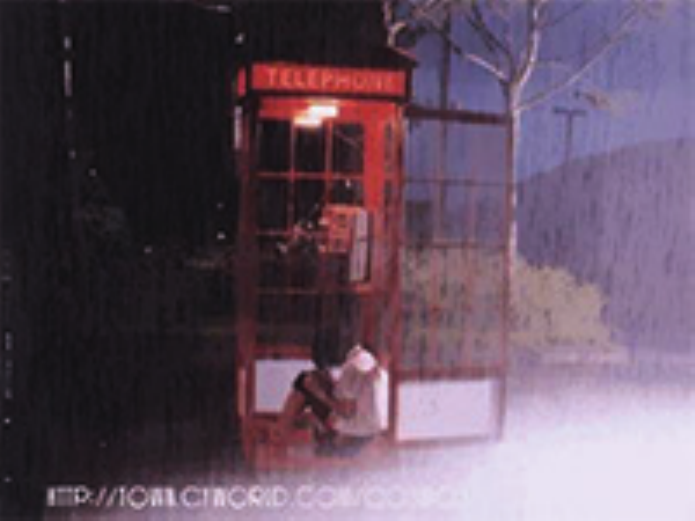}}
\end{minipage}
\hfill
\begin{minipage}{.115\linewidth}
\centering{\includegraphics[width=.995\linewidth]{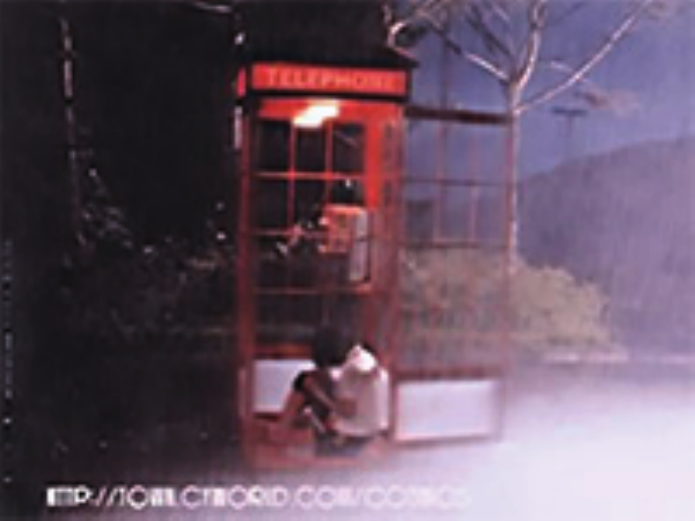}}
\end{minipage}
\hfill
\begin{minipage}{.115\linewidth}
\centering{\includegraphics[width=.995\linewidth]{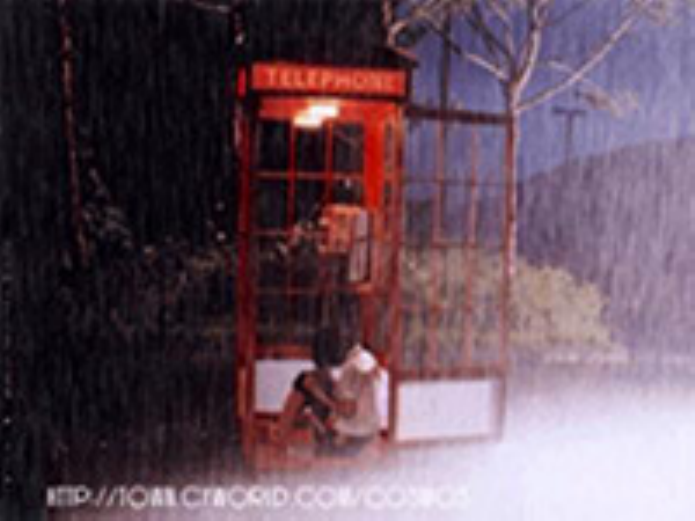}}
\end{minipage}
\hfill
\begin{minipage}{.115\linewidth}
\centering{\includegraphics[width=.995\linewidth]{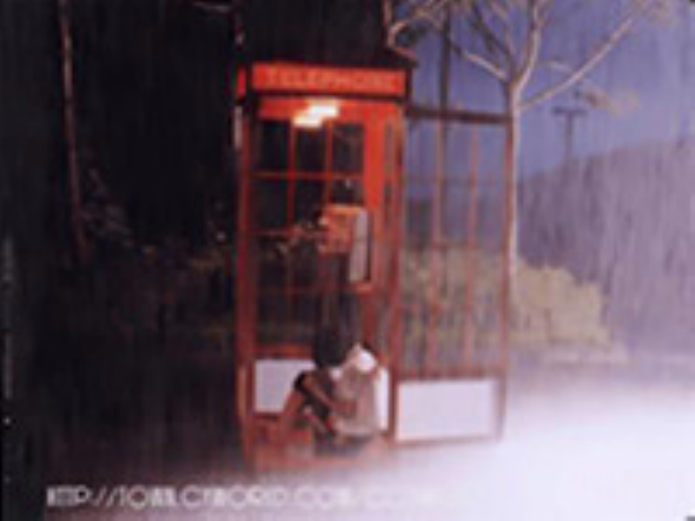}}
\end{minipage}
\hfill
\begin{minipage}{.115\linewidth}
\centering{\includegraphics[width=.995\linewidth]{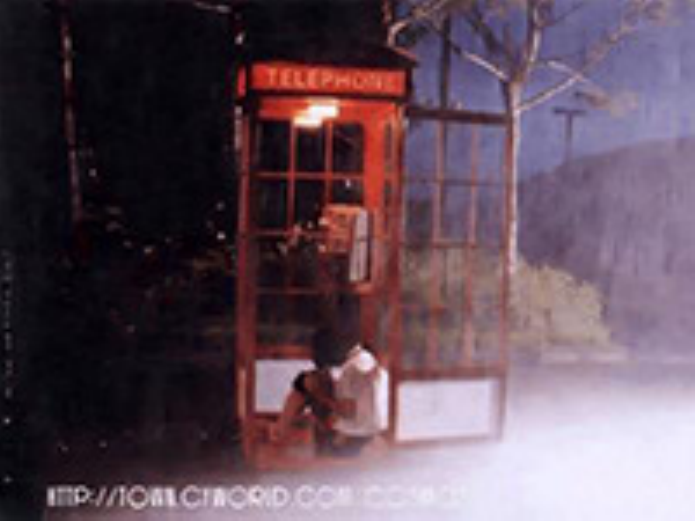}}
\end{minipage}
\hfill
\begin{minipage}{.115\linewidth}
\centering{\includegraphics[width=.995\linewidth]{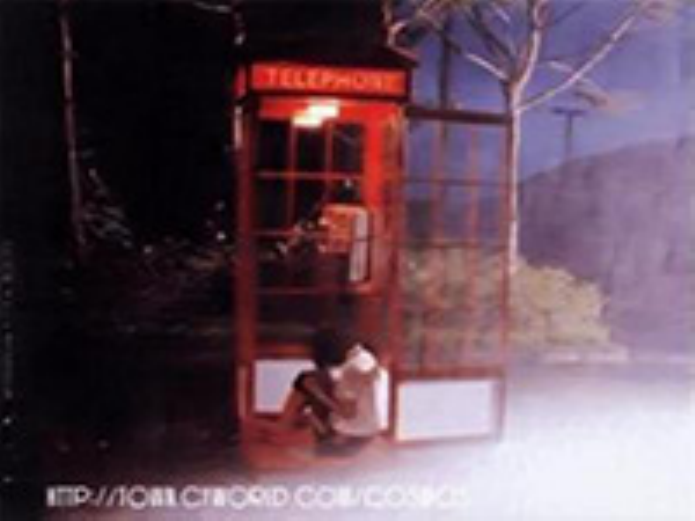}}
\end{minipage}
\hfill
\begin{minipage}{.115\linewidth}
\centering{\includegraphics[width=.995\linewidth]{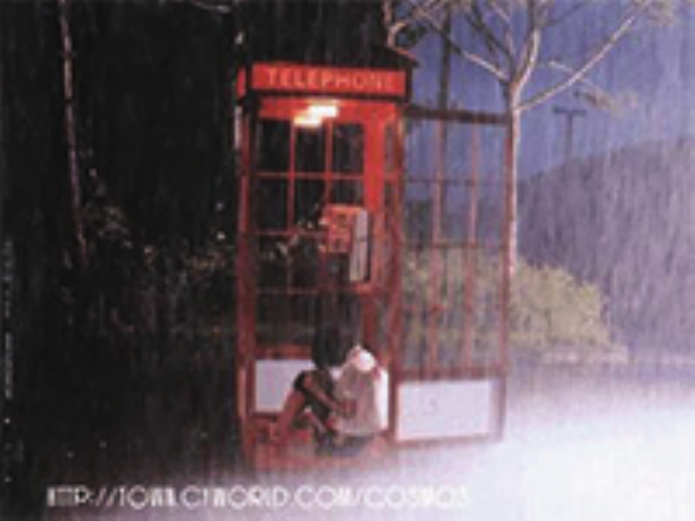}}
\end{minipage}
\vspace{0.5mm}
\vfill
\begin{minipage}{0.115\linewidth}
\centering{\includegraphics[width=.995\linewidth]{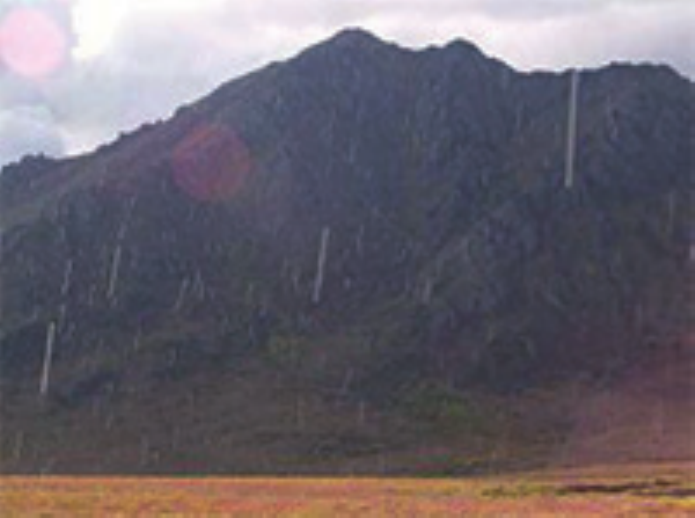}}
\end{minipage}
\hfill
\begin{minipage}{.115\linewidth}
\centering{\includegraphics[width=.995\linewidth]{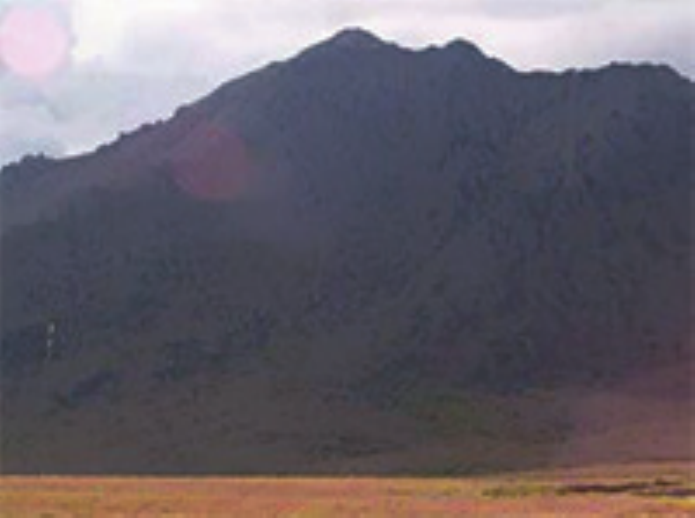}}
\end{minipage}
\hfill
\begin{minipage}{.115\linewidth}
\centering{\includegraphics[width=.995\linewidth]{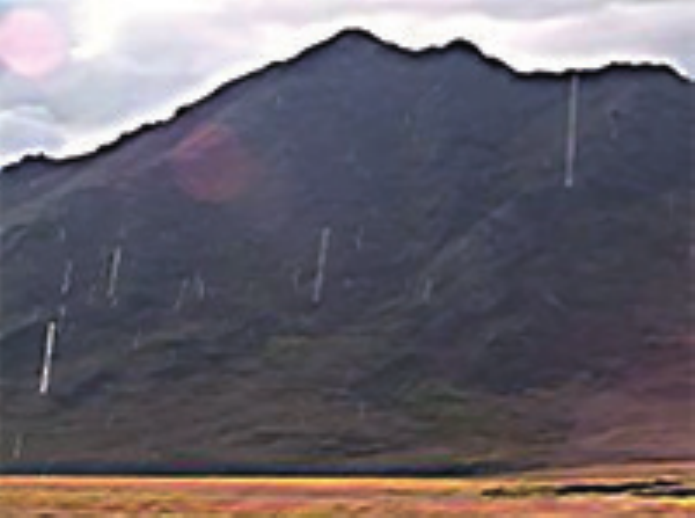}}
\end{minipage}
\hfill
\begin{minipage}{.115\linewidth}
\centering{\includegraphics[width=.995\linewidth]{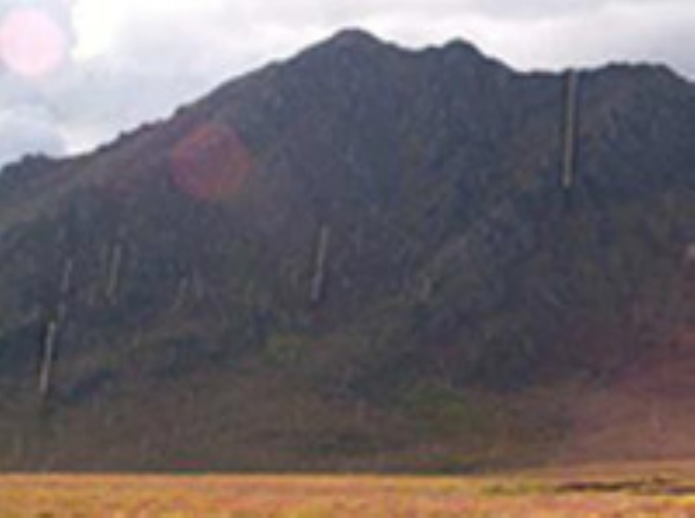}}
\end{minipage}
\hfill
\begin{minipage}{.115\linewidth}
\centering{\includegraphics[width=.995\linewidth]{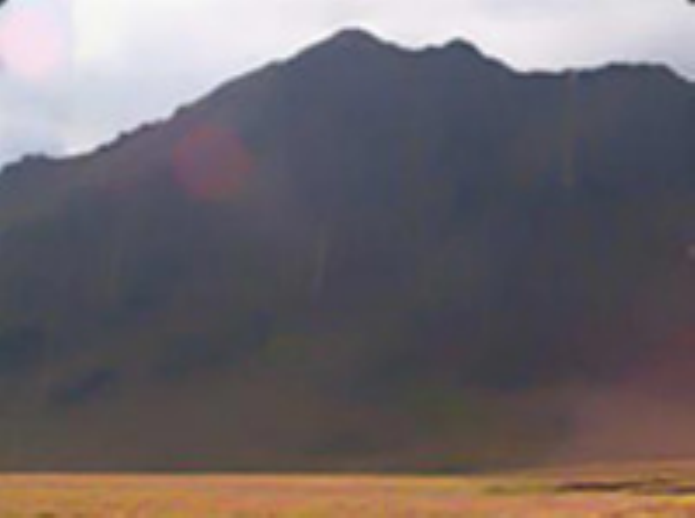}}
\end{minipage}
\hfill
\begin{minipage}{.115\linewidth}
\centering{\includegraphics[width=.995\linewidth]{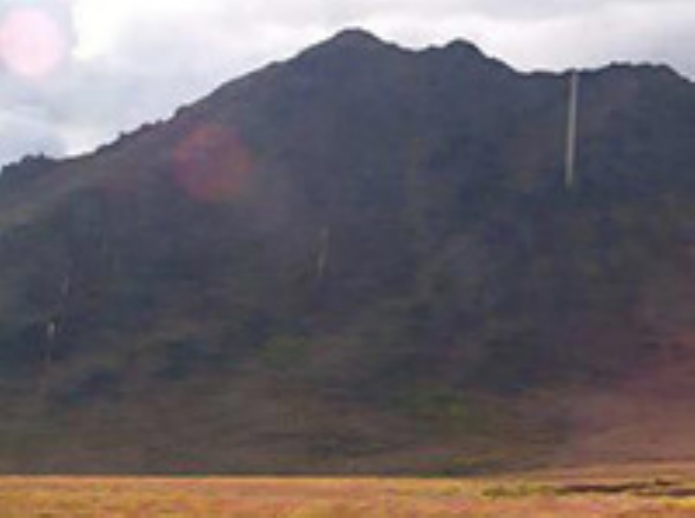}}
\end{minipage}
\hfill
\begin{minipage}{.115\linewidth}
\centering{\includegraphics[width=.995\linewidth]{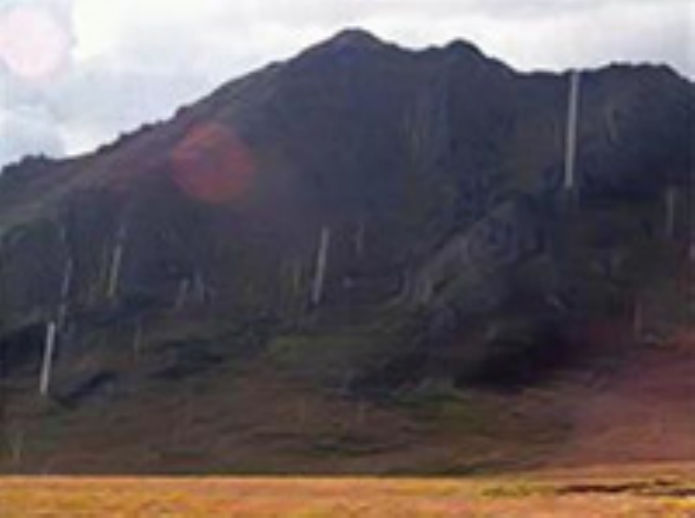}}
\end{minipage}
\hfill
\begin{minipage}{.115\linewidth}
\centering{\includegraphics[width=.995\linewidth]{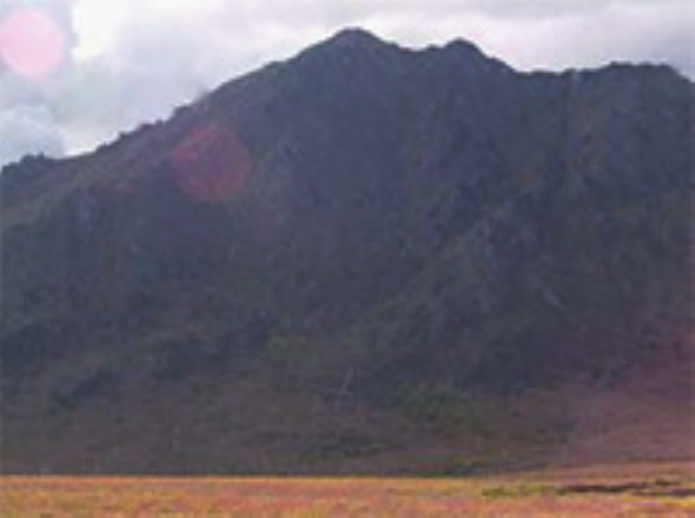}}
\end{minipage}
\vspace{0.5mm}
\vfill
\begin{minipage}{0.115\linewidth}
\centering{\includegraphics[width=.995\linewidth]{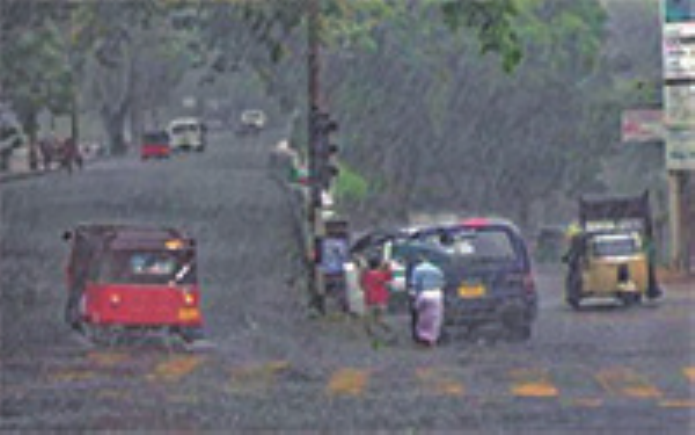}}
\end{minipage}
\hfill
\begin{minipage}{.115\linewidth}
\centering{\includegraphics[width=.995\linewidth]{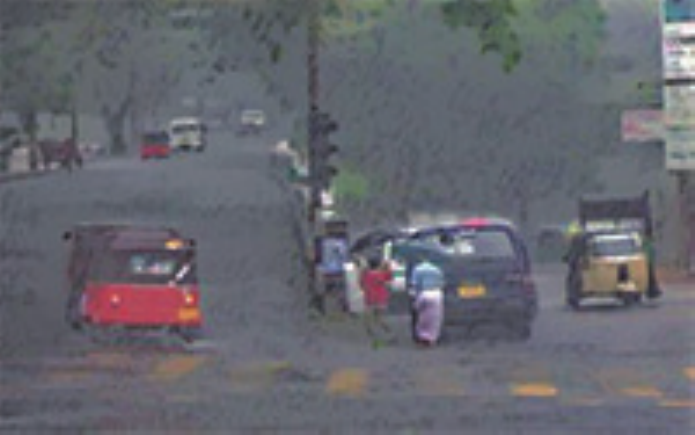}}
\end{minipage}
\hfill
\begin{minipage}{.115\linewidth}
\centering{\includegraphics[width=.995\linewidth]{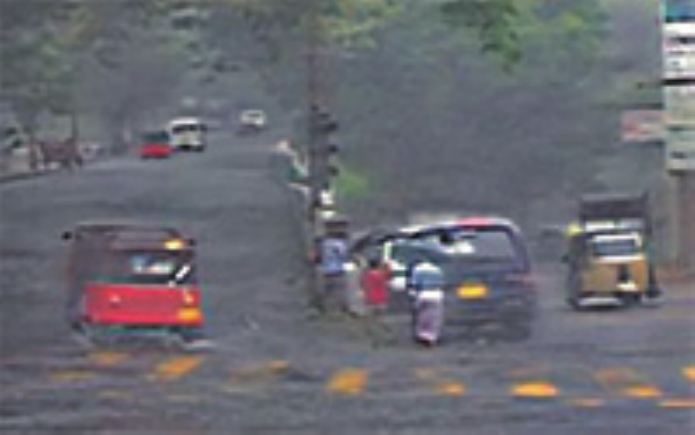}}
\end{minipage}
\hfill
\begin{minipage}{.115\linewidth}
\centering{\includegraphics[width=.995\linewidth]{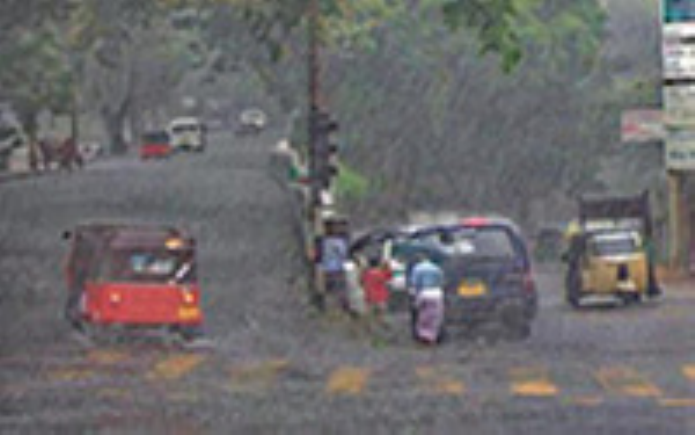}}
\end{minipage}
\hfill
\begin{minipage}{.115\linewidth}
\centering{\includegraphics[width=.995\linewidth]{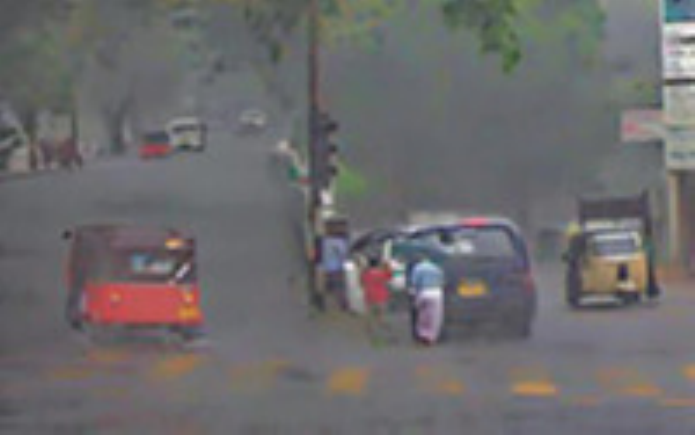}}
\end{minipage}
\hfill
\begin{minipage}{.115\linewidth}
\centering{\includegraphics[width=.995\linewidth]{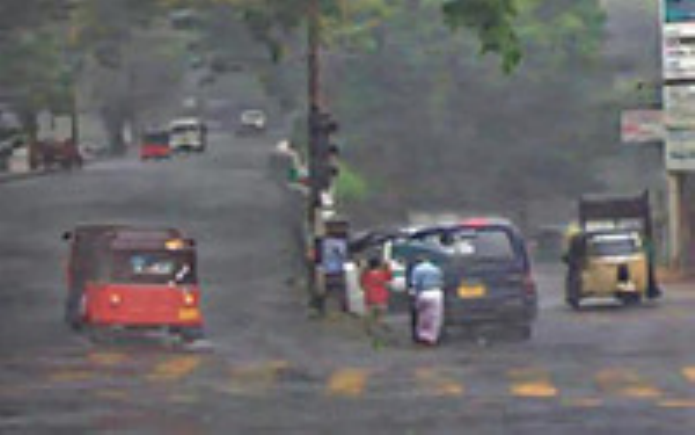}}
\end{minipage}
\hfill
\begin{minipage}{.115\linewidth}
\centering{\includegraphics[width=.995\linewidth]{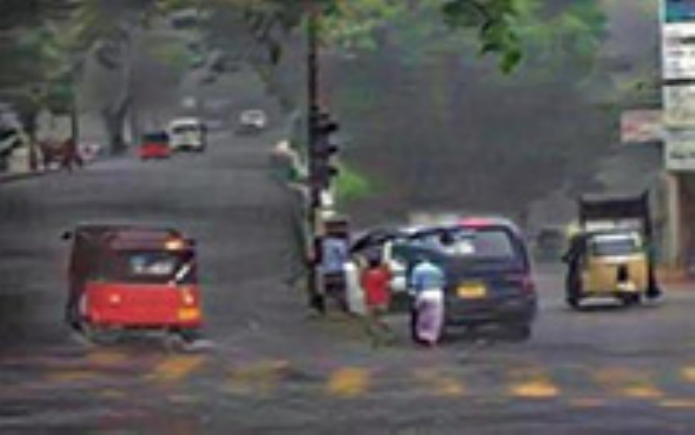}}
\end{minipage}
\hfill
\begin{minipage}{.115\linewidth}
\centering{\includegraphics[width=.995\linewidth]{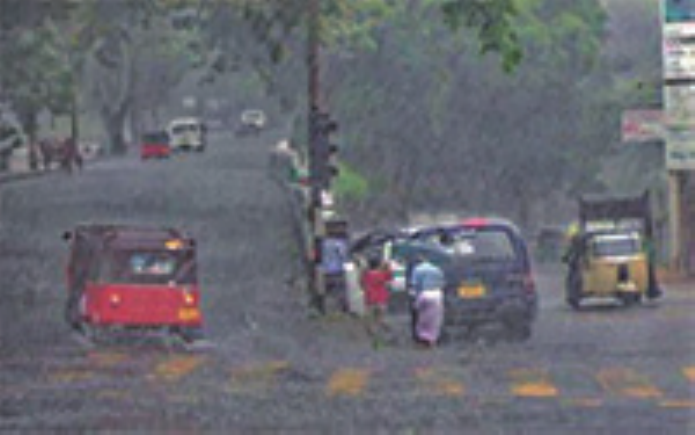}}
\end{minipage}
\vspace{0.5mm}
\vfill
\begin{minipage}{0.115\linewidth}
\centering{\includegraphics[width=.995\linewidth]{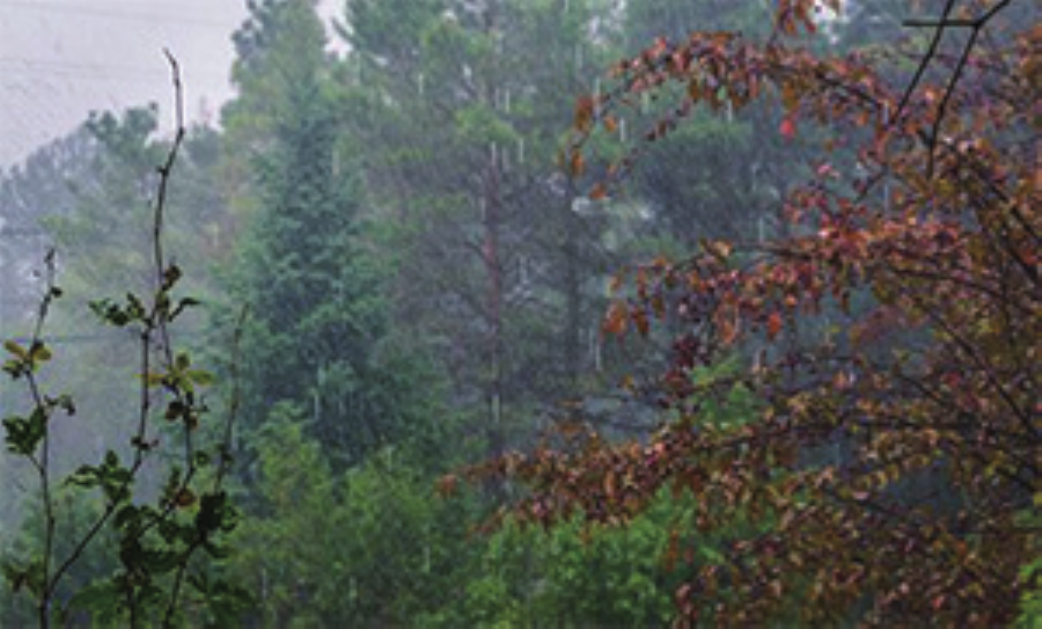}}
\centerline{(a)}
\end{minipage}
\hfill
\begin{minipage}{.115\linewidth}
\centering{\includegraphics[width=.995\linewidth]{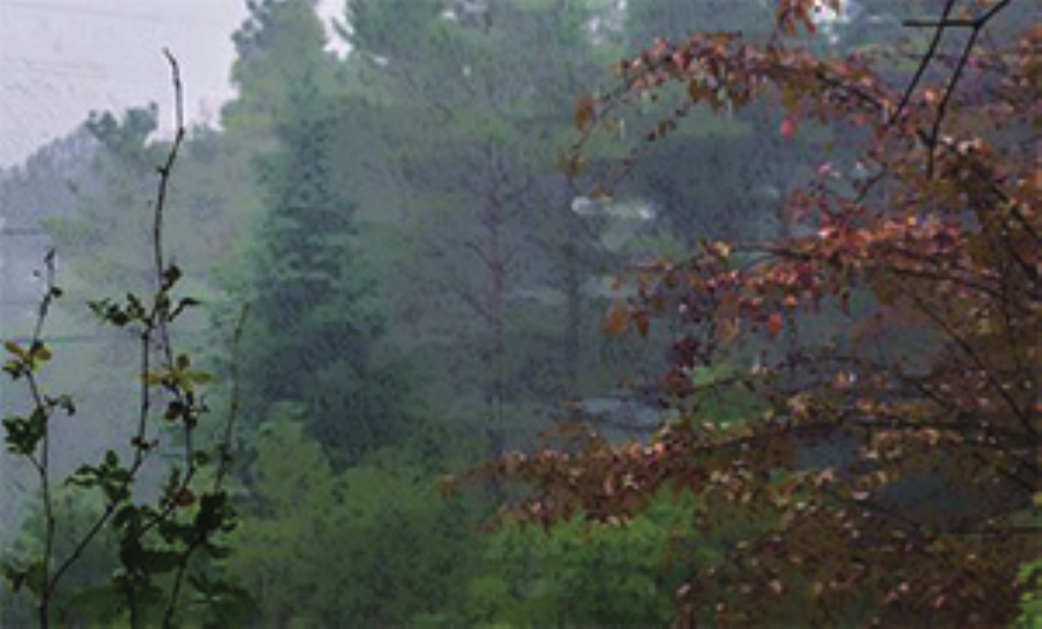}}
\centerline{(b)}
\end{minipage}
\hfill
\begin{minipage}{.115\linewidth}
\centering{\includegraphics[width=.995\linewidth]{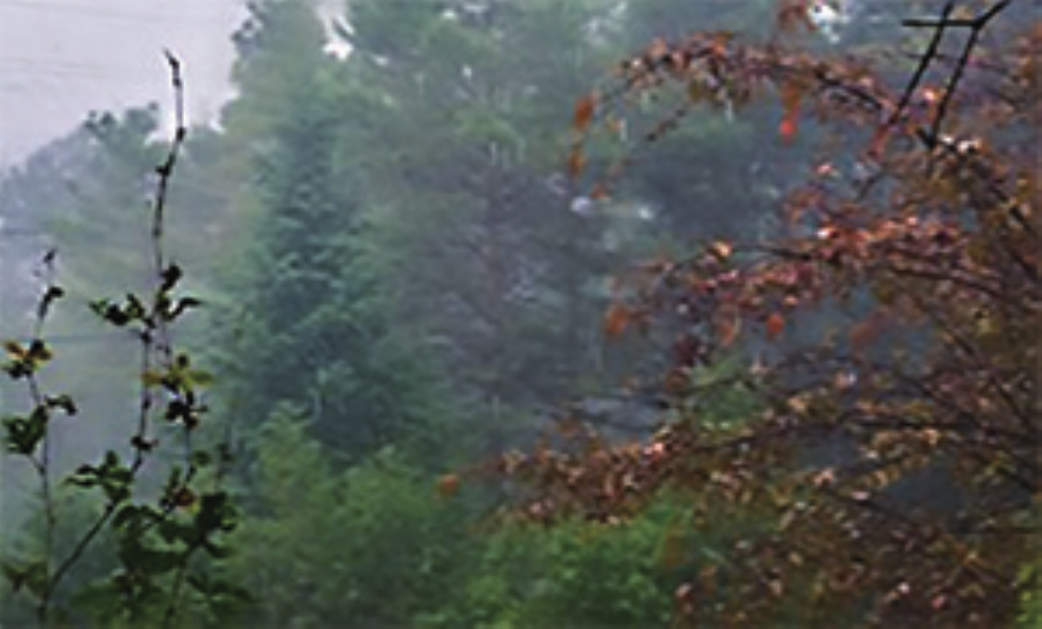}}
\centerline{(c)}
\end{minipage}
\hfill
\begin{minipage}{.115\linewidth}
\centering{\includegraphics[width=.995\linewidth]{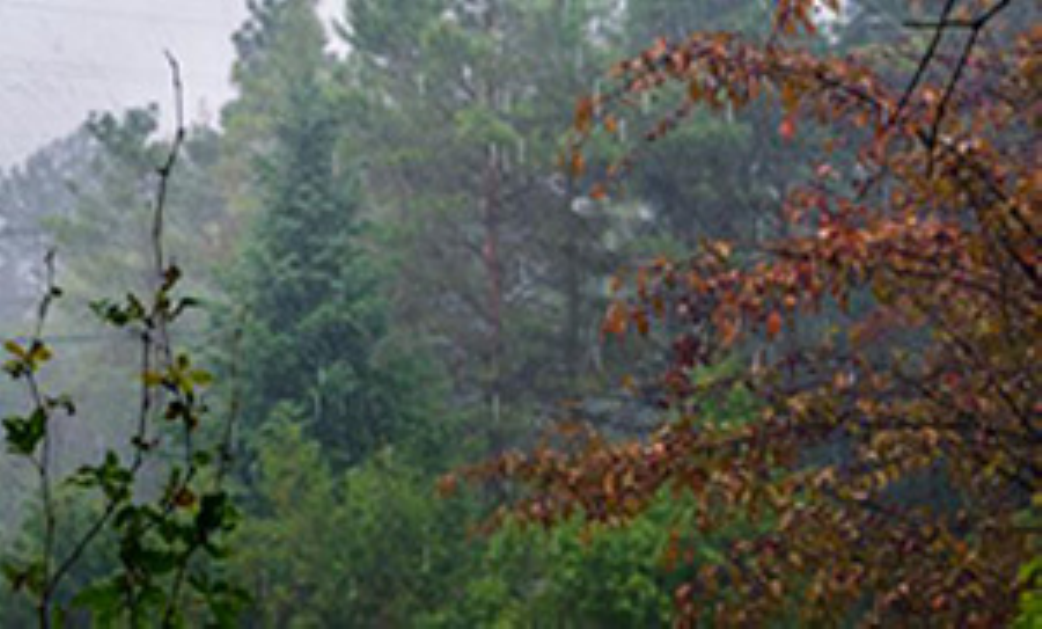}}
\centerline{(d)}
\end{minipage}
\hfill
\begin{minipage}{.115\linewidth}
\centering{\includegraphics[width=.995\linewidth]{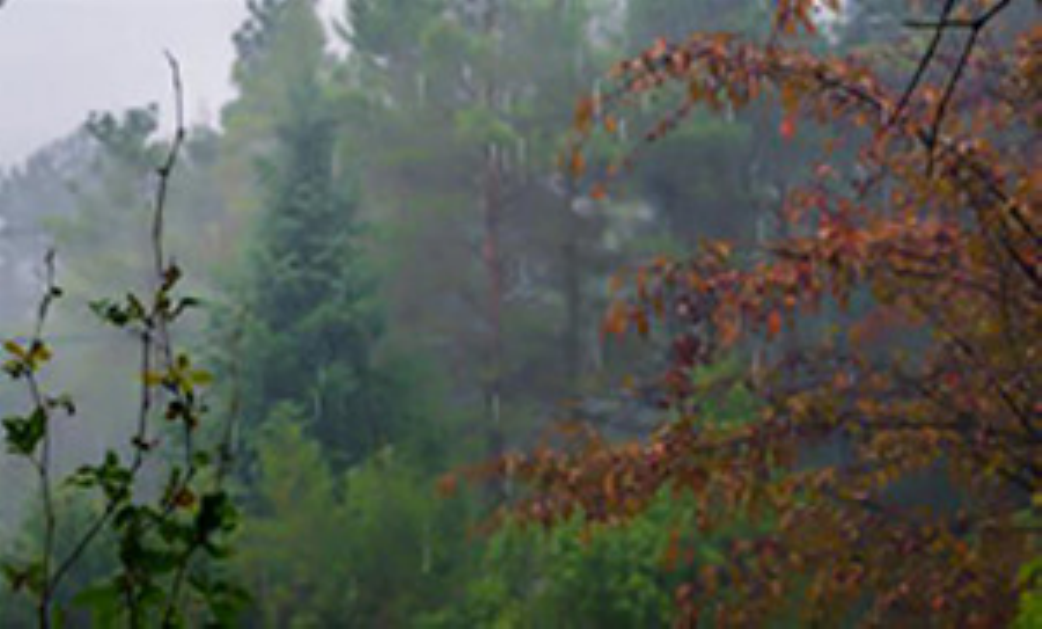}}
\centerline{(e)}
\end{minipage}
\hfill
\begin{minipage}{.115\linewidth}
\centering{\includegraphics[width=.995\linewidth]{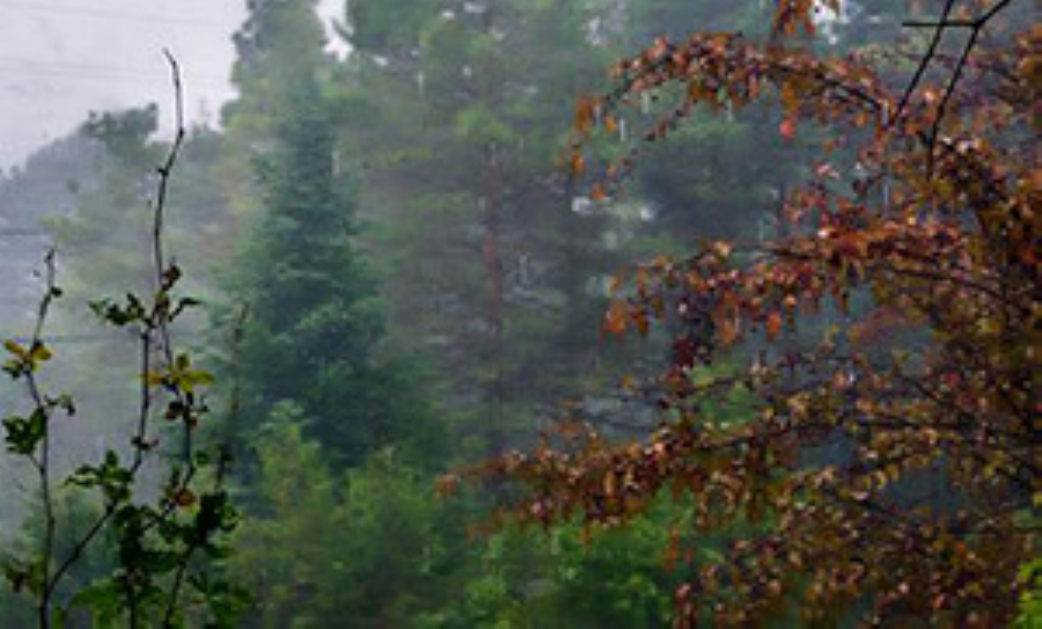}}
\centerline{(f)}
\end{minipage}
\hfill
\begin{minipage}{.115\linewidth}
\centering{\includegraphics[width=.995\linewidth]{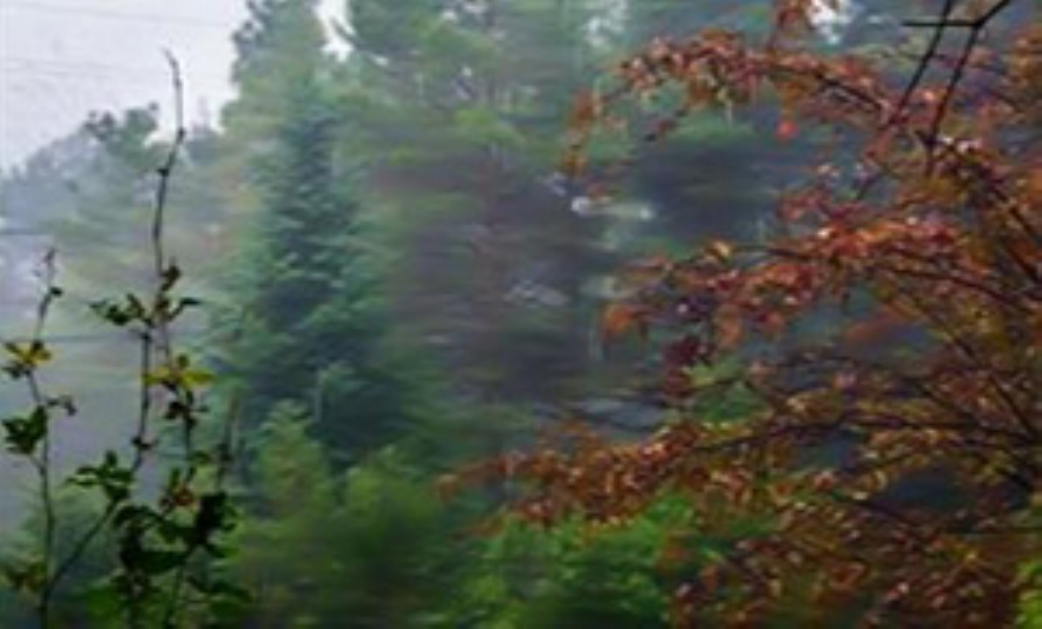}}
\centerline{(g)}
\end{minipage}
\hfill
\begin{minipage}{.115\linewidth}
\centering{\includegraphics[width=.995\linewidth]{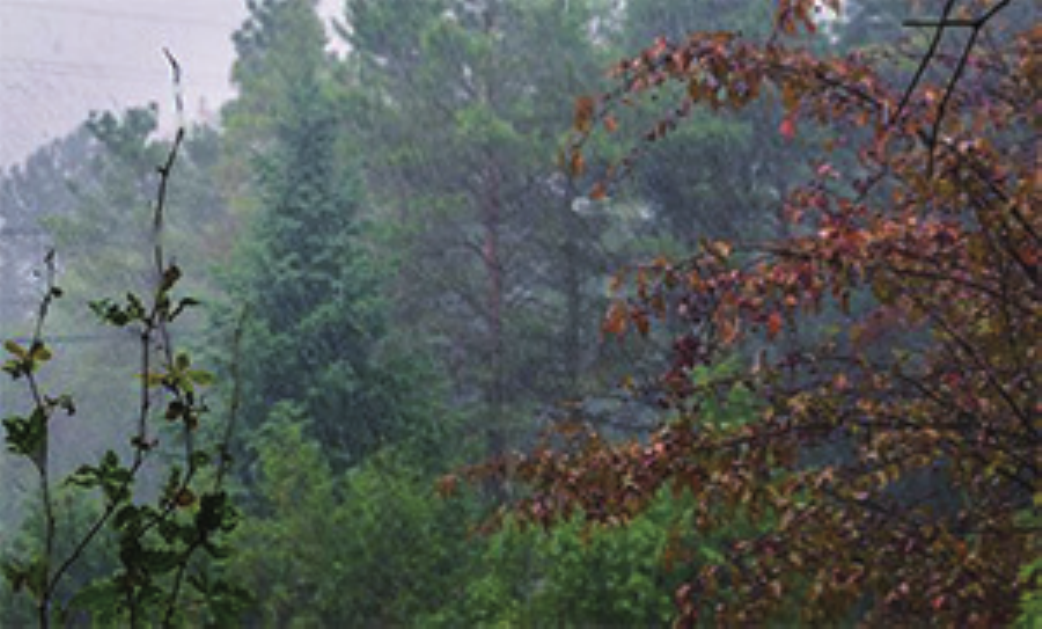}}
\centerline{(h)}
\end{minipage}
\vspace{0.5mm}
\caption{(a) Original rain images. (b) Results by Ding \emph{et al.} in \cite{Ding_2015_MTA}.
(c) Results by Chen \emph{et al.} in \cite{Chen_2014_CSVT}. (d) Results by Luo \emph{et al.} in \cite{Luo_2015_ICCV}.
(e) Results by Li \emph{et al.} in \cite{Li_2016_CVPR}.
(f) Results by Fu \emph{et al.} in \cite{Fu_2017_CVPR}. (g) Results by Zhang \emph{et al.} in \cite{Zhang_2018_CVPR}.
(h) Results by our proposed method.}
\label{fig:result_compare1}
\end{figure*}

\subsection{Results analysis}

In this subsection, we try to analyze the rain removal effectiveness
of different methods. The advantages or disadvantages of different
methods are discussed according to the rain-removed results that
is obtained by applying the selected methods on practical rain images. Notice that some images employed in these experiments have a large size so that the rain streaks look tenuous.

\textbf{Method by Ding \emph{et al.}:} The first row of Fig. \ref{fig:result_compare} shows a rain image with slight rain streaks. The result by Ding \emph{et al.}, as shown in the second column, seems to have removed the rain streaks quite well at the first glance. However, when the picture is zoomed in, it is found that a lot of non-rain details are lost. To verify this point more clearly, a small part of the rain picture and its corresponding rain-removed results by the selected methods are shown in the second row of Fig. \ref{fig:result_compare}. Now, it becomes obvious that some details of tree leaves have been removed together with the rain streaks. This is due to the threshold of the $L_0$ filters used in \cite{Ding_2015_MTA}: some non-rain objects whose size is relatively small would be mistreated as rain streaks and get removed.

The third row is still a slight rain image, but the rain streaks are denser. When zoomed in, the details-losing becomes more apparent. For heavy rain streaks in images shown in the sixth, seventh, eighth rows of Fig. \ref{fig:result_compare}, they can not be removed by the method of Ding \emph{et al.}. This is because the size of rain streaks in these images is beyond the preset threshold of $L_0$ filters. If we set the threshold larger, the rain streaks with wide size will be removed. However more details in the images will also be removed at the same time. For the light rain images which have less tenuous details (the third, forth and sixth rows in Fig. \ref{fig:result_compare1}), this method has satisfactory rain removal effectiveness.

\textbf{Method by Chen \emph{et al.}:} The results by Chen \emph{et al.} are shown in the third column. For the light rain images that have less subtle details (such as the image in the fifth row of Fig. \ref{fig:result_compare}, the third, forth and sixth rows in Fig. \ref{fig:result_compare1}), this method can obtain good rain removal results. However, if the rain images possess subtle details (such as the first, third and forth rows of Fig. \ref{fig:result_compare}), the detail-losing and image-blurring are inevitable. The reason is that the HOG descriptor used here cannot separate rain streaks and subtle details well.
The lost details can be seen clearly in third image of the second row of Fig. \ref{fig:result_compare}, which is obtained by zooming in a part of the image in the first row. Moreover, low-pass filter cannot filter bright
rain streaks completely. Consequently, the method by Chen \emph{et al.} can not deal with heavy rain images (such as the images in the sixth and seventh rows of Fig. \ref{fig:result_compare}).

\textbf{Method by Luo \emph{et al.}:} The results by Luo \emph{et al.} are in the forth column of Fig. \ref{fig:result_compare} and \ref{fig:result_compare1}. Obviously, this method can not remove rain streaks well. This is due to the discrimination of the sparse code used in this work, which is not good to separate a rain image into the rain layer
and non-rain layer. However, this method can make the intensity of rain streaks a little weaker. Hence, for tenuous rain streaks considered in their work, their method seems to have removed rain well. When rain steaks become brighter or wider, they can not be removed well.

\textbf{Method by Li \emph{et al.}:} Li \emph{et al.} used priors for both background and rain layer (which are based on Gaussian mixture models) to remove rain streaks. We show the results by this method in the fifth column. For the images that have little subtle details (the fifth, seventh, ninth, and tenth image in Fig. \ref{fig:result_compare}, as well as
the third, forth, and sixth images in Fig. \ref{fig:result_compare1}),
this method can obtain good rain-removal effectiveness. However, for rain images that have subtle details (e.g., the first, third and forth of Fig. \ref{fig:result_compare}), many subtle details are lost. This point can be seen clearly in the fifth image of the second row of Fig. \ref{fig:result_compare}.
As mentioned above, this image is part of the image in the first row that is zoomed in to see the details more clearly.

\textbf{Method by Fu \emph{et al.}:} For the majority of selected practical images, this work can achieve good results.
But there are still some defects. The first apparent one is that this method can cause slight blur for some rainy images,
such as the second and eighth images in Fig. \ref{fig:result_compare}. That is also the reason that this method has
lower PNSR/SSIM values than ours for the images in Fig. \ref{fig:result_render_compare}. The second is the generalization.
This method can not handle some rain images. For example, the seventh and eighth images in Fig. \ref{fig:result_compare}, the rain streaks are left in the results.

\textbf{Method by Zhang \emph{et al.}:} The work by Zhang \emph{et al.} is the most recent work
published on CVPR. We can see that this method faces the similar problems as the work by Fu \emph{et al.} \cite{Fu_2017_CVPR}.
The details lose seriously for some practical images, especially, the images with slim details (the last one in Fig.
\ref{fig:result_compare} and \ref{fig:result_compare1} separately, you can enlarge the images in this paper to see clear).
This method also can not deal with some rainy images, and some apparent rain streaks are left in some rain-removed results.

\textbf{Our work:} The results by our proposed method are shown in the eighth column. Compared with other traditional rain removal works, our proposed approach achieves better rain removal results. When compared with deep learning based works,
our method produces comparable results for majority of rain images. But for some other rain images, the selected deep learning
based methods can not handle well and better rain-removed result are obtained by our method.
Because our method acquires relatively more accurate locations, the remaining image details can be preserved well. Besides, the image quasi-sparsity prior offers a robust tool to the image recovery. Hence, better PSNR/SSIM values and good visual quality have been achieved in our proposed method.

\subsection{Limitations}

By experiments, our method can deal with majority of rain images.
However, every algorithm has its drawbacks, so does our method. For some images with non-rain objects that are very similar to the shape and color of rain streaks, some mis-detections are inevitable. This will result in the loss of some useful information. Besides, when the rain is very heavy, the rain streak will be combined to produce fog. A shallow thought for this situation is that we can remove the rain streaks by our method first, and a dehaze method can be used to remove haze which is caused by heavy rain. We note that this situation has been discussed in a very recent work in \cite{Li_2016_CVPR}. We will continue to work on this situation in our future work. Another future work is to further improve the rain detection.

\section{Conclusions}
\label{sec:Conclusion}
In this paper, we have proposed a new rain streaks detection and removal method from a single color image. Our results suggest that using a morphological image processing to extract connected components and quantifying the characteristics of extracted connected components by
the principal component analysis (PCA) are effective in detecting rain streaks. Once rain streaks are detected, we employ an image sparsity prior to accurately decompose a rain image into the rain layer and non-rain layer, which has also been proven to be effective. In addition, quantitative (objective) evaluations and an user study (subjective) validate the overall rain removal effectiveness of our method, which outperforms four selected traditional methods and is comparable to the most recent deep learning based works that are all proposed very recently and widely regarded as the state-of-the-art.


%

%
%
%
%
%

\ifCLASSOPTIONcaptionsoff
  \newpage
\fi

\end{document}